%% file: main.tex
\documentclass{article}
\usepackage{graphicx,color} 

\title{Low-dimensional adaptation of diffusion models: \\ Convergence 
in total variation}
\author{%
Jiadong Liang\thanks{Department of Statistics and Data Science, the Wharton School, University of Pennsylvania; email: \texttt{\{jdl97,zhihanh,yuxinc\}@wharton.upenn.edu}.} 
\and
Zhihan Huang\footnotemark[1] 
\and
 Yuxin Chen\footnotemark[1]
}
\date{January, 2025;~~Revised: July 2026}

\usepackage{geometry}

\usepackage[unicode=true,
 bookmarks=false,
 breaklinks=false,pdfborder={0 0 1},colorlinks=false]
 {hyperref}

\hypersetup{colorlinks,citecolor=blue,filecolor=blue,linkcolor=blue,urlcolor=blue}

\usepackage{fullpage,psfrag,enumerate,color}
\usepackage{natbib}
\usepackage{ulem}
\usepackage{amsmath,mathtools,amssymb}
\usepackage{amsfonts,comment}
\usepackage[ruled,vlined]{algorithm2e}
\usepackage{enumitem}
\usepackage{enumerate}
\usepackage{subfiles}
\usepackage{subfigure,float,bbm,cleveref}
\usepackage{multirow}
\usepackage{makecell}
\usepackage{rotating}
\usepackage{cases}
\usepackage{tablefootnote}
\usepackage{ulem}
\usepackage{algorithmic}
\usepackage{cleveref}
\usepackage{xcolor}
\usepackage{colortbl}
\usepackage{appendix}
\usepackage{caption}
\usepackage{booktabs}
\usepackage{tabularx} 
\usepackage{tikz}
\usetikzlibrary{positioning}
\usetikzlibrary{arrows.meta}

\allowdisplaybreaks

\newcolumntype{Y}{>{\centering\arraybackslash}X}

\providecommand{\U}[1]{\protect\rule{.1in}{.1in}}
\providecommand{\U}[1]{\protect\rule{.1in}{.1in}}
\newtheorem{thm}{Theorem}

\newtheorem{proposition}{Proposition}
\newtheorem{lemma}{Lemma}
\newtheorem{corollary}{Corollary}
\newtheorem{assumption}{Assumption}
\newtheorem{remark}{Remark}
\newtheorem{definition}{Definition}
\newcommand{\mymid}{\,|\,}
\newcommand{\EE}{\mathbb{E}}
\newcommand{\PP}{\mathbb{P}}

\input{math_command}

\newtheorem{proof}{Proof}

\usepackage{pifont}
\newcommand{\Checkmark}{\ding{51}} 
\newcommand{\XSolidBrush}{\ding{55}} 

\usepackage{array}

\definecolor{yxc}{RGB}{255,0,0}

\allowdisplaybreaks

\begin{document}

\maketitle

\input{abstract.tex}
\renewcommand{\thefootnote}{}
\footnotetext{Accepted for presentation at the Conference on Learning Theory (COLT) 2025.}
\addtocounter{footnote}{-1}

\setcounter{tocdepth}{2}
\tableofcontents

\input{intro.tex}

\input{preliminary.tex}
\input{main_results.tex}

\subsection{Other alternatives of coefficient design?}
\label{sec:necessity_opt_step_size}
\input{Necessity_of_opt_step_size}

\input{related_works}

\input{discussion}

\section*{Acknowledgments}

Y.~Chen is supported in part by the Alfred P.~Sloan Research Fellowship, the AFOSR grant FA9550-22-1-0198, the ONR grants N00014-22-1-2354 and N00014-25-1-2344,   the NSF grants 2221009 and 2218773, and the Amazon Research Award. 
We thank Yuting Wei and Yuchen Wu for their helpful discussions.

\normalem

\appendix



\input{technical_lemmas}

\input{DDIM_Analysis}

\input{DDPM_Analysis}

\input{equivalence_to_the_Denoising_Diffusion_Implicit_Model}

\input{proof_propositions}

\input{proof_lower_bound}

\input{Auxiliary_lemmas}

\input{training_complexity}

\bibliographystyle{apalike}
\bibliography{reference}
\end{document}

%% file: math_command.tex
\def\rd{{\textnormal{d}}}

\DeclareMathAlphabet{\mathsfit}{\encodingdefault}{\sfdefault}{m}{sl}
\SetMathAlphabet{\mathsfit}{bold}{\encodingdefault}{\sfdefault}{bx}{n}

\def\gA{{\mathcal{A}}}
\def\gB{{\mathcal{B}}}
\def\gC{{\mathcal{C}}}

\def\gE{{\mathcal{E}}}

\def\gG{{\mathcal{G}}}
\def\gH{{\mathcal{H}}}
\def\gI{{\mathcal{I}}}
\def\gJ{{\mathcal{J}}}
\def\gK{{\mathcal{K}}}
\def\gL{{\mathcal{L}}}
\def\gM{{\mathcal{M}}}
\def\gN{{\mathcal{N}}}

\def\gR{{\mathcal{R}}}
\def\gS{{\mathcal{S}}}
\def\gT{{\mathcal{T}}}

\def\gX{{\mathcal{X}}}

\def\0{{\bf 0}}
\def\1{{\bf 1}}

\def\EB{{\mathbb E}}

\def\PB{{\mathbb P}}

\def\RB{{\mathbb R}}

\def\tr{\mathrm{tr}}

\newcommand{\R}{\mathbb{R}}

\newcommand{\Cov}{\mathrm{Cov}}

\newcommand{\KL}{\mathsf{KL}}
\newcommand{\ssum}[3]{\sum\limits_{{#1}={#2}}^{#3}}

\newcommand{\TV}{\mathsf{TV}}

\newcommand{\pdata}{p_{\mathsf{data}}}

\def\inner#1#2{\left\langle #1, #2 \right\rangle}
\newcommand{\norm}[1]{\left\|{#1}\right\|}
\newcommand{\abs}[1]{\left|{#1}\right|}

%% file: abstract.tex
\begin{abstract}
This paper investigates how diffusion generative models leverage (unknown) low-dimensional structure to accelerate sampling. Focusing on two mainstream samplers --- the denoising diffusion implicit model (DDIM) and the denoising diffusion probabilistic model (DDPM), we prove that their iteration complexities under exact score functions are at most the order of $k/\varepsilon$ (up to log factor), 
where $\varepsilon$ is the precision in total variation distance and $k$ is some intrinsic dimension of the target distribution. We further extend these convergence guarantees to the setting in which the score functions are learned from data rather than known exactly,  showing that the convergence performance degrades gracefully under suitable score estimation assumptions. We then show that these assumptions are attainable via kernel-based score estimators with finite-sample guarantees that also adapt to the low-dimensional structure.  
Our results apply to a broad family of target distributions without requiring smoothness or log-concavity.  
Our findings provide the first rigorous evidence for the adaptivity of the DDIM-type samplers to unknown low-dimensional structure, and 
 improve over the state-of-the-art DDPM theory regarding total variation convergence. 
\end{abstract}


%% file: intro.tex
\section{Introduction}

As a cornerstone of the rapidly evolving field of generative AI, diffusion generative models have driven mind-blowing progress across a diverse range of applications, such as image and video generation, medical image analysis, and time-series forecasting, to name just a few \citep{ramesh2022hierarchical,croitoru2023diffusion,kazerouni2023diffusion}. 
The remarkable effectiveness of diffusion models has inspired a recent wave of activity aimed at developing and strengthening their theoretical underpinnings.

\subsection{Score-based generative modeling: DDPM and DDIM}
At their core, diffusion models seek to gradually transform pure noise into new samples that emulate a $d$-dimensional target distribution $\pdata$, accomplished by learning to reverse a forward stochastic process that progressively converts data into noise, detailed below.

\medskip
\noindent
{\bf Forward process.} 
A common choice of the forward process with  horizon $T$ is given by 
\begin{align}
    \label{eq:forward}
    X_0 \sim \pdata, 
    \qquad  
    X_t = \sqrt{\alpha_{t}}\, X_{t-1} + \sqrt{1-\alpha_t}\, W_t, \quad t = 1, \cdots, T,
\end{align}
where $\{W_t\}_{t=1}^T$ are independent noise vectors obeying $W_i\overset{\mathrm{i.i.d.}}{\sim} \gN(0, I_d)$, 
and the sequence  \(\{\alpha_t\}_{t=1}^T \subseteq (0,1) \) 
controls the variance of the Gaussian noise injected in each step. 
Informally, as $T$ grows,
the distribution of $X_t$ typically converges to the standard Gaussian $\mathcal{N}(0,I_d)$.

\medskip
\noindent
{\bf Reverse process and  diffusion-based samplers.} 
As it turns out, 
the forward Markov process \eqref{eq:forward} is reversible in general,  a property that follows from classical results in the stochastic differential equation (SDE) literature \citep{anderson1982reverse,haussmann1986time}. 
This intriguing property underpins the data generation process of diffusion models, 
which involves constructing a reverse process $Y_T\rightarrow \dots \rightarrow Y_1\rightarrow Y_0$ 
that closely mimics the forward process \eqref{eq:forward} in the sense that $Y_t \overset{\mathrm{d}}{\approx} X_t$ for each step $t$. 
Crucially, the reversal of the forward process hinges upon access to the so-called (Stein) score function
\begin{align}
    s_t^{\star}(X) \coloneqq \nabla \log p_{X_t}(X)  
    \label{eq:defn-score}
\end{align}
--- hence the term ``score-based generative modeling.''  
To formalize the sampling process,  
one needs to specify the initialization and iterative steps of the reverse process. 
The initialization step is straightforward: 
given that $X_T$ is approximately Gaussian for large enough $T$, one generic choice is to draw $Y_T$ as  pure noise $\mathcal{N}(0,I_d)$.  
As such, a key step underlying the design of the sampling process boils down to how to update $Y_t$ at  each step while maintaining the desired distributional proximity.
In what follows, we single out two mainstream paradigms, assuming availability of an estimate $s_t$ of the true score function $s_t^{\star}$ at each $t$:  
\begin{itemize}

    \item {\it Denoising Diffusion Implicit Model (DDIM).} 
    The DDIM sampler (or the probability flow ODE sampler) \citep{song2020denoising} adopts a deterministic update rule below: 
    \begin{align}\label{eq:DDIM-update}
    Y_T \sim \mathcal{N}(0,I_d), \qquad 
    Y_{t-1} = \frac{1}{\sqrt{\alpha_t}}\big(Y_t + \eta_t^{\mathsf{ddim}} s_t(Y_t)\big),    
    \qquad t = T,\cdots, 1,
    \end{align}
where $\{\eta_t^{\mathsf{ddim}}\}$ represents some suitably chosen coefficients. 
    In words, each step \eqref{eq:DDIM-update} computes $Y_{t-1}$ as a weighted sum of $Y_t$ and its score estimate.

    \item {\it Denoising Diffusion Probabilistic Model (DDPM).} Originally proposed by \citet{ho2020denoising} as a way to optimize certain variational lower bounds on the log-likelihood, DDPM employs the following stochastic iterative updates: 
    \begin{align}
    Y_T \sim \mathcal{N}(0,I_d), \qquad 
    Y_{t-1} = \frac{1}{\sqrt{\alpha_t}}\big(Y_t + \eta_t^{\mathsf{ddpm}} s_t(Y_t) + \sigma_t^{\mathsf{ddpm}} Z_t\big),
    \quad t = T,\cdots, 1,
    \label{eq:DDPM-update}
    \end{align}
    where the $Z_t$'s are independently generated obeying $Z_t\sim \mathcal{N}(0,I_d)$, and $\{\eta_t^{\mathsf{ddpm}}\}$ and $\{\sigma_t^{\mathsf{ddpm}}\}$ are properly chosen coefficients. 
A key distinction from DDIM is that the iterative updates \eqref{eq:DDPM-update} inject additional stochastic noise at each step.

\end{itemize}

\subsection{Harnessing low-dimensional structure?}

Motivated by the practical efficacy of diffusion models, the past few years have witnessed a flurry of activity towards establishing convergence theory for both DDPM and DDIM \citep{lee2022convergence,chen2022sampling,chen2023improved,chen2023restoration,chen2024probability,benton2024nearly,benton2023error,li2024towards,li2024sharp,li2024unified,gao2024convergence,huang2024convergence,li2024d,liang2024non,li2024improved}. For a fairly general family of target distributions $\pdata$ (without assuming smoothness and log-concavity), the state-of-the-art theory \citet{li2024adapting,li2024sharp} demonstrated that for both DDPM and DDIM, it takes at most the order of (modulo some log factor)
\begin{align}
    d/\varepsilon~\text{ iterations }
    \label{eq:state-of-the-art-convergence}
\end{align}
to yield a sample whose distribution is $\varepsilon$-close in total variation (TV) distance to the target distribution,  
provided that perfect score function estimates are available. 

Nevertheless, even linear scaling in $d$ can be prohibitively expensive for many contemporary applications.  
Take the ImageNet dataset \citep{deng2009imagenet} for instance: each image might contain 150,528 pixels, 
while its intrinsic dimension is estimated to be 43 or less \citep{pope2021intrinsic}. 
As a result, applying the state-of-the-art theory  \eqref{eq:state-of-the-art-convergence} could suggest an iteration complexity that exceeds one million, 
even though practical implementations of DDIM and DDPM often produce high-quality samples in just a few hundred (or even a few ten) iterations. 
The discrepancy between theory and practice suggests that worst-case bounds, such as \eqref{eq:state-of-the-art-convergence}, may be overly conservative. 
To reconcile this discrepancy, it is crucial to bear in mind the intrinsic dimension of the target data distribution and explore whether and how diffusion models can harness this potentially low-dimensional structure.

The development of diffusion model theory that can effectively account for low dimensionality is, however, still in its early stages. For example, the ability of DDIM to adapt to low-dimensional structure was previously completely out of reach in theory, despite its widespread use.  
The situation for DDPM is more advanced:  recent work 
\citet{li2024adapting,azangulov2024convergence,potaptchik2024linear,huang2024denoising}   explored its low-dimensional adaptation capability, assuming that the target data distribution is supported on some low-dimensional structure like a manifold. These studies focused primarily on convergence in Kullback–Leibler (KL) divergence, 
which
is known to yield loose results when directly translated into  convergence guarantees based on other metrics like the TV distance.

\subsection{This paper}

\medskip
\noindent
{\bf An overview of our contributions.} 
In this paper, we develop a new suite of total-variation-based theoretical guarantees for the DDIM and DDPM samplers to uncover how they leverage low-dimensional structure to accelerate sampling. Concretely, 
consider a general definition of  intrinsic dimension for the target distribution $\pdata$, such that the intrinsic dimension is $k$ if the logarithm of the covering number of the support of $\pdata$ is on the order of $k$ (up to some log factor).  Under this notion, we first prove in Theorem~\ref{thm:true-score-convergence} that, with exact score functions, both DDPM and DDIM require at most the order of 
\begin{align}
    k/\varepsilon~\text{ iterations ~~(up to log factor)}
    \label{eq:state-of-the-art-convergence-low-dim}
\end{align}
to generate a sample that is $\varepsilon$-close in TV distance to the target distribution. This result applies to a broad class of target distributions without requiring restrictive assumptions such as smoothness or log-concavity. For those applications where $k\ll d$ --- a situation that is prevalent in many modern-day applications --- our theory underscores the striking capability of diffusion models to automatically exploit the favorable intrinsic structure of $\pdata$ without explicitly modeling the low-dimensional structure or altering the algorithms.

Next, we extend these guarantees to the more realistic setting in which the score functions are learned from data rather than known exactly. Theorems~\ref{thm:ddim_tv_conv}, \ref{thm:vanilla_ddpm} and \ref{thm:ddpm_tv_conv} show that, under suitable score estimation assumptions, 
 the convergence performance of both DDIM and DDPM degrade gracefully as the score estimation accuracy worsens. We further establish that these score estimation assumptions are achievable by kernel-based score estimators, deriving finite-sample guarantees that automatically adapt to the underlying low-dimensional structure of \(\pdata\) (see Theorems~\ref{thm:ddim-training-guarantee} and \ref{thm:ddpm-training-guarantee}).

Furthermore, we illuminate the specific coefficient choices of the DDIM/DDPM samplers,  by linking them with reverse-time differential equations with specific  discretization to exploit low dimensionality. 
We also develop a lower bound for a single step of the discretized reverse process, which unravels the necessity and optimality of the coefficient designs proposed originally by \cite{ho2020denoising,song2020denoising}.

It is worth emphasizing that  
our results provide the first theory justifying the low-dimensional adaptation ability of the DDIM-type samplers, and significantly improve over the state-of-the-art DDPM theory regarding total variation convergence;  see Table~\ref{tab:DDIM} and Table~\ref{tab:DDPM} for detailed comparisons with prior DDIM and DDPM theory, respectively.\footnote{Note that in a large fraction of prior DDPM theory, the bound based on the TV distance is obtained by applying Pinsker's inequality (i.e., $\mathsf{TV}(p_{X_1},q_{Y_1})\leq \sqrt{2\mathsf{KL}(p_{X_1} \parallel p_{Y_1})}$\,). } 
%


\newcommand{\topsepremove}{\aboverulesep = 0mm \belowrulesep = 0mm} \topsepremove 

\input{tabular_comparison_ddim}

\medskip
\noindent
{\bf Notation.} 
For any positive integer $n$, let $[n]=\{1,\dots,n\}$. 
For any $x\in\mathbb{R}^d$ and $r>0$, define $\mathcal{B}(x,r)\coloneqq \{z\in\mathbb{R}^d:\|z-x\|_2\le r\}$ as the closed $\ell_2$-neighborhood of radius $r$ centered at $x$.
For two functions $f$ and $g$, we employ the  notation $f = O(g)$ or $f\lesssim g$ to mean that there exists some universal constant $C > 0$ such that $f \le C g$. 
The notation $f = \widetilde{O}(g)$ is defined analogously except that the log dependency is hidden. Additionally, $f\gtrsim g$ means $g\lesssim f$, and $f\asymp g$ means  $f\lesssim g$ and $g \lesssim f$ hold at once. 
For any two distributions $p$ and $q$, we denote by $\mathsf{TV}(p, q)$ (resp.~$\mathsf{KL}(p \parallel q)$) the TV distance (resp.~KL divergence) between $p$ and $q$. 
We denote by $p_{X_t}$ and $p_{Y_t}$ the probability density function of $X_t$ and $Y_t$, respectively.
For any matrix $A$, we denote by $\|A\|$ (resp. $\|A\|_{\mathrm{F}}$) the spectral norm (resp. Frobenius norm) of $A$, and $\mathrm{tr}(A)$ the trace of $A$. For any vector $A$, write $A^{\otimes 2}\coloneqq AA^\top$.
For any vector-valued function $f(x)$ , we let $\frac{\partial f}{ \partial x}$ represent the Jacobian matrix of $f(x)$; for any real-valued function $g(x)$, we let $\nabla g(x)$ represent the gradient of $g(x)$. Also, for any random object $X$, we denote by $\mathsf{supp}(X)$ the support of $X$. 
For any two $a,b\in \mathbb{R}$, define $a \wedge b \coloneqq \min\{a,b\}$.

\input{tabular_comparison_ddpm}

%% file: tabular_comparison_ddim.tex
\begin{table}[t]
    \centering
    \renewcommand\arraystretch{1.4} 
    \fontsize{7}{8}\selectfont
    \begin{tabular}
    {>{\centering\arraybackslash}m{2.1cm}|>{\centering\arraybackslash}m{2.3cm}|>
    {\centering\arraybackslash}m{2.5cm}|>
    {\centering\arraybackslash}m{2.8cm}|>
    {\centering\arraybackslash}m{2.3cm}|>
    {\centering\arraybackslash}m{1.8cm}}
    \toprule
    \textbf{paper} & \textbf{smoothness of scores} &
    \textbf{score matching assumption}
    &\textbf{convergence rate (in total variation)} &
    \textbf{iteration complexity}
    &\textbf{adaptation to low dimension}\\
    \hline
    \citet{chen2023restoration}& $L$-Lipschitz & $s_t = s_t^{\star}$ &$\mathrm{poly}(Ld)\big/\sqrt{T}$ &
    $\mathrm{poly}(Ld)\big/\varepsilon^2$
    &{\centering\XSolidBrush}\\
    \hline
    \citet{li2023towards}& no requirement & $s_t \approx s_t^{\star}$, $\frac{\partial s_t}{\partial x} \approx \frac{\partial s_t^{\star}}{\partial x}$
    &$d^2/T + d^6/T^2$&
    $d^2/\varepsilon + d^3/\sqrt{\varepsilon}$
    & {\centering\XSolidBrush} \\
    \hline
    \citet{huang2024convergence} &$L$-Lipschitz&
    $s_t {\approx} s_t^{\star}$
    & $L^2d^{2}/T$&
    $L^2d^2\big/ \varepsilon$
    & {\centering\XSolidBrush}\\
    \hline
    \citet{li2024sharp}& no requirement &
    $s_t \approx s_t^{\star}$, $\frac{\partial s_t}{\partial x} \approx \frac{\partial s_t^{\star}}{\partial x}$
    &$ d/T $ when $T> d^2$&
    $d/\varepsilon + d^2$
    & {\centering\XSolidBrush}\\
    \hline
    \citet{li2024unified}& $L$-Lipschitz & $s_t \approx s_t^{\star}$, $\frac{\partial s_t}{\partial x} \approx \frac{\partial s_t^{\star}}{\partial x}$ & $Ld(L+d)\big/ T$ &
    $Ld(L+d)\big/\varepsilon$
    &{\centering\XSolidBrush}\\
    \hline
    \rowcolor{gray!20}
    \makecell{Our work\\ (Theorem~\ref{thm:true-score-convergence},~\ref{thm:ddim_tv_conv})}& no requirement &
    $s_t \approx s_t^{\star}$, $\frac{\partial s_t}{\partial x} \approx \frac{\partial s_t^{\star}}{\partial x}$, $\nabla \tr(\frac{\partial s_t}{\partial x}){\approx} \nabla \tr(\frac{\partial s_t^{\star}}{\partial x})$
    &$k/T$&
    $k/\varepsilon$
    & {\centering\Checkmark}\\
    \bottomrule
    \end{tabular}
    \caption{
    Comparison with prior DDIM theory. The convergence rates and iteration complexities provided here assume accurate scores and ignore $\log$ factors,
    where
    the iteration complexity refers to the number of iterations needed to yield $\varepsilon$ precision in TV distance. \label{tab:DDIM}
    }
\end{table}

%% file: tabular_comparison_ddpm.tex
\begin{table}[!htb]
    \centering
    \renewcommand\arraystretch{1.15} 
    \fontsize{7}{8}\selectfont
    \begin{tabular}
    {>{\centering\arraybackslash}m{2.2cm}|>{\centering\arraybackslash}m{2.3cm}|>
    {\centering\arraybackslash}m{2.5cm}|>
    {\centering\arraybackslash}m{2.2cm}|>
    {\centering\arraybackslash}m{2.0cm}|>{\centering\arraybackslash}m{1.8cm}}
    \toprule
    \textbf{paper} & \textbf{smoothness of scores} &
    \textbf{score matching assumption}
    &\textbf{convergence rate (in total variation)} &
    \textbf{iteration complexity}
    &\textbf{adaptation to low dimension}\\
    \hline
    \citet{chen2022sampling} & $L$-Lipschitz &
    $s_t \approx s_t^{\star}$
    & $ L\sqrt{d/T}$&
    $L^2 d/\varepsilon^2$
    & {\centering\XSolidBrush}\\
    \hline
    \citet{lee2022convergence}& no requirement &
    $s_t \approx s_t^{\star}$
    & $\sqrt{d^{3}\big/ T}$
    & $d^3\big/ \varepsilon^2$
    &{\centering\XSolidBrush}\\
    \hline
    \citet{chen2023improved} & no requirement &
    $s_t \approx s_t^{\star}$
    &$\sqrt{d^2/T}$&
    $d^2/\varepsilon^2$
    & {\centering\XSolidBrush}\\
    \hline
    \citet{benton2024nearly}&
    no requirement&
    $s_t \approx s_t^{\star}$
    &$\sqrt{d/T}$ &
    $d/\varepsilon^2$
    & {\centering\XSolidBrush}\\
    \hline
    \citet{liang2024non} & no requirement &
    $s_t \approx s_t^{\star}$, ${\nabla s_t} \approx {\nabla s_t^{\star}}$ &
    $d^{3/2}\big/T$ & $d^{3/2}\big/\varepsilon$& {\centering\XSolidBrush}\\
    \hline
    \citet{li2024adapting} & no requirement &
    $s_t \approx s_t^{\star}$
    &$ k^2/\sqrt{T} $ &
    $k^4/\varepsilon^2$
    &{\centering\Checkmark}\\
    \hline
    \citet{li2024d}& no requirement&
    $s_t \approx s_t^{\star}$
    &$d/T$ &
    $d/\varepsilon$
    & {\centering\XSolidBrush}\\
     \hline
    \citet{azangulov2024convergence}& no requirement&
    $s_t \approx s_t^{\star}$
    &$\sqrt{k^3/T}$&
    $k^3/\varepsilon^2$
    & {\centering\Checkmark}
    \\
    \hline
    \citet{potaptchik2024linear}& no requirement &
    $s_t \approx s_t^{\star}$
    &$k/\varepsilon^2$
    &$\sqrt{k/T}$ & {\centering\Checkmark}
    \\
    \hline
    \citet{huang2024denoising}& no requirement &
    $s_t \approx s_t^{\star}$
    &$k/\varepsilon^2$
    &$\sqrt{k/T}$ & {\centering\Checkmark}
    \\
    \hline
    \rowcolor{gray!20}
    \makecell{Our work\\ (Theorems~\ref{thm:true-score-convergence}, \ref{thm:vanilla_ddpm}, \ref{thm:ddpm_tv_conv})}& no requirement&
    $s_t \approx s_t^{\star}$
    &$k/T$&
    $k/\varepsilon$  
    & {\centering\Checkmark}\\
    \bottomrule
    \end{tabular}
    \caption{
    Comparison with prior DDPM theory. The convergence rates and iteration complexities here assume accurate scores and ignore $\log$ factors, where the iteration complexity refers to the number of iterations needed to yield $\varepsilon$ accuracy in TV distance.\label{tab:DDPM}
    }
\end{table}

%% file: preliminary.tex
\section{Preliminaries}\label{sec:pre}
%
Before proceeding to our formal theory and analysis, we briefly overview some basics and the operational mechanism of diffusion models, covering both DDIM and DDPM.  


\medskip
\noindent
{\bf Forward process and noise schedule.} 
As previously described in \eqref{eq:forward}, 
the forward process progressively injects Gaussian noise to transform the target distribution $\pdata$ into a pure noise distribution that is easy to sample from. The Gaussian nature of the injected noise allows for a more direct relation between \(X_0\) and \(X_t\) as follows:
\begin{equation}
    \label{eq:forward-marginal}
    X_t = \sqrt{\overline{\alpha}_t}X_0 + \sqrt{1 - \overline{\alpha}_t}\,\overline{W}_t 
    \qquad \text{  with  } \overline{W}_t \sim \gN(0,I_d),
\end{equation}
where we introduce the following parameters for any $1\leq t\leq T$: 
\begin{align}
\label{eq:bar-alphat-discrete}
\overline{\alpha}_t \coloneqq \prod_{i=1}^t \alpha_i.  
\end{align}
%

As it turns out, the choices of the coefficients $\{\alpha_t\}$ play an important role in determining the convergence properties of diffusion models. 
Here and throughout, we adopt the choices used in the  previous work~\citep{li2024sharp,li2024adapting,li2024d}:
\begin{subequations}
    \label{eq:noise-schedule}
\begin{align}
    \beta_1 &\coloneqq 1-\alpha_1= \frac{1}{T^{c_0}},\\
    \beta_{t+1} &\coloneqq 1 - \alpha_{t+1} =\frac{c_1 \log T}{T}\min\left\{\beta_1\left(1 + \frac{c_1 \log T}{T}\right)^t, 1\right\},\quad 1\leq t < T,
\end{align}
\end{subequations}
where \(c_0, c_1>0\) are some large enough numerical constants. 
In words, this noise variance schedule (as $\beta_t$ is the variance of the noise injected at step $t$) contains two phases: it grows exponentially at the beginning, and then stays flat after reaching the order of $\frac{ \log T }{ T} $, 
which is consistent with the state-of-the-art diffusion model theory (e.g., \citet{benton2024nearly,potaptchik2024linear,huang2024denoising,li2024sharp}).

\medskip
\noindent
{\bf Score-based generative models.}
Next, we describe the precise update rules for both DDIM-type and DDPM-type samplers. 

%
\begin{enumerate}[label = $\bullet$]

    \item  \textit{DDIM-type samplers.} 
    As mentioned previously, a DDIM-type sampler starts with $Y_T\sim \mathcal{N}(0,I_d)$ and adopts the following deterministic update rule: 
    \begin{align}
    Y_{t-1} = \frac{1}{\sqrt{\alpha_t}}\big(Y_t + \eta_t^{\mathsf{ddim}} s_t(Y_t)\big),
    \qquad t= T,\cdots, 1.
    \label{eq:DDIM-update-prelim}
    \end{align}
    Here, \(\eta_t^{\mathsf{ddim}}\) is a design parameter that admits multiple alternatives, and 
    we list a couple of choices used in previous literature: 
    \begin{subequations}
    \label{eq:pars-DDIM-all}
    \begin{numcases}{\eta_t^{\mathsf{ddim}} 
    =}
        \frac{1 - \alpha_t}{1 + \sqrt{\frac{\alpha_t - \overline{\alpha}_t}{1 - \overline{\alpha}_t}}} & \text{(original DDIM    \citep{song2020denoising}; \underline{this work})} \label{eq:pars-DDIM-original}\\
         \frac{1-\alpha_t}{2}& \text{\citep{li2024sharp,li2024unified}}\\
        \frac{-1+4\sqrt{\alpha_t}-3\alpha_t}{2\sqrt{\alpha_t}} & \text{\citep{song2020score}}
    \end{numcases}
    \end{subequations}
   Importantly, all of these parameter choices lead to samplers that are asymptotically consistent --- meaning that the distribution of the sampling output converges to the target data distribution as $T$ grows --- under mild conditions on the target data distribution. 
   In this paper, we concentrate on the parameter schedule \eqref{eq:pars-DDIM-original}  proposed in the original DDIM paper \citep{song2020denoising}.
   

    \item \textit{DDPM-type samplers.} 
    A DDPM-type sampler adopts the initialization $Y_T\sim \mathcal{N}(0,I_d)$ and implements the stochastic update rule: 
    \begin{align}
    Y_{t-1} = \frac{1}{\sqrt{\alpha_t}}\big(Y_t + \eta_t^{\mathsf{ddpm}} s_t(Y_t) + \sigma_t^{\mathsf{ddpm}} Z_t\big),
    \qquad t = T,\cdots, 1,
    \label{eq:DDPM-update-prelim}
    \end{align}
    with independent noise $Z_t\overset{\mathrm{i.i.d.}}{\sim} \mathcal{N}(0,I_d)$. 
    Here, \(\eta_t^{\mathsf{ddpm}}\) and \(\sigma_t^{\mathsf{ddpm}}\) are design parameters, with several choices listed below:
    {\footnotesize
    \begin{subequations}
    \begin{numcases}{(\eta_t^{\mathsf{ddpm}}, \sigma_t^{\mathsf{ddpm}}) 
    =}
    \left(1-\alpha_t, \sqrt{\frac{(1-\alpha_t)(\alpha_t - \overline{\alpha}_t)}{1-\overline{\alpha}_t}}\right)  &
    \parbox{20em}{\(\left(
    \begin{array}{c}
    \text{original DDPM \citep{ho2020denoising};} \label{eq:pars-DDPM-original}\\
    
    \text{\citet{potaptchik2024linear};} \\
    \text{\citet{huang2024denoising};}\\
    
    \text{a special case of \underline{this work}}
    \end{array}
    \right)\)} \label{eq:eta-sigma-DDPM-original}\\
    \big(\,2(1-\sqrt{\alpha}_t), \sqrt{1-\alpha_t}\,\big) & \text{\citep{benton2024nearly,chen2023improved}}\\
    \left(1-\alpha_t, \sqrt{1-\alpha_t}\right) & \text{\citep{li2023towards,li2024d}}
   \end{numcases}
   \end{subequations}
   }
    All of the above choices come with convergence theory guaranteeing asymptotic consistency. In this work, we would like to accommodate a range of parameter schedules that subsumes as a special case the one \eqref{eq:pars-DDPM-original}  proposed in the original DDPM paper \citep{ho2020denoising}.
    

\end{enumerate}

\medskip
\noindent
{\bf ODE and SDE perspectives.}
To shed light on the rationale and feasibility of the  DDIM-type and DDPM-type samplers,  it is helpful to look at the continuous-time analogs of both forward and backward processes and resort to the toolbox of ordinary differential equations (ODEs) and stochastic differential equations (SDEs). We briefly review some basics in the sequel, and will illuminate deeper connections in Section~\ref{sec:interpretation}. 
\begin{enumerate}[label = $\bullet$]
\item \textit{Forward SDE.}
The forward process~\eqref{eq:forward-marginal} is intimately connected with the following continuous-time process with some specific choice of $\beta(t)$:
\begin{equation}
\label{eq:forward-SDE}
\rd X_t = -\beta(t) X_t \rd t + \sqrt{2\beta(t)} \,\rd B_t,
\end{equation}
where $(B_t)$ represents a standard Brownian motion in $\mathbb{R}^d$. 
In fact,  standard SDE theory reveals that SDE~\eqref{eq:forward-SDE} admits the following characterization 
\begin{equation}
X_t = \exp\left(-\int_0^t \beta(s) \rd s\right) X_0 + \sqrt{1-\exp\bigg(-2\int_0^t \beta(s) \rd s\bigg)}\overline{W}_t
\end{equation}
for some $\overline{W}_t\sim \mathcal{N}(0,I_d)$, 
whose marginal distribution coincides with that of \eqref{eq:forward-marginal} if
\begin{align}\label{eq:alpha-beta}
    \alpha_t = \exp\left(-2\int_{t-1}^t \beta(s) \rd s\right),
    \qquad t= 1,\cdots, T.
\end{align}

    \item \textit{Probability flow ODE or diffusion ODE.} One way to reverse the forward process is through the probability flow ODE  \citep{song2020score} (also known as diffusion ODE):
     \begin{equation}\label{eq:continue-time-ode}
     \rd Y_t = \left(Y_t + s^{\star}_{T-t}(Y_t)\right)\beta(T-t) \rd t,
    \qquad t \in [0, T],
     \end{equation}
     which enjoys matching marginal distribution $Y_{T-t} \overset{\mathrm{d}}{=} X_t$ for all $0 \le t \le T$ as long as we generate $Y_0\sim p_{X_T}$. 
     To approximately simulate this reverse ODE in practice and obtain a tractable sampler, a common strategy is to perform time discretization of ODE~\eqref{eq:continue-time-ode}. 
     Note that different discretization schemes can result in different design coefficients $\eta_t^{\mathsf{ddim}}$ as in the DDIM-type update rule \eqref{eq:DDIM-update-prelim}.
     \item \textit{Reverse-time SDE.}
     \quad
     An alternative way to reverse the forward process is via a properly chosen SDE.
     In view of the classical results in the SDE literature \citep{anderson1982reverse,haussmann1986time}, the following SDE,  
\begin{equation}\label{eq:continue-time-sde} 
     \rd Y_t = \left(Y_t + 2 s^{\star}_{T-t}(Y_t)\right)\beta(T-t) \rd t + \sqrt{2\beta(T-t)} \, \rd W_t,
    \qquad t \in [0,T] 
     \end{equation}
     with $(W_t)$ a standard Brownian motion in $\mathbb{R}^d$, 
     reverses the forward process \eqref{eq:forward-SDE} in the sense that $Y_{T-t} \overset{\mathrm{d}}{=} X_t$ for all $0 \le t \le T$ as long as $Y_0 \sim p_{X_T}$. 
     Akin to the DDIM counterpart, the DDPM-type samplers can often be viewed as time discretization of SDE~\eqref{eq:continue-time-sde}, 
     and different discretization schemes correspond to different coefficient choices of $(\eta_t^{\mathsf{ddpm}},\sigma_t^{\mathsf{ddpm}})$ as in \eqref{eq:DDPM-update-prelim}.
     %

\end{enumerate}
\begin{enumerate}[label = $\bullet$]
     \item \textit{Generalized reverse-time ODE/SDE.}
     We also consider the more general form:  \begin{equation}\label{eq:gen-continue-time}
     \rd Y_t = \left(Y_t + \big(1 + \xi(T-t)\big) s^{\star}_{T-t}(Y_t)\right)\beta(T-t) \rd t + \sqrt{2\xi(T-t)\beta(T-t)} \, \rd W_t,
     \end{equation}
     $t \in [0,T]$, 
     for some general function $\xi(t) \ge 0$ for all $0 \le t \le T$, where $(W_t)$ again represents a standard Brownian motion in $\mathbb{R}^d$. We shall formally demonstrate the desired distributional property of this family of differential equations in Appendix~\ref{sec:gen-reverse}.  
     When $\xi(t) = 0$ (resp.~$\xi(t) = 1$) for all $t \in [0,T]$, \eqref{eq:gen-continue-time} reduces to ODE~\eqref{eq:continue-time-ode} (resp.~SDE~\eqref{eq:continue-time-sde}).
     For a general $\xi(t)$, suitable time discretizationschemes of \eqref{eq:gen-continue-time} can lead to new samplers other than  the original DDIM and DDPM. 
\end{enumerate}


%


%% file: main_results.tex
\section{Main results}

This section presents our main theory. 
We begin in Section~\ref{sec:assumption} by establishing performance guarantees for both DDIM and DDPM under the idealized setting where exact scores are available. We then turn to the more practical setting in which only approximate score estimates are accessible. Sections~\ref{sec:conv_DDIM_in_TV} and~\ref{sec:conv_DDPM_in_TV} analyze DDIM and DDPM, respectively,  establishing convergence guarantees for data generation under approximate scores, accompanied by adaptive,  finite-sample guarantees for score estimation.  Section~\ref{sec:necessity_opt_step_size}
develops a lower bound  that demonstrates the optimality of the coefficients in DDIM and DDPM. 

\subsection{Theoretical guarantees under exact scores}\label{sec:assumption}

%
%
%
We first study the idealized setting in which the ground-truth score functions are available, i.e., \(s_t=s_t^\star\), and establish convergence guarantees for both DDIM and DDPM samplers.  

\subsubsection{Assumptions}
Our theory relies on two assumptions about the target distribution $\pdata$. 
The first assumption requires the support of $\pdata$ --- denoted by 
$\mathcal{X}_{\mathsf{data}}\in \R^d$ ---  to be (polynomially) bounded. 
\begin{assumption}[Polynomially bounded support]\label{ass:bd_supp}
    Suppose that there exists a universal constant $c_R > 0$ such that
    \[
    \sup_{x \in \mathcal{X}_{\mathsf{data}}} \norm{x}_2 \leq R_0~~~ \text{ for some } 1 \le R_0 \le T^{c_R}.
    \]
\end{assumption}
\begin{remark}
Note that the size of the support $\mathcal{X}_{\mathsf{data}}$ is allowed to scale polynomially (with an arbitrarily large constant degree) in the number of iterations of the sampler, which accommodates a very wide range of practical applications like image generation.\end{remark}

Next, we formalize the notion of \textit{low-dimensional structure} of $\pdata$.
To rigorously define its ``intrinsic dimension,'' we introduce the following definition of covering number \citep[Chapter 5]{wainwright2019high}, a standard and widely used measure of the complexity of a set $\mathcal{X}$.

%
%
\begin{definition}[Covering number]
    For any set $\mathcal{X}\subseteq \mathbb{R}^d$,
    the (Euclidean) covering number at scale $\epsilon_0 > 0$, denoted by $N_{\epsilon_0}(\mathcal{X})$,  is defined as the smallest integer $n$ such that there exist points $x_1, \dots , x_n$ obeying
    \begin{align*}
        \mathcal{X} \subseteq \bigcup_{i=1}^n \mathcal{B}(x_i,\epsilon_0).
    \end{align*}
\end{definition}
The covering number in turn enables a flexible characterization of the complexity of $\pdata$.
\begin{assumption}[Intrinsic dimension]\label{ass:low_dim}
    For any $\epsilon_0>0$, the covering number of the support $\mathcal{X}_{\mathsf{data}}$ of $\pdata$ is assumed to satisfy
    \[
    \log N_{\epsilon_0}(\mathcal{X}_{\mathsf{data}})
    \leq C_{\mathsf{cover}} k \log\left(\frac{R_0}{\epsilon_0}\right)
    \]
    for some universal constant $C_{\mathsf{cover}} > 0$, where \(R_0\) is the support radius in Assumption~\ref{ass:bd_supp}.
    Here and throughout, we shall refer to $k$ as the intrinsic dimension of $\pdata$.
\end{assumption}
The intrinsic dimension defined above is fairly generic, facilitating studies of a number of important low-dimensional structures.
Partial examples that satisfy Assumption~\ref{ass:low_dim} include $k$-dimensional linear subspace in $\mathbb{R}^d$ and
$k$-dimensional manifolds
(provided that $\mathcal{X}_{\mathsf{data}}$ is polynomially bounded as in Assumption~\ref{ass:bd_supp}), as well as structures with doubling dimension $k$ \citep{dasgupta2008random}.
See \citet[Section 4.1]{huang2024denoising} for more detailed discussion.

\subsubsection{Convergence theory under exact scores}

Armed with the above assumptions --- which encompass a broad class of target distributions  --- we now establish the performance guarantees of DDIM and DDPM under exact scores. 
\begin{thm}\label{thm:true-score-convergence}
Suppose that Assumptions~\ref{ass:bd_supp} and~\ref{ass:low_dim} hold, and that the score estimates are exact (i.e., \(s_t=s_t^\star\) for all \(t=1,\ldots,T\)). Then
the following results hold.
\begin{itemize}
\item
\textbf{DDIM.}
The DDIM sampler \eqref{eq:DDIM-update} with
\(
\eta_t^{\mathsf{ddim}}
=
\frac{1-\alpha_t}
{1+\sqrt{\frac{\alpha_t-\overline{\alpha}_t}{1-\overline{\alpha}_t}}}
\)
satisfies
\[
\TV(p_{X_1},p_{Y_1})
\lesssim
\frac{k\log^3T}{T}.
\]

\item
\textbf{DDPM.}
The DDPM sampler \eqref{eq:DDPM-update} with
\(
\eta_t^{\mathsf{ddpm}}=1-\alpha_t,\,
\sigma_t^{\mathsf{ddpm}}
=
\sqrt{\frac{(\alpha_t-\overline{\alpha}_t)(1-\alpha_t)}
{1-\overline{\alpha}_t}}
\)
satisfies
\[
\TV(p_{X_1},p_{Y_1})
\lesssim
\frac{k\log^3T}{T}.
\]
\end{itemize}
\end{thm}

Encouragingly, Theorem~\ref{thm:true-score-convergence} offers several key insights into the remarkable efficacy of DDIM and DDPM when the target distribution exhibits \textit{unknown} low dimensionality.
\begin{itemize}
\item \textit{Adaptation to intrinsic dimension.}
Theorem~\ref{thm:true-score-convergence} shows that both DDIM and DDPM can
adapt to the low-dimensional structure of the data distribution when the true scores
are accessible. In particular, both samplers achieve an iteration complexity of 
\(\widetilde O(k/\varepsilon)\)  for reaching
\(\TV(p_{X_1},p_{Y_1})\le\varepsilon\), exhibiting a \textit{linear} dependence on the intrinsic dimension $k$. For DDIM, to the best of our
knowledge, this is the first result showing adaptation to unknown
low-dimensional structure. For DDPM, such low-dimensional adaptation was first
studied by \citet{li2024adapting}; in comparison, our TV-based theory establishes
a linear scaling on the intrinsic dimension \(k\) that significantly improves upon \citet{li2024adapting}. See
Tables~\ref{tab:DDIM} and~\ref{tab:DDPM} for comparisons with prior results.

\item \textit{Minimal distributional assumptions.} The low-dimensional adaptation capability established by our theory holds under remarkably broad distributional assumptions, requiring neither smoothness nor log-concavity. This broad applicability is aligned with the
practical success of diffusion models, which  are routinely used to model and generate samples from rich and
highly complex data distributions.

\item \textit{No burn-in cost.}
We also compare Theorem~\ref{thm:ddim_tv_conv} with the state-of-the-art DDIM theory \citep{li2024sharp} for the case with $k=d$ and exact scores.  Recall that \citet{li2024sharp} established an \( \widetilde{O}(d/T) \) convergence rate,
which is consistent with Theorem~\ref{thm:ddim_tv_conv}. Nevertheless, the  theory therein requires a burn-in cost \( T \gtrsim d^2 \log^5T \); in comparison, our theorem does not impose such a burn-in requirement.

\item \textit{Coefficient choices.}
As it turns out, the remarkable low-dimensional adaptation capability of diffusion models is achieved using the
classical coefficient choices. For DDIM, the coefficient
\(
\eta_t^{\mathsf{ddim}}
=
\frac{1-\alpha_t}
{1+\sqrt{\frac{\alpha_t-\overline{\alpha}_t}{1-\overline{\alpha}_t}}}
\)
matches exactly that proposed in the original DDIM sampler
\citep{song2020denoising}. Likewise, for DDPM, the coefficients
\(
\eta_t^{\mathsf{ddpm}}=1-\alpha_t,\,
\sigma_t^{\mathsf{ddpm}}
=
\sqrt{\frac{(\alpha_t-\overline{\alpha}_t)(1-\alpha_t)}
{1-\overline{\alpha}_t}}
\)
coincide with those introduced in the original DDPM paper \citet{ho2020denoising}. Our results therefore provide a theoretical explanation for the practical success of these classical coefficient choices: they are naturally aligned with the low-dimensional structure of the target distribution, enabling both samplers to adapt automatically to such favorable structure.
\end{itemize}

\subsection{Theoretical guarantees under noisy score estimates}

We now turn to the more realistic setting in which the scores are estimated from data (and hence noisy) rather than known exactly. To characterize the impact of score estimation accuracy on the performance of diffusion samplers, we introduce additional assumptions that quantify the quality of the estimated scores, and establish convergence guarantees for both DDIM and DDPM accordingly. We further show how such assumptions can be attained by statistical estimation procedures and derive finite-sample guarantees that also adapt to the underlying low-dimensional structure.

\subsubsection{DDIM theory under noisy scores}\label{sec:conv_DDIM_in_TV}

\medskip
\noindent {\bf Score estimation assumptions.} 
We first analyze the TV-based performance of the DDIM sampler in the presence of noisy score estimates. Our theory relies on the following assumptions on the estimated scores. Compared with those required for DDPM (see Section~\ref{sec:conv_DDPM_in_TV}), these assumptions are more stringent due to the deterministic nature of DDIM, but, as we will shortly justify, they are attainable by kernel-based score estimators with finite-sample and adaptive guarantees.
\setcounter{assumption}{2}
\begin{assumption}[Score estimates for DDIM]\label{ass:ddim_score_matching}
    Suppose that each of the estimated score functions $\{s_t(\cdot)\}_{t=1}^T$ is twice continuously differentiable. For each $1\leq t \le T$, assume that
    \begin{subequations}
    \begin{align}
    \eta_t^{\mathsf{ddim}} v^\top \nabla s_t(x) v \ge -\frac{1}{4}\norm{v}_2^2\qquad  \text{for all } v,x \in \RB^d.
    \end{align}
    Further, we suppose that
        \begin{align}
        \frac{1}{T}\ssum{t}{1}{T}(1-\overline{\alpha}_t)^2\varepsilon_{\mathsf{score},t}^2 &\le \varepsilon_{\mathsf{ddim}\text{-}\mathsf{sc}}^2
        &&\text{ with }\varepsilon_{\mathsf{score},t}^2 \coloneqq \EB \big[ \norm{s_t(X_t) - s_t^{\star}(X_t)}_2^2 \big],\\
        \frac{1}{T}\ssum{t}{1}{T} (1-\overline{\alpha}_t)^2\varepsilon_{\mathsf{Jacobi},1,t}^2 &\le \varepsilon_{\mathsf{Jacobi},1}^2
        &&\text{ with } \varepsilon_{\mathsf{Jacobi},1,t}^2 \coloneqq \EB\bigg[\norm{ \frac{\partial s_t (X_t)}{\partial x} - \frac{\partial s_t^{\star}(X_t)}{\partial x}  }_{\mathrm{F}}^2\bigg],  \\
        \frac{1}{T}\ssum{t}{1}{T} (1-\overline{\alpha}_t)^2\varepsilon_{\mathsf{Jacobi},2,t}^2 &\le \varepsilon_{\mathsf{Jacobi},2}^2 &&\text{ with }
\varepsilon_{\mathsf{Jacobi},2,t}^2 \coloneqq \EB \bigg[ \tr\bigg(\frac{\partial s_t (X_t)}{\partial x} - \frac{\partial s_t^{\star}(X_t)}{\partial x} \bigg)^2 \bigg], \\
        \frac{1}{T}\ssum{t}{1}{T} (1-\overline{\alpha}_t)^2\varepsilon_{\mathsf{Hess},t}^2 &\le \varepsilon_{\mathsf{Hess}}^2
&&\text{ with }\varepsilon_{\mathsf{Hess},t}^{2} \coloneqq \EB\bigg[\norm{\nabla\tr\bigg(\frac{\partial s_t (X_t)}{\partial x} - \frac{\partial s_t^{\star}(X_t)}{\partial x}\bigg)}_2^{2}\bigg].
        \end{align}
        \end{subequations}
\end{assumption}
In essence,  Assumption~\ref{ass:ddim_score_matching} is concerned with the accuracy of the estimated score functions under a few metrics,
including time-averaged errors for the score estimates themselves as well as their associated Jacobians and Hessians. 
Intuitively, given
that DDIM is deterministic without injecting extra random noise to smooth the
trajectory, additional assumptions like higher-order score estimation accuracy
are needed in order to mitigate the propagation of estimation errors in each
backward step; see, e.g.,  \citet{li2024sharp,huang2024convergence}.  Moreover, we will show that the above assumptions are attainable via kernel-based score estimation procedures with adaptive finite-sample guarantees.

\medskip
\noindent {\bf Convergence guarantees for DDIM.} 
We now present performance guarantees for the DDIM sampler when exact scores are inaccessible. The proof is postponed to Appendix~\ref{sec:k/T_DDIM_Analysis}.
\begin{thm}\label{thm:ddim_tv_conv}
 Under Assumptions~\ref{ass:bd_supp}, \ref{ass:low_dim}, and~\ref{ass:ddim_score_matching}, the DDIM sampler \eqref{eq:DDIM-update} with the coefficients $\eta_t^{\mathsf{ddim}} = \frac{1 - \alpha_t}{1 + \sqrt{\frac{\alpha_t - \overline{\alpha}_t}{1 - \overline{\alpha}_t}}}$ yields
    \begin{align}
    \TV(p_{X_1}, p_{Y_1}) \lesssim
    \frac{k\log^3 T}{T} + (\varepsilon_{\mathsf{ddim}\text{-}\mathsf{sc}} + \varepsilon_{\mathsf{Jacobi},1} + \varepsilon_{\mathsf{Jacobi},2} + \varepsilon_{\mathsf{Hess}})\sqrt{\log T}.
    \end{align}
    %
\end{thm}


Compared with the exact-score setting, the TV distance of interest continues to converge at a rate  $\widetilde{O}(k/T)$ until it reaches an error floor determined by the error metrics in Assumption~\ref{ass:ddim_score_matching}.  
Unlike previous analysis for DDIM,  Theorem~\ref{thm:ddim_tv_conv} makes an assumption about the second-order approximation of \( s_t(\cdot) \) to \( s_t^{\star}(\cdot) \), i.e., the additional error term \( \varepsilon_{\mathsf{Hess}} \) in Assumption~\ref{ass:ddim_score_matching}. This arises because, in prior studies, the score  error terms in the convergence rate of DDIM were dependent on the ambient dimension \( d \). For instance, in \citet{huang2024convergence,li2024sharp}, the estimation error terms in their respective convergence rates are given by $d^{\frac{3}{4}}L^{\frac{1}{2}}\varepsilon_{\mathsf{score}}$ and $\sqrt{d}\varepsilon_{\mathsf{score}} + d \varepsilon_{\mathsf{Jacobi}}$, where $\varepsilon_{\mathsf{score}}$ and $L$ denote  certain $\ell_2$ score error and smoothness parameter, respectively. In comparison, in our Theorem~\ref{thm:ddim_tv_conv}, the score error term in the sampling error is nearly dimension-free (except for log dependency), meaning  that this error almost does not amplify when the intrinsic and ambient dimensions increase.


\medskip
\noindent {\bf Adaptive score estimation guarantees.} 
While Assumption~\ref{ass:ddim_score_matching} enables the convergence theory in
Theorem~\ref{thm:ddim_tv_conv}, a natural question is whether it can be satisfied with finite samples, and if so, by what statistical procedures. We now address this question by showing that Assumption~\ref{ass:ddim_score_matching} is attainable with finite-sample guarantees through estimators that adapt to low-dimensional structure. Although these results are primarily included to complement our convergence theory of the DDIM sampler, they provide a rigorous statistical justification for the attainability of our score estimation assumptions.


Setting the stage, suppose we observe a dataset $X^{\mathsf{sample}}$ containing $n$ i.i.d.~samples, namely, 
\begin{equation}
X^{\mathsf{sample}}
\coloneqq
\big\{X_i^{\mathsf{sample}} \big\}_{i=1}^n \qquad 
\text{with }X_i^{\mathsf{sample}}\overset{\mathrm{i.i.d.}}{\sim} p_{\mathsf{data}}.
\label{eq:samples-score-learning}
\end{equation}
The goal is to use these
samples to construct a collection of estimated score functions that approximate the true score functions
\(\{s_t^\star(\cdot)\}_{t=1}^T\). 
We now show that score estimates satisfying Assumption~\ref{ass:ddim_score_matching} can indeed be achieved with finite-sample and adaptive guarantees. 

\begin{thm}
\label{thm:ddim-training-guarantee}
Suppose Assumptions \ref{ass:bd_supp} and \ref{ass:low_dim} hold. Let
\(k^\star\coloneqq C_{\mathsf{cover}}k\), with the constant $C_{\mathsf{cover}}$ defined in Assumption~\ref{ass:low_dim}. Let
\(\alpha_t=1-\beta_t\) be chosen according to the schedule
\eqref{eq:noise-schedule}, and recall the constant $c_0>0$ in \eqref{eq:noise-schedule}. Then there exists a score estimator
\(s_t(\cdot)\) such that
\begin{subequations}
\begin{align}
\varepsilon_{\mathsf{ddim}\text{-}\mathsf{sc}}^2
&\lesssim
\frac{R_0^{k^\star+\frac{4}{k^\star+2}}\log^2(dR_0nT^{c_0})}
{n}
T^{c_0\left(\frac{k^\star}{2}+\frac{2}{k^\star+2}\right)},
\\
\varepsilon_{\mathsf{Jacobi},1}^2,
\varepsilon_{\mathsf{Jacobi},2}^2
&\lesssim
\frac{R_0^{k^\star+\frac{16}{k^\star+4}}\log^2(dR_0nT^{c_0})}
{n}
T^{c_0\left(\frac{k^\star}{2}+\frac{8}{k^\star+4}\right)},
\\
\varepsilon_{\mathsf{Hess}}^2
&\lesssim
\frac{R_0^{k^\star+\frac{36}{k^\star+6}}\log^2(dR_0nT^{c_0})}
{n}
T^{c_0\left(\frac{k^\star}{2}+1+\frac{18}{k^\star+6}\right)}, \\
\eta_t^{\mathsf{ddim}} v^\top \nabla s_t(x) v &\ge -\frac{1}{4}\|v\|_2^2 \qquad  \text{for all } v, x\in \RB^d.
\end{align}
\end{subequations}
\end{thm}
At a high level, the score estimator \(s_t\) in the above theorem is the plug-in score induced by the
Gaussian kernel density estimator. 
The detailed construction of \(s_t\) and the proof of
Theorem~\ref{thm:ddim-training-guarantee} are postponed to
Appendix~\ref{prf:thm:ddim-training-guarantee}.

Regarding finite-sample guarantees for score estimation tailored to DDIM, \citet{cai2025minimax} established the first sample complexity result for the full-dimensional setting, showing a rate that scales as $T^{\Theta(d)}$ and is nearly minimax optimal. To the best of our knowledge, however, no prior work has established a sample complexity guarantee for DDIM score estimation that accounts for the low-dimensional structure of $\pdata$.  
When $\pdata$ has intrinsic dimension $k$, it is natural to expect a minimax lower bound of order \(T^{\Theta(k)}\) \citep[e.g.,][]{tang2022minimax,berenfeld2019density}. Our result matches this dependence in the exponent in an orderwise sence, thereby providing the first finite-sample guarantee that provably reveals low-dimensional adaptation in the score learning stage of DDIM.

Compared with prior score learning theory for DDPM, the main technical challenge is that our DDIM convergence analysis requires controlling higher-order score approximation errors, whose finite-sample analysis is substantially more involved than that of the standard \(\ell_2\) score estimation error. In addition, unlike \citet{cai2025minimax}, we need to establish the Hessian-based approximation guarantee required by our convergence analysis, which is essential for eliminating the dependence on the ambient dimension \(d\).


 %



\subsubsection{DDPM theory under noisy scores}\label{sec:conv_DDPM_in_TV}

Turning attention to the DDPM-type samplers, this subsection establishes TV-based
convergence guarantees for DDPM in the absence of exact score functions. Compared with the DDIM case, we are in need of much simpler score estimation assumptions.
\setcounter{assumption}{3}
\begin{assumption}[Score estimates for DDPM]\label{ass:ddpm_score_matching} Let \(\varepsilon_{\mathsf{score},t}^2 \coloneqq \EB \big[ \norm{s_t(X_t) - s_t^{\star}(X_t)}_2^2 \big]\), and 
    suppose that the estimated score functions $\{s_t(\cdot)\}_{t=1}^T$ obey
    \begin{align}
    \label{eq:defn-L2-error}
    \frac{1}{T}\ssum{t}{1}{T}(1-\overline{\alpha}_t)\varepsilon_{\mathsf{score},t}^2 \le \varepsilon_{\mathsf{ddpm}\text{-}\mathsf{sc}}^2.
    \end{align}
\end{assumption}
Note that $\ell_2$ score estimation accuracy is commonly used in diffusion model theory (e.g., \citet{chen2022improved,benton2024nearly}).
This form of estimation error also aligns with practical training procedures such as score matching  \citep{hyvarinen2005estimation,vincent2011connection}.

\medskip
\noindent {\bf Convergence guarantees for DDPM.} 
We present below our TV-based convergence guarantees for the original DDPM sampler proposed by \citet{ho2020denoising} under noisy score estimates.
The proof can be found in   Appendix~\ref{sec:k/T_DDPM_Analysis}.
\begin{thm}\label{thm:vanilla_ddpm}
Under Assumptions~\ref{ass:bd_supp}, \ref{ass:low_dim}, and~\ref{ass:ddpm_score_matching}, the DDPM sampler \eqref{eq:DDPM-update} with the coefficients  $\eta_t^{\mathsf{ddpm}} = 1-\alpha_t$ and $\sigma_t^{\mathsf{ddpm}} = \sqrt{\frac{(\alpha_t - \overline{\alpha}_t)(1-\alpha_t)}{1 - \overline{\alpha}_t}}$ achieves
\begin{align}
    \TV(p_{X_1}, p_{Y_1}) \lesssim \frac{k\log^3 T}{T} + \varepsilon_{\mathsf{ddpm}\text{-}\mathsf{sc}}\sqrt{\log T}.
\end{align}
\end{thm}
When score estimation is imperfect, our TV-based convergence guarantees degrade gracefully, with the bounds scaling linearly in $\varepsilon_{\mathsf{ddpm}\text{-}\mathsf{sc}}$ (a metric that measures the $\ell_2$ estimation error).
In stark contrast to our DDIM theory in Theorem~\ref{thm:ddim_tv_conv},   convergence of DDPM can be established under fewer assumptions; for instance, there is no need of imposing assumptions on the Jacobian or Hessian of score estimates as in Assumption~\ref{ass:ddim_score_matching}.
This favorable feature of DDPM arises since its stochastic update rule helps smooth the sampling trajectory, thereby eliminating the need to cope with many boundary cases.

As it turns out, the coefficients \eqref{eq:pars-DDPM-original} are not the only choice of DDPM-type samplers that enable the desirable adaptation.
Our convergence theory can be extended to accommodate a broader set of coefficients, as stated below.
The proof is deferred to  Appendix~\ref{sec:k/T_DDPM_Analysis}.


\begin{thm}\label{thm:ddpm_tv_conv}

Suppose that the coefficients $\eta_t^{\mathsf{ddpm}}$ and  $\sigma_t^{\mathsf{ddpm}}$ satisfy
\begin{equation}\label{eq:ddpm_gen_stp_choice}
(1 - \overline{\alpha}_t)\bigg(1 - \frac{\eta_t^{\mathsf{ddpm}}}{1 - \overline{\alpha}_t}\bigg)^2 = \alpha_t - \overline{\alpha}_t - \big(\sigma_t^{\mathsf{ddpm}}\big)^2,
\qquad t = 1,\dots, T.
\end{equation}
Also, assume that there exist some universal constants $C_1, C_2 \geq 1/2$ such that
\begin{equation}\label{eq:ddpm_gen_stp_choice1}
\eta_t^{\mathsf{ddpm}} \le \min\left\{C_1(1 - \alpha_t), C_2\sqrt{1-
\alpha_t}\sigma_t^{\mathsf{ddpm}},\frac{1}{2}(1 - \overline{\alpha}_t)\right\},\qquad  t = 1,\cdots, T.
\end{equation}
%

%
Then under Assumptions~\ref{ass:bd_supp}, \ref{ass:low_dim}, and~\ref{ass:ddpm_score_matching},  the DDPM sampler \eqref{eq:DDPM-update} yields
    \[
    \TV(p_{X_1}, p_{Y_1}) \lesssim \frac{k\log^3 T}{T} + \varepsilon_{\mathsf{ddpm}\text{-}\mathsf{sc}}\sqrt{\log T}.
    \]

\end{thm}

Let us take a moment to discuss the range of coefficients satisfying relation~\eqref{eq:ddpm_gen_stp_choice}.
Interestingly, this relation between $\eta_t^{\mathsf{ddpm}}$ and $\sigma_t^{\mathsf{ddpm}}$ aligns perfectly with the set of coefficients discussed in \citet[Section~4.1]{song2020denoising}.
More specifically, \citet[Eq.~(12)]{song2020denoising} singled out the update rule below:
\begin{equation}\label{eq:reverse_of_original_ddim}
    Y_{t-1} = \frac{1}{\sqrt{\alpha_t}}\Big(Y_t - \sqrt{1 - \overline{\alpha}_t}\epsilon^{\mathsf{noise}}_{t}(Y_t) + \sqrt{\alpha_t - \overline{\alpha}_t - \alpha_t\varsigma_t^2}\epsilon_{t}^{\mathsf{noise}}(Y_t) + \sqrt{\alpha_t}\varsigma_tZ_t \Big)
\end{equation}
for some coefficient $\varsigma_t$,
where $\epsilon_{t}^{\mathsf{noise}}(Y_t) = - \sqrt{1 - \overline{\alpha}_t}s_t(Y_t)$ serves as an estimate of the noise injected in the forward process.
One can demonstrate its equivalence with \eqref{eq:ddpm_gen_stp_choice}. The interested reader is referred to Appendix~\ref{sec:eqv_to_ddim} for more details. 

\begin{remark}
The differential-equation viewpoint explains why the DDIM and
generalized DDPM updates inherit low-dimensional adaptivity:
 after rewriting the reverse ODE/SDE through Tweedie's formula,
  the essential drift term is governed by the posterior mean of
  \(X_0\), which pulls the dynamics toward the
  low-dimensional data support. We defer the full derivation
  and the exact equivalence statements to
  Appendix~\ref{sec:interpretation}.
\end{remark}

\medskip
\noindent {\bf Adaptive score estimation guarantee.} 
We now study the feasibility of constructing score estimators to satisfy Assumption~\ref{ass:ddpm_score_matching}, again under the data generating model \eqref{eq:samples-score-learning}. Unlike DDIM, whose analysis requires higher-order conditions, the DDPM theory above only relies on $\ell_2$ score matching errors.  As a
result, we obtain a sharper finite-sample rate, as stated in the
following result.

\begin{thm}
\label{thm:ddpm-training-guarantee}
Suppose Assumptions~\ref{ass:bd_supp} and~\ref{ass:low_dim} hold. Let
\(k^\star\coloneqq C_{\mathsf{cover}}k\), with the constant $C_{\mathsf{cover}}$ defined in Assumption~\ref{ass:low_dim}. Let
\(\alpha_t=1-\beta_t\) be chosen according to the schedule
\eqref{eq:noise-schedule}, and recall the constant $c_0>0$ in \eqref{eq:noise-schedule}. Then there exists a score estimator
\(s_t(\cdot)\) such that
\[
\varepsilon_{\mathsf{ddpm}\text{-}\mathsf{sc}}^2
\lesssim
\frac{R_0^{k^\star}\log(dR_0nT^{c_0})}
{n}
\min\left\{
d^{\frac{2}{k^\star+2}}T^{c_0k^\star/2},
R_0^{\frac{4}{k^\star+2}}
T^{c_0\left(\frac{k^\star}{2}+\frac{2}{k^\star+2}\right)}
\right\}.
\]
In particular, since \(0\le 1-\overline{\alpha}_t\le1\), the same upper bound
also applies to \(\varepsilon_{\mathsf{ddim}\text{-}\mathsf{sc}}^2\).
\end{thm}
The score estimator in this theorem is a thresholded kernel-based score estimator. 
Its detailed construction and the proof of
Theorem~\ref{thm:ddpm-training-guarantee} are given in
Appendix~\ref{prf:thm:ddpm-training-guarantee}.


In contrast to the DDIM setting, where finite-sample approximation guarantees
that adapt to low-dimensional structure appear to be previously unavailable, several recent papers have studied the sample complexity of score learning for DDPM under
low-dimensional structure. Examples include
\citet{azangulov2024convergence} and \citet{yakovlev2025generalization}. The
former assumes that the data distribution is supported on a low-dimensional
manifold embedded in a high-dimensional ambient space, while the latter assumes
that samples are generated by first drawing from a low-dimensional cube and then
mapping the sample to the ambient space through a latent function. In contrast,
our result is built upon the more general intrinsic-dimension assumption (see Assumption~\ref{ass:low_dim}). Moreover, their approximation
guarantees contain terms that scale with
\(d^{O(k)}\), whereas the dependence on the ambient dimension \(d\) in our result is only
\(d^{2/(k^\star+2)}\). Consequently, when \(k\gtrsim \log d\), this factor is of constant order, showing that, according to our theory, the score learning phase --- and hence DDPM --- fully adapts to the underlying low-dimensional structure in this regime.

%% file: Necessity_of_opt_step_size.tex
Thus far, our main theorems (i.e., Theorems~\ref{thm:ddim_tv_conv}-\ref{thm:ddpm_tv_conv}) focus attention on specific coefficient choices as in the original DDIM and DDPM samplers. One might naturally wonder whether other coefficient choices could also facilitate low-dimensional adaptation capabilities. 
As it turns out, these particular coefficients --- or those exceedingly close to them --- are almost necessary to achieve adaptivity, as explained in this subsection.


For simplicity, consider the case with accurate score estimates (i.e., $s_t^{\star}=s_t$ for all $t$), and let us look at the following mapping: 
\begin{equation}
\Phi_t^{\star}(x,z) \coloneqq \frac{1}{\sqrt{\alpha_t}}\big(x + \eta_t s_t^{\star}(x) + \sigma_t z\big).
\end{equation}
Clearly, both the DDIM update rule \eqref{eq:DDIM-update-prelim} and DDPM update rule \eqref{eq:DDPM-update-prelim} in the $t$-th iteration can be described as $Y_{t-1}= \Phi_t^{\star}(Y_t,Z_t)$ for some choices of $\eta_t$ and $\sigma_t$ (i.e., $\sigma_t=0$ for DDIM and $\sigma_t\neq 0$ for DDPM), where $Z_t$ is an independent standard Gaussian vector.  
To evaluate how well the efficacy of DDIM-type and DDPM-type samplers, we propose to perform a sort of \textit{one-step analysis} as follows: 
\begin{itemize}
\item[1)] Start the sampler from $X_t$ of the forward process \eqref{eq:forward};  

\item[2)] Compute one iteration $Y_{t-1}=\Phi_t^{\star}(Y_t,Z_t)$ with an independent vector $Z_t\sim \mathcal{N}(0,I_d)$; 

\item[3)] Evaluate the TV distance between $Y_t$ and $X_t$ and see whether it is well-controlled. 

\end{itemize}
An ideal sampler that can effectively adapt to unknown low dimensionality would not incur a TV distance blowing up with the ambient dimension $d$.






As it turns out, in order for the TV distance between $X_t$ and $Y_t$ to be well-controlled, 
the coefficients $(\eta_t,\sigma_t)$ must  be carefully chosen, as revealed by the following lower bound. 
The proof of this lower bound is provided in Appendix~\ref{sec:proof:thm:lower-bound}. 


\begin{thm}\label{thm:lower_bound}
    Consider any $k\le d/2$, and 
    take the target distribution $\pdata$ to be $\gN\left(0, {\footnotesize\left[\begin{array}{cc}
I_k\\
 & 0
\end{array}\right]}\right)$. Then for arbitrary choices of $(\eta_t, \sigma_t)$, we have
    \begin{align}
    \label{eq:TV-lower-bound-d}
    \TV\big(\Phi_t^{\star}(X_t,Z_t), X_{t-1}\big) \ge 
    \frac{1}{100}\min\left\{ \sqrt{\frac{d}{2}} \left| \frac{1-\overline{\alpha}_t}{\alpha_t - \overline{\alpha}_t}\left(1 - \frac{\eta_t}{1-\overline{\alpha}_t}\right)^2 + \frac{\sigma_t^2}{\alpha_t - \overline{\alpha}_t} - 1\right|, \, 1\right\}.
    \end{align}
    %
\end{thm}
In words, Theorem~\ref{thm:lower_bound} asserts that even when initialized from a point from the true forward process,  performing one iteration of DDIM/DDPM updates might already incur a TV distance that scales polynomially in the ambient dimension,  
unless the coefficients are chosen to obey
\begin{equation}
\label{eq:general-constraint-LB}
\frac{1-\overline{\alpha}_t}{\alpha_t - \overline{\alpha}_t}\left(1 - \frac{\eta_t}{1-\overline{\alpha}_t}\right)^2 + \frac{\sigma_t^2}{\alpha_t - \overline{\alpha}_t} - 1 \approx 0. 
\end{equation}
\begin{itemize}
    \item Consider the DDIM-type sampler \eqref{eq:DDIM-update-prelim}, which has $\sigma_t=0$. The requirement \eqref{eq:general-constraint-LB} then simplifies to
    \[
\eta_{t}\approx1-\overline{\alpha}_{t}-\sqrt{(\alpha_{t}-\overline{\alpha}_{t})(1-\overline{\alpha}_{t})}=\frac{1-\alpha_{t}}{1+\sqrt{\frac{\alpha_{t}-\overline{\alpha}_{t}}{1-\overline{\alpha}_{t}}}},
\]
the right-hand side of which is precisely the choice \eqref{eq:pars-DDIM-original} of the original DDIM sampler. 

    \item The DDPM-type sampler \eqref{eq:DDPM-update-prelim} then corresponds to the case with $\sigma_t>0$. Clearly, all coefficient choices studied in Theorem~\ref{thm:ddpm_tv_conv} satisfy this requirement,  subsuming the choice \eqref{eq:eta-sigma-DDPM-original} of the original DDPM sampler as a special case. 
\end{itemize}
Somewhat surprisingly, while the original DDIM and DDPM update rules \citep{song2020denoising,ho2020denoising} were derived heuristically (namely, by maximizing some variational lower bounds on the log-likelihoods) without any explicit consideration of the low-dimensional structure, the coefficients of the resulting algorithms prove to be  nearly essential for adaptation to low dimensionality.

%% file: related_works.tex
\section{Related work}

\noindent
{\bf General convergence analysis of diffusion models.}
Take the DDPM for instance, the work \citet{chen2022sampling} established convergence analysis (based on Girsanov's theorem) without assuming log-concavity; the smoothness assumption was further relaxed by \citet{lee2023convergence,chen2023improved}.
A recent strand of work has been devoted to analyzing the convergence behavior of diffusion models~\citep{chen2022sampling,lee2022convergence,liu2022let,lee2023convergence,chen2023improved,benton2024nearly,chen2023restoration,li2023towards,pedrotti2023improved,cheng2023convergence,huang2024convergence,liang2024non,li2024d,li2024unified,gao2024convergence,gentiloni2025beyond}; see \citet{tang2024score} for a tutorial.
Regarding the use of DDPM for a general class of non-smooth and non-log-concave distributions, \citet{benton2024nearly} established the best-known  KL-based convergence guarantees, whereas the state-of-the-art TV-based convergence was derived by  \citet{li2024d}. Turning attention to the DDIM, \citet{chen2023restoration} derived the first polynomial-time analysis,
while \citet{chen2023probability} provided improved analysis for a variation of the probability flow ODE (by adding an additional stochastic step).
\citet{li2023towards}
improved the TV-based  iteration complexity of the DDIM to $\widetilde{O}(d^2/\varepsilon)$, which was subsequently improved by  \citet{li2024sharp} to $\widetilde{O}(d/\varepsilon + d^2)$.
Additionally, higher-order samplers tailored to solving the reverse-time SDE or probability flow ODE (e.g., \citet{lu2022dpm,lu2022dpmp}) have been proven to achieve faster convergence
\citep{li2024accelerating,wu2024stochastic,li2024provable,huang2024convergence}. Randomized midpoint methods have also been leveraged to provably speed up convergence \citep{gupta2024faster,li2024improved}. The convergence behavior of conditional diffusion models (or diffusion guidance) is another important topic that has been studied by several recent work (e.g., \cite{wu2024theoretical,chidambaram2024does,chen2024overview,fu2024unveil}).

\medskip
\noindent
{\bf Score matching.}
An important stage of score-based generative modeling is score matching or score learning \citep{hyvarinen2005estimation,hyvarinen2007some,vincent2011connection,song2019generative,ho2020denoising},
which aims to learn the score functions (typically using deep neural networks or transformers).
From the statistical perspectives,
\cite{koehler2022statistical} characterized the asymptotic statistical efficiency of score matching,
while \cite{oko2023diffusion,wibisono2024optimal,zhang2024minimax,han2024neural,dou2024optimal} derived the statistical error rates and sample complexity for score matching.
Another recent work \cite{feng2024optimal} leveraged some idea from score matching to tackle convex M-estimation. We omit further details, as score matching is not our focus.


%



\medskip
\noindent
{\bf Diffusion models in the presence of low-dimensional structure.}
Given the ubiquity of low-dimensional structure in practice~\citep{pope2021intrinsic},
a recent line of research sought to unveil the role of low dimensionality in enabling more efficient data generation~\citep{li2024adapting,azangulov2024convergence,potaptchik2024linear,huang2024denoising}.
More concretely, \citet{li2024adapting} established the first  iteration complexity upper bound for the DDPM  that is adaptive to unknown low-dimensional structures, without the need of modifying the algorithm; the iteration complexity therein is proportional to $k^4$, with $k$ the intrinsic dimension. This $k$-dependency was subsequently improved by \citet{azangulov2024convergence} to $k^3$ and then tightened by \citet{potaptchik2024linear,huang2024denoising} to linear scaling. However, existing results establish only KL-based convergence, which can be overly conservative for TV guarantees via Pinsker’s inequality. Moreover, prior work does not address how ODE-based samplers adapt to unknown low-dimensional structure.




Apart from the above-mentioned convergence analysis for the sampling stage, the interplay between diffusion models and low-dimensional structure has been investigated from other perspectives as well~\citep{chen2023score,wang2024diffusion,stanczuk2024diffusion,mei2023deep,li2023generalization,azangulov2024convergence,li2024understanding,li2024shallow,cui2025precise}.
For instance, \citet{chen2023score} considered the case where the target distribution lies on a linear subspace and developed sample complexity bounds for score matching that are independent of the ambient dimension. \citet{wang2024diffusion} assumed the target distribution to be a mixture of low-rank Gaussians and explored the equivalence between score matching in this setting and subspace clustering.
\citet{tang2024adaptivity} studied the case when the data are supported on low-dimensional manifolds, and provided explicit convergence rates highlighting the importance of score estimation methods in such settings. \citet{stanczuk2024diffusion} showed that diffusion models encode the data manifold by approximating its
normal bundle.
Moreover, \citet{li2023generalization} established theoretical estimates of the generalization gap that evolves with the training dynamics of score-based diffusion models, suggesting a polynomially small generalization error that evades the curse of dimensionality.

%% file: discussion.tex
\section{Discussion}

 We have made progress towards understanding how diffusion models harness (unknown) low-dimensional structure to accelerate data generation. 
 For the DDIM sampler, 
we have provided the first theoretical guarantee demonstrating its ability to adapt to low-dimensional structure. 
 When it comes to the DDPM sampler, we have improved the TV-based iteration complexity from $\widetilde{O}(k/\varepsilon^2)$ \citep{potaptchik2024linear,huang2024denoising} to  $\widetilde{O}(k/\varepsilon)$ under exact scores. We have further extended these guarantees to the setting with noisy score estimates under suitable score estimation assumptions. Moreover, we have demonstrated that these assumptions can be achieved by score estimators with finite-sample guarantees that adapt to the underlying low-dimensional structure. Notably, our analysis covers the standard coefficient choices of \citet{ho2020denoising,song2020denoising}, and our lower bounds highlight how these designs are essential for enabling low-dimensional adaptation.

Before concluding, we mention two directions for future work. First, the DDIM
coefficients in Theorem~\ref{thm:ddim_tv_conv} are a special case of the
coefficient class in \eqref{eq:ddpm_gen_stp_choice} used in
Theorem~\ref{thm:ddpm_tv_conv}. However, our DDPM analysis does not yet cover
the regime \(\eta_t \gtrsim (1-\alpha_t)\sigma_t^2\), leaving a gap between
the DDIM and DDPM theories. Closing this gap could lead to a more unified
framework for both samplers. Second, Theorem~\ref{thm:lower_bound} gives a
single-step lower bound for the discretized reverse process. Extending it to
multi-step lower bounds for broader SDE/ODE-based samplers is another
interesting direction.



%
%

%% file: technical_lemmas.tex
\section{Interpretation from the lens of differential equations}
\label{sec:interpretation}


In order to help elucidate why DDIM and DDPM are adaptive to low dimensionality,
we take a moment to derive their exact correspondence to reversed differential equations.
This viewpoint unearths the underlying forces that steer their trajectories toward the low-dimensional structure of interest, despite the effects of time discretization.
For convenience of presentation, we overload the notation by setting
\begin{align}
    \overline{\alpha}_t = \exp\left(-2\int_{0}^t \beta(s) \rd s\right),
    \qquad \text{for all }t\in [0,T],
\end{align}
where $\beta(t)$ denotes the coefficient schedule in the forward SDE~\eqref{eq:forward-SDE}; as alluded to previously, this function $\overline{\alpha}_t$ coincides with  $\{\overline{\alpha}_t\}_{t=1}^T$ defined in \eqref{eq:bar-alphat-discrete} for the discrete-time process. In addition,
we recall from the Tweedie formula \citep{efron2011tweedie} that
\begin{subequations}
\begin{align}
    \mu_t^{\star}(x) \coloneqq \mathbb{E}[X_0 \mymid  X_t = x] = \frac{1}{\sqrt{\overline{\alpha}_t}}\big(x + (1-\overline{\alpha}_t)s^{\star}_t(x)\big),
    \label{eq:defn-mut-star}
\end{align}
and also introduce the noisy counterpart:
\begin{align}
    \mu_t(x) \coloneqq \frac{1}{\sqrt{\overline{\alpha}_t}}\big(x + (1-\overline{\alpha}_t)s_t(x)\big).
    \label{eq:defn-mut}
\end{align}
\end{subequations}



In the sequel, we isolate several discretized differential equations that correpond exactly with the DDIM and DDPM samplers considered in the present paper.
\begin{enumerate}[label = $\bullet$]

    \item \textit{DDIM sampler.}
    The probability flow ODE~\eqref{eq:continue-time-ode} can be reparametrized by $\mu_t^{\star}$ as follows using Tweedie's formula \eqref{eq:defn-mut-star}:
    \begin{align*}
        \rd Y_t = \left(-\frac{\overline{\alpha}_{T-t}}{1-\overline{\alpha}_{T-t}}Y_t + \frac{\sqrt{\overline{\alpha}_{T-t}}}{1-\overline{\alpha}_{T-t}}\mu^*_{T-t}(Y_{t})\right)\beta(T-t)\rd t ,
    \end{align*}
    where the drift term exhibits a semi-linear structure. To approximately solve this ODE, one can apply the exponential integrator scheme on the estimated semi-linear structure and select time discretization points as $t_n = n$ for all $n=0,1,\cdots, T$,  leading to the discretized dynamics below:
    \begin{align}\label{eq:ddim-dynamic}
        \rd \widetilde{Y}_t = \left(-\frac{\overline{\alpha}_{T-t}}{1-\overline{\alpha}_{T-t}}\widetilde{Y}_t + \frac{\sqrt{\overline{\alpha}_{T-t}}}{1-\overline{\alpha}_{T-t}}\mu_{T-t_n}(\widetilde{Y}_{t_n})\right)\beta(T-t)\rd t, \quad t \in [t_n, t_{n+1}).
    \end{align}
    The DDIM sampler is intimately connected with this discretized dynamic, as asserted by the following proposition, whose proof can be found in Appendix~\ref{sec:prf-ode}.
    \begin{proposition}\label{prop:ode}
        The discretized process~\eqref{eq:ddim-dynamic} is solved exactly by the DDIM update rule~\eqref{eq:DDIM-update} with coefficient \eqref{eq:pars-DDIM-original} in the sense that $\widetilde{Y}_{n}=Y_{T-n}$ for all $n=0,1,\dots,T$, provided that $\widetilde{Y}_0=Y_T$.
    \end{proposition}


    \item \textit{(Generalized) DDPM sampler.}
    Similarly, the DDPM-type sampler --- with the coefficients chosen as in  Theorem~\ref{thm:ddpm_tv_conv} --- can be exactly mapped to certain discretized differential equations.
    %
%
    %
    More precisely, consider the generalized semi-linear SDE/ODE~\eqref{eq:gen-continue-time}, which can be reparametrized via $\mu_t^{\star}$ through Tweedie's formula \eqref{eq:defn-mut-star}:
    \begin{align*}
        \rd Y_t = \biggl(-\frac{\xi(T-t) + \overline{\alpha}_{T-t}}{1-\overline{\alpha}_{T-t}}Y_t
        + \frac{\left(1+\xi(T-t)\right)\sqrt{\overline{\alpha}_{T-t}}}{1-\overline{\alpha}_{T-t}}\mu^{\star}_{T-t}(Y_{t})\biggr)\beta(T-t)\rd t\\ + \sqrt{2\xi(Y-t)\beta(T-t)}\,\rd W_t.
    \end{align*}
    Adopting similar discretization scheme as in \eqref{eq:ddim-dynamic}, we arrive at the following discretized process:
    \begin{align}\label{eq:gen-dynamic}
        \rd \widetilde{Y}_t = \biggl(-\frac{\xi(T-t_n)+\overline{\alpha}_{T-t}}{1-\overline{\alpha}_{T-t}}\widetilde{Y}_t &+ \frac{\left(1+\xi(T-t_n)\right)\sqrt{\overline{\alpha}_{T-t}}}{1-\overline{\alpha}_{T-t}}\mu_{T-t_n}(\widetilde{Y}_{t_n})\biggr)\beta(T-t)\rd t \notag\\
        & + \sqrt{2\xi(T-t_n)\beta(T-t)}\,\rd W_t, \qquad t \in [t_n, t_{n+1}),
    \end{align}
    where we recall that $t_n=n$.
    Interestingly, the DDPM-type samplers considered in Theorem~\ref{thm:ddpm_tv_conv} correspond exactly to \eqref{eq:gen-dynamic} with suitably chosen $\xi(t)$, as stated below. The proof of Proposition~\ref{prop:sde} can be found in Appendix~\ref{sec:prf_of_prop_sde}.
    \begin{proposition}\label{prop:sde}
        The discretized process~\eqref{eq:gen-dynamic} can be solved exactly by the (generalized) DDPM update rule~\eqref{eq:DDPM-update} --- with coefficients satisfying \eqref{eq:ddpm_gen_stp_choice} --- in the sense that $\widetilde{Y}_{n}=Y_{T-n}$ for all $n=0,1,\dots,T$, provided that $\widetilde{Y}_0=Y_T$ and that the standard Gaussian vectors $\{Z_t\}$ are chosen properly based on $(W_t)$.
    \end{proposition}
    %

    \begin{remark}
        In the special case with coefficients \eqref{eq:eta-sigma-DDPM-original},
        the precise connection between such a discretized differential equation and the original DDPM sampler \citep{ho2020denoising} has been discussed and utilized in the recent work \citet{azangulov2024convergence,potaptchik2024linear,huang2024denoising}.
    \end{remark}

\end{enumerate}

With the equivalent description \eqref{eq:ddim-dynamic} (resp.~\eqref{eq:gen-dynamic}) of the DDIM (resp.~DDPM) sampler,
one can already gain insight into how these samplers adapt to unknown low-dimensional structure.
Suppose that we have access to accurate scores, so that $\mu^{\star}_t=\mu_t$. A closer inspection of \eqref{eq:ddim-dynamic} and \eqref{eq:gen-dynamic} reveals that: the nonlinear components of the drift terms of both processes are proportional to $\mu^{\star}_{T-t_n}$,
which is defined as the conditional expectation of $X_0$ (cf.~\eqref{eq:defn-mut-star}). In other words, the most critical drift components take the form of conditional expectation of $X_0$, which inherently capture the low-dimensional structure of $\pdata$ and steer the sampling dynamics towards this inherent structure.


\section{Technical lemmas}

In this section, we present several technical lemmas that prove useful for establishing our main theorems, with their proofs deferred to Appendix~\ref{sec:aux_lem&prf}.
For simplicity of presentation, we assume without loss of generality that \( k \geq \log d \) throughout the proof.

Before proceeding, let us introduce several notation that will be useful throughout.
\begin{itemize}
\item
For any covering scale \(\epsilon_0>0\) such that
\[
\epsilon_0 \le c_{\epsilon}\,
\frac{1\wedge \sqrt{k\log T/d}}{T}
\]
for a sufficiently small universal constant \(c_{\epsilon}>0\).
Let \( \{x_i^\star\}_{1 \leq i \leq N_{\epsilon_0}} \) be an \(\epsilon_0\)-net of \( \mathcal{X}_{\mathsf{data}} \), with $N_{\epsilon_0}$ denoting its cardinality.
Let \( \{ \mathcal{B}_i \}_{1 \leq i \leq N_{\epsilon_0}} \) be a disjoint \(\epsilon_0\)-cover for \( \mathcal{X}_{\mathsf{data}} \) such that \( x_i^\star \in \mathcal{B}_i \) for each $i$.
See, e.g., \cite{vershynin2018high}, for the definition of epsilon-net and epsilon-cover.

\item
Define the following two sets:
\begin{align}
\mathcal{I} \coloneqq \{ 1 \leq i \leq N_{\epsilon_0} : \mathbb{P}(X_0 \in \mathcal{B}_i) \geq \exp(-C_1 k \log (R_0/\epsilon_0)) \}
\label{eq:defn-I-set}
\end{align}
and
\begin{equation}
\begin{aligned}
\label{eq:defn-G-set-construct}
\mathcal{G} \coloneqq& \biggl\{ \omega \in \mathbb{R}^d : \|\omega\|_2 \leq 2 \sqrt{d} + \sqrt{C_1 k \log (R_0/\epsilon_0)},\\
&\hspace{2em}\abs{(x_i^\star - x_j^\star)^\top \omega} \leq \sqrt{C_1 k \log (R_0/\epsilon_0)}~ \|x_i^\star - x_j^\star\|_2,~\forall 1 \leq i,j \leq N_{\epsilon_0} \biggr\}.
\end{aligned}
\end{equation}
for some sufficiently large universal constant \( C_1 > 0 \). As it turns out,  \( \bigcup_{i \in \mathcal{I}} \mathcal{B}_i \) and \( \mathcal{G} \) form certain high-probability sets related to the random vector \( X_0 \sim \pdata \) and the standard Gaussian random vector in \( \mathbb{R}^d \), respectively.

\item
As mentioned previously in \eqref{eq:forward-marginal}, we can  express
\begin{equation}
X_t = \sqrt{\overline{\alpha}_t}X_0 + \sqrt{1 - \overline{\alpha}_t}Z
\end{equation}
for some random vector $Z\sim \mathcal{N}(0,I_d)$. By introducing
\begin{equation}
V_\alpha \coloneqq \sqrt{\alpha}V_1 + \sqrt{1 - \alpha} Z \qquad \text{with }  V_1 \coloneqq X_0
\end{equation}
for any $\alpha \in [0,1]$,
we can write
\begin{equation}
X_t = V_{\overline{\alpha}_t}
\end{equation}
for any \( t \).
For every $\alpha \in [0,1 - 1/T]$, define a typical set for \( V_{\alpha} \) as follows:
\begin{equation}
\label{eq:defn-T-alpha-appendix}
\mathcal{T}_{\alpha} \coloneqq \bigg\{ \sqrt{\alpha} v_1 + \sqrt{1 - \alpha} \,\omega : v_1 \in \bigcup_{i \in \mathcal{I}} \mathcal{B}_i, \omega \in \mathcal{G} \bigg\}.
\end{equation}

\item Next, we turn to the posterior distribution of \( V_1 \) given \( V_\alpha \), which dictates the performance of DDIM and DDPM samplers. Let us introduce the following shorthand notation:
\begin{subequations}
\label{eq:posterior_mu_cov}
\begin{equation}
\begin{aligned}
\mu_{V_1|V_\alpha}(v) &\coloneqq \mathbb{E}[V_1\mid V_\alpha = v]
,\\
\Cov_{V_1|V_\alpha}(v) &\coloneqq \EB [V_1V_1^\top \mid V_\alpha = v] - \mu_{V_1|V_\alpha}(v)\mu_{V_1|V_\alpha}(v)^\top.
\end{aligned}
\end{equation}
Given that the random objects $\mu_{V_1\mid V_{\overline{\alpha}_t}}(V_{\overline{\alpha}_t})$ and $\Cov_{V_1\mid V_{\overline{\alpha}_t}}(V_{\overline{\alpha}_t})$ will be used frequently throughout the proof, we shall often employ the following shorthand notation
\begin{align}
\mu_{0|t} \coloneqq \mu_{V_1|V_{\overline{\alpha}_t}}
\qquad \text{and} \qquad \Cov_{0|t}\coloneqq \Cov_{V_1|V_{\overline{\alpha}_t}}
\end{align}
\end{subequations}
as long as it is clear from the context.

\item In addition, we find it convenient to define
\begin{equation}
\varepsilon^{\mathsf{sc}}_{t}(x) \coloneqq s_t(x) - s_t^{\star}(x)
\qquad \text{and}\qquad
\varepsilon^{\mathsf{J}}_{t}(x) \coloneqq \frac{\partial s_t(x)}{\partial x} - \frac{\partial s_t^{\star}(x)}{\partial x}.
\label{eq:defn-eps-t-sc-J}
\end{equation}

\end{itemize}


With the above set of notation in place, let us proceed to present a couple of technical lemmas.
 While some of these proofs can be found in \citet{li2024adapting,huang2024denoising}, we provide the proofs here for the sake of completeness.

The first lemma demonstrates that, for any \( \alpha \in [0, 1 - 1/T] \), \( \mathcal{T}_\alpha \) (cf.~\eqref{eq:defn-T-alpha-appendix}) is a high-probability set for \( V_\alpha \). The proof of this result is deferred to Appendix~\ref{sec:prf_of_gT_alpha_typical}.
\begin{lemma}\label{lem:gT_alpha_typical}
    There exists some universal constant \( C_1 \ge 16C_{\mathsf{cover}}\) such that for any \( \alpha \in [0,1-1/T] \)
    and any covering scale \(\epsilon_0\),
    the set \( \mathcal{T}_\alpha \) defined in \eqref{eq:defn-T-alpha-appendix} satisfies
\[
\mathbb{P}\left( V_\alpha \notin \mathcal{T}_\alpha \right) \leq \exp\left( -\frac{C_1}{4} k \log (R_0/\epsilon_0) \right).
\]
\end{lemma}

Next, we develop a lemma that characterizes the concentration property of the point $V_1$ given the observation $V_\alpha$; the proof can be found in Appendix~\ref{sec:prf_of_posterior_norm}.

\begin{lemma}\label{lem:posterior_norm} For any given covering scale \(\epsilon_0\), 
    consider any $v \in \gT_{\alpha}$, and write  $v = \sqrt{\alpha}v_1^{\star} + \sqrt{1 - \alpha}\,\omega$ for some $v_1^{\star} \in \bigcup_{i\in \gI}\gB_i$ and $\omega \in \gG$ (cf.~\eqref{eq:defn-T-alpha-appendix}). Suppose that $v_1^{\star} \in \gB_{i(v)}$ for some $i(v)\in \mathcal{I}$. Then there exists some universal constant $C_2>0$ such that for any quantity $C \ge C_2$, one has
    $$
    \PB\left( \sqrt{\alpha}\,\big\| V_1 - x^{\star}_{i(v)} \big\| \ge \sqrt{C k (1-\alpha)\log (R_0/\epsilon_0)}
     \mid  V_\alpha = v \right) \le \exp\left(-\frac{C}{20}k \log (R_0/\epsilon_0)\right).
    $$
\end{lemma}

The above lemma has an immediate consequence that upper bounds on the moments of $V_1$ under the posterior distribution $\PB(\cdot \mid V_\alpha = v)$,  provided that $v\in \gT_\alpha$.
This is stated in the following corollary, whose proof (which is fairly straightforward) is omitted for the sake of brevity.
\begin{corollary}\label{cor:posterior_mmt}
There exists a universal constant $C_3>0$, such that for any $\alpha\in[0, 1 - 1/T]$, the following inequalities hold for any point $v\in \gT_\alpha$:
\begin{equation}
\begin{aligned}\label{eq:posterior_mmt}
    \EB \left[\norm{V_1 - \mu_{V_1|V_\alpha}(v)}_2^l ~\big|~ V_\alpha = v\right] \le C_3 \left(\frac{1-{\alpha}}{{\alpha}}k\log T\right)^{l/2},
    \qquad l = 1,2,3,4.
\end{aligned}
\end{equation}
\end{corollary}

Moreover, we single out the lemma below that can help control the posterior covariance of interest. The proof is postponed to Appendix~\ref{sec:proof:lem:bound_CoV_1|t_F^2}.
\begin{lemma}\label{lem:bound_CoV_1|t_F^2}
Suppose that Assumptions~\ref{ass:low_dim} and \ref{ass:bd_supp} hold. Denote $\widetilde{\sigma}_t^2=\frac{\overline{\alpha}_t(1 - \alpha_t)}{(\alpha_t - \overline{\alpha}_t)(1 - \overline{\alpha}_t)}$. Then for any $t  \ge 2$, the posterior covariance defined in \eqref{eq:posterior_mu_cov} satisfies
    \begin{align*}
        \widetilde{\sigma}_t^2\EB_{X_t} \Big[\norm{\Cov_{0|t}(X_t)}_{\mathrm{F}}^2 \Big] \le 3\Big\{\EB \big[\tr\big(\Cov_{0|t}(X_t)\big) \big] - \EB \big[\tr\big(\Cov_{0|t-1}(X_{t-1})\big) \big]\Big\} + \frac{3}{T^{10}}.
    \end{align*}
\end{lemma}

 We also make note of the following basic property about $\{\alpha_t\}$ (see \citet[Section~5.1]{li2024sharp}):
    \begin{equation}
    \label{eq:basic-alphat-property}
    \frac{1}{2} \frac{1 - \alpha_t}{1 - \overline{\alpha}_t}
\leq \frac{1}{2} \frac{1 - \alpha_t}{\alpha_t - \overline{\alpha}_t}
\leq \frac{1 - \alpha_t}{1 - \overline{\alpha}_{t-1}}
\leq \frac{4c_1 \log T}{T}
\qquad \text{for any }2\leq t\leq T.
    \end{equation}
    The lemma below is a consequence of this property, whose proof can be found in Appendix~\ref{sec:proof:lem:step_size}.
%
%
\begin{lemma}\label{lem:step_size}
There exists some universal constant $C_6>0$ such that for any $t \ge 1$,
    $$
    \frac{\overline{\alpha}_t(1 - \alpha_t)}{(\alpha_t - \overline{\alpha}_t)(1 - \overline{\alpha}_t)} -
      \frac{\overline{\alpha}_{t+1}(1 - \alpha_{t+1})}{(\alpha_{t+1} - \overline{\alpha}_{t+1})(1 - \overline{\alpha}_{t+1})}
     \le
     \frac{C_6 \log^2 T}{T^2}\frac{\overline{\alpha}_{t}}{1 - \overline{\alpha}_t}.
    $$
\end{lemma}

In addition, the lemma below helps one control the tightness of the  Taylor expansion of a certain log-determinant function. Its proof is deferred to Appendix~\ref{sec:proof:lem:logdet_expand}.
\begin{lemma}\label{lem:logdet_expand}
    Let \(  A \in \mathbb{R}^{d \times d} \) be any positive semidefinite matrix, and let \( \Delta \in \mathbb{R}^{d \times d} \) be any square matrix. Suppose that \( \eta \| \Delta \| \le 1/4 \),
where \( 0 < \eta < 1 \). Then it holds that
    \[
    \log \det\left( I + \eta  A + \eta \Delta\right) \ge \eta\big(\tr( A) + \tr(\Delta)\big) - 4\eta^2\left(\norm{ A}_{\mathrm{F}}^2 + \norm{\Delta}_{\mathrm{F}}^2\right).
    \]
\end{lemma}

Finally, we state below the Tweedie formula \citep{efron2011tweedie}, which establishes the intimate connection between the score function (resp.~its corresponding Jacobian matrix) and the posterior mean (resp.~posterior covariance) of \( X_0 \) given \( X_t \). For its proof, one can refer to~\citet{robbins1992empirical}.
\begin{equation}\label{eq:score_to_posterior_new}
\begin{aligned}
    s_t^{\star}(x_t) &= \frac{\sqrt{\overline{\alpha}_t}}{1 - \overline{\alpha}_t} \mu_{0|t}(x_t) - \frac{1}{1-\overline{\alpha}_t}x_t,
    \\
    \frac{\partial s_t^{\star}(x_t)}{\partial x_t} &=
    \frac{\overline{\alpha}_t}{(1-\overline{\alpha}_t)^2}\Cov_{0|t}(x_t) - \frac{1}{1 - \overline{\alpha}_t} I.
\end{aligned}
\end{equation}
where
\begin{subequations}\label{eq:def_mu_cov}
\begin{align}
    \mu_{0|t}(x_t) &= \mathbb{E}[X_0 \mid X_t= x_t], \\
    \Cov_{0|t}(x_t) &= \mathbb{E}[X_0X_0^{\top} \mid X_t= x_t]
    - \mathbb{E}[X_0 \mid X_t= x_t] \,\mathbb{E}[X_0 \mid X_t= x_t]^{\top}.
\end{align}
\end{subequations}

%% file: DDIM_Analysis.tex
\section{Analysis for DDIM (proof of Theorem~\ref{thm:ddim_tv_conv})}\label{sec:k/T_DDIM_Analysis}

In this section, we establish our convergence guarantees for the DDIM sampler as stated in Theorem~\ref{thm:ddim_tv_conv}.

\subsection{Main steps for proving Theorem~\ref{thm:ddim_tv_conv}}

Our proof consists of several steps, which we present below.

\medskip
\noindent {\bf Preparation.}
Before proceeding to the proof, let us introduce a set of useful notation, to be used throughout this section. First, we define the following (deterministic) mapping:
\begin{align}
\Phi_t(x) = \frac{1}{\sqrt{{\alpha}_t}}\big(x+\eta_t s_t(x)\big)
\qquad \text{with }\eta_t = \frac{1-\alpha_t}{1 + \sqrt{\frac{\alpha_t - \overline{\alpha}_t}{1-\overline{\alpha}_t}}}.
\label{eq:defn-Phit-DDIM-etat}
\end{align}
%
%

\begin{lemma}\label{lem:inverse_mapping}
   For any $t = 1,\cdots , T$, the mapping ${\Phi}_t$ defined by~\eqref{eq:defn-Phit-DDIM-etat} is a $C^1-$diffeomorphism on $\RB^d$.
\end{lemma}
The proof of Lemma~\ref{lem:inverse_mapping} is deferred to Appendix~\ref{sec:proof:lem:inverse_mapping}.

\medskip
\noindent {\bf Step 1: linking $\TV(p_{X_{t-1}}, p_{Y_{t-1}})$ with $\TV(p_{X_{t}}, p_{Y_{t}})$.}
To begin with, we single out
the following recursion that plays a pivotal role in our analysis: for any $t \ge 2$,
\begin{align}
    \TV(p_{X_{t-1}}, p_{Y_{t-1}}) &{=} \sup\limits_{\gA}\left\{\PB_{X_{t-1}}(\gA) - \PB_{Y_{t-1}}(\gA)\right\}
    = \sup\limits_{\gA}\left\{
    \PB_{X_{t-1}}(\gA) - \PB_{Y_{t}}\big(\Phi_t^{-1}(\gA)\big)
    \right\} \notag\\
    &{\le}
    \sup\limits_{\gA}\left\{\PB_{X_{t-1}}(\gA) {-} \PB_{X_t}\big(\Phi_t^{-1}(\gA)\big)\right\} {+} \sup\limits_{\gA}\left\{
    \PB_{X_{t}}\big(\Phi_t^{-1}(\gA)\big) {-} \PB_{Y_t}\big(\Phi_t^{-1}(\gA)\big)
    \right\} \notag\\
    &\le \sup\limits_{\gA}\left\{\PB_{X_{t-1}}(\gA) - \PB_{\Phi_t(X_t)}(\gA)\right\} + \TV(p_{X_t}, p_{Y_t}) \notag\\
    &= \TV(p_{X_{t-1}}, p_{\Phi_t(X_t)}) + \TV(p_{X_t}, p_{Y_t}),
    \label{eq:ddim_tv_xy*_eq1}
\end{align}
where the first identity arises from the basic property of the TV distance \citep{tsybakov2009introduction}.
This relation \eqref{eq:ddim_tv_xy*_eq1} underscores the importance of controlling
$\TV(p_{X_{t-1}}, p_{\Phi_t(X_t)})$  when linking the TV distances of interest across two adjacent steps. Notably, in this extra TV distance term, the randomness of both $X_{t-1}$ and $ {\Phi}_t(X_t)$ comes only from the forward process.

\medskip
\noindent {\bf Step 2: identifying a crucial relation on the difference between $p_{X_{t-1}}$ and $ p_{{\Phi}_t(X_t)}$.}
In order to bound $\TV(p_{X_{t-1}}, p_{\Phi_t(X_t)})$, we need to examine the difference between $p_{X_{t-1}}$ and $ p_{\Phi_t(X_t)}$. As it turns out, it would be helpful to first control the discrepancy between $p_{X_{t-1}}$ and $ p_{{\Phi}_t(X_t)}$, given that ${\Phi}_t$ exhibits some invertibility property when restricted to the set $\mathcal{E}_t$.

In view of  Lemma~\ref{lem:inverse_mapping}
for any $x_{t-1} \in \RB^d$, there exists a unique $x_t \in \RB^d$ obeying  ${\Phi}_t(x_t) = x_{t-1}$,
which in turn allows us to write
\begin{align}
p_{{\Phi}_t(X_t)}(x_{t-1}) = p_{X_t}\big({\Phi}_{t}^{-1}(x_{t-1})\big)\cdot \det\left(\frac{\partial {{\Phi}_t^{-1}}(x_{t-1})}{\partial x_{t-1}}\right) = p_{X_t}(x_t)\cdot \det\left(\frac{\partial x_t}{\partial x_{t-1}}\right).
\label{eq:p-Phit-tilde-t-1}
\end{align}
Consequently, we can demonstrate that: for any $t\geq 2$ and any $x_{t-1} \in \RB^d$, it holds that
\begin{align}
     p_{{\Phi}_t(X_t)}&(x_{t-1}) - p_{X_{t-1}}(x_{t-1})
=p_{X_t}(x_t) \det\left(\frac{\partial x_t}{\partial x_{t-1}}\right) - p_{X_{t-1}}(x_{t-1}) \notag\\
&=\int \left\{ p_{X_t\mymid  X_0}(x_t\mymid  x_0) \det\left(\frac{\partial x_t}{\partial x_{t-1}}\right) - p_{X_{t-1}\mymid  X_0}(x_{t-1}\mymid  x_0)
\right\}p_{X_0}(x_0)\rd x_0 \notag\\
&=
\int \Bigg \{1 - \frac{p_{X_{t-1}\mymid  X_0}(x_{t-1}\mymid  x_0)}{p_{X_t\mymid  X_0}(x_t\mymid  x_0)}
\det\left(\frac{\partial x_{t-1}}{\partial x_{t}}\right)\Bigg\}p_{X_t,X_0}(x_t,x_0)
\det\left(\frac{\partial x_{t}}{\partial x_{t-1}}\right)\rd x_0 .\label{eq:p_Xt-1-p_Phi_X_t}
\end{align}
Moreover, recalling how $X_{t-1}$ and $X_t$ are generated, we can decompose
\begin{equation}\label{eq:gT=gT_1xgT_2}
\begin{aligned}
& \frac{p_{X_{t-1}\mymid  X_0}(x_{t-1}\mymid  x_0)}{p_{X_t\mymid  X_0}(x_t\mymid  x_0)}
\det\left(\frac{\partial x_{t-1}}{\partial x_{t}}\right)
=
\frac{\left(\frac{1}{ 1{-}\overline{\alpha}_{t-1}}\right)^{\frac{d}{2}}\exp\Big\{
    \frac{{-}\|x_{t-1} {-} \sqrt{\overline{\alpha}_{t-1}}x_0\|_2^2}{2(1-\overline{\alpha}_{t-1})}
    \Big\}}{\left(\frac{1}{ 1{-}\overline{\alpha}_{t}}\right)^{\frac{d}{2}}\exp\Big\{
    \frac{{-}\|x_{t} {-} \sqrt{\overline{\alpha}_{t}}x_0\|_2^2}{2(1-\overline{\alpha}_{t})}
    \Big\}}
    \det\left(\frac{\partial x_{t-1}}{\partial x_{t}}\right)\\
    &\qquad\quad =
    \underbrace{\left(\frac{1-\overline{\alpha}_t}{1-\overline{\alpha}_{t-1}}\right)^{\frac{d}{2}}\det\left(\frac{\partial x_{t-1}}{\partial x_t}\right)}_{
    \eqqcolon \gT_1(x_t,x_0)
    }
    \underbrace{\exp\left\{
    \frac{\norm{x_t -\sqrt{\overline{\alpha}_t}x_0}_2^2}{2(1-\overline{\alpha}_t)} -
    \frac{\norm{x_{t-1} - \sqrt{\overline{\alpha}_{t-1}}x_0}_2^2}{2(1-\overline{\alpha}_{t-1})}
    \right\}}_{\eqqcolon \gT_2(x_t,x_0)}.
\end{aligned}
\end{equation}
As a result, for any $x_{t-1} \in \RB^d$ we have
\begin{align}
     p_{{\Phi}_t(X_t)}(x_{t-1}) &- p_{X_{t-1}}(x_{t-1})\notag\\
&= \int \big( 1 - \gT_1(x_t,x_0)\gT_2(x_t,x_0)\big) p_{X_t,X_0}(x_t,x_0)
\det\left(\frac{\partial x_{t}}{\partial x_{t-1}}\right)\rd x_0.
\label{eq:p_Xt-1-p_Phi_X_t-Gamma12}
\end{align}
%


\medskip
\noindent {\bf Step 3: calculating  $\gT_1(x_t,x_0)$ and $\gT_2(x_t,x_0)$.}
In order to take advantage of the above relation \eqref{eq:p_Xt-1-p_Phi_X_t-Gamma12},
an important task is to quantify the two terms $\gT_1(x_t,x_0)$ and $\gT_2(x_t,x_0)$. For notational convenience, define
    \begin{align}
    \xi_t(x_t,x_0) &\coloneqq
    \left(1 {-} \frac{\eta_t}{1-\overline{\alpha}_t}\right)\frac{\sqrt{\overline{\alpha}_t}\eta_t}{(\alpha_t - \overline{\alpha}_t)(1 - \overline{\alpha}_t)}(x_t - \sqrt{\overline{\alpha}_t}x_0)^\top \big(x_0 - \mu_{0|t}(x_t)\big)\notag\\
&\qquad {-} \frac{\overline{\alpha}_t\eta_t^2}{2(\alpha_t - \overline{\alpha}_t)(1 - \overline{\alpha}_t)^2}\norm{x_0 - \mu_{0\mymid  t}(x_t)}_2^2\notag{+}
\frac{\eta_t}{\alpha_t - \overline{\alpha}_t}\big(\mu_{0|t} (x_t) - x_0 \big)^\top \varepsilon^{\mathsf{sc}}_{t}(x_t)\notag\\
&\qquad{+}\left(1 - \frac{\eta_t}{2(1 - \overline{\alpha}_t)}\right)\frac{\overline{\alpha}_t \eta_t}{(\alpha_t - \overline{\alpha}_t)(1 - \overline{\alpha}_t)}\tr\big(\Cov_{0\mymid  t}(x_t)\big) ;
\label{defn:xi-xt-x0-DDIM}  \\
W_t(x_t)& \coloneqq \log \det\left( I + \sqrt{\frac{1 - \overline{\alpha}_t}{\alpha_t - \overline{\alpha}_t}}\frac{\overline{\alpha}_t \eta_t}{(1-\overline{\alpha}_t)^2}\Cov_{0|t}(x_t) + \sqrt{\frac{1 - \overline{\alpha}_t}{\alpha_t - \overline{\alpha}_t}}\eta_t \varepsilon^{\mathsf{J}}_{t}(x_t)\right)
\notag\\
&\qquad
- \frac{\eta_t^2}{2(\alpha_t - \overline{\alpha}_t)}\norm{\varepsilon^{\mathsf{sc}}_{t}(x_t)}_2^2
-\sqrt{\frac{1- \overline{\alpha}_t}{\alpha_t - \overline{\alpha}_t}}\eta_t s_t^{\star}(x_t)^\top \varepsilon^{\mathsf{sc}}_{t}(x_t)\notag\\
&\qquad
-\frac{\overline{\alpha}_t \eta_t}{(\alpha_t - \overline{\alpha}_t)(1 - \overline{\alpha}_t)}\left(1 - \frac{\eta_t}{2(1 - \overline{\alpha}_t)}\right)\tr\left(\Cov_{0|t}(x_t)\right).
\label{eq:defn-Wt-DDIM}
\end{align}
The following lemma, whose proof is deferred to Appendix~\ref{sec:proof:lem:bounds-T1-T2-DDIM}, establishes an intimate connection between $(\mathcal{T}_1, \mathcal{T}_2)$ and $(\xi_t,W_t)$.
\begin{lemma}
    \label{lem:bounds-T1-T2-DDIM}
    The quantities $\gT_1(x_t,x_0)$ and $\gT_2(x_t,x_0)$ defined in~\eqref{eq:gT=gT_1xgT_2} satisfy
    \begin{align}
    \label{eq:prod-T1-T2-xit-Wt-DDIM}
\gT_1(x_t,x_0)\gT_2(x_t,x_0)
        = e^{\xi_t(x_t,x_0)}\cdot e^{W_t(x_t)}.
    \end{align}
Further, it can be derived that
\begin{align}
\int_{x_0}\xi_t(x_t,x_0)p_{X_0\mymid  X_t}(x_0\mymid  x_t)\rd x_0 = 0 \qquad \text{for all } x_t \in \R^d.
\label{eq:int-xi-xt-x0-zero}
\end{align}
\end{lemma}
%
%
%

\medskip
\noindent {\bf Step 4: bounding $p_{{\Phi}_t(X_t)}- p_{X_{t-1}}$.}
We can now invoke the preceding results to upper bound $p_{{\Phi}_t(X_t)}- p_{X_{t-1}}$.  Focusing on any  $\gA\subseteq \RB^d$, one can put
\eqref{eq:p_Xt-1-p_Phi_X_t-Gamma12} and Lemma~\ref{lem:bounds-T1-T2-DDIM} together to show that
\begin{align}
    \PB_{{\Phi}_t(X_t)}(\gA) -& \PB_{X_{t-1}}(\gA)  = \int_{\gA}\left\{ p_{{\Phi}_t(X_t)}(x_{t-1}) - p_{X_{t-1}}(x_{t-1})\right\}\rd x_{t-1} \notag\\
    \overset{\text{(a)}}{=} & \int_{\gA\times \R^d} \big\{ 1 - \gT_1(x_t,x_0)\gT_2(x_t,x_0) \big\}p_{X_t, X_0}(x_t,  x_0)\det\left(\frac{\partial x_t}{\partial x_{t-1}}\right)\rd x_0 \rd x_{t-1} \notag\\
    \overset{\text{(b)}}{=}&
    \int_{{\Phi}_t^{-1}(\gA)\times \R^d}\left\{1 - e^{\xi_t(x_t,x_0)}\cdot e^{W_t(x_t)}\right\}p_{X_t, X_0}(x_t, x_0) \rd x_0 \rd x_t \notag\\
    =&
    \int_{{\Phi}_t^{-1}(\gA)\times \R^d} \left\{\big(1 - e^{\xi_t(x_t,x_0)}\big)e^{W_t(x_t)} + \big(1 - e^{W_t(x_t)}\big)\right\}p_{X_t,X_0}(x_t,x_0)\rd x_0\rd x_t \notag\\
    \overset{\text{(c)}}{\le}&
    \int_{x_t\in {\Phi}_t^{-1}(\gA)}\left\{
    1 - e^{W_t(x_t)}
    \right\}p_{X_t}(x_t)\rd x_t
    \le -\int_{x_t\in {\Phi}_t^{-1}(\gA)}
    W_t(x_t)p_{X_t}(x_t) \rd x_t.
    \label{eq:PB_X_t-1_gA-PB_Phi_X_t_gA_eq1}
\end{align}
%
Here, (a) arises from \eqref{eq:p_Xt-1-p_Phi_X_t-Gamma12} and uses bijection of ${\Phi}_t$ over $\RB^d$ (so that $x_t$ is well-defined given $x_{t-1}\in \RB^d$),  (b) results from the bijection of ${\Phi}_t$ over $\RB^d$ as well as Lemma~\ref{lem:bounds-T1-T2-DDIM},
while the last inequality comes from the elementary inequality $1 - e^x \le -x$.  Also, to explain why (c) is valid,
it suffices to see that for any $x_t \in \R^d$,
\begin{equation}
\label{eq:exp_xi-1_posi}
\begin{aligned}
    \int_{x_0}\left( 1 - e^{\xi_t(x_t,x_0)} \right)e^{W_t(x_t)}&p_{X_0\mymid  X_t}(x_0\mymid  x_t)\rd x_0\\
    &\le -e^{W_t(x_t)}\int_{x_0}\xi_t(x_t,x_0)p_{X_0\mymid  X_t}(x_0\mymid  x_t)\rd x_0 = 0,
\end{aligned}
\end{equation}
where the inequality holds since $1-e^x\leq -x$ for all $x\in \mathbb{R}$, and the last relation is due to \eqref{eq:int-xi-xt-x0-zero}.


\medskip
\noindent {\bf Step 5: bounding the last term in \eqref{eq:PB_X_t-1_gA-PB_Phi_X_t_gA_eq1}.}
Inequality~\eqref{eq:PB_X_t-1_gA-PB_Phi_X_t_gA_eq1} makes apparent the need to control $W_t(x_t)$. The following lemma provided an upper bound on $W_t(x_t)$, whose proof can be found in Appendix~\ref{sec:control-Wx_t}.
\begin{lemma}
\label{lemma:control-Wx_t-DDIM}
For any $x_t \in \RB^d$, the quantity $W_t(x_t)$ defined in \eqref{eq:defn-Wt-DDIM} obeys
\begin{align}
    -W_t(x_t) &\le
    \frac{4\overline{\alpha}_t^2 \eta_t^2}{(\alpha_t -
    \overline{\alpha}_t)(1 - \overline{\alpha}_t)^3}\norm{\Cov_{0|t}(x_t)}_{\mathrm{F}}^2+ \frac{4(1-\overline{\alpha}_t)\eta_t^2}{\alpha_t - \overline{\alpha}_t}\norm{\varepsilon^{\mathsf{J}}_{t}(x_t)}_{\mathrm{F}}^2 + \frac{\eta_t^2}{2(\alpha_t - \overline{\alpha}_t)}\norm{\varepsilon^{\mathsf{sc}}_{t}(x_t)}_2^2\notag\\
    &\qquad + \frac{\overline{\alpha}_t \eta_t^2}{2(\alpha_t - \overline{\alpha}_t)(1 - \overline{\alpha}_t)^2}\tr\left(\Cov_{0|t}(x_t)\right)
    + \sqrt{\frac{1- \overline{\alpha}_t}{\alpha_t - \overline{\alpha}_t}}\eta_t
    \Delta\big(\varepsilon^{\mathsf{sc}}_{t}(x_t), \varepsilon^{\mathsf{J}}_{t}(x_t)\big),\label{eq:EB-W_t_eq2}
\end{align}
where we define
\begin{align}
\Delta\big(\varepsilon^{\mathsf{sc}}_{t}(x_t), \varepsilon^{\mathsf{J}}_{t}(x_t)\big) \coloneqq s_t^{\star}(x_t)^\top \varepsilon^{\mathsf{sc}}_{t}(x_t) - \tr\left(\varepsilon^{\mathsf{J}}_{t}(x_t)\right).
\end{align}
\end{lemma}

In order to take advantage of \eqref{eq:EB-W_t_eq2}, it is also crucial to bound the quantity $\Delta\big(\varepsilon^{\mathsf{sc}}_{t}(x_t), \varepsilon^{\mathsf{J}}_{t}(x_t)\big)$, which we study in the following lemma. The proof is provided in Appendix~\ref{sec:control_est_err}.
%
%
\begin{lemma}\label{lem:bd_Delta_epsi}
 For any set $\gA \subseteq \RB^d$, we have
 \begin{align*}
      \int_{x_t \in \gA}\Delta\big(\varepsilon^{\mathsf{sc}}_{t}(x_t), \varepsilon^{\mathsf{J}}_{t}(x_t)\big) p_{X_t}(x_t) \rd x_t \le \frac{2}{\sqrt{1 - \overline{\alpha}_t}}
     \varepsilon_{\mathsf{score},t} + \varepsilon_{\mathsf{Jacobi},1,t} + \varepsilon_{\mathsf{Jacobi},2,t} + \varepsilon_{\mathsf{Hess},t}.
 \end{align*}
\end{lemma}

With the preceding two lemmas in place, we can establish an upper bound on the integral in \eqref{eq:PB_X_t-1_gA-PB_Phi_X_t_gA_eq1}, as stated in the lemma below.   The proof is deferred to Appendix~\ref{sec:proof:lem:integral-Wt-Et}.
\begin{lemma}
\label{lem:integral-Wt-Et}
The quantity $W_t(x_t)$ defined in \eqref{eq:defn-Wt-DDIM} obeys
\begin{align}
    \int_{\Phi_t^{-1}(\gA)} - W_t(x_t) p_{X_t}(x_t) \rd x_t
    &\le
    \frac{\widetilde{\sigma}_t^2 \eta_t}{2(1 - \overline{\alpha}_t)}\EB_{X_t} \big[\tr\big(\Cov_{0|t}(X_t)\big)\big] + \frac{4(\alpha_t - \overline{\alpha}_t)}{1 - \overline{\alpha}_t}\widetilde{\sigma}_t^4 \EB_{X_t}\big[\norm{\Cov_{0|t}(X_t)}_{\mathrm{F}}^2\big]\notag\\
    &\quad +
    \frac{4(1 - \overline{\alpha}_t)\eta_t^2}{\alpha_t - \overline{\alpha}_t}\varepsilon_{\mathsf{Jacobi},1,t}^2 +
    \frac{\eta_t^2}{2(\alpha_t -\overline{\alpha}_t)}\varepsilon_{\mathsf{score},t}^2 \label{eq:int-W_eq1}\\
    &\quad +\frac{2\eta_t\sqrt{1-\overline{\alpha}_t}}{\sqrt{\alpha_t - \overline{\alpha}_t}}
    \bigg\{\frac{\varepsilon_{\mathsf{score},t}}{
        \sqrt{1-\overline{\alpha}_t}
    } + \varepsilon_{\mathsf{Jacobi},1,t} + \varepsilon_{\mathsf{Jacobi},2,t} + \varepsilon_{\mathsf{Hess},t}\bigg\}.\notag
\end{align}
\end{lemma}



%
%

\medskip
\noindent {\bf Step 6: bounding $\TV(p_{X_{t-1}}, p_{\Phi_t(X_t)})$. }
Putting Lemma~\ref{lem:integral-Wt-Et} and \eqref{eq:PB_X_t-1_gA-PB_Phi_X_t_gA_eq1} together with the definition of the total variation yields
\begin{align}
&\TV(p_{X_{t-1}}, p_{\Phi_t(X_t)}) = \sup\limits_{\gA\subseteq \R^d}\left\{\PB_{X_{t-1}}(\gA) - \PB_{\Phi_t(X_t)}(\gA)\right\}
\le \frac{\widetilde{\sigma}_t^2 \eta_t}{2(1 - \overline{\alpha}_t)}\EB_{X_t} \big[\tr
    \big(\Cov_{0|t}(X_t)\big)\big]
\notag\\
     &+ \frac{4(\alpha_t - \overline{\alpha}_t)}{1 - \overline{\alpha}_t}\widetilde{\sigma}_t^4 \EB_{X_t}\left[\norm{\Cov_{0|t}(X_t)}_{\mathrm{F}}^2\right]
    {+}
    \frac{4(1 - \overline{\alpha}_t)\eta_t^2}{\alpha_t - \overline{\alpha}_t}\varepsilon_{\mathsf{Jacobi},1,t}^2 {+}
    \frac{\eta_t^2}{2(\alpha_t -\overline{\alpha}_t)}\varepsilon_{\mathsf{score},t}^2 \notag\\
    &+\frac{2\eta_t\sqrt{1-\overline{\alpha}_t}}{\sqrt{\alpha_t - \overline{\alpha}_t}}
    \bigg\{\frac{\varepsilon_{\mathsf{score},t}}{
        \sqrt{1-\overline{\alpha}_t}
    } + \varepsilon_{\mathsf{Jacobi},1,t} + \varepsilon_{\mathsf{Jacobi},2,t} + \varepsilon_{\mathsf{Hess},t}\bigg\}.
    \label{eq:tv_x_t-1&phi*x_t}
\end{align}

In addition, Lemma~\ref{lem:bound_CoV_1|t_F^2} tells us that
\begin{align}\label{eq:EB_Cov_F^2}
\frac{\alpha_t - \overline{\alpha}_t}{1 - \overline{\alpha}_t}{\widetilde{\sigma}_t^4}\EB\left[\norm{\Cov_{0|t}(X_t)}_{\mathrm{F}}^2\right] &\le
\frac{3(\alpha_t - \overline{\alpha}_t)}{1 - \overline{\alpha}_t}{\widetilde{\sigma}_t^2} \biggl\{
    \EB \big[\tr\big(\Cov_{X_0\mymid  X_t}(X_t)\big)\big]\notag\\
    &\qquad - \EB \big[\tr\big(\Cov_{X_0|X_{t-1}}(X_{t-1})\big)\big]
    \biggr\} + \frac{1}{T^{10}}.
\end{align}
Taking \eqref{eq:ddim_tv_xy*_eq1}, \eqref{eq:tv_x_t-1&phi*x_t} and \eqref{eq:EB_Cov_F^2} collectively, we can demonstrate that
\begin{align}
    &\TV(p_{X_{t-1}}, p_{Y_{t-1}}) \le \TV(p_{X_{t-1}},p_{\Phi_t(X_t)}) + \TV(p_{X_t}, p_{Y_t})\le
    \TV(p_{X_t}, p_{Y_t})\notag\\
    &{+} \underbrace{\frac{\widetilde{\sigma}_t^2(\eta_t + 3\alpha_t - 3\overline{\alpha}_t)}{1 - \overline{\alpha}_t}\EB\left[\tr\left(
    \Cov_{0|t}(X_t)
    \right)\right] {-} \frac{3(\alpha_t - \overline{\alpha}_t)}{1 - \overline{\alpha}_t}\widetilde{\sigma}_t^2 \EB \left[\tr\left(\Cov_{0|t-1}(X_{t-1})\right)\right] + \frac{1}{T^{10}}}_{\eqqcolon\, \gS_{t,1}} \label{eq:ddim_tv_xy_eq2}\\
    & +\underbrace{\frac{2\eta_t\sqrt{1-\overline{\alpha}_t}}{\sqrt{\alpha_t - \overline{\alpha}_t}}
    \bigg\{\frac{\varepsilon_{\mathsf{score},t}}{
        \sqrt{1-\overline{\alpha}_t}
    } + \varepsilon_{\mathsf{Jacobi},1,t} + \varepsilon_{\mathsf{Jacobi},2,t} + \varepsilon_{\mathsf{Hess},t}\bigg\} {+}
    \frac{4(1 - \overline{\alpha}_t)\eta_t^2}{\alpha_t - \overline{\alpha}_t}\varepsilon_{\mathsf{Jacobi},1,t}^2 +
    \frac{\eta_t^2}{2(\alpha_t -\overline{\alpha}_t)}\varepsilon_{\mathsf{score},t}^2}_{\eqqcolon\, \gS_{t,2}}. \notag
\end{align}

Here, we divide the residual terms into two parts, \( \gS_{t,1} \) and \( \gS_{t,2} \): the term \( \gS_{t,1} \) reflects the discretization error, while \( \gS_{t,2} \) is associated with the score estimation error.
The following lemma helps control the two sums in \eqref{eq:ddim_tv_xy_eq2}, with the proof postponed to  Appendix~\ref{sec:proof-lem:sum-S1-sum-S2-DDIM}.
\begin{lemma}
    \label{lem:sum-S1-sum-S2-DDIM}
    There exist some universal constants $C_{10},C_{11}>0$ such that
    \begin{align}
    \ssum{t}{2}{T}\gS_{t,1} + \ssum{t}{2}{T}\gS_{t,2}
    \leq C_{10}\frac{k\log^3 T}{T}
    + C_{11} \left(\varepsilon_{\mathsf{score}} + \varepsilon_{\mathsf{Jacobi},1} + \varepsilon_{\mathsf{Jacobi},2} +
    \varepsilon_{\mathsf{Hess}}
    \right)\log T.
    \end{align}
\end{lemma}

\medskip
\noindent {\bf Step 7: putting all pieces together.}
To finish up, applying inequality~\eqref{eq:ddim_tv_xy_eq2} recursively from $1$ to $T$, and combining \eqref{eq:ddim_tv_xy_eq3} and \eqref{eq:ddim_tv_xy_eq4}, we reach
\begin{align*}
    \TV(p_{X_1}, p_{Y_1}) &\le \TV(p_{X_T}, p_{Y_T}) + \ssum{t}{2}{T}\gS_{t,1} + \ssum{t}{2}{T}\gS_{t,2}\\
    &\le
    C_{10}\frac{k\log^3 T}{T} + C_{11}\left(\varepsilon_{\mathsf{score}} + \varepsilon_{\mathsf{Jacobi},1} + \varepsilon_{\mathsf{Jacobi},2} +
    \varepsilon_{\mathsf{Hess}}
    \right)\log T + \TV(p_{X_T}, p_{Y_T})\\
    &\le
    C_{10}\frac{k\log^3 T}{T} + C_{11}\left(\varepsilon_{\mathsf{score}} + \varepsilon_{\mathsf{Jacobi},1} + \varepsilon_{\mathsf{Jacobi},2} +
    \varepsilon_{\mathsf{Hess}}
    \right)\log T + \frac{1}{T^{10}},
\end{align*}
%
where the last inequality arises from \citet[Lemma 10]{li2024adapting}. This concludes the proof of Theorem~\ref{thm:ddim_tv_conv}.

\subsection{Proof of Lemma~\ref{lem:inverse_mapping}}\label{sec:proof:lem:inverse_mapping}
To prove this lemma, we make use of the following global inverse function theorem, which is presented in \citep{ruzhansky2015global}
\begin{lemma}[Theorem 2.2 in \citep{ruzhansky2015global}]\label{lem:global_inverse}
    A $C^1$-map $f : \mathbb{R}^d \to \mathbb{R}^d$ is a $C^1$-diffeomorphism iff the Jacobian $\det \left(\nabla f(x)\right)$ never vanishes and $\norm{f(x)}_2 \to \infty$ whenever $\norm{x}_2 \to \infty$.
\end{lemma}

For any $x \in \RB^d$ and any nonzero vector $v \in \RB^d$, Tweedie's formula~\eqref{eq:score_to_posterior_new} tells us that the mapping ${\Phi}_t$ (cf.~\eqref{eq:defn-Phit-DDIM-etat}) obeys
\begin{align*}
v^\top \frac{\partial{\Phi}_t(x)}{\partial x} v &=
v^\top \bigg(I + \eta_t \frac{\partial s_t(x)}{\partial x}\bigg) v\\
&=
\norm{v}_2^2 + \eta_t v^\top \frac{\partial s_t (x)}{\partial x}v \ge \norm{v}_2^2 - \frac{1}{4}\norm{v}_2^2 > 0.
\end{align*}
Here, the penultimate relation follows from Assumption~\ref{ass:ddim_score_matching}.
This result implies that $\frac{\partial{\Phi}_t(x)}{\partial x}$ never vanishes at all points $x \in \RB^d$ uniformly.
On the other hand, for any $x\in \RB^d$, it holds that
\begin{align*}
    \norm{\Phi_t(x)}_2^2 & = \norm{x + \eta_t s_t(x)}_2^2 = \norm{x}_2^2 + 2\eta_t x^\top s_t(x) + \eta_t^2 \norm{s_t(x)}_2^2\\
    &\ge \norm{x}_2^2 + 2\eta_t x^\top s_t(x)
    \overset{(a)}{=}
    \norm{x}_2^2 + 2\eta_t x^\top \frac{\partial s_t (\vartheta)}{\partial \vartheta} x\\
    &\overset{(b)}{\ge} \norm{x}_2^2 - \frac{1}{2}\norm{x}_2^2 = \frac{1}{2}\norm{x}_2^2.
\end{align*}
Here $(a)$ holds by making use of the Lagrange mean value theorem, $(b)$ follows from Assumption~\ref{ass:ddim_score_matching}. Based on this result, we know that $\lim\limits_{x\to \infty} \norm{\Phi_t(x)}_2 = \infty$. This completes the proof of the lemma.

\subsection{Proof of Lemma~\ref{lem:bounds-T1-T2-DDIM}}
\label{sec:proof:lem:bounds-T1-T2-DDIM}

In what follows, we would like to establish the following identities:
\begin{subequations}
    \label{eq:bounds-T1-T2-DDIM}
    \begin{align}
    \gT_1(x_t,x_0) &=
 \det\left( I + \sqrt{\frac{1 - \overline{\alpha}_t}{\alpha_t - \overline{\alpha}_t}}\frac{\overline{\alpha}_t \eta_t}{(1-\overline{\alpha}_t)^2}\Cov_{0|t}(x_t) +
\sqrt{\frac{1 - \overline{\alpha}_t}{\alpha_t - \overline{\alpha}_t}}\eta_t \varepsilon^{\mathsf{J}}_{t}(x_t)
\right), \label{eq:bounds-T1-T2-DDIM-T1}\\
 \log\gT_2(x_t,x_0)
     &=
     \xi_t(x_t,x_0) - \frac{\overline{\alpha}_t \eta_t}{(\alpha_t - \overline{\alpha}_t)(1 - \overline{\alpha}_t)}\left(1 - \frac{\eta_t}{2(1 - \overline{\alpha}_t)}\right)\tr\left(\Cov_{0|t}(x_t)\right)
     \notag\\
     &\quad\quad -\sqrt{\frac{1-\overline{\alpha}_t}{\alpha_t - \overline{\alpha}_t}}\eta_t s_t^{\star}(x_t)^\top \varepsilon^{\mathsf{sc}}_{t}(x_t)
     - \frac{\eta_t^2}{2(\alpha_t - \overline{\alpha}_t)}\norm{\varepsilon^{\mathsf{sc}}_{t}(x_t)}_2^2.
     \label{eq:bounds-T1-T2-DDIM-T2}
    \end{align}
    \end{subequations}
The advertised relation \eqref{eq:prod-T1-T2-xit-Wt-DDIM} then follows immediately from \eqref{eq:bounds-T1-T2-DDIM}. The remainder of this proof thus comes down to establishing \eqref{eq:bounds-T1-T2-DDIM}.

\subsubsection{Controlling the term \texorpdfstring{$\gT_1(x_t,x_0)$}{gT1xtx0}}
Let us first look at the term $\gT_1(x_t,x_0)$, which satisfies
\begin{align}
    \gT_1(x_t,x_0) &= \left(\frac{1-\overline{\alpha}_t}{1-\overline{\alpha}_{t-1}}\right)^{\frac{d}{2}}\det\left(\frac{\partial \big(x_{t}+ \eta_t s_t(x_t) \big)/\sqrt{{\alpha}_t}}{\partial x_t}\right)
    = \left(\frac{1-\overline{\alpha}_t}{\alpha_t -\overline{\alpha}_{t}}\right)^{\frac{d}{2}}\det\left(\frac{\partial \big(x_{t}+ \eta_t s_t(x_t) \big)}{\partial x_t}\right) \notag\\
    &=
    \left(\frac{1-\overline{\alpha}_t}{\alpha_t -\overline{\alpha}_{t}}\right)^{\frac{d}{2}}\det\left(  I + \eta_t \frac{\partial}{\partial x_t}s_t^{\star}(x_t) + \eta_t\left(\frac{\partial}{\partial x_t} s_t(x_t) - \frac{\partial}{\partial x_t} s_t^{\star}(x_t)\right)\right) \notag\\
    &\overset{\text{(a)}}{=} \left(\frac{1-\overline{\alpha}_t}{\alpha_t -\overline{\alpha}_{t}}\right)^{\frac{d}{2}}
    \det\left( I + \eta_t
    \left\{\frac{\overline{\alpha}_t}{(1-\overline{\alpha}_t)^2}\Cov_{0|t}(x_t) - \frac{1}{1-\overline{\alpha}_t} I\right\} + \eta_t \varepsilon^{\mathsf{J}}_{t}(x_t)
    \right) \notag\\
    &= \det\left(
    \sqrt{\frac{1 - \overline{\alpha}_t}{\alpha_t - \overline{\alpha}_t}}\left(1 - \frac{\eta_t}{1-\overline{\alpha}_t}\right) I
    + \sqrt{\frac{1 - \overline{\alpha}_t}{\alpha_t - \overline{\alpha}_t}}\frac{\overline{\alpha}_t \eta_t}{(1-\overline{\alpha}_t)^2}\Cov_{0|t}(x_t) + \sqrt{\frac{1 - \overline{\alpha}_t}{\alpha_t - \overline{\alpha}_t}}\eta_t \varepsilon^{\mathsf{J}}_{t}(x_t)
    \right) \notag\\
    &\overset{\text{(b)}}{=} \det\left( I + \sqrt{\frac{1 - \overline{\alpha}_t}{\alpha_t - \overline{\alpha}_t}}\frac{\overline{\alpha}_t \eta_t}{(1-\overline{\alpha}_t)^2}\Cov_{0|t}(x_t) +
\sqrt{\frac{1 - \overline{\alpha}_t}{\alpha_t - \overline{\alpha}_t}}\eta_t \varepsilon^{\mathsf{J}}_{t}(x_t)
\right)
\label{eq:gT_1_form}
\end{align}
%
as claimed. Here, (a) arises from Tweedie's formula~\eqref{eq:score_to_posterior_new},
whereas (b) follows since (see \eqref{eq:defn-Phit-DDIM-etat})
\begin{equation}
\label{eq:ddim_step_size}
\begin{aligned}
    \sqrt{\frac{1 - \overline{\alpha}_t}{\alpha_t - \overline{\alpha}_t}}\left(1 - \frac{\eta_t}{1-\overline{\alpha}_t}\right) = \sqrt{\frac{1 - \overline{\alpha}_t}{\alpha_t - \overline{\alpha}_t}}\frac{(1-\overline{\alpha}_t) - \eta_t }{1-\overline{\alpha}_t} =
    \sqrt{\frac{1 - \overline{\alpha}_t}{\alpha_t - \overline{\alpha}_t}} \frac{\sqrt{(\alpha_t - \overline{\alpha}_t)(1 - \overline{\alpha}_t)}}{1 - \overline{\alpha}_t} = 1.
\end{aligned}
\end{equation}

\subsubsection{Controlling the term \texorpdfstring{$\gT_2(x_t,x_0)$}{gT2xtx0}}
Next, we turn attention to the term $\gT_2(x_t, x_0)$, which obeys
\begin{equation}\label{eq:loggT_2_eq1}
\begin{aligned}
    \log \gT_2(x_t,x_0) &= \frac{\norm{x_t -\sqrt{\overline{\alpha}_t}x_0}_2^2}{2(1-\overline{\alpha}_t)} -
    \frac{\norm{\big(x_t + \eta_t s_t(x_t)\big)/\sqrt{\alpha_t} - \sqrt{\overline{\alpha}_{t-1}}x_0}_2^2}{2(1-\overline{\alpha}_{t-1})}\\
    &=
    \frac{\norm{x_t -\sqrt{\overline{\alpha}_t}x_0}_2^2}{2(1-\overline{\alpha}_t)} -
    \frac{\norm{x_t + \eta_t s_t(x_t) - \sqrt{\overline{\alpha}_{t}}x_0}_2^2}{2(\alpha_t -\overline{\alpha}_{t})}.
\end{aligned}
\end{equation}
Regarding the term $x_t + \eta_t s_t(x_t) - \sqrt{\overline{\alpha}_t}x_0$, we can apply Tweedie's formula \eqref{eq:score_to_posterior_new} to show that
\begin{align*}
    &x_t + \eta_t s_t(x_t) - \sqrt{\overline{\alpha}_t}x_0 = \left(1 - \frac{\eta_t}{1 - \overline{\alpha}_t}\right)x_t + \frac{\sqrt{\overline{\alpha}_t}\eta_t}{1-\overline{\alpha}_t}\mu_{0|t}(x_t) - \sqrt{\overline{\alpha}_t}x_0 + \eta_t\big(s_t(x_t) - s_t^{\star}(x_t)\big)\\
    &= \left(1 - \frac{\eta_t}{1 - \overline{\alpha}_t}\right)(x_t - \sqrt{\overline{\alpha}_t}x_0) + \frac{\sqrt{\overline{\alpha}_t}\eta_t}{1-\overline{\alpha}_t}\mu_{0|t}(x_t)
     - \frac{\eta_t}{1 - \overline{\alpha}_t}\sqrt{\overline{\alpha}_t}x_0
     + \eta_t\big(s_t(x_t) - s_t^{\star}(x_t)\big)\\
    &= \left(1 - \frac{\eta_t}{1 - \overline{\alpha}_t}\right)(x_t - \sqrt{\overline{\alpha}_t}x_0)+ \frac{\sqrt{\overline{\alpha}_t}\eta_t}{1 - \overline{\alpha}_t}\big(\mu_{0|t}(x_t) - x_0 \big) + \eta_t\big(s_t(x_t) - s_t^{\star}(x_t)
    \big)\\
    &= \left(1 - \frac{\eta_t}{1 - \overline{\alpha}_t}\right)(x_t - \sqrt{\overline{\alpha}_t}x_0)+ \frac{\sqrt{\overline{\alpha}_t}\eta_t}{1 - \overline{\alpha}_t}\big(\mu_{0|t}(x_t) - x_0 \big) + \eta_t\varepsilon^{\mathsf{sc}}_{t}(x_t).
\end{align*}
Substitution into \eqref{eq:loggT_2_eq1} yields
\begin{equation}\label{eq:loggT_2_eq2}
\begin{aligned}
    &\log \gT_2(x_t,x_0) = \frac{\norm{x_t -\sqrt{\overline{\alpha}_t}x_0}_2^2}{2(1-\overline{\alpha}_t)} -
    \frac{1}{2(\alpha_t - \overline{\alpha}_t)}\left(1 - \frac{\eta_t}{1 - \overline{\alpha}_t}\right)^2\norm{x_t - \sqrt{\overline{\alpha}_t}x_0}_2^2\\
    &+ \left(1 - \frac{\eta_t}{1 - \overline{\alpha}_t}\right)\frac{\sqrt{\overline{\alpha}_t}\eta_t}{(\alpha_t - \overline{\alpha}_t)(1 - \overline{\alpha}_t)}
    (x_t - \sqrt{\overline{\alpha}_t}x_0)^\top \big(x_0 - \mu_{0|t}(x_t) \big)
    - \frac{\overline{\alpha}_t \eta_t^2}{2(\alpha_t - \overline{\alpha}_t)(1 - \overline{\alpha}_t)^2}\norm{x_0 - \mu_{0|t}(x_t)}_2^2\\
    &+ \frac{\eta_t}{\alpha_t - \overline{\alpha}_t}\left(1 {-} \frac{\eta_t}{1 - \overline{\alpha}_t}\right)\big(x_t - \sqrt{\overline{\alpha}_t}\mu_{0|t}(x_t)\big)^\top \varepsilon^{\mathsf{sc}}_{t}(x_t)\\
    &+ \frac{\eta_t}{\alpha_t - \overline{\alpha}_t}\big(\mu_{0|t}(x_t) {-} x_0\big)^\top \varepsilon^{\mathsf{sc}}_{t}(x_t) + \frac{\eta_t^2}{2(\alpha_t - \overline{\alpha}_t)}\norm{\varepsilon^{\mathsf{sc}}_{t}(x_t)}_2^2.
\end{aligned}
\end{equation}
In the sequel, we control each term of the above display separately.
\begin{itemize}
\item
Firstly, it follows from \eqref{eq:ddim_step_size} that
\[
\frac{1}{2(1-\overline{\alpha}_t)} - \frac{1}{2(\alpha_t - \overline{\alpha}_t)}\left(1 - \frac{\eta_t}{1 - \overline{\alpha}_t}\right)^2 = \frac{1}{2(1-\overline{\alpha}_t)} -
\frac{1}{2(\alpha_t - \overline{\alpha}_t)}\frac{\alpha_t - \overline{\alpha}_t}{1 - \overline{\alpha}_t} = 0
,\]
thus implying that
\[
\frac{\norm{x_t -\sqrt{\overline{\alpha}_t}x_0}_2^2}{2(1-\overline{\alpha}_t)} -
    \frac{1}{2(\alpha_t - \overline{\alpha}_t)}\left(1 - \frac{\eta_t}{1 - \overline{\alpha}_t}\right)^2\norm{x_t - \sqrt{\overline{\alpha}_t}x_0}_2^2 = 0.
\]
\item Secondly, invoking Tweedie's formula \eqref{eq:score_to_posterior_new} once again yields
\begin{align*}
\frac{\eta_t}{\alpha_t - \overline{\alpha}_t}\left(1 {-} \frac{\eta_t}{1 - \overline{\alpha}_t}\right)\big(x_t - \sqrt{\overline{\alpha}_t}\mu_{0|t} (x_t) \big)^\top \varepsilon^{\mathsf{sc}}_{t}(x_t) =
- \frac{\eta_t (1 - \overline{\alpha}_t - \eta_t)}{\alpha_t - \overline{\alpha}_t}s_t^{\star}(x_t)^\top\varepsilon^{\mathsf{sc}}_{t}(x_t).
\end{align*}

\item
Thirdly, consider the following component of \( \log \gT_2(x_t, x_0) \):
\begin{align}
\mathcal{T}_{23}(x_t,x_0) &\coloneqq
\left(1 - \frac{\eta_t}{1-\overline{\alpha}_t}\right)\frac{\sqrt{\overline{\alpha}_t}\eta_t}{(\alpha_t - \overline{\alpha}_t)(1 - \overline{\alpha}_t)}(x_t - \sqrt{\overline{\alpha}_t}x_0)^\top \big(x_0 - \mu_{0|t}(x_t) \big) \notag\\
&\qquad - \frac{\overline{\alpha}_t\eta_t^2}{2(\alpha_t - \overline{\alpha}_t)(1 - \overline{\alpha}_t)^2}\norm{x_0 - \mu_{0\mymid  t}(x_t)}_2^2 +
\frac{\eta_t}{\alpha_t - \overline{\alpha}_t}(\mu_{0|t} (x_t) - x_0)^\top \varepsilon^{\mathsf{sc}}_{t}(x_t).
\label{eq:3rd-term-log-Gamma2}
\end{align}
Taking the expectation of the above term \eqref{eq:3rd-term-log-Gamma2} under the conditional distribution \( p_{X_0 \mymid  X_t} \), we find that
\begin{align*}
\mathbb{E}_{X_0 \sim p_{X_0|X_t=x_t}}\big[\mathcal{T}_{23}(x_t,X_0) \big] &=
    -\left(1 - \frac{\eta_t}{1 - \overline{\alpha}_t}\right)\frac{\overline{\alpha}_t \eta_t}{(\alpha_t  - \overline{\alpha}_t)(1 - \overline{\alpha}_t)}\tr\big(\Cov_{0\mymid  t}(x_t)\big)
    \notag \\
    &\qquad- \frac{\overline{\alpha}_t \eta_t^2}{2(\alpha_t - \overline{\alpha}_t)(1 - \overline{\alpha}_t)^2}\tr\big(\Cov_{0\mymid  t}(x_t)\big)\\
    &= -\left(1 - \frac{\eta_t}{2(1 - \overline{\alpha}_t)}\right)\frac{\overline{\alpha}_t \eta_t}{(\alpha_t - \overline{\alpha}_t)(1 - \overline{\alpha}_t)}\tr\big(\Cov_{0\mymid  t}(x_t)\big).
\end{align*}
Therefore, the quantity $\xi_t(x_t,x_0)$ defined in \eqref{defn:xi-xt-x0-DDIM} obeys
\begin{align*}
    \xi_t(x_t,x_0)
& = \mathcal{T}_{23}(x_t,x_0)  - \mathbb{E}_{X_0 \sim p_{X_0|X_t=x_t}}\big[\mathcal{T}_{23}(x_t,X_0) \big],
\end{align*}
it can be easily verified that equation~\eqref{eq:int-xi-xt-x0-zero} holds.
\end{itemize}
Thus, substituting the preceding relations back into Eqn.~\eqref{eq:loggT_2_eq2}, we can establish \eqref{eq:bounds-T1-T2-DDIM-T2} as follows:
\begin{align}
    \log\gT_2(x_t,x_0)
     &= \xi_t(x_t,x_0) - \frac{\overline{\alpha}_t \eta_t}{(\alpha_t - \overline{\alpha}_t)(1 - \overline{\alpha}_t)}\left(1 - \frac{\eta_t}{2(1 - \overline{\alpha}_t)}\right)\tr\left(\Cov_{0|t}(x_t)\right) \notag\\
     &\quad\quad -\frac{\eta_t (1 - \overline{\alpha}_t - \eta_t)}{\alpha_t - \overline{\alpha}_t} s_t^{\star}(x_t)^\top \varepsilon^{\mathsf{sc}}_{t}(x_t)
     - \frac{\eta_t^2}{2(\alpha_t - \overline{\alpha}_t)}\norm{\varepsilon^{\mathsf{sc}}_{t}(x_t)}_2^2 \notag\\
     &=
     \xi_t(x_t,x_0) - \frac{\overline{\alpha}_t \eta_t}{(\alpha_t - \overline{\alpha}_t)(1 - \overline{\alpha}_t)}\left(1 - \frac{\eta_t}{2(1 - \overline{\alpha}_t)}\right)\tr\left(\Cov_{0|t}(x_t)\right)
     \notag\\
     &\quad\quad -\sqrt{\frac{1-\overline{\alpha}_t}{\alpha_t - \overline{\alpha}_t}}\eta_t s_t^{\star}(x_t)^\top \varepsilon^{\mathsf{sc}}_{t}(x_t)
     - \frac{\eta_t^2}{2(\alpha_t - \overline{\alpha}_t)}\norm{\varepsilon^{\mathsf{sc}}_{t}(x_t)}_2^2,
     \label{eq:loggT_2_eq3}
\end{align}
where the last equality holds since
\[
\frac{1 - \overline{\alpha}_t - \eta_t}{\alpha_t - \overline{\alpha}_t}
= \frac{\sqrt{(1 - \overline{\alpha}_t)(\alpha_t - \overline{\alpha}_t)}}{\alpha_t - \overline{\alpha}_t} = \sqrt{\frac{1- \overline{\alpha}_t}{\alpha_t - \overline{\alpha}_t}}.
\]

\subsection{Proof of Lemma~\ref{lemma:control-Wx_t-DDIM}}\label{sec:control-Wx_t}

For any $x_t \in \RB^d$,
applying Lemma~\ref{lem:logdet_expand} reveals that
\begin{align*}
    -\log \det&\left( I + \sqrt{\frac{1 - \overline{\alpha}_t}{\alpha_t - \overline{\alpha}_t}}\frac{\overline{\alpha}_t \eta_t}{(1-\overline{\alpha}_t)^2}\Cov_{0|t}(x_t) + \sqrt{\frac{1 - \overline{\alpha}_t}{\alpha_t - \overline{\alpha}_t}}\eta_t \varepsilon^{\mathsf{J}}_{t}(x_t)\right)\\
    &\le
    \frac{4\overline{\alpha}_t^2 \eta_t^2}{(\alpha_t -
    \overline{\alpha}_t)(1 - \overline{\alpha}_t)^3}\norm{\Cov_{0|t}(x_t)}_{\mathrm{F}}^2
    + \frac{4(1-\overline{\alpha}_t)\eta_t^2}{\alpha_t - \overline{\alpha}_t}\norm{\varepsilon^{\mathsf{J}}_{t}(x_t)}_{\mathrm{F}}^2\\
    &\quad - \sqrt{\frac{1 - \overline{\alpha}_t}{\alpha_t - \overline{\alpha}_t}}\frac{\overline{\alpha}_t \eta_t}{(1-\overline{\alpha}_t)^2}\tr\left(\Cov_{0|t}(x_t)\right)
    - \sqrt{\frac{1 - \overline{\alpha}_t}{\alpha_t - \overline{\alpha}_t}}\eta_t
    \tr\big(\varepsilon^{\mathsf{J}}_{t}(x_t)\big),
\end{align*}
provided that $\sqrt{\frac{1 - \overline{\alpha}_t}{\alpha_t - \overline{\alpha}_t}}\eta_t\norm{\varepsilon^{\mathsf{J}}_{t}(x_t)}_2 \le \frac{1}{4}$.
Combining this with the definition \eqref{eq:defn-Wt-DDIM} of $W_t$ results in
\begin{align}
    -W_t(x_t) &=
    \frac{\overline{\alpha}_t \eta_t}{(\alpha_t - \overline{\alpha}_t)(1 - \overline{\alpha}_t)}\left(1 - \frac{\eta_t}{2(1 - \overline{\alpha}_t)}\right)\tr\left(\Cov_{0|t}(x_t)\right)
    + \frac{\eta_t^2}{2(\alpha_t - \overline{\alpha}_t)}\norm{\varepsilon^{\mathsf{sc}}_{t}(x_t)}_2^2 \notag\\
& +
\sqrt{\frac{1- \overline{\alpha}_t}{\alpha_t - \overline{\alpha}_t}}\eta_t s_t^{\star}(x_t)^\top \varepsilon^{\mathsf{sc}}_{t}(x_t)
-\log \det\left( I + \sqrt{\frac{1 - \overline{\alpha}_t}{\alpha_t - \overline{\alpha}_t}}\frac{\overline{\alpha}_t \eta_t}{(1-\overline{\alpha}_t)^2}\Cov_{0|t}(x_t) {+} \sqrt{\frac{1 - \overline{\alpha}_t}{\alpha_t - \overline{\alpha}_t}}\eta_t \varepsilon^{\mathsf{J}}_{t}(x_t)\right) \notag\\
&\le
\frac{4\overline{\alpha}_t^2 \eta_t^2}{(\alpha_t -
    \overline{\alpha}_t)(1 - \overline{\alpha}_t)^3}\norm{\Cov_{0|t}(x_t)}_{\mathrm{F}}^2
    + \frac{4(1-\overline{\alpha}_t)\eta_t^2}{\alpha_t - \overline{\alpha}_t}\norm{\varepsilon^{\mathsf{J}}_{t}(x_t)}_{\mathrm{F}}^2 + \frac{\eta_t^2}{2(\alpha_t - \overline{\alpha}_t)}\norm{\varepsilon^{\mathsf{sc}}_{t}(x_t)}_2^2 \notag\\
    &\qquad + \left[
    \frac{\overline{\alpha}_t \eta_t}{(\alpha_t - \overline{\alpha}_t)(1 - \overline{\alpha}_t)}\left(1 - \frac{\eta_t}{2(1 - \overline{\alpha}_t)}\right) - \sqrt{\frac{1 - \overline{\alpha}_t}{\alpha_t - \overline{\alpha}_t}}\frac{\overline{\alpha}_t \eta_t}{(1 - \overline{\alpha}_t)^2}
    \right]\tr\left(\Cov_{0|t}(x_t)\right) \notag\\
    &\qquad + \sqrt{\frac{1- \overline{\alpha}_t}{\alpha_t - \overline{\alpha}_t}}\eta_t \Big(
    s_t^{\star}(x_t)^\top \varepsilon^{\mathsf{sc}}_{t}(x_t) - \tr\left(\varepsilon^{\mathsf{J}}_{t}(x_t)\right)\Big).
    \label{eq:EB-W_t_eq1}
\end{align}
Recalling our choice of the coefficient $\eta_t = (1 - \alpha_t)\big/ \big(1 + \sqrt{\frac{\alpha_t - \overline{\alpha}_t}{1 - \overline{\alpha}_t}} \big)$, we have
\begin{align*}
&\frac{\overline{\alpha}_t \eta_t}{(\alpha_t - \overline{\alpha}_t)(1 - \overline{\alpha}_t)}\left(1 - \frac{\eta_t}{2(1 - \overline{\alpha}_t)}\right) - \sqrt{\frac{1 - \overline{\alpha}_t}{\alpha_t - \overline{\alpha}_t}}\frac{\overline{\alpha}_t \eta_t}{(1 - \overline{\alpha}_t)^2}\\
&\qquad = \frac{\overline{\alpha}_t \eta_t}{(\alpha_t - \overline{\alpha}_t)(1- \overline{\alpha}_t)}\left(
1 - \frac{\eta_t}{2(1 - \overline{\alpha}_t)} - \sqrt{\frac{\alpha_t - \overline{\alpha}_t}{\alpha_t - \overline{\alpha}_t}}
\right)\\
&\qquad = \frac{\overline{\alpha}_t \eta_t}{(\alpha_t - \overline{\alpha}_t)(1- \overline{\alpha}_t)}\left(
1 - \frac{\eta_t}{1 - \overline{\alpha}_t} - \sqrt{\frac{\alpha_t - \overline{\alpha}_t}{\alpha_t - \overline{\alpha}_t}} + \frac{\eta_t}{2(1 - \overline{\alpha}_t)}
\right)
\\
&\qquad = \frac{\overline{\alpha}_t\eta_t^2}{2(\alpha_t - \overline{\alpha}_t)(1 - \overline{\alpha}_t)^2},
\end{align*}
where the penultimate equality holds due to  Eqn.~\eqref{eq:ddim_step_size}. Substituting this result into \eqref{eq:EB-W_t_eq1} yields
\begin{equation}
\begin{aligned}
    -W_t(x_t) &\le
    \frac{4\overline{\alpha}_t^2 \eta_t^2}{(\alpha_t -
    \overline{\alpha}_t)(1 - \overline{\alpha}_t)^3}\norm{\Cov_{0|t}(x_t)}_{\mathrm{F}}^2
    + \frac{4(1-\overline{\alpha}_t)\eta_t^2}{\alpha_t - \overline{\alpha}_t}\norm{\varepsilon^{\mathsf{J}}_{t}(x_t)}_{\mathrm{F}}^2 + \frac{\eta_t^2}{2(\alpha_t - \overline{\alpha}_t)}\norm{\varepsilon^{\mathsf{sc}}_{t}(x_t)}_2^2\\
    &\qquad + \frac{\overline{\alpha}_t \eta_t^2}{2(\alpha_t - \overline{\alpha}_t)(1 - \overline{\alpha}_t)^2}\tr\left(\Cov_{0|t}(x_t)\right)
    + \sqrt{\frac{1- \overline{\alpha}_t}{\alpha_t - \overline{\alpha}_t}}\eta_t \underbrace{\left(
    s_t^{\star}(x_t)^\top \varepsilon^{\mathsf{sc}}_{t}(x_t) - \tr\left(\varepsilon^{\mathsf{J}}_{t}(x_t)\right)\right)}_{\eqqcolon  \Delta(\varepsilon^{\mathsf{sc}}_{t}(x_t), \varepsilon^{\mathsf{J}}_{t}(x_t))}.
\end{aligned}
\end{equation}

\subsection{Proof of Lemma~\ref{lem:bd_Delta_epsi}
}\label{sec:control_est_err}

    To begin with, consider the inner product term $s_t^{\star}(x_t)^\top\varepsilon^{\mathsf{sc}}_{t}(x_t)$. For any set $\gA$, one can derive
    \begin{align}
    \int_{x_t \in \gA}  s_t^{\star}(x_t)^\top \varepsilon^{\mathsf{sc}}_{t}(x_t) &p_{X_t}(x_t)\rd x_t  =  \int_{x_t \in \gA}\frac{1}{{1 - \overline{\alpha}_t}}\big(x_t - \sqrt{\overline{\alpha}_t}\mu_{0|t}(x_t)\big)^\top \varepsilon^{\mathsf{sc}}_{t}(x_t) p_{X_t}(x_t) \rd x_t \notag\\
        =&
        \int_{\gA\times \gX_{\mathsf{data}}} \frac{1}{{1 - \overline{\alpha}_t}}(x_t - \sqrt{\overline{\alpha}_t}x_0)^\top \varepsilon^{\mathsf{sc}}_{t}(x_t)p_{X_t}(x_t)p_{X_0\mymid  X_t}(x_0\mymid  x_t)\rd x_t \rd x_0 \notag\\
        =&
        \int_{\gA\times\gX_{\mathsf{data}}}\frac{1}{{1 - \overline{\alpha}_t}}(x_t - \sqrt{\overline{\alpha}_t}x_0)^\top \varepsilon^{\mathsf{sc}}_{t}(x_t)p_{X_t\mymid  X_0}(x_t\mymid  x_0) p_{X_0}(x_0) \rd x_t \rd x_0 \notag\\
        \le&
        \int_{\R^d\times\gX_{\mathsf{data}}} \abs{
        \frac{1}{{1 - \overline{\alpha}_t}}(x_t - \sqrt{\overline{\alpha}_t}x_0)^\top \varepsilon^{\mathsf{sc}}_{t}(x_t)
        }p_{X_t\mymid  X_0}(x_t\mymid  x_0)p_{X_0}(x_0) \rd x_t \rd x_0,
        \label{eq:Delta_inner_eq1}
    \end{align}
    where the first identity follows from Tweedie's formula~\eqref{eq:score_to_posterior_new}.
    For any given point $x_0\in \gX_{\mathsf{data}}$, applying the Cauchy-Schwarz inequality gives
    \begin{equation}\label{eq:Delta_inner_eq2}
    \begin{aligned}
        \int_{\R^d}&\abs{
        \frac{1}{{1 - \overline{\alpha}_t}}(x_t - \sqrt{\overline{\alpha}_t}x_0)^\top \varepsilon^{\mathsf{sc}}_{t}(x_t)
        }p_{X_t\mymid  X_0}(x_t\mymid  x_0) \rd x_t\\
        \le&
        \left(
        \int_{\R^d}\left(\frac{1}{{1 - \overline{\alpha}_t}}(x_t - \sqrt{\overline{\alpha}_t}x_0)^\top \varepsilon^{\mathsf{sc}}_{t}(x_t)\right)^2 p_{X_t\mymid  X_0}(x_t\mymid  x_0) \rd x_t
        \right)^{\frac{1}{2}}
        \\
        =&
        \left(
        \int_{\R^d} \frac{1}{(1 - \overline{\alpha}_t)^2}\Big\langle\varepsilon^{\mathsf{sc}}_{t}(x_t)\varepsilon^{\mathsf{sc}}_{t}(x_t)^\top ,\, (x_t - \sqrt{\overline{\alpha}_t}x_0)(x_t - \sqrt{\overline{\alpha}_t}x_0)^\top\Big\rangle p_{X_t\mymid  X_0}(x_t\mymid  x_0) \rd x_t
        \right)^{\frac{1}{2}}.
    \end{aligned}
    \end{equation}
    Note that for given $x_0$, we know that $X_t\mymid  X_0=x_0 $ has distribution $ \gN\big(\sqrt{\overline{\alpha}_t} x_0, (1 - \overline{\alpha}_t)^{-1} I\big)$. As a result,
    \begin{align*}
        \nabla^2_{x_t} p_{X_t\mymid  X_0}(x_t\mymid  x_0) &= \nabla^2_{x_t} \left\{\left(\frac{1}{2\pi(1 - \overline{\alpha}_t)}\right)^{\frac{d}{2}}\exp\left(- \frac{\norm{x_t - \sqrt{\overline{\alpha}_t} x_0}_2^2}{2(1 - \overline{\alpha}_t)}\right)\right\}\\
        &=
        \left(\frac{1}{2\pi(1 - \overline{\alpha}_t)}\right)^{\frac{d}{2}} \nabla_{x_t} \left\{-
        \frac{x_t - \sqrt{\overline{\alpha}_t}x_0}{1 - \overline{\alpha}_t}\exp\left(- \frac{\norm{x_t - \sqrt{\overline{\alpha}_t} x_0}_2^2}{2(1 - \overline{\alpha}_t)}\right)
        \right\}\\
        &=
        \left(\frac{1}{2\pi(1 - \overline{\alpha}_t)}\right)^{\frac{d}{2}}e^{- \frac{\|x_t - \sqrt{\overline{\alpha}_t} x_0\|_2^2}{2(1 - \overline{\alpha}_t)}}\left\{
        \frac{(x_t - \sqrt{\overline{\alpha}_t}x_0)(x_t - \sqrt{\overline{\alpha}_t}x_0)^\top}{(1 - \overline{\alpha}_t)^2} - \frac{1}{1 - \overline{\alpha}_t} I
        \right\}\\
        &= p_{X_t\mymid  X_0}(x_t\mymid  x_0)\left\{
        \frac{(x_t - \sqrt{\overline{\alpha}_t}x_0)(x_t - \sqrt{\overline{\alpha}_t}x_0)^\top}{(1 - \overline{\alpha}_t)^2} - \frac{1}{1 - \overline{\alpha}_t} I
        \right\},
    \end{align*}
    which in turn gives
\[
\frac{(x_t - \sqrt{\overline{\alpha}_t}x_0)(x_t - \sqrt{\overline{\alpha}_t}x_0)^\top}{(1-\overline{\alpha}_t)^2}p_{X_t\mymid  X_0}(x_t\mymid  x_0) = \nabla^2_{x_t} p_{X_t\mymid  X_0}(x_t\mymid  x_0) + p_{X_t\mymid  X_0}(x_t\mymid  x_0)\frac{1}{1-\overline{\alpha}_t}I.
\]
Substituting this into \eqref{eq:Delta_inner_eq2} yields
    \begin{align}
        &\int_{\R^d}\abs{
        \frac{1}{{1 - \overline{\alpha}_t}}(x_t - \sqrt{\overline{\alpha}_t}x_0)^\top \varepsilon^{\mathsf{sc}}_{t}(x_t)
        }p_{X_t\mymid  X_0}(x_t\mymid  x_0) \rd x_t \notag\\
        &\le
        \left(
        \int_{\R^d}\inner{\varepsilon^{\mathsf{sc}}_{t}(x_t)\varepsilon^{\mathsf{sc}}_{t}(x_t)^\top}{
        \nabla^2_{x_t} p_{X_t\mymid  X_0}(x_t\mymid  x_0)
        + \frac{1}{(1 - \overline{\alpha}_t)}p_{X_t\mymid  X_0}(x_t\mymid  x_0)  I
        } \rd x_t
        \right)^{\frac{1}{2}} \notag\\
        &\le
        \left(
        \int_{\R^d} \inner{\varepsilon^{\mathsf{sc}}_{t}(x_t)\varepsilon^{\mathsf{sc}}_{t}(x_t)^\top}{\nabla^2_{x_t} p_{X_t\mymid  X_0}(x_t\mymid  x_0)}\rd x_t
        \right)^{\frac{1}{2}} + \frac{1}{\sqrt{1 - \overline{\alpha}_t}}\left(\int_{\R^d}\norm{\varepsilon^{\mathsf{sc}}_{t}(x_t)}_2^2 p_{X_t\mymid  X_0}(x_t\mymid  x_0) \rd x_t\right)^{\frac{1}{2}},
        \label{eq:Delta_inner_eq3}
    \end{align}
    where the last inequality follows since $\sqrt{a + b} \le \sqrt{a} + \sqrt{b}$ for all $a,b \ge 0$. This leaves us with two terms to control.

    With regards to the first term of the above bound \eqref{eq:Delta_inner_eq3}, we make the observation that
    \[
    \inner{\varepsilon^{\mathsf{sc}}_{t}(x_t)\varepsilon^{\mathsf{sc}}_{t}(x_t)^\top}{\nabla^2 p_{X_t\mymid  X_0}(x_t\mymid  x_0)} = \ssum{i}{1}{d}\ssum{j}{1}{d} [\varepsilon^{\mathsf{sc}}_{t}(x_t)]_i[\varepsilon^{\mathsf{sc}}_{t}(x_t)]_j \frac{\partial^2}{\partial x_i \partial x_j}p_{X_t\mymid  X_0}(x_t\mymid  x_0).
    \]
    Here and throughout,  $[v]_i$ represents the $i$-coordinate of the vector $v$.
    We can start by analyzing each \((i, j)\) component. In fact, for any $1 \le i,j \le d$, it holds that
    \begin{align}
        \int_{\R^d} [\varepsilon^{\mathsf{sc}}_{t}(x_t)]_i[\varepsilon^{\mathsf{sc}}_{t}(x_t)]_j \frac{\partial^2}{\partial x_i \partial x_j}p_{X_t\mymid  X_0}(x_t\mymid  x_0) \rd x_t
        \overset{\text{(a)}}{=}&
        {-}\int_{\R^d}\frac{\partial}{\partial x_i}\big([\varepsilon^{\mathsf{sc}}_{t}(x_t)]_i[\varepsilon^{\mathsf{sc}}_{t}(x_t)]_j\big)\frac{\partial}{\partial x_j}p_{X_t\mymid  X_0}(x_t\mymid  x_0) \rd x_t \notag\\
        \overset{\text{(b)}}{=}&
        \int_{\R^d}\frac{\partial^2}{\partial x_i \partial x_j}\big([\varepsilon^{\mathsf{sc}}_{t}(x_t)]_i[\varepsilon^{\mathsf{sc}}_{t}(x_t)]_j\big) p_{X_t\mymid  X_0}(x_t\mymid  x_0) \rd x_t,
        \label{eq:Delta_inner_eq4}
    \end{align}
    where (a) and (b) apply the integration by parts formula with respect to \( x_i \) and \( x_j \), respectively.  Denoting by $[ A]_{ij}$ the $(i,j)$-th element of the matrix $ A$, we have
    \begin{align*}
        \frac{\partial^2}{\partial x_i \partial x_j}\big([\varepsilon^{\mathsf{sc}}_{t}(x_t)]_i[\varepsilon^{\mathsf{sc}}_{t}(x_t)]_j\big) &= \left(\frac{\partial^2}{\partial x_i \partial x_j}[\varepsilon^{\mathsf{sc}}_{t}(x_t)]_i\right)[\varepsilon^{\mathsf{sc}}_{t}(x_t)]_j + [\varepsilon^{\mathsf{sc}}_{t}(x_t)]_i\left(\frac{\partial^2}{\partial x_i \partial x_j}[\varepsilon^{\mathsf{sc}}_{t}(x_t)]_j\right)\\
        &+ [\varepsilon^{\mathsf{J}}_{t}(x_t)]_{ij}[\varepsilon^{\mathsf{J}}_{t}(x_t)]_{ji} + [\varepsilon^{\mathsf{J}}_{t}(x_t)]_{ii} [\varepsilon^{\mathsf{J}}_{t}(x_t)]_{jj},
    \end{align*}
    where we recall the definition of $\varepsilon_t^{\mathsf{J}}$ in \eqref{eq:defn-eps-t-sc-J}.
    Substitution into \eqref{eq:Delta_inner_eq4} yields
    \begin{equation}\label{eq:inner_score_Hessian}
    \begin{aligned}
        \int_{\R^d}&\inner{\varepsilon^{\mathsf{sc}}_{t}(x_t)\varepsilon^{\mathsf{sc}}_{t}(x_t)^\top}{\nabla^2 p_{X_t\mymid  X_0}(x_t\mymid  x_0)} \rd x_t = \ssum{i}{1}{d}\ssum{j}{1}{d} \int_{\R^d}[\varepsilon^{\mathsf{sc}}_{t}(x_t)]_i[\varepsilon^{\mathsf{sc}}_{t}(x_t)]_j \frac{\partial^2}{\partial x_i \partial x_j}p_{X_t\mymid  X_0}(x_t\mymid  x_0)\rd x_t\\
        =& \ssum{i}{1}{d}\ssum{j}{1}{d}\int_{\R^d}\frac{\partial^2}{\partial x_i \partial x_j}\left([\varepsilon^{\mathsf{sc}}_{t}(x_t)]_i[\varepsilon^{\mathsf{sc}}_{t}(x_t)]_j\right) p_{X_t\mymid  X_0}(x_t\mymid  x_0) \rd x_t\\
        =& \ssum{i}{1}{d}\ssum{j}{1}{d}\int_{\R^d}\left\{
        \left(\frac{\partial^2}{\partial x_i \partial x_j}[\varepsilon^{\mathsf{sc}}_{t}(x_t)]_i\right) [\varepsilon^{\mathsf{sc}}_{t}(x_t)]_j + [\varepsilon^{\mathsf{sc}}_{t}(x_t)]_i\left(\frac{\partial^2}{\partial x_i \partial x_j}[\varepsilon^{\mathsf{sc}}_{t}(x_t)]_j\right)
        \right\} p_{X_t\mymid  X_0}(x_t\mymid  x_0)\rd x_t\\
        &+ \ssum{i}{1}{d}\ssum{j}{1}{d}\int_{\R^d}\left\{
        [\varepsilon^{\mathsf{J}}_{t}(x_t)]_{ij} [\varepsilon^{\mathsf{J}}_{t}(x_t)]_{ji} +
        [\varepsilon^{\mathsf{J}}_{t}(x_t)]_{ii} [\varepsilon^{\mathsf{J}}_{t}(x_t)]_{jj}
        \right\} p_{X_t\mymid  X_0}(x_t\mymid  x_0)\rd x_t.
    \end{aligned}
    \end{equation}
We now proceed to investigate each term in the above expression.
\begin{itemize}
    \item For any two indices $i$ and $j$, we have
    \[
    \left(\frac{\partial^2}{\partial x_i \partial x_j}\left[\varepsilon^{\mathsf{sc}}_{t}(x_t)\right]_i \right)[\varepsilon^{\mathsf{sc}}_{t}(x_t)]_j = \left(\frac{\partial}{\partial x_j}\left[\varepsilon^{\mathsf{J}}_{t}(x_t)\right]_{ii}\right)[\varepsilon^{\mathsf{sc}}_{t}(x_t)]_j,
    \]
    and consequently,
    \begin{align*}
        \ssum{i}{1}{d}\ssum{j}{1}{d}\int_{\R^d}&\left\{
        \left(\frac{\partial^2}{\partial x_i \partial x_j}[\varepsilon^{\mathsf{sc}}_{t}(x_t)]_i\right) [\varepsilon^{\mathsf{sc}}_{t}(x_t)]_j + [\varepsilon^{\mathsf{sc}}_{t}(x_t)]_i\left(\frac{\partial^2}{\partial x_i \partial x_j}[\varepsilon^{\mathsf{sc}}_{t}(x_t)]_j\right)
        \right\} p_{X_t\mymid  X_0}(x_t\mymid  x_0)\rd x_t\\
        &= 2\ssum{j}{1}{d}\int_{\R^d}\frac{\partial}{\partial x_j}\left(\ssum{i}{1}{d}[\varepsilon^{\mathsf{J}}_{t}(x_t)]_{ii}\right)[\varepsilon^{\mathsf{sc}}_{t}(x_t)]_j p_{X_t\mymid  X_0}(x_t\mymid  x_0)\rd x_t\\
        &= \int_{\R^d}2\inner{\nabla \tr\big(\varepsilon^{\mathsf{J}}_{t}(x_t)\big)}{\varepsilon^{\mathsf{sc}}_{t}(x_t)}p_{X_t\mymid  X_0}(x_t\mymid  x_0)\rd x_t.
    \end{align*}
    \item Next, for terms of the form $[\varepsilon^{\mathsf{J}}_{t}(x_t)]_{ij}[\varepsilon^{\mathsf{J}}_{t}(x_t)]_{ji}$ and $[\varepsilon^{\mathsf{J}}_{t}(x_t)]_{ii}[\varepsilon^{\mathsf{J}}_{t}(x_t)]_{jj}$, simple calculations yield
    \begin{align*}
        \Bigg| \ssum{i}{1}{d}\ssum{j}{1}{d}[\varepsilon^{\mathsf{J}}_{t}(x_t)]_{ij}[\varepsilon^{\mathsf{J}}_{t}(x_t)]_{ji} \Bigg| &  
        \leq  \norm{\varepsilon^{\mathsf{J}}_{t}(x_t)}_{\mathrm{F}}^2;\\
        \ssum{i}{1}{d}\ssum{j}{1}{d}[\varepsilon^{\mathsf{J}}_{t}(x_t)]_{ii}[\varepsilon^{\mathsf{J}}_{t}(x_t)]_{jj}&=
        \left(\ssum{i}{1}{d}[\varepsilon^{\mathsf{J}}_{t}(x_t)]_{ii}\right)\left(\ssum{j}{1}{d}[\varepsilon^{\mathsf{J}}_{t}(x_t)]_{jj}\right) = \big(\tr(\varepsilon^{\mathsf{J}}_{t}(x_t))\big)^2.
    \end{align*}
    \item
    Substituting these results into~\eqref{eq:inner_score_Hessian}, we obtain
    \begin{align*}
        \int_{\R^d}&\inner{\varepsilon^{\mathsf{sc}}_{t}(x_t)\varepsilon^{\mathsf{sc}}_{t}(x_t)^\top}{\nabla^2 p_{X_t\mymid  X_0}(x_t\mymid  x_0)} \rd x_t\\
        &= \int_{\R^d}\left\{2\inner{\nabla \tr\left(\varepsilon^{\mathsf{J}}_{t}(x_t)\right)}{\varepsilon^{\mathsf{sc}}_{t}(x_t)} + \big(\tr(\varepsilon^{\mathsf{J}}_{t}(x_t))\big)^2  + \norm{\varepsilon^{\mathsf{J}}_{t}(x_t)}_{\mathrm{F}}^2\right\}
        p_{X_t\mymid  X_0}(x_t\mymid  x_0) \rd x_t\\
        &\le \int_{\R^d}\left\{\norm{\nabla \tr(\varepsilon^{\mathsf{J}}_{t}(x_t))}_2^2 +
        \norm{\varepsilon^{\mathsf{sc}}_{t}(x_t)}_2^2 + \big(\tr(\varepsilon^{\mathsf{J}}_{t}(x_t))\big)^2  + \norm{\varepsilon^{\mathsf{J}}_{t}(x_t)}_{\mathrm{F}}^2
        \right\}p_{X_t\mymid  X_0}(x_t\mymid  x_0) \rd x_t.
    \end{align*}
    As a result, one can further deduce that
    \begin{align}
 & \int_{\mathcal{X}_{\mathsf{data}}}\bigg(\int_{\R^{d}}\inner{\varepsilon_{t}^{\mathsf{sc}}(x_{t})\varepsilon_{t}^{\mathsf{sc}}(x_{t})^{\top}}{\nabla^{2}p_{X_{t}\mymid  X_{0}}(x_{t}\mymid  x_{0})}\rd x_{t}\bigg)^{\frac{1}{2}}p_{X_{0}}(x_{0})\rd x_{0}\notag\\
 & \le\int_{\mathcal{X}_{\mathsf{data}}}\bigg(\int_{\R^{d}}\Big\{ \norm{\nabla\tr(\varepsilon_{t}^{\mathsf{J}}(x_{t}))}_2^{2}+\norm{\varepsilon_{t}^{\mathsf{sc}}(x_{t})}_2^{2}+\big(\tr(\varepsilon_{t}^{\mathsf{J}}(x_{t}))\big)^{2}\notag\\
 &\qquad +\norm{\varepsilon_{t}^{\mathsf{J}}(x_{t})}_{\mathrm{F}}^{2}\Big\} p_{X_{t}\mymid  X_{0}}(x_{t}\mymid  x_{0})\rd x_{t}\bigg)^{\frac{1}{2}}p_{X_{0}}(x_{0})\rd x_{0}\notag\\
 & \leq\bigg(\int_{\mathcal{X}_{\mathsf{data}}}\int_{\R^{d}}\left\{ \norm{\nabla\tr(\varepsilon_{t}^{\mathsf{J}}(x_{t}))}_2^{2}+\norm{\varepsilon_{t}^{\mathsf{sc}}(x_{t})}_2^{2}+\big(\tr(\varepsilon_{t}^{\mathsf{J}}(x_{t}))\big)^{2}+\norm{\varepsilon_{t}^{\mathsf{J}}(x_{t})}_{\mathrm{F}}^{2}\right\} p_{X_{t}\mymid  X_{0}}(x_{t}\mymid  x_{0})\rd x_{t}\rd x_{0}\bigg)^{\frac{1}{2}} \notag\\
 & =\bigg(\int_{\R^{d}}\left\{ \norm{\nabla\tr(\varepsilon_{t}^{\mathsf{J}}(x_{t}))}_2^{2}+\norm{\varepsilon_{t}^{\mathsf{sc}}(x_{t})}_2^{2}+\big(\tr(\varepsilon_{t}^{\mathsf{J}}(x_{t}))\big)^{2}+\norm{\varepsilon_{t}^{\mathsf{J}}(x_{t})}_{\mathrm{F}}^{2}\right\} p_{X_{t}}(x_{t})\rd x_{t}\bigg)^{\frac{1}{2}} \notag\\
  & \leq\bigg(\int_{\R^{d}}\norm{\nabla\tr(\varepsilon_{t}^{\mathsf{J}}(x_{t}))}_2^{2}p_{X_{t}}(x_{t})\rd x_{t}\bigg)^{\frac{1}{2}}+\bigg(\int_{\R^{d}}\norm{\varepsilon_{t}^{\mathsf{sc}}(x_{t})}_2^{2}p_{X_{t}}(x_{t})\rd x_{t}\bigg)^{\frac{1}{2}} \notag\\
  &\qquad +\bigg(\int_{\R^{d}}\big(\tr(\varepsilon_{t}^{\mathsf{J}}(x_{t}))\big)^{2}p_{X_{t}}(x_{t})\rd x_{t}\bigg)^{\frac{1}{2}}+\bigg(\int_{\R^{d}}\norm{\varepsilon_{t}^{\mathsf{J}}(x_{t})}_{\mathrm{F}}^{2}p_{X_{t}}(x_{t})\rd x_{t}\bigg)^{\frac{1}{2}},
\end{align}
where the second inequality comes from Jensen's inequality.
\end{itemize}

With the above result in place,  one can readily combine it with \eqref{eq:Delta_inner_eq1} and \eqref{eq:Delta_inner_eq3} to reach
    \begin{equation}
    \begin{aligned}
        \int_{x_t \in \gA}s_t^{\star}(x_t)^\top &\varepsilon^{\mathsf{sc}}_{t}(x_t) p_{X_t}(x_t) \rd x_t
        \le \frac{1}{\sqrt{1 - \overline{\alpha}_t}}\int_{\gX_{\mathsf{data}}}\left(\int_{\R^d} \norm{\varepsilon^{\mathsf{sc}}_{t}(x_t)}_2^2 p_{X_t\mymid  X_0}(x_t\mymid  x_0)\rd x_t\right)^{\frac{1}{2}}p_{X_0}(x_0) \rd x_0\\
        &\quad\quad +
        \int_{x_0\in \gX_{\mathsf{data}}}\left(\int_{\R^d}\inner{\varepsilon^{\mathsf{sc}}_{t}(x_t)\varepsilon^{\mathsf{sc}}_{t}(x_t)^\top}{\nabla^2 p_{X_t\mymid  X_0}(x_t\mymid  x_0)}\rd x_t\right)^{\frac{1}{2}}p_{X_0}(x_0)\rd x_0\\
        \le&
        \left(1 + \frac{1}{\sqrt{1 - \overline{\alpha}_t}}\right)\left(\int_{\R^d} \norm{\varepsilon^{\mathsf{sc}}_{t}(x_t)}_2^2 p_{X_t}(x_t) \rd x_t\right)^{\frac{1}{2}} + \left(\int_{\R^d} \norm{\nabla \tr\big(\varepsilon^{\mathsf{J}}_{t}(x_t)\big)}_2^2 p_{X_t}(x_t) \rd x_t\right)^{\frac{1}{2}}\\
        &\quad\quad +
        \left(\int_{\R^d}\norm{\varepsilon^{\mathsf{J}}_{t}(x_t)}_{\mathrm{F}}^2 p_{X_t}(x_t) \rd x_t\right)^{\frac{1}{2}} +
        \left(\int_{\R^d} \tr\big(\varepsilon^{\mathsf{J}}_{t}(x_t)\big)^2 p_{X_t}(x_t) \rd x_t\right)^{\frac{1}{2}}\\
        \le& \frac{2}{\sqrt{1 - \overline{\alpha}_t}}\varepsilon_{\mathsf{score},t} + \varepsilon_{\mathsf{Jacobi},1,t} + \varepsilon_{\mathsf{Jacobi},2,t} + \varepsilon_{\mathsf{Hess},t}.
    \end{aligned}
    \end{equation}
    Thus, we can apply Jensen's inequality once again to arrive at
    \begin{align*}
        \int_{x_t\in \gA} \Delta\big(\varepsilon^{\mathsf{sc}}_{t}(x_t), \varepsilon^{\mathsf{J}}_{t}(x_t) \big)& p_{X_t}(x_t) \rd x_t = \int_{x_t\in \gA} \left\{ s_t^{\star}(x_t)^\top \varepsilon^{\mathsf{sc}}_{t}(x_t)  - \tr\big(\varepsilon^{\mathsf{J}}_{t}(x_t)\big)\right\} p_{X_t}(x_t) \rd x_t\\
        \le&
        \frac{2}{\sqrt{1 - \overline{\alpha}_t}}\varepsilon_{\mathsf{score},t} + \varepsilon_{\mathsf{Jacobi},1,t} + \varepsilon_{\mathsf{Jacobi},2,t} + \varepsilon_{\mathsf{Hess},t}
        + \int_{\R^d}\abs{\tr\big(\varepsilon^{\mathsf{J}}_{t}(x_t)\big)}p_{X_t}(x_t)\rd x_t\\
        \le& \frac{2}{\sqrt{1 - \overline{\alpha}_t}}
        \varepsilon_{\mathsf{score},t} + \varepsilon_{\mathsf{Jacobi},1,t} + \varepsilon_{\mathsf{Jacobi},2,t} + \varepsilon_{\mathsf{Hess},t}.
    \end{align*}

\subsection{Proof of Lemma~\ref{lem:integral-Wt-Et}}
\label{sec:proof:lem:integral-Wt-Et}


Before applying Lemma~\ref{lemma:control-Wx_t-DDIM} to control \( W_t(x_t) \), let us first look at some key coefficients in the inequality \eqref{eq:EB-W_t_eq2} therein. As in Lemma~\ref{lem:bound_CoV_1|t_F^2},  define \( \widetilde{\sigma}_t^2 \coloneq \frac{\overline{\alpha}_t(1 - \alpha_t)}{(\alpha_t - \overline{\alpha}_t)(1 - \overline{\alpha}_t)} \). For the DDIM coefficient choice~\eqref{eq:pars-DDIM-original}, it holds that \( \eta_t = \frac{1 - \alpha_t}{1 + \sqrt{\frac{\alpha_t - \overline{\alpha}_t}{1 - \overline{\alpha}_t}}} \leq 1 - \alpha_t \), and hence we can derive
\begin{align*}
\frac{\overline{\alpha}_t^2 \eta_t^2}{(\alpha_t - \overline{\alpha}_t)(1 - \overline{\alpha}_t)^3} &\le \frac{\alpha_t - \overline{\alpha}_t}{1 - \overline{\alpha}_t}\cdot \left(\frac{\overline{\alpha}_t (1 - \alpha_t)}{(\alpha_t - \overline{\alpha}_t)(1 - \overline{\alpha}_t)}\right)^2=\frac{\alpha_t - \overline{\alpha}_t}{1 - \overline{\alpha}_t}\widetilde{\sigma}_t^4;\\
\frac{\overline{\alpha}_t \eta_t^2}{(\alpha_t - \overline{\alpha}_t)(1 - \overline{\alpha}_t)^2} &\le
\frac{\eta_t}{1 - \overline{\alpha}_t}\cdot
\frac{\overline{\alpha}_t (1 - \alpha_t)}{(\alpha_t - \overline{\alpha}_t)(1 - \overline{\alpha}_t)}
=\frac{\widetilde{\sigma}_t^2\eta_t}{1 - \overline{\alpha}_t}.
\end{align*}
Substitution into \eqref{eq:EB-W_t_eq2} allows one to control the term $\int_{x_t\in \Phi_t^{-1}(\gA)}W_t(x_t)p_{X_t}(x_t)\rd x_t$ as follows:
\begin{align*}
    \int_{\Phi_t^{-1}(\gA)} &- W_t(x_t) p_{X_t}(x_t) \rd x_t\\
    &\le \int_{\Phi_t^{-1}(\gA)}\left\{\frac{\widetilde{\sigma}_t^2 \eta_t}{2(1 - \overline{\alpha}_t)}\tr\left(\Cov_{0|t}(x_t)\right) + \frac{4(\alpha_t - \overline{\alpha}_t)}{1 - \overline{\alpha}_t}\widetilde{\sigma}_t^4 \norm{\Cov_{0|t}(x_t)}_{\mathrm{F}}^2\right\} p_{X_t}(x_t)\rd x_t\\
    &\qquad +
    \int_{\Phi_t^{-1}(\gA)}\left\{
    \frac{4(1-\overline{\alpha}_t)\eta_t^2}{\alpha_t - \overline{\alpha}_t}\norm{\varepsilon^{\mathsf{J}}_{t}(x_t)}_{\mathrm{F}}^2 + \frac{\eta_t^2}{2(\alpha_t - \overline{\alpha}_t)}\norm{\varepsilon^{\mathsf{sc}}_{t}(x_t)}_2^2
    \right\}p_{X_t}(x_t)\rd x_t\\
    &\qquad+ \sqrt{\frac{1 - \overline{\alpha}_t}{\alpha_t - \overline{\alpha}_t}}\eta_t \int_{\Phi_t^{-1}(\gA)}\Delta\big(\varepsilon^{\mathsf{sc}}_{t}(x_t),\varepsilon^{\mathsf{J}}_{t}(x_t)\big) p_{X_t}(x_t)\rd x_t\\
    &\le \int_{\R^d}\left\{\frac{\widetilde{\sigma}_t^2 \eta_t}{2(1 - \overline{\alpha}_t)}\tr\left(\Cov_{0|t}(x_t)\right) + \frac{4(\alpha_t - \overline{\alpha}_t)}{1 - \overline{\alpha}_t}\widetilde{\sigma}_t^4 \norm{\Cov_{0|t}(x_t)}_{\mathrm{F}}^2\right\} p_{X_t}(x_t)\rd x_t\\
    &\qquad+
    \int_{\R^d}\left\{
    \frac{4(1-\overline{\alpha}_t)\eta_t^2}{\alpha_t - \overline{\alpha}_t}\norm{\varepsilon^{\mathsf{J}}_{t}(x_t)}_{\mathrm{F}}^2 + \frac{\eta_t^2}{2(\alpha_t - \overline{\alpha}_t)}\norm{\varepsilon^{\mathsf{sc}}_{t}(x_t)}_2^2
    \right\}p_{X_t}(x_t)\rd x_t\\
    &\qquad+
    \sqrt{\frac{1 - \overline{\alpha}_t}{\alpha_t - \overline{\alpha}_t}}\eta_t
    \sup\limits_{\gA\subseteq \R^d} \int_{\gA}\Delta\big(\varepsilon^{\mathsf{sc}}_{t}(x_t),\varepsilon^{\mathsf{J}}_{t}(x_t)\big) p_{X_t}(x_t)\rd x_t.
\end{align*}

Additionally, in view of Lemma~\ref{lem:bd_Delta_epsi}, we know that for any measurable set $\gA \subseteq \R^d$,
\begin{align*}
    \sqrt{\frac{1 - \overline{\alpha}_t}{\alpha_t - \overline{\alpha}_t}}\eta_t\int_{\gA}&
    \Delta\big(\varepsilon^{\mathsf{sc}}_{t}(x_t), \varepsilon^{\mathsf{J}}_{t}(x_t)\big)p_{X_t}(x_t)\rd x_t\\
    &\le \frac{2\eta_t\sqrt{1-\overline{\alpha}_t}}{\sqrt{\alpha_t - \overline{\alpha}_t}}
    \left\{\frac{\varepsilon_{\mathsf{score},t}}{\sqrt{1-\overline{\alpha}_t}}
     + \varepsilon_{\mathsf{Jacobi},1,t} + \varepsilon_{\mathsf{Jacobi},2,t} + \varepsilon_{\mathsf{Hess},t}\right\}.
\end{align*}
Taking the above pieces together, we can readily conclude the proof of the advertised result.
%
%

\subsection{Proof of Lemma~\ref{lem:sum-S1-sum-S2-DDIM}}
\label{sec:proof-lem:sum-S1-sum-S2-DDIM}

Let us first cope with  $\ssum{t}{1}{T}\gS_{t,1}$, which concerns the accumulated discretization error. A little algebra gives
\begin{align*}
    \ssum{t}{1}{T}\gS_{t,1} &=  \ssum{t}{1}{T-1}\left(\frac{\widetilde{\sigma}_t^2 \eta_t}{1 - \overline{\alpha}_t} + \frac{3(\alpha_t - \overline{\alpha}_t)\widetilde{\sigma}_t^2}{1 - \overline{\alpha}_t} - \frac{3(\alpha_{t+1} - \overline{\alpha}_{t+1})\widetilde{\sigma}_{t+1}^2}{1 - \overline{\alpha}_{t+1}}\right)\EB \left[\tr\left(\Cov_{0|t}(X_t)\right)\right]\\
    &\quad +
    \frac{\widetilde{\sigma}_T^2(\eta_T + \alpha_T - \overline{\alpha}_T)}{1 - \overline{\alpha}_T}\EB \left[\tr\left(\Cov_{0|T}(X_T)\right)\right] + \frac{1}{T^{9}}.
\end{align*}
Applying Lemma~\ref{lem:step_size} and the basic property \eqref{eq:basic-alphat-property},
%
we can show that
\begin{align*}
    \frac{\widetilde{\sigma}_t^2 \eta_t}{1 - \overline{\alpha}_t} {+} \frac{3(\alpha_t - \overline{\alpha}_t)\widetilde{\sigma}_t^2}{1 - \overline{\alpha}_t} {-}& \frac{3(\alpha_{t+1} - \overline{\alpha}_{t+1})\widetilde{\sigma}_{t+1}^2}{1 - \overline{\alpha}_{t+1}}
    {=} \frac{\widetilde{\sigma}_t^2 \eta_t}{1 - \overline{\alpha}_t} {+} 3\left(1 - \frac{1-\alpha_t}{1 - \overline{\alpha}_t}\right)\widetilde{\sigma}_{t}^2 {-} 3\left(1 - \frac{1 - \alpha_{t+1}}{1 - \overline{\alpha}_{t+1}}\right)\widetilde{\sigma}_{t+1}^2\\
    &\le \frac{3(1-\alpha_t)\widetilde{\sigma}_t^2}{1 - \overline{\alpha}_t} - \frac{3(1 - \alpha_t)\widetilde{\sigma}_t^2}{1 - \overline{\alpha}_t} + \frac{3(1 - \alpha_{t+1})\widetilde{\sigma}_{t+1}^2}{1 - \overline{\alpha}_{t+1}} + 3(\widetilde{\sigma}_t^2 - \widetilde{\sigma}_{t+1}^2)\\
    &\le 3\left(\frac{1-\alpha_{t+1}}{1 - \overline{\alpha}_{t+1}}\right)^2\frac{\overline{\alpha}_{t+1}^2}{1 - \overline{\alpha}_{t+1}} +  \frac{C_6\log^2 T}{T^2}\frac{\overline{\alpha}_t}{1 - \overline{\alpha}_t} \le
    \frac{2C_6 \log^2 T}{T^2}\frac{\overline{\alpha}_t}{1 - \overline{\alpha}_t}.
\end{align*}
As a consequence, we can demonstrate that
\begin{equation}\label{eq:ddim_tv_xy_eq3}
\begin{aligned}
    \ssum{t}{1}{T} \gS_{t,1} &{\le}
    \left(\frac{2C_6\log T}{T}\right)^2 \ssum{t}{1}{T-1}\frac{\overline{\alpha}_t}{1 - \overline{\alpha}_t}\EB \left[\tr\left(\Cov_{0|t}(X_t)\right)\right] +
    \frac{2C_6\log T}{T} \frac{\overline{\alpha}_T}{1 - \overline{\alpha}_T}\EB \left[\tr\left(\Cov_{0|T}(X_T)\right)\right] + \frac{1}{T^9}\\
    &\overset{\text{(a)}}{\le}
    C_3kT\log T\left(\frac{2C_6 \log T}{T}\right)^2 + \frac{2C_3C_6 k \log^2 T}{T} + \frac{1}{T^9}
    \le C_9\frac{k\log^3 T}{T} + \frac{1}{T^9} \le C_{10}\frac{k\log^3 T}{T},
\end{aligned}
\end{equation}
where (a) applies the moment inequality \eqref{eq:posterior_mmt} with $l=2$.

Next, we turn to \( \sum_{t=1}^{T} \gS_{t,2} \), which concerns the cumulative estimation error. Given that
\(\frac{\eta_t}{\sqrt{\alpha_t - \overline{\alpha}_t}}
\le \frac{\eta_t}{\alpha_t - \overline{\alpha}_t} \sqrt{1-\overline{\alpha}_t}
\le \frac{1 - \alpha_t}{1 - \overline{\alpha}_t} \sqrt{1-\overline{\alpha}_t}
\le \frac{8c_1\log T}{T}\sqrt{1-\overline{\alpha}_t}\), we can derive
\begin{equation}\label{eq:ddim_tv_xy_eq4}
\begin{aligned}
    \ssum{t}{2}{T}\gS_{t,2} &= \ssum{t}{2}{T}\frac{2\eta_t\sqrt{1-\overline{\alpha}_t}}
    {\sqrt{\alpha_t - \overline{\alpha}_t}}
    \bigg\{\frac{\varepsilon_{\mathsf{score},t}}{
        \sqrt{1-\overline{\alpha}_t}
    } + \varepsilon_{\mathsf{Jacobi},1,t} + \varepsilon_{\mathsf{Jacobi},2,t} + \varepsilon_{\mathsf{Hess},t}\bigg\}\\
    &\qquad +
    \ssum{t}{2}{T}\frac{68(1 - \overline{\alpha}_t)\eta_t^2}{\alpha_t - \overline{\alpha}_t}\varepsilon_{\mathsf{Jacobi},1,t}^2 +
    \ssum{t}{2}{T}\frac{\eta_t^2}{2(\alpha_t -\overline{\alpha}_t)}\varepsilon_{\mathsf{score},t}^2\\
    &\le
    \frac{8c_1 \log T}{T}\bigg\{\ssum{t}{1}{T}\sqrt{1-\overline{\alpha}_t}\varepsilon_{\mathsf{score},t} {+}
    \ssum{t}{2}{T} (1-\overline{\alpha}_t)\varepsilon_{\mathsf{Jacobi},1,t} \\
    &\hspace{7em}+ \ssum{t}{2}{T}(1-\overline{\alpha}_t)\varepsilon_{\mathsf{Jacobi},2,t} {+}
    \ssum{t}{2}{T}(1-\overline{\alpha}_t)\varepsilon_{\mathsf{Hess},t}\bigg\}\\
    &\qquad +
    \frac{C_{10}\log^2 T}{T^2}\left\{
    \ssum{t}{1}{T}(1-\overline{\alpha}_t)^2\varepsilon_{\mathsf{Jacobi},1,t}^2 +
     \ssum{t}{1}{T}(1-\overline{\alpha}_t)^2\varepsilon_{\mathsf{score},t}^2
    \right\}\\
    &\overset{\text{(a)}}{\le}
    8c_1 \left(\varepsilon_{\mathsf{score}} + \varepsilon_{\mathsf{Jacobi},1} + \varepsilon_{\mathsf{Jacobi},2} +
    \varepsilon_{\mathsf{Hess}}
    \right)\log T + \frac{C_{10}\log^2 T}{T}\left(\varepsilon_{\mathsf{score}}^2 + \varepsilon_{\mathsf{Jacobi},1}^2\right)\\
    &\le C_{11} \left(\varepsilon_{\mathsf{score}} + \varepsilon_{\mathsf{Jacobi},1} + \varepsilon_{\mathsf{Jacobi},2} +
    \varepsilon_{\mathsf{Hess}}
    \right)\log T,
\end{aligned}
\end{equation}
where (a) results from the Cauchy-Schwarz inequality and Assumption~\ref{ass:ddim_score_matching}, and the last inequality holds provided that $\frac{\log T}{T}(\varepsilon_{\mathsf{score}} + \varepsilon_{\mathsf{Jacobi},1})\le 1$.

%% file: DDPM_Analysis.tex
\section{Analysis for DDPM (proof of Theorem~\ref{thm:ddpm_tv_conv})}\label{sec:k/T_DDPM_Analysis}

Given that Theorem~\ref{thm:vanilla_ddpm} is a special case of Theorem~\ref{thm:ddpm_tv_conv},
we shall focus on proving Theorem~\ref{thm:ddpm_tv_conv} in this section.

\subsection{Preparation}
Before proceeding, let us introduce several convention and auxiliary objects that will be useful throughout.

\medskip
\noindent {\bf Random vectors in the extended space.}
Firstly,
the random vectors in this proof are allowed to take values in the extended space $\RB^d\cup \{\infty\}$ that covers the point $\infty$ (think about it as infinity in $d$ dimension). Namely, they can be constructed in the following way:
\[
X = \begin{cases}
X^\prime, & \text{with probability } \theta, \\
\infty, & \text{with probability } 1 - \theta,
\end{cases}
\]
where $\theta \in [0,1]$ and $X^\prime$
is a random vector in $\RB^d$ in the usual sense. If $X^\prime$ has a density $p_{X^\prime}$,
then the generalized density of $X$ is
\[
p_X(x) = \theta p_{X^\prime}(x)\mathbbm{1}\{x\in \RB^d\} + (1 - \theta)\delta_\infty,
\]
where $\delta_{\infty}$ indicates the Dirac measure at $\infty$.

\medskip
\noindent {\bf Introducing auxiliary sequences.}
 Secondly, let us introduce several auxiliary sequences that shall play a pivotal role in our analysis. Here and throughout, the notation $X \mymid  Y \sim \widetilde{X} \mymid  \widetilde{Y}$ means that the conditional density of $X$ given $ Y = y $ and that of $\widetilde{X}$ given $ \widetilde{Y}=y$ coincide for any $y$.
 \begin{itemize}
 \item
First, we define a discrete-time reverse process $\{Y_t^\star\}_{t=T}^1$ by
\begin{align}
Y_T^{\star} =Y_T \sim \gN(0,I_d),\qquad Y_{t-1}^\star = \frac{1}{\sqrt{\alpha}_t}\big(Y_t^\star + \eta_t s_t^\star(Y_t^\star) + \sigma_t W_t\big).
\label{eq:defn-Yt-star-DDPM}
\end{align}
In short, this auxiliary process $\{Y^{\star}\}$ implements DDPM using exact score functions.

\item Based on the above process, we construct an auxiliary reverse process $\overline{Y}_t$ that follows the same transition dynamics as $Y_t^\star$ in the absence of score estimation errors:
\begin{align}
    \overline{Y}_{t-1}^-\mymid  \overline{Y}_{t} \sim Y_{t-1}^{\star}\mymid  Y_{t}^{\star},\qquad
    \overline{Y}_t\mymid  \{\overline{Y}_{t}^- = y_t^-\} = \begin{cases}
        y_t^-, &\text{with prob. }\frac{p_{X_t}(y_t^-)}{p_{\overline{Y}_t^-}(y_t^-)}\wedge 1 \\
        \infty, &\text{otherwise}
    \end{cases}
    \label{eq:defn-Yt-1-minus-Yt-bar}
\end{align}
for any $y_t^- \neq \infty$,
where we recall that $a\wedge b \coloneqq \min\{a,b\}$.
It is straightforward to show that
\begin{equation}\label{eq:p_barY_le_p_X}
p_{\overline{Y}_{t}}(y_{t})= \int_{\RB^{d}}\left(p_{X_{t}}(y_{t}^{-})\wedge p_{\overline{Y}_{t}^{-}}(y_{t}^{-})\right)\delta(y_{t}-y_{t}^{-})\rd y_{t}^{-}=p_{X_{t}}(y_{t})\wedge p_{\overline{Y}_{t}^{-}}(y_{t})
\end{equation}
for any $y_t \neq \infty$,
%
%
where $\delta(\cdot)$ denotes the Dirac measure.

\item
To account for the score estimation error, we introduce another auxiliary reverse process $\widehat{Y}_t$ based on the dynamics of $Y_t$:
\begin{align}
    \widehat{Y}_{t-1}^-\mymid \widehat{Y}_{t} \sim Y_{t-1}\mymid  Y_{t},\qquad
    \widehat{Y}_t\mymid  \{\widehat{Y}_{t}^- = y_t^-\} = \begin{cases}
        y_t^-, &\text{with prob. }\frac{p_{X_t}(y_t^-)}{p_{\overline{Y}_t^-}(y_t^-)}\wedge 1, \\
        \infty, &\text{otherwise}.
    \end{cases}
    \label{eq:defn-Yt-1-minus-Yt-bar-error}
\end{align}
It is seen that the probability densities of $Y_t$ and $\widehat{Y}_t$ satisfy the properties stated in the following lemma, whose proof can be found in Appendix~\ref{sec:proof:lem:p_Y_ge_p_hat_Y}.
\begin{lemma}\label{lem:p_Y_ge_p_hat_Y}
    For all $t = 1,\cdots, T$, it holds that
    \begin{equation}\label{eq:p_Y_ge_p_hat_Y}
p_{Y_t}(x) \ge p_{\widehat{Y}_t}(x), \qquad \text{for all } x\in \RB^d.
\end{equation}
\end{lemma}
\end{itemize}

\subsection{Main steps for proving Theorem~\ref{thm:ddpm_tv_conv}}

We are now in a position to present the main steps of our proof.

\medskip
\noindent {\bf Step 1: linking the TV distances between adjacent steps.}
Define the following set
\begin{align}
\label{eq:defn-At-DDPM}
\gA_{t} \coloneqq \Big\{x: p_{X_t}(x) \ge p_{\overline{Y}_t^-}(x)\Big\}.
\end{align}
In view of \eqref{eq:p_barY_le_p_X}, the condition $x\in \mathcal{A}_t$ is equivalent to $p_{\overline{Y}_t}(x) \leq p_{X_t}(x)$, which together with some well-known property of the TV distance yields
\begin{align}
\TV(p_{X_t}, p_{\overline{Y}_t}) = \int_{x:p_{\overline{Y}_t}(x) \leq p_{X_t}(x)}\big(p_{X_t}(x) - p_{\overline{Y}_t}(x)\big)\rd x
= \int_{{\mathcal{A}_t}}\big(p_{X_t}(x) - p_{\overline{Y}_t}(x)\big)\rd x.
\label{eq:TV-Xt-Ytbar-eqn-DDPM}
\end{align}
To link the iterates in step $t$ and step $t-1$, we observe that the density $p_{\overline{Y}_{t-1}^-}(\cdot)$ satisfies
\begin{align}
&p_{\overline{Y}_{t-1}^-}(x_{t-1}) = \int_{\RB^d}p_{\overline{Y}_{t-1}^-\mymid  \overline{Y}_t}(x_{t-1}\mymid  x_t) p_{\overline{Y}_t}(x_t)\rd x_t
= \int_{\RB^d}p_{{Y}_{t-1}^{\star}|Y_{t}^{\star}}(x_{t-1}\mymid  x_t) p_{\overline{Y}_t}(x_t)\rd x_t\notag\\
&\quad =  \int_{\RB^d}p_{Y_{t-1}^{\star}\mymid  Y_{t}^{\star}}(x_{t-1}\mymid  x_t) p_{X_t}(x_t)\rd x_t +
\int_{\RB^d}p_{Y_{t-1}^{\star}\mymid  Y_{t}^{\star}}(x_{t-1}\mymid  x_t) \big(p_{\overline{Y}_t}(x_t) - p_{X_t}(x_t) \big)\rd x_t,
\end{align}
where we have utilized the construction in \eqref{eq:defn-Yt-1-minus-Yt-bar}.
With the preceding two identities
in place, we can further derive the following recursion for all $t \ge 2$:
\begin{equation}\label{eq:tv_xy*_eq1}
\begin{aligned}
    \TV(p_{X_{t-1}}, p_{\overline{Y}_{t-1}}) &= \int_{\gA_{t-1}}\big(p_{X_{t-1}}(x_{t-1}) {-} p_{\overline{Y}_{t-1}}(x_{t-1}) \big) \rd x_{t-1} \\
    &\overset{\text{(a)}}{=} \int_{\gA_{t-1}}\big(p_{X_{t-1}}(x_{t-1}) {-} p_{\overline{Y}_{t-1}^-}(x_{t-1})\big) \rd x_{t-1}\\
    &= \underbrace{\int_{\gA_{t-1}}p_{X_{t-1}}(x_{t-1}) - \int_{\gA_{t-1}\times \RB^d}p_{Y_{t-1}^{\star}\mymid  Y_{t}^{\star}}(x_{t-1}\mymid  x_t)p_{X_t}(x_t)\rd x_{t-1}\rd x_t}_{\eqqcolon\, \gR_{t-1}}\\
    &\quad + \int_{\gA_{t-1}\times \RB^d}p_{Y_{t-1}^{\star}\mymid  Y_{t}^{\star}}(x_{t-1}\mymid  x_t) \big(p_{X_t}(x_t) - p_{\overline{Y}_t}(x_t) \big)\rd x_{t-1} \rd x_t\\
    &\overset{\text{(b)}}{\le} \gR_{t-1} + \int_{\gA_{t-1}\times \gA_t}p_{Y_{t-1}^{\star}\mymid  Y_{t}^{\star}}(x_{t-1}\mymid  x_t)\big(p_{X_t}(x_t) - p_{\overline{Y}_t}(x_t) \big)\rd x_{t-1}\rd x_t\\
    &= \gR_{t-1} + \int_{\gA_{t}}\PB_{Y_{t-1}^{\star}\mymid  Y_{t}^{\star}}(\gA_{t-1}\mymid  x_t)\big(p_{X_t}(x_t) - p_{\overline{Y}_t}(x_t)\big)\rd x_t \\
    &\overset{\text{(c)}}{\le} \gR_{t-1} + \TV(p_{X_t}, p_{\overline{Y}_t}).
\end{aligned}
\end{equation}
Here, (a) follows since $p_{\overline{Y}_{t-1}}=p_{\overline{Y}_{t-1}^-}$ on $\gA_{t-1}$ (see \eqref{eq:p_barY_le_p_X}), (b) holds since $
p_{X_t}(x_t) - p_{\overline{Y}_{t}}(x_t)
\leq 0$ on $\gA_t^{\mathrm{c}}$, while (c) is valid due to \eqref{eq:TV-Xt-Ytbar-eqn-DDPM} and the fact that $\PB_{Y_{t-1}^{\star}\mymid  Y_{t}^{\star}}(\gA_{t-1}\mymid  x_t)\le 1$ for all $x_t \in \RB^d$.
Importantly, the recursion \eqref{eq:tv_xy*_eq1} indicates that each iteration of DDPM can increase the TV distance of interest by at most $\mathcal{R}_{t-1}$.

It then boils down to controlling $\mathcal{R}_{t-1}$. Towards this end,
we would like to decompose
\begin{align}
\label{eq:Rt-minus-1-Rt-integral}
\gR_{t-1} = \int_{\gA_{t-1}}\gR_{t-1}(x_{t-1})\rd x_{t-1},
\end{align}
where we define
\begin{align}
\label{eq:defn-Rt-minus-1-x-t-minus-1}
\gR_{t-1}(x_{t-1}) \coloneqq p_{X_{t-1}}(x_{t-1}) - \int_{\mathbb{R}^d}p_{Y_{t-1}^{\star}\mymid  Y_{t}^{\star}}(x_{t-1}\mymid  x_t)p_{X_t}(x_t)\rd x_{t}.
\end{align}
In the ensuing steps,
we shall bound $\gR_{t-1}(x_{t-1})$ for  $x_{t-1} \in \gA_{t-1}$, which in turn facilitates analysis for $\gR_{t-1}$.

\medskip
\noindent {\bf Step 2: decomposing and calculating \texorpdfstring{$\gR_{t-1}(x_{t-1})$}{gRx at t minus 1}.}
In this step, we intend to calculate the function $\gR_{t-1}(x_{t-1})$ defined in \eqref{eq:defn-Rt-minus-1-x-t-minus-1}.
For notational convenience, for any vector $x\in \mathbb{R}^d$ we shall denote
\begin{align}
u_t(x) \coloneqq x + \eta_t s_t^{\star}(x)
\label{eq:defn-ut-DDPM-analysis}
\end{align}
in the sequel.
Further, we present the following useful lemma, whose proof can be found in Appendix~\ref{sec:proof:lem:u_t_bijection}.
\begin{lemma}\label{lem:u_t_bijection}
    For any $t = 1,\cdots, T$, the mapping $u_t: \RB^d \to \RB^d$ is a $C^1-$diffeomorphism.
\end{lemma}
For convenience in the following discussion, we denote the \textbf{inverse} of $u_t$ by $x_t(\cdot)$.

Equipped with the quantities $\{\overline{\alpha}_t\}$ (cf.~\eqref{eq:bar-alphat-discrete}) and the update rule \eqref{eq:defn-Yt-star-DDPM} of  $\{Y_{t}^{\star} \}$, we can demonstrate that, for each $t \ge 2$,
\begin{align*}
    p_{X_{t-1}\mymid  X_0}(x_{t-1}\mymid  x_0) &= \left(\frac{1}{2\pi (1-\overline{\alpha}_t)}\right)^{\frac{d}{2}}\exp\left(
    - \frac{\norm{x_{t-1} - \sqrt{\overline{\alpha}_{t-1}}x_0}_2^2}{2(1-\overline{\alpha}_{t-1})}
    \right),\\
    p_{Y_{t-1}^{\star}\mymid  Y_{t}^{\star}}(x_{t-1}\mymid  x_t) &= \left(\frac{1}{2\pi \sigma_t^2}\right)^{\frac{d}{2}}\exp\left(
    -\frac{\norm{\sqrt{\alpha_t} x_{t-1} - u_t}_2^2}{2\sigma_t^2}
    \right).
\end{align*}
Clearly, both $p_{X_{t-1}\mymid X_0}$ and $p_{Y_{t-1}^{\star}\mymid Y_{t}^{\star}}$ are density functions of conditional Gaussians. In light of this, it turns out that we can find another conditional Gaussian distribution $\widetilde{p}_{U_t|X_0}(u_t\,|\,x_0)$ satisfying the following convolution formula:
\begin{equation}\label{eq:gR_x_t-1_eq0}
p_{X_{t-1}\mymid  X_0}(x_{t-1}\mymid  x_0) = \int_{x_t} p_{Y_{t-1}^{\star}\mymid  Y_{t}^{\star}}\big(x_{t-1}\mymid  x_t(u_t)\big)\widetilde{p}_{U_t|X_0}(u_t\mymid  x_0) \rd u_t,\quad \forall x_{t-1}\in \gA_{t-1},~ x_0 \in \gX_{\mathsf{data}},
\end{equation}
with the exact form of $\widetilde{p}_{U_t\mymid  X_0}$ provided in the following lemma. The proof is deferred to Appendix~\ref{sec:proof:lem:determine_p_U_t_X_0}.
\begin{lemma}\label{lem:determine_p_U_t_X_0}
    The probability density function $\widetilde{p}_{U_t\mymid  X_0}(u_t\mymid  x_0)$ is given by
    \[
    \widetilde{p}_{U_t\mymid  X_0}(u_t\mymid  x_0) = \left(\frac{1}{2\pi(\alpha_t - \overline{\alpha}_t - \sigma_t^2)}\right)^{d/2}
    \exp\left\{
    -\frac{\norm{u_t - \sqrt{\overline{\alpha}_t}x_0}_2^2}{2(\alpha_t - \overline{\alpha}_t - \sigma_t^2)}
    \right\}.
    \]
\end{lemma}

Armed with the above density function and \eqref{eq:gR_x_t-1_eq0}, we can deduce the following lemma whose proof is deferred to Appendix~\ref{sec:proof:lem:gR_x_t-1_lem1}.
\begin{lemma}\label{lem:gR_x_t-1_lem1}
    For any $t = 2, \cdots, T$ and any $x_{t-1} \in \gA_{t-1}$, $\gR_{t-1}(x_{t-1})$ can be expressed as
    \[
    \gR_{t-1}(x_{t-1}) = \int p_{Y_{t-1}^\star\mid Y_t^\star}(x_{t-1}\mid x_t)p_{X_t\mid X_0}(x_t\mid x_0)\big(\gG(x_t, x_0) - 1\big)p_{X_0}(x_0)\rd x_0 \rd x_t.
    \]
with the definition of $\gG(x_t, x_0)$ given by
\[
\gG(x_t,  x_0) \coloneqq \frac{\widetilde{p}_{U_t\mymid  X_0}(u_t(x_t)\mymid  x_0)}{p_{X_t\mymid  X_0}(x_t\mymid  x_0)}\det\left(\frac{\rd u_t}{\rd x_{t}}\right)
\]
\end{lemma}

Given that both $\widetilde{p}_{U_t|X_0}$ and $p_{X_t\mymid  X_0}$ represent Gaussian distributions, we can obtain that
\begin{equation}\label{eq:G=G1xG2}
\begin{aligned}
    \gG(x_t,x_0) &= \det\left(\frac{\rd \big(x_t + \eta_t s_{t}^{\star}(x_t)\big)}{\rd x_t}\right)\cdot \frac{\left(\overline{\sigma}_t^2\right)^{-d/2}\exp\Big\{-
    \frac{
    \|u_t - \sqrt{\overline{\alpha}_t}x_0\|_2^2
    }{
    2\overline{\sigma}_t^2
    }\Big\}}
    {(1 - \overline{\alpha}_t)^{-d/2}\exp\Big\{-\frac{\|x_t - \sqrt{\overline{\alpha}_t}x_0\|_2^2}{2(1 - \overline{\alpha}_t)}\Big\}}\\
    &= \underbrace{\det\left(\frac{\rd \big(x_t + \eta_t s_{t}^{\star}(x_t)\big)}{\rd x_t}\right)\cdot
    \frac{(1 - \overline{\alpha}_t)^{d/2}}{\overline{\sigma}_t^d}}_{\eqqcolon \, \gG_1(x_t,x_0)}\cdot
    \underbrace{\exp\left\{
    \frac{\norm{x_t - \sqrt{\overline{\alpha}_t}x_0}_2^2}{2(1 - \overline{\alpha}_t)} -
    \frac{\norm{u_t - \sqrt{\overline{\alpha}_t}x_0}_2^2}{2\overline{\sigma}_t^2}
    \right\}}_{\eqqcolon \, \gG_2(x_t, x_0)}.
\end{aligned}
\end{equation}
Taken Lemma~\ref{lem:gR_x_t-1_lem1} and \eqref{eq:G=G1xG2} collectively, the above results demonstrate that
\begin{equation}\label{eq:gR_x_t-1_eq1}
    \gR_{t-1}(x_{t-1}) =
    \int p_{Y_{t-1}^{\star}\mymid  Y_{t}^{\star}}(x_{t-1}\mymid  x_t)p_{X_t\mymid  X_0}(u_t\mymid  x_0)
    \big(\gG_1(x_t, x_0)\gG_2(x_t, x_0) -1 \big)
    p_{X_0}(x_0)\rd x_0\rd x_t.
\end{equation}

\input{general_step_size_analysis}

\medskip
\noindent {\bf Step 5: establishing recursions for $\TV(p_{X_t}, p_{\overline{Y}_t})$ with the aid of conditional covariances.}
From Eqn.~\eqref{eq:gR_x_t-1_eq3_new}, we know that $\gR_{t-1}$ is upper bounded by $ \EB\big[\big(-Z_t(X_t)\big)_+\big]$, an object that can be further controlled through
the following lemma. The proof can be found in Appendix~\ref{sec:proof:lem:bd_EB-ZX_t_+}.
\begin{lemma}\label{lem:bd_EB-ZX_t_+}
For any iteration $t$, one has
\begin{equation}\label{eq:EB-Zx_t_+_eq2}
    \begin{aligned}
    \EB\big[\big(-Z_t(X_t)\big)_+\big] &\le \frac{8\overline{\alpha}_t \eta_t^2}{(1 - \overline{\alpha}_t)^3}\EB \left[\tr\big(\Cov_{0\mymid  t}(X_t)\big)\right]\\
    &+ C\widetilde{\sigma}_t^2
    \Big\{
    \EB \left[\tr\big(\Cov_{X_0\mymid  X_t}(X_t)\big)\right] - \EB \left[\tr\big(\Cov_{X_0|X_{t-1}}(X_{t-1})\big)\right]
    \Big\} + \frac{C}{T^{10}},
    \end{aligned}
    \end{equation}
    where $C>0$ is some universal constant.
\end{lemma}
Taking \eqref{eq:tv_xy*_eq1}, \eqref{eq:gR_x_t-1_eq3_new} and \eqref{eq:EB-Zx_t_+_eq2} collectively yields
\begin{equation}\label{eq:rec_TV_XbarY}
\begin{aligned}
    \TV(p_{X_{t-1}}, p_{\overline{Y}_{t-1}}) &\le \gR_{t-1} + \TV(p_{X_{t}}, p_{\overline{Y}_{t}}) \le
    \TV(p_{X_{t}}, p_{\overline{Y}_{t}}) + \EB\big[(-Z_t(X_t))_+\big]\\
    &\le \TV(p_{X_{t}}, p_{\overline{Y}_{t}})
    {+} \left(C\widetilde{\sigma}_t^2 {+} \frac{\overline{\alpha}_t\eta_t^2}{(1 - \overline{\alpha}_t)^3}\right)\EB\left[\tr\big(\Cov_{0\mid t}(X_t)\big)\right]\\
    &\quad -{C\widetilde{\sigma}_t^2}\EB\left[\tr\left(\Cov_{0|t-1}(X_{t-1})\right)\right] {+} \frac{C}{T^{10}}.
\end{aligned}
\end{equation}

\medskip
\noindent {\bf Step 6: controlling the effect of score estimation errors.}
Recall that the process $\{\overline{Y}_t\}$ is constructed without incorporating score matching errors, and hence we still need to quantify the influence of inexact scores upon convergence.
Towards this, apply the data processing inequality w.r.t.~KL divergence to obtain
\begin{align*}
    &\KL(p_{\overline{Y}_1}\parallel p_{\widehat{Y}_1}) \le \KL\left(p_{\overline{Y}_1,\overline{Y}_1^-,\cdots,\overline{Y}_T,\overline{Y}_T^-}~\big\|~p_{\widehat{Y}_1,\widehat{Y}_1^-,\cdots,\widehat{Y}_T,\widehat{Y}_T^-}\right)\\
    &\overset{\text{(a)}}{=} \KL(p_{\overline{Y}_T^-}\parallel p_{\widehat{Y}_T^-}) + \ssum{t}{2}{T}\EB_{x_t\sim p_{\overline{Y}_t}}\left[\KL\left(p_{\overline{Y}_{t-1}^-\mymid  \overline{Y}_t=x_t}~\big\|~ p_{\widehat{Y}_{t-1}^-|\widehat{Y}_t=x_t}\right)\right]\\
    &\qquad + \ssum{t}{2}{T}\EB_{x_t\sim p_{\overline{Y}_t^-}}\left[\KL\left(p_{\overline{Y}_{t-1}|\overline{Y}_t^-=x_t}~\big\|~ p_{\widehat{Y}_{t-1}|\widehat{Y}_t^-=x_t}\right)\right]\\
    &\overset{\text{(b)}}{=} \ssum{t}{2}{T}\EB_{x_t\sim p_{\overline{Y}_t}}\left[\KL\left(p_{\overline{Y}_{t-1}^-\mymid  \overline{Y}_t=x_t}~\big\|~ p_{\widehat{Y}_{t-1}^-|\widehat{Y}_t=x_t}\right) \right]
    \overset{\text{(c)}}{=}
    \ssum{t}{2}{T}\EB_{x_t\sim p_{\overline{Y}_t}}\left[\KL\left(p_{{Y}_{t-1}^{\star}|Y_{t}^{\star}=x_t}~\big\|~ p_{{Y}_{t-1}|{Y}_t=x_t}\right)\right].
\end{align*}
Here, (a) follows from the chain rule of KL divergence, (b) holds since the conditional distribution of $\widehat{Y}_t$ given $\widehat{Y}_t^- = x$ and that of $\overline{Y}_t$ given $\overline{Y}_t^- = x$ are identical, while (c) arises from the construction of  $\overline{Y}_{t-1}^-\mymid  \overline{Y}_t$ and $\widehat{Y}_{t-1}^-\mymid  \widehat{Y}_t$ (see \eqref{eq:defn-Yt-1-minus-Yt-bar} and \eqref{eq:defn-Yt-1-minus-Yt-bar-error}).

Recall that \( Y_{t-1}^{\star} \mymid  Y_{t}^{\star} = x_t \) (see \eqref{eq:defn-Yt-star-DDPM}) and \( Y_{t-1} \mymid  Y_t = x_t \) are two Gaussian distributions given by
\[
Y_{t-1}^{\star}\mymid  Y_{t}^{\star}=x_t~\sim \gN\left(\frac{x_t + \eta_t s_t^{\star}(x_t)}{\sqrt{\alpha_t}}, \sigma_t^2 I_d\right),\quad
Y_{t-1}\mymid  Y_t=x_t~\sim \gN\left(\frac{x_t + \eta_t s_t(x_t)}{\sqrt{\alpha_t}}, \sigma_t^2 I_d\right)
\]
with $\eta_t,~\sigma_t^2$ satisfying $\eta_t^2 \le C(1 - \alpha_t)\sigma_t^2$. Further, the KL divergence between two Gaussian measures admits a closed-form expression, i.e.,
\begin{align*}
&\KL\left(p_{Y_{t-1}^{\star}\mymid  Y_{t}^{\star}}(\cdot|x_t)\parallel p_{Y_{t-1}|Y_t}(\cdot|x_t)\right) = \KL\left(\gN\left(\frac{x_t + \eta_t s_t^{\star}(x_t)}{\sqrt{\alpha_t}}, \sigma_t^2 I_d\right)~\bigg\|~
\gN\left(\frac{x_t + \eta_t s_t(x_t)}{\sqrt{\alpha_t}}, \sigma_t^2 I_d\right)
\right)\\
&= \frac{\eta_t^2/\alpha_t}{2\sigma_t^2}\norm{s_t(x_t) - s_t^{\star}(x_t)}_2^2 = \frac{\eta_t^2/\alpha_t}{2\sigma_t^2}\norm{\varepsilon^{\mathsf{sc}}_{t}(x_t)}_2^2
= \frac{C(1-\alpha_t)}{2\alpha_t}\norm{\varepsilon^{\mathsf{sc}}_{t}(x_t)}_2^2 \le C(1 - \alpha_t)\norm{\varepsilon^{\mathsf{sc}}_{t}(x_t)}_2^2.
\end{align*}
Therefore, we can control the KL divergence between $\overline{Y}_1$ (the auxiliary process without score errors) and $\widehat{Y}_1$ (the auxiliary process with score errors) as follows:
\begin{equation}\label{eq:kl_pbarY_0&phatY_0}
\begin{aligned}
    \KL(p_{\overline{Y}_1}\parallel p_{\widehat{Y}_1}) &\le
    \ssum{t}{2}{T}\EB_{x_t\sim p_{\overline{Y}_t}}
    \left[\KL\left(p_{{Y}_{t-1}^{\star}\mymid  Y_{t}^{\star}=x_t}~\big\|~ p_{{Y}_{t-1}\mymid  {Y}_t=x_t}\right)\right]\\
    &\le \ssum{t}{2}{T} C(1 - \alpha_t)\EB_{x_t\sim p_{\overline{Y}_t}}\left[\norm{\varepsilon^{\mathsf{sc}}_{t}(x_t)}_2^2\right]
    \overset{\text{(a)}}{\le}
    \ssum{t}{2}{T} C(1 - \alpha_t)\EB_{x_t\sim p_{X_t}}\left[\norm{\varepsilon^{\mathsf{sc}}_{t}(x_t)}_2^2\right]\\
    &\le \frac{c_1C\log T}{T}\ssum{t}{2}{T}\varepsilon_{\mathsf{score},t}^2 \le c_1C(\log T) \varepsilon_{\mathsf{score}}^2,
\end{aligned}
\end{equation}
where (a) follows from \eqref{eq:p_barY_le_p_X}.

\medskip
\noindent {\bf Step 7: putting all pieces together.}
Applying inequality~\eqref{eq:rec_TV_XbarY} from $1$ to \(T\) recursively, we arrive at
\begin{equation}\label{eq:tv_px_0_p_barY_0}
\begin{aligned}
    \TV(p_{X_1}, p_{\overline{Y}_1}) &\le
    \TV(p_{X_T}, p_{Y_{t}^{\star}}) + \ssum{t}{2}{T-1}\left(
    \frac{8\overline{\alpha}_t \eta_t^2}{(1 - \overline{\alpha}_t)^3} + C(\widetilde{\sigma}_t^2 - \widetilde{\sigma}_{t+1}^2)
    \right)\EB \left[\tr\left(\Cov_{0\mid t}(X_t)\right)\right]\\
    &\quad\quad +
    \left(C\widetilde{\sigma}_t^2 + \frac{8\overline{\alpha}_t \eta_t^2}{(1 - \overline{\alpha}_t)^3} \right)\EB \left[\tr\left(\Cov_{0\mid T}(X_T)\right)\right] + \frac{C}{T^9}\\
    &\overset{\text{(a)}}{\le}  \TV(p_{X_T}, p_{\overline{Y}_T}) {+}
    \left(\frac{C\log T}{T}\right)^2 \ssum{t}{1}{T-1}\frac{\overline{\alpha}_t}{1-\overline{\alpha}_t}\EB\left[\tr\left(\Cov_{0\mid t}(X_t)\right)\right]\\
    &\quad\quad
    + \frac{C \log T}{T}\frac{\overline{\alpha}_T}{1 - \overline{\alpha}_T}\EB\left[\tr\left(\Cov_{0\mid T}(X_T)\right)\right] + \frac{1}{T^9}\\
    &\overset{\text{(b)}}{\le}
    \TV(p_{X_T}, p_{\overline{Y}_T}) + C_3kT\log T\left(\frac{C\log T}{T}\right)^2 + \frac{C_3C k \log^2 T}{T} + \frac{C}{T^9}\\
    &\le \frac{1}{T^{10}} + C^2C_3\frac{k\log^3 T}{T} + \frac{C}{T^9} \le C_8\frac{k\log^3 T}{T}.
\end{aligned}
\end{equation}
Here, (a) applies Lemma~\ref{lem:step_size} and the basic property \eqref{eq:basic-alphat-property},
  (b) makes use of the moment inequality \eqref{eq:posterior_mmt} with $l=2$, whereas the penultimate inequality results from \citet[Lemma 10]{li2024adapting}.

It remains to bound the TV distance between $p_{Y_1}$ and $p_{\overline{Y}_1}$. Towards this, observe that
\begin{equation}\label{eq:tv_pY0&pbarY0}
\begin{aligned}
\TV(p_{Y_1}, p_{\overline{Y}_1}) &= \int_{\mathbb{R}^d} \left( p_{\overline{Y}_1}(x) - p_{{Y}_1}(x) \right) \mathbbm{1} \{ p_{\overline{Y}_1}(x) > p_{Y_1}(x) \}  \rd x + \mathbb{P}(\overline{Y}_1 = \infty) \\
&\overset{\text{(a)}}{\le} \int_{\mathbb{R}^d} \left( p_{\overline{Y}_1}(x) - p_{\widehat{Y}_1}(x) \right) \mathbbm{1} \{ p_{\overline{Y}_1}(x) > p_{\widehat{Y}_1}(x) \} \rd x + \mathbb{P}(\overline{Y}_1 = \infty) \\
&\overset{\text{(b)}}{\le} \TV( p_{\overline{Y}_1}, p_{\widehat{Y}_1}) + \TV(p_{X_1}, p_{\overline{Y}_1})
\overset{\text{(c)}}{\le} \sqrt{\KL( p_{\overline{Y}_1} \parallel p_{\widehat{Y}_1})} + C \frac{k \log^3 T}{T},
\end{aligned}
\end{equation}
where (a) holds due  to Lemma~\ref{lem:p_Y_ge_p_hat_Y}, (b) follows since $\PB(\overline{Y}_1 = \infty) \le \TV(X_1, \overline{Y}_1)$, and (c) invokes Pinsker's inequality and \eqref{eq:tv_px_0_p_barY_0}.
Combine \eqref{eq:kl_pbarY_0&phatY_0}, \eqref{eq:tv_px_0_p_barY_0} and \eqref{eq:tv_pY0&pbarY0} to reach
\begin{align*}
    \TV(p_{X_1},p_{Y_1}) &\le \TV(p_{X_1},p_{\overline{Y}_1}) + \TV(p_{\overline{Y}_1}, p_{Y_1})\\
    &\le C\frac{k\log^3 T}{T} + C\frac{k\log^3 T}{T} + \sqrt{\KL( p_{\overline{Y}_1}\parallel p_{\widehat{Y}_1})}\\
    &\le 2C\frac{k\log^3 T}{T} + \sqrt{c_1 \varepsilon_{\mathsf{score}}^2\log T} = C\frac{k\log^3 T}{T} + \sqrt{c_1C\log T}\varepsilon_{\mathsf{score}},
\end{align*}
thereby concluding the proof of  Theorem~\ref{thm:ddpm_tv_conv}.

\subsection{Proof of Lemma~\ref{lem:p_Y_ge_p_hat_Y}}\label{sec:proof:lem:p_Y_ge_p_hat_Y}
We can invoke induction to establish this result. To begin with, it can be easily checked that $p_{Y_T} = p_{\widehat{Y}_T}$. Next, suppose that the claim~\eqref{eq:p_Y_ge_p_hat_Y} holds for $t+1$, then it follows that%
\begin{align*}
    p_{\widehat{Y}_t}(x) &= \int_{\RB^d} p_{\widehat{Y}_t\mymid \widehat{Y}_t^-}(x\mymid  x^\prime)p_{\widehat{Y}_t^-}(x^\prime)\rd x^\prime = \left(\frac{p_{X_t}(x)}{p_{\overline{Y}_t^-}(x)}\wedge 1\right)p_{\widehat{Y}_t^-}(x) \le p_{\widehat{Y}_t^-}(x)\\
    &= \int_{\RB^d}p_{\widehat{Y}_t^-\mymid \widehat{Y}_{t+1}}(x\mymid  x^\prime)p_{\widehat{Y}_{t+1}}(x^\prime)\rd x^\prime
    \le \int_{\RB^d}p_{Y_t\mymid Y_{t+1}}(x\mymid  x^\prime)p_{Y_{t+1}}(x^\prime)\rd x^\prime = p_{Y_{t}}(x),
\end{align*}
thus validating the claim~\eqref{eq:p_Y_ge_p_hat_Y} for $t$. This immediately concludes the proof by induction.

\subsection{Proof of Lemma~\ref{lem:u_t_bijection}}\label{sec:proof:lem:u_t_bijection}
We still use Lemma~\ref{lem:global_inverse} to prove this lemma.
In fact, what we need to do is verify that $\det \left(\frac{\partial u_t(x)}{\partial x}\right) \neq 0, ~\forall x \in \RB^d$ and $\lim\limits_{x\to \infty}\norm{u_t(x)}_2 = \infty$.
On the one hand, owing to Tweedie's formula~\eqref{eq:score_to_posterior_new}, it can be derived that
\begin{align*}
    \frac{\partial u_t(x)}{\partial x} &= \frac{\partial (x + \eta_t s_t^\star(x))}{\partial x} = I + \eta_t \left(\frac{\overline{\alpha}_t}{(1 - \overline{\alpha}_t)^2}\Cov_{0\mid t}(x) - \frac{1}{1-\overline{\alpha}_t}I\right)\\
    &= \left(1 - \frac{\eta_t}{1 - \overline{\alpha}_t}\right)I + \frac{\overline{\alpha}_t \eta_t}{(1 - \overline{\alpha}_t)^2}\Cov_{0\mid t}(x) \succeq  \left(1 - \frac{\eta_t}{1 - \overline{\alpha}_t}\right)I \succeq \frac{1}{2}I.
\end{align*}
Here, the last inequality is a consequence of the step size condition $\eta_t \le \frac{1}{2}(1 - \overline{\alpha}_t)$. According to above derivation, we know that $\det\left(\frac{\partial u_t(x)}{\partial x}\right) \neq 0$ for any $x\in \RB^d$.

On the other hand, Assumption~\ref{ass:bd_supp} says that $\sup\limits_{x\in \gX_{\mathsf{data}}}\norm{x}_2 \le T^{c_R}$. Hence, $\mu_{0\mid t}(x)$ defined by~\eqref{eq:def_mu_cov} satisfies that
\[
\norm{\mu_{0\mid t}(x)}_2 = \norm{\EB[X_0\mid X_t = x]}_2 \le \EB[\norm{X_0}_2\mid X_t = x] \le T^{c_R},~ \forall x \in \RB^d.
\]
Combining this with Tweedie's formula~\eqref{eq:score_to_posterior_new} yields,
\begin{align*}
    \lim\limits_{x\to \infty}\norm{u_t(x)}_2 &= \lim\limits_{x\to \infty}\norm{x+ \eta_t s_t^\star(x)}_2 = \lim\limits_{x\to \infty}\norm{\left(1 - \frac{\eta_t}{1 - \overline{\alpha}_t}\right)x + \frac{\sqrt{\overline{\alpha}_t}}{1 - \overline{\alpha}_t}\mu_{0\mid t}(x)}_2\\
    &\ge \left(1 - \frac{\eta_t}{1 - \overline{\alpha}_t}\right)\lim\limits_{x\to \infty}\norm{x}_2 - \frac{\sqrt{\overline{\alpha}_t}}{1 - \overline{\alpha}_t}T^{c_R} \ge \frac{1}{2}\lim\limits_{x\to \infty}\norm{x}_2 - \frac{\sqrt{\overline{\alpha}_t}}{1 - \overline{\alpha}_t}T^{c_R} = \infty.
\end{align*}
This completes the proof of the lemma.

\subsection{Proof of Lemma~\ref{lem:determine_p_U_t_X_0}}\label{sec:proof:lem:determine_p_U_t_X_0}
Suppose that conditional on $X_0=x_0$, one has $U_t \sim \gN(\lambda_t x_0, \overline{\sigma}_t^2 I)$ for some quantities $\lambda_t$ and $\overline{\sigma}_t>0$.
Denoting by $\widetilde{p}_{U_t|X_0}$ the conditional density of $U_t$ given $X_0$, one can easily see that: the distribution associated with the pdf $\widetilde{q}_t(x_{t-1})=\int_{x_t} p_{Y_{t-1}^{\star}\mymid  Y_{t}^{\star}}(x_{t-1}|x_t)\widetilde{p}_{U_t|X_0}(u_t|x_0) \rd u_t$ is
\[
\gN\left(\frac{\lambda_t}{\sqrt{\alpha_t}}x_0,~ \frac{\sigma_t^2 + \overline{\sigma}_t^2}{\alpha_t}\right).
\]
By taking
\[
\lambda_t = \sqrt{\alpha_t}\cdot \sqrt{\overline{\alpha}_{t-1}} = \sqrt{\overline{\alpha}_t}\qquad \text{and} \qquad
\overline{\sigma}_t^2 = \alpha_t(1 - \overline{\alpha}_{t-1}) - {\sigma_t^2},
\]
we see that the above Gaussian distribution coincides with $\gN\left(\sqrt{\overline{\alpha}_{t-1}}x_0, 1 - \overline{\alpha}_{t-1}\right)$, which is precisely the distribution of $X_{t-1}$ given $X_0=x_0$.
%

%

\subsection{Proof of Lemma~\ref{lem:gR_x_t-1_lem1}}\label{sec:proof:lem:gR_x_t-1_lem1}
By combining Lemma~\ref{lem:u_t_bijection} and the basic change of variable formula, we can do the following computation,
\begin{align}
    &\gR_{t-1}(x_{t-1}) = p_{X_{t-1}}(x_{t-1}) - \int p_{Y_{t-1}^{\star}\mymid  Y_{t}^{\star}}(x_{t-1}\mymid  x_t)p_{X_t\mymid  X_0}(x_t\mymid  x_0)p_{X_0}(x_0)\rd x_0 \rd x_t \notag\\
    &= p_{X_{t-1}}(x_{t-1}) - \int p_{Y_{t-1}^{\star}\mymid  Y_{t}^{\star}}(x_{t-1}\mymid  x_t)\widetilde{p}_{U_t\mymid  X_0}\big(u_t(x_t)\mymid  x_0\big){\frac{p_{X_t\mymid  X_0}\big(x_t\mymid  x_0\big)}{\widetilde{p}_{U_t\mymid  X_0}\big(u_t(x_t)\mymid  x_0\big)}}p_{X_0}(x_0)\rd x_0\rd x_t \notag\\
    &= p_{X_{t-1}}(x_{t-1}) - \int p_{Y_{t-1}^{\star}\mymid  Y_{t}^{\star}}\big(x_{t-1}\mymid  x_t(u_t)\big)\widetilde{p}_{U_t\mymid  X_0}(u_t\mymid  x_0){\frac{p_{X_t\mymid  X_0}\big(x_t(u_t)\mymid  x_0\big)}{\widetilde{p}_{U_t\mymid  X_0}(u_t\mymid  x_0)}}\det\left(\frac{\rd x_{t}}{\rd u_t}\right)p_{X_0}(x_0)\rd x_0\rd u_t \notag\\
    &= \underbrace{\left(\int p_{X_{t-1}\mymid  X_0}(x_{t-1}\mymid  x_0)p_{X_0}(x_0)\rd x_0 -
    \int p_{Y_{t-1}^{\star}\mymid  Y_{t}^{\star}}\big(x_{t-1}\mymid  x_t(u_t)\big)\widetilde{p}_{U_t\mymid  X_0}(u_t\mymid  x_0)p_{X_0}(x_0)\rd x_0\rd u_t\right)}_{\eqqcolon\, \gR_{t-1}^\prime(x_{t-1})} \notag\\
    & - \int p_{Y_{t-1}^{\star}\mymid  Y_{t}^{\star}}\big(x_{t-1}\mymid  x_t(u_t)\big)\widetilde{p}_{U_t\mymid  X_0}(u_t\mymid  x_0){\left(\frac{p_{X_t\mymid  X_0}(x_t(u_t)\mymid  x_0)}{\widetilde{p}_{U_t\mymid  X_0}(u_t\mymid  x_0)}\det\left(\frac{\rd x_{t}}{\rd u_t}\right) - 1\right)}p_{X_0}(x_0)\rd x_0\rd u_t.
    \label{eq:Rt-1-x-t-1-equiva-DDPM}
\end{align}
Regarding the first term in \eqref{eq:Rt-1-x-t-1-equiva-DDPM}, it is readily seen from~\eqref{eq:gR_x_t-1_eq0} that $$\gR_{t-1}^{\prime}(x_{t-1}) = 0.$$
When it comes to the second term in \eqref{eq:Rt-1-x-t-1-equiva-DDPM}, it follows from the definition of $u_t$ (cf.~\eqref{eq:defn-ut-DDPM-analysis}) that
\begin{align*}
    &- \int p_{Y_{t-1}^{\star}\mymid  Y_{t}^{\star}}\big(x_{t-1}\mymid  x_t(u_t)\big)\widetilde{p}_{U_t\mymid  X_0}(u_t\mymid  x_0){\left(\frac{p_{X_t\mymid  X_0}\big(x_t(u_t)\mymid  x_0\big)}{\widetilde{p}_{U_t\mymid  X_0}(u_t\mymid  x_0)}\det\left(\frac{\rd x_{t}}{\rd u_t}\right) - 1\right)}p_{X_0}(x_0)\rd x_0\rd u_t\\
    &= - \int p_{Y_{t-1}^{\star}\mymid  Y_{t}^{\star}}\big(x_{t-1}\mymid  x_t(u_t)\big)p_{X_t\mymid  X_0}\big(x_t(u_t)\mymid  x_0\big)
    \det\left(\frac{\rd x_{t}}{\rd u_t}\right)\\
    &\quad\quad\quad\quad\quad\quad\quad\quad\quad\quad\left(1 - \frac{\widetilde{p}_{U_t\mymid  X_0}(u_t\mymid  x_0)}{p_{X_t\mymid  X_0}\big(x_t(u_t)\mymid  x_0\big)}\det\left(\frac{\rd u_t}{\rd x_{t}}\right) \right)
    p_{X_0}(x_0)\rd x_0\rd u_t\\
    &=\int p_{Y_{t-1}^{\star}\mymid  Y_{t}^{\star}}(x_{t-1}\mymid  x_t)p_{X_t\mymid  X_0}(x_t\mymid  x_0)
    \left(\frac{\widetilde{p}_{U_t\mymid  X_0}\big(u_t(x_t)\mymid  x_0\big)}{p_{X_t\mymid  X_0}(x_t\mymid  x_0)}\det\left(\frac{\rd u_t}{\rd x_{t}}\right) -1 \right)
    p_{X_0}(x_0)\rd x_0\rd x_t\\
    &= \int p_{Y_{t-1}^\star\mid Y_t^\star}(x_{t-1}\mid x_t)p_{X_t\mid X_0}(x_t\mid x_0)\big(\gG(x_t, x_0) - 1\big)p_{X_0}(x_0)\rd x_0 \rd x_t.
\end{align*}
This completes the proof of the lemma.

\subsection{Proof of Lemma~\ref{lem:bounds-G1-G2-DDPM}}\label{sec:proof:lem:bounds-G1-G2-DDPM}
From Tweedie's formula~\eqref{eq:score_to_posterior_new}, we can further simplify $\gG_1(x_t,x_0)$ as follows:
\begin{equation}\label{eq:expr_gG_1x_tx_0}
\begin{aligned}
    \gG_1(x_t, x_0) &= \det\left( I + \eta_t \frac{\partial s_{t}^{\star}(x_t)}{\partial x_t}\right)\cdot \frac{(1 - \overline{\alpha}_t)^{d/2}}{\overline{\sigma}_t^d}\\
    &\overset{\text{(a)}}{=}
    \det\left(
     I + \eta_t\left\{\frac{\overline{\alpha}_t}{(1-\overline{\alpha}_t)^2}\Cov_{0\mid t}(x_t) - \frac{1}{1-\overline{\alpha}_t} I\right\}
    \right)\cdot \frac{(1 - \overline{\alpha}_t)^{d/2}}{\overline{\sigma}_t^d}\\
    &= \det\left(
    \left(1 - \frac{\eta_t}{1 - \overline{\alpha}_t}\right) I + \frac{\overline{\alpha}_t\eta_t}{(1 - \overline{\alpha}_t)^2}\Cov_{0\mid t}(x_t)
    \right)\cdot \frac{(1 - \overline{\alpha}_t)^{d/2}}{\overline{\sigma}_t^d}\\
    &\overset{\text{(b)}}{=} \det\left( I + \frac{\overline{\alpha}_t\eta_t}{(1 - \overline{\alpha}_t)(1 - \overline{\alpha}_t - \eta_t)}\Cov_{0\mid t}(x_t)\right) = \det\left( I + \overline{\alpha}_t\eta^{(1)}_t \Cov_{0\mid t}(x_t)\right),
\end{aligned}
\end{equation}
where we define $\eta^{(1)}_t \coloneqq \frac{\eta_t}{(1 - \overline{\alpha}_t)(1 - \overline{\alpha}_t - \eta_t)}$. Here, (a) follows from \eqref{eq:score_to_posterior_new}, whereas (b) can be shown by combining $\det(\lambda  A) = \lambda^d \det( A)$ and the relation~\eqref{eq:ddpm_gen_stp_choice}.

We then move on to $\gG_2(x_t, x_0)$. Towards this end, we make the observation that
\begin{equation}\label{eq:loggG_2_eq1_new}
\begin{aligned}
    \log \gG_2(x_t, x_0) &= \frac{\norm{x_t - \sqrt{\overline{\alpha}_t}x_0}_2^2}{2(1 - \overline{\alpha}_t)} -
    \frac{\norm{u_t - \sqrt{\overline{\alpha}_t}x_0}_2^2}{2\overline{\sigma}_t^2}
    =
    \frac{\norm{x_t - \sqrt{\overline{\alpha}_t}x_0}_2^2}{2(1 - \overline{\alpha}_t)} -
    \frac{\norm{x_t - \sqrt{\overline{\alpha}_t}x_0 + \eta_t s_{t}^{\star}(x_t)}_2^2}{2\overline{\sigma}_t^2}\\
    &=
    \left(\frac{1}{2(1-\overline{\alpha}_t)} {-} \frac{1}{2\overline{\sigma}_t^2}\right)\norm{x_t {-} \sqrt{\overline{\alpha}_t}x_0}_2^2 {-} \frac{\eta_t}{\overline{\sigma}_t^2}(x_t {-} \sqrt{\overline{\alpha}_t}x_0)^\top s_{t}^{\star}(x_t) {-} \frac{\eta_t^2}{2\overline{\sigma}_t^2}\norm{s_{t}^{\star}(x_t)}_2^2\\
    &\overset{\text{(a)}}{=}
    -\frac{\eta_t}{\overline{\sigma}_t^2(1 - \overline{\alpha}_t)}(x_t - \sqrt{\overline{\alpha}_t}x_0)^\top \big(\sqrt{\overline{\alpha}_t}\mu_{0|t}(x_t) - x_t \big) -
    \frac{\eta_t^2}{2\overline{\sigma}_t^2(1 - \overline{\alpha}_t)^2}\norm{\sqrt{\overline{\alpha}_t}\mu_{0|t} (x_t)- x_t}_2^2\\
    &\quad + \left(\frac{1}{2(1-\overline{\alpha}_t)} {-} \frac{1}{2\overline{\sigma}_t^2}\right)\norm{x_t {-} \sqrt{\overline{\alpha}_t}x_0}_2^2\\
    &=
    \left(\frac{1}{2(1-\overline{\alpha}_t)} {-} \frac{1}{2\overline{\sigma}_t^2}\right)\norm{x_t {-} \sqrt{\overline{\alpha}_t}x_0}_2^2 +
    \left(\frac{\eta_t}{\overline{\sigma}_t^2(1 - \overline{\alpha}_t)} - \frac{\eta_t^2}{2\overline{\sigma}_t^2(1 - \overline{\alpha}_t)^2}\right)\norm{\sqrt{\overline{\alpha}_t}\mu_{0|t} (x_t) - x_t}_2^2\\
    &\quad +
    \frac{\eta_t}{\overline{\sigma}_t^2(1 - \overline{\alpha}_t)}\big(\sqrt{\overline{\alpha}_t}x_0 - \sqrt{\overline{\alpha}_t}\mu_{0|t}(x_t)\big)^\top \big(\sqrt{\overline{\alpha}_t}\mu_{0|t}(x_t) - x_t\big),
\end{aligned}
\end{equation}
where (a) follows since $s_{t}^{\star}(x_t) = \frac{\sqrt{\overline{\alpha}_t}}{1-\overline{\alpha}_t}\mu_{0|t} (x_t) - \frac{1}{1-\overline{\alpha}_t}x_t$ (see~\eqref{eq:score_to_posterior_new}). Further, in view of~\eqref{eq:ddpm_gen_stp_choice}, one has
\begin{align*}
    \frac{1}{2(1-\overline{\alpha}_t)} - \frac{1}{2\overline{\sigma}_t^2} &= \frac{\overline{\sigma}_t^2 - (1 - \overline{\alpha}_t)}{2(1 - \overline{\alpha}_t)\overline{\sigma}_t^2} = \frac{(1-\overline{\alpha}_t)\left(1 - \frac{\eta_t}{1 - \overline{\alpha}_t}\right)^2 - (1 - \overline{\alpha}_t)}{2(1 - \overline{\alpha}_t)\overline{\sigma}_t^2}\\
    &= -\frac{2(1 - \overline{\alpha}_t)\eta_t - \eta_t^2}{2(1 - \overline{\alpha}_t)^2 \overline{\sigma}_t^2} = -\left(\frac{\eta_t}{(1 - \overline{\alpha}_t)\overline{\sigma}_t^2} - \frac{\eta_t^2}{2(1 - \overline{\alpha}_t)^2\overline{\sigma}_t^2}\right).
\end{align*}
Let us denote $\eta^{(2)}_t\coloneqq \frac{\eta_t}{(1 - \overline{\alpha}_t)\overline{\sigma}_t^2} - \frac{\eta_t^2}{2(1 - \overline{\alpha}_t)^2\overline{\sigma}_t^2}$.
Substituting these equations into \eqref{eq:loggG_2_eq1_new} yields
\begin{equation}\label{eq:loggG_2_eq2_new}
\begin{aligned}
    \log \gG_2(x_t, x_0) & = \eta^{(2)}_t\left\{
    \norm{\sqrt{\overline{\alpha}_t}\mu_{0|t}(x_t) {-} x_t}_2^2 {-} \norm{\sqrt{\overline{\alpha}_t}x_0 {-} x_t}_2^2
    \right\} \\
    &\qquad +
    \frac{\sqrt{\overline{\alpha}_t}\eta_t}{\overline{\sigma}_t^2(\alpha_t - \overline{\alpha}_t)^2}\big(x_0 {-} \mu_{0|t}(x_t) \big)^\top \big(\sqrt{\overline{\alpha}_t}\mu_{0|t} (x_t){-} x_t\big)\\
    &=
    \zeta_t(x_t,x_0)+ \int_{x_0} \log \gG_2(x_t,x_0)p_{X_0\mymid  X_t}(x_0\mymid  x_t)\rd x_0
    =
    \zeta_t(x_t,x_0) {-} \overline{\alpha}_t\eta^{(2)}_t \tr\left(\Cov_{0\mid t}(x_t)\right).
\end{aligned}
\end{equation}
Here, it can be easily verified that $\zeta_t(x_t,x_0)$ satisfies $\int_{x_0}\zeta_t(x_t,x_0)p_{X_0\mymid  X_t}(x_0\mymid  x_t)\rd x_0 = 0$ for all $x_t$.

To finish up, combining~\eqref{eq:expr_gG_1x_tx_0} and \eqref{eq:loggG_2_eq2_new}  concludes the proof.

\subsection{Proof of Lemma~\ref{lem:bd_EB-ZX_t_+}}\label{sec:proof:lem:bd_EB-ZX_t_+}
By setting the square matrix $\Delta$ in Lemma~\ref{lem:logdet_expand} to $0$, we can show that for any positive semi-definite matrix $ A\in \RB^{d\times d}$,
    \begin{equation}\label{eq:logdet_I+A}
    \log \det\left( I +  A\right) \ge \tr( A) - \norm{ A}_{\mathrm{F}}^2.
    \end{equation}
Equipped with this result, we can derive the following inequality:
\begin{equation}\label{eq:EB-Zx_t_+_eq1}
\begin{aligned}
    \EB_{X_t}\big[(-Z_t(X_t))_+\big] &= \EB_{X_t}\left[\left(
    \overline{\alpha}_t{\eta}_t^\prime
    \tr\left(\Cov_{0\mid t}(X_t)\right)
    - \log \det \big( I + \overline{\alpha}_t \eta^{(1)}_t \Cov_{0\mid t}(X_t)\big)
    \right)_+\right]\\
    &\overset{(a)}{\le}
    \EB_{X_t}\left[\left(
    \overline{\alpha}_t\left(\eta^{(2)}_t - \eta^{(1)}_t\right)\tr\left(\Cov_{0\mid t}(X_t)\right)
    + {\overline{\alpha}_t^2 \big(\eta^{(1)}_t\big)^2} \norm{\Cov_{0\mid t}(X_t)}_{\mathrm{F}}^2
    \right)_+\right]\\
    &\le
    \overline{\alpha}_t \left(\eta^{(2)}_t - \eta^{(1)}_t\right)_+
    \EB\left[\tr\left(\Cov_{0\mid t}(X_t)\right)\right] + {\overline{\alpha}_t^2 \big(\eta^{(1)}_t\big)^2} \EB\left[\norm{\Cov_{0\mid t}(X_t)}_{\mathrm{F}}^2\right].
\end{aligned}
\end{equation}
Here,  (a) holds by combining \eqref{eq:logdet_I+A} and the fact that the function $(\cdot)_+$ is non-decreasing and the last inequality follows from $(a + b)_+ \le (a)_+ + (b)_+$.

In order to further bound \eqref{eq:EB-Zx_t_+_eq1}, let us inspect the coefficients  $\overline{\alpha}_t\big(\eta^{(2)}_t - \eta^{(1)}_t\big)_+$ and ${\overline{\alpha}_t^2 \big(\eta^{(1)}_t\big)^2}$. Combining the definitions of $\eta^{(1)}_t,~ \eta^{(2)}_t$ and the relation~\eqref{eq:ddpm_gen_stp_choice} results in
\begin{align*}
    \overline{\alpha}_t\left(\eta^{(2)}_t - \eta^{(1)}_t\right)_+ &= \overline{\alpha}_t\left(\frac{\eta_t}{(1 -\overline{\alpha}_t)\overline{\sigma}_t^2} - \frac{\eta_t^2}{2(1 - \overline{\alpha}_t)^2\overline{\sigma}_t^2}
    - \frac{\eta_t}{(1 - \overline{\alpha}_t)(1 - \overline{\alpha}_t - \eta_t)}\right)_+\\
    &\le  \frac{\overline{\alpha}_t\eta_t}{1-\overline{\alpha}_t}\cdot \frac{\left(1 - \overline{\alpha}_t - \eta_t - \overline{\sigma}_t^2\right)_+}{\overline{\sigma}_t^2(1 - \overline{\alpha}_t - \eta_t)}\\ &=
     \frac{\overline{\alpha}_t\eta_t}{1-\overline{\alpha}_t}\cdot \frac{\left(1 - \overline{\alpha}_t - \eta_t - (1 - \overline{\alpha}_t)\left(1 - \frac{2\eta_t}{1 - \overline{\alpha}_t} + \frac{\eta_t^2}{(1 - \overline{\alpha}_t)^2}\right)\right)_+}{(1 - \overline{\alpha}_t - \eta_t)\overline{\sigma}_t^2}\\
    &= \frac{\overline{\alpha}_t\eta_t}{1-\overline{\alpha}_t}\cdot \frac{\left(\eta_t - \eta_t^2\big/(1 - \overline{\alpha}_t)\right)_+}{(1 - \overline{\alpha}_t - \eta_t)\overline{\sigma}_t^2}
    \le \frac{8\overline{\alpha}_t \eta_t^2}{(1 - \overline{\alpha}_t)^3},
\end{align*}
where the last inequality holds due to $\eta_t \le \frac{1}{2}(1-\overline{\alpha}_t)$ and $\overline{\sigma}_t^2 = (1 - \overline{\alpha}_t)\left(1 - \frac{\eta_t}{1 - \overline{\alpha}_t}\right)^2 \ge (1 - \overline{\alpha}_t)\left(1 - \frac{1}{2}\right)^2 = 4(1 - \overline{\alpha}_t)$. Applying similar arguments, we can also derive
\begin{align*}
    {\overline{\alpha}_t^2 \big(\eta^{(1)}_t\big)^2} = \frac{\overline{\alpha}_t^2\eta_t^2}{(1 - \overline{\alpha}_t)^2(1 - \overline{\alpha}_t - \eta_t)^2} \le \frac{4\overline{\alpha}_t^2\eta_t^2}{(1 - \overline{\alpha}_t)^2(\alpha_t - \overline{\alpha}_t)^2} \le \frac{4C^2 (1 - \alpha_t)^2}{(1 - \overline{\alpha}_t)^2(\alpha_t - \overline{\alpha}_t)^2} = 4C^2 \widetilde{\sigma}_t^4,
\end{align*}
where $\widetilde{\sigma}_t^2 = \frac{\overline{\alpha}_t (1 - \alpha_t)}{(\alpha_t - \overline{\alpha}_t)(1 - \overline{\alpha}_t)}$.

Additionally,  Lemma~\ref{lem:bound_CoV_1|t_F^2} tells us that
\[
\widetilde{\sigma}_t^4\EB\left[\norm{\Cov_{0\mid t}(X_t)}_{\mathrm{F}}^2\right] \le
3\widetilde{\sigma}_t^2 \left\{
    \EB \left[\tr\big(\Cov_{0\mid t}(X_t)\big)\right] - \EB \left[\tr\big(\Cov_{0\mid {t-1}}(X_{t-1})\big)\right]
    \right\} + \frac{3}{T^{10}}.
\]
Plugging the above results into~\eqref{eq:EB-Zx_t_+_eq1}, we complete the proof.

%% file: general_step_size_analysis.tex
\medskip
\noindent {\bf Step 3: determining the exponent of the product $\gG_1(x_t,x_0)\gG_2(x_t,x_0)$.}
In order to control
\eqref{eq:gR_x_t-1_eq1} further, one needs to cope with the product $\gG_1(x_t,x_0) \gG_2(x_t,x_0)$. Towards this end, we find it helpful to introduce the following functions:
\begin{subequations}
    \begin{align}
    \zeta_t(x_t,x_0) &\coloneqq
   \left(\frac{\eta_t}{(1-\overline{\alpha}_t)\overline{\sigma}_t^2} - \frac{\eta_t^2}{2(1 - \overline{\alpha}_t)^2 \overline{\sigma}_t^2}\right)\left\{
   \norm{\sqrt{\overline{\alpha}_t}\mu_{0\mymid  t}(x_t) - x_t}_2^2 - \norm{\sqrt{\overline{\alpha}_t} x_0 - x_t}_2^2
   \right\}\notag\\
&\qquad + \frac{\sqrt{\overline{\alpha}_t}\eta_t}{\overline{\sigma}_t^2(\alpha_t - \overline{\alpha}_t)^2}\big(x_0 - \mu_{0\mymid  t}(x_t)\big)^\top \big(\sqrt{\overline{\alpha}_t}\mu_{0\mymid  t}(x_t) - x_t\big)\notag\\
&\qquad -
 \left(\frac{\overline{\alpha}_t\eta_t}{(1-\overline{\alpha}_t)\overline{\sigma}_t^2} - \frac{\eta_t^2}{2(1 - \overline{\alpha}_t)^2 \overline{\sigma}_t^2}\right)\tr\big(\Cov_{0\mymid  t}(x_t)\big),
\label{defn:zeta-xt-x0-DDPM}  \\
Z_t(x_t)& \coloneqq \log \det\left( I + \frac{\eta_t}{(1 - \overline{\alpha}_t)(1 - \overline{\alpha}_t - \eta_t)}\Cov_{0\mymid  t}(x_t)\right)\notag\\
&\qquad - \left(\frac{\overline{\alpha}_t\eta_t}{(1-\overline{\alpha}_t)\overline{\sigma}_t^2} - \frac{\eta_t^2}{2(1 - \overline{\alpha}_t)^2 \overline{\sigma}_t^2}\right)\tr\big(\Cov_{0\mymid  t}(x_t)\big),
\label{eq:defn-Zt-DDPM}
\end{align}
\end{subequations}
where we recall the definition of $\mu_{0\mymid  t}$ and $\Cov_{0\mymid  t}$ in \eqref{eq:posterior_mu_cov}.  The following lemma asserts that these two functions can be harnessed to represent $\gG_1(x_t,x_0)\gG_2(x_t,x_0)$.
\begin{lemma}\label{lem:bounds-G1-G2-DDPM}
    The quantities $\gG_i(x_t,x_0),~ i=1,2$ defined in~\eqref{eq:G=G1xG2} satisfy
    \[
\gG_1(x_t,x_0)\gG_2(x_t,x_0) = e^{\zeta_t(x_t,x_0)}\cdot e^{Z_t(x_t)}.
    \]
Further, it holds that
\[
\int \zeta_t(x_t,x_0)p_{X_0\mymid  X_t}(x_0\mymid  x_t) \rd x_0 = 0 \qquad \text{ for all } x_t \in \RB^d.
\]
\end{lemma}
In short, Lemma~\ref{lem:bounds-G1-G2-DDPM} determines the exponent of $\gG_1(x_t,x_0)\gG_2(x_t,x_0)$, where one of the two components satisfies $\mathbb{E}[\zeta_t(X_t,X_0)\mymid X_t=x_t]=0$ for an arbitrary $x_t$.
The proof of this lemma is postponed to Appendix~\ref{sec:proof:lem:bounds-G1-G2-DDPM}.

\medskip
\noindent {\bf Step 4: bounding $\gR_{t-1}$ using $Z_t(X_t)$.}
With the above calculations of $\gR_{t-1}(x_{t-1})$ in place, we are now ready to bound $\gR_{t-1}$.
In view of Lemma~\ref{lem:bounds-G1-G2-DDPM}, for any $x_t \in \mathbb{R}^d$ one has
\begin{equation}\label{eq:exp_zeta-1_posi_new}
\int \left(e^{\zeta_t(x_t,x_0)} - 1\right)e^{Z_t(x_t)}p_{X_0\mymid  X_t}(x_0\mymid  x_t)\rd x_0 \ge e^{Z_t(x_t)}\int \zeta_t(x_t,x_0)p_{X_0\mymid  X_t}(x_0\mymid  x_t)\rd x_0 = 0,
\end{equation}
where we have invoked the elementary inequality $e^x -1 \geq x$.
With \eqref{eq:gR_x_t-1_eq1} and \eqref{eq:exp_zeta-1_posi_new} in place, we can show that
\begin{equation}\label{eq:gR_x_t-1_eq2_new}
\begin{aligned}
    \gR_{t-1} &= \int_{\gA_{t-1}}\gR_{t-1}(x_{t-1})\rd x_{t-1} {=} \int_{\gA_{t-1}\times \RB^d\times \gX_{\mathsf{data}}}p_{Y_{t-1}^{\star}\mymid  Y_{t}^{\star}}(x_{t-1}\mymid x_t)p_{X_t,X_0}(x_t,x_0)\big(
    \gG(x_t,x_0) - 1
    \big) \rd x_0 \rd x_t \rd x_{t-1}\\
    &=
    \int_{\gA_{t-1}\times \RB^d\times \gX_{\mathsf{data}}}p_{Y_{t-1}^{\star}\mymid  Y_{t}^{\star}}(x_{t-1}\mymid x_t)p_{X_t,X_0}(x_t,x_0)\big\{
    \gG_1(x_t,x_0)\gG_2(x_t,x_0) - 1
    \big\} \rd x_0 \rd x_t \rd x_{t-1}\\
    &\overset{\text{(a)}}{=}
    \int_{\gA_{t-1}\times \RB^d\times \gX_{\mathsf{data}}}p_{Y_{t-1}^{\star}\mymid  Y_{t}^{\star}}(x_{t-1}\mymid x_t)p_{X_t,X_0}(x_t,x_0)
    \left\{e^{\zeta_t(x_t,x_0) + Z_t(x_t)} - 1\right\}\rd x_0 \rd x_t \rd x_{t-1}\\
    &\le
    \int \PB_{Y_{t-1}^{\star}\mymid  Y_{t}^{\star}}\left(\gA_{t-1}\mymid x_t\right)p_{X_t}(x_t)\left\{
    \int (e^{\zeta_t(x_t,x_0)} {-} 1)e^{Z_t(x_t)}p_{X_0\mymid  X_t}(x_0\mymid  x_t)\rd x_0 {+} \big|e^{Z_t(x_t)} {-} 1 \big|
    \right\}\rd x_t\\
    &\overset{\text{(b)}}{\le}
    \int p_{X_t}(x_t)\left\{
    \int (e^{\zeta_t(x_t,x_0)} {-} 1)e^{Z_t(x_t)}p_{X_0\mymid  X_t}(x_0\mymid  x_t)\rd x_0 {+} \big|e^{Z_t(x_t)} {-} 1\big|
    \right\}\rd x_t\\
    &=
    \int p_{X_t,X_0}(x_t,x_0)e^{\zeta_t(x_t,x_0)+ Z_t(x_t)}\rd x_t \rd x_0 - \int p_{X_t}(x_t)e^{Z_t(x_t)}\rd x_t + \int p_{X_t}(x_t)\big| e^{Z_t(x_t)} - 1\big| \rd x_t.
\end{aligned}
\end{equation}
Here $(a)$ follows from Lemma~\ref{lem:bounds-G1-G2-DDPM}, and $(b)$ holds according to $\PB_{Y_{t-1}^\star\mid Y_t^\star}(\gA_{t-1}\mid x_t)$.
Note that $e^{\zeta_t(x_t,x_0)+Z_t(x_t)} = \gG(x_t,x_0) = \frac{\widetilde{p}_{U_t|X_0}(u_t\mymid x_0)}{p_{X_t\mymid  X_0}(x_t\mymid  x_0)}\det\big(\frac{\rd u_t}{\rd x_t}\big)$, and as a result,
\[
\int p_{X_t,X_0}(x_t,x_0)e^{\zeta_t(x_t,x_0)+ Z_t(x_t)}\rd x_t \rd x_0 = \int \widetilde{p}_{U_t|X_0}(u_t\mymid x_0)p_{X_0}(x_0)\rd u_t \rd x_0 = 1.
\]
Substitution into \eqref{eq:gR_x_t-1_eq2_new} yields
\begin{equation}\label{eq:gR_x_t-1_eq3_new}
\begin{aligned}
    \gR_{t-1} &\le 1 - \int p_{X_t}(x_t)e^{Z_t(x_t)}\rd x_t + \int p_{X_t}(x_t)\big|e^{Z_t(x_t)} - 1\big|\rd x_t\\
    &= 2\int p_{X_t}(x_t)\big(1- e^{Z_t(x_t)}\big)_+\rd x_t
    \le 2\EB \big[\left(-Z_t(X_t)\right)_+\big] ,
\end{aligned}
\end{equation}
where the equality in the last line arises from the elementary inequality $1-z+|1-z|=2(1-z)_+$, and the last inequality holds by combining $1 - e^x \le -x$ and the non-decreasing property of the function $(\cdot)_+$.

%% file: equivalence_to_the_Denoising_Diffusion_Implicit_Model.tex
\section{Equivalence between relation~\eqref{eq:ddpm_gen_stp_choice} and \citet[Eq.~(12)]{song2020denoising}}\label{sec:eqv_to_ddim}

Recall that
$
\epsilon_t^{\mathsf{noise}}(Y_t) = - \sqrt{1 - \overline{\alpha}_t}s_t(Y_t).
$
Substituting this expression into \eqref{eq:reverse_of_original_ddim} (i.e., \citet[Eq.~(12)]{song2020denoising}) results in:
\begin{align*}
Y_{t-1} = \frac{1}{\sqrt{\alpha_t}}\Big(
Y_t + (1 - \overline{\alpha}_t)s_t(Y_t) - \sqrt{(1 - \overline{\alpha}_t)(\alpha_t - \overline{\alpha}_t - \alpha_t\varsigma_t^2)}\, s_t(Y_t) + \sqrt{\alpha_t}\varsigma_t Z_t
\Big).
\end{align*}
By taking
\[
\eta_t^{\mathsf{ddpm}} = (1 - \overline{\alpha}_t) - \sqrt{(1 - \overline{\alpha}_t)(\alpha_t - \overline{\alpha}_t - \alpha_t \varsigma_t^2)}
\qquad \text{and} \qquad  \sigma_t^{\mathsf{ddpm}} = \sqrt{\alpha_t}\varsigma_t,
\]
we can derive
\begin{align*}
    (1 - \overline{\alpha}_t)\left(1 - \frac{\eta_t^{\mathsf{ddpm}}}{1 - \overline{\alpha}_t}\right)^2 &= (1 - \overline{\alpha}_t)\Bigg(1 - \frac{(1 - \overline{\alpha}_t) - \sqrt{(1 - \overline{\alpha}_t)\big(\alpha_t - \overline{\alpha}_t - \big(\sigma_t^{\mathsf{ddpm}}\big)^2\big)}}{1 - \overline{\alpha}_t}\Bigg)^2\\
    &= (1 - \overline{\alpha}_t)\left(\sqrt{\frac{\alpha_t - \overline{\alpha}_t - \big(\sigma_t^{\mathsf{ddpm}}\big)^2}{1 - \overline{\alpha}_t}}\right)^2 = \alpha_t - \overline{\alpha}_t - \big(\sigma_t^{\mathsf{ddpm}}\big)^2,
\end{align*}

which is precisely the relation in \eqref{eq:ddpm_gen_stp_choice}.

%% file: proof_propositions.tex
\section{Proofs about reverse-time differential equations}\label{sec:prop-prf}
\subsection{Generalized reverse-time differential equations}\label{sec:gen-reverse}
We formally state the time-reversal property of the generalized reverse-time differential equation introduced in Section~\ref{sec:pre}.
\begin{proposition}\label{prop:gen-reverse}
    Suppose the generalized reverse-time differential equation
    \begin{align}
        \rd Y_t = \left(Y_t + \big(1 + \xi(T-t)\big) s^*_{T-t}(Y_t)\right)\beta(T-t) \rd t + \sqrt{2\xi(T-t)\beta(T-t)} \, \rd W_t,
    \qquad t \in [0,T]
    \label{eq:defn-Yt-generalized-proof}
    \end{align}
    has a unique strong solution, where $(W_t)$ represents a standard Brownian motion in $\mathbb{R}^d$.
    Then under the boundary condition $Y_0 \overset{\mathrm{d}}= X_T$, it satisfies  $Y_{T-t} \overset{\mathrm{d}}{=} X_t$ for all $0 \le t \le T$.
\end{proposition}

\medskip
\noindent {\bf Proof of Proposition~\ref{prop:gen-reverse}.}
Recall that the continuous-time forward process is given by
\begin{equation*}
\rd X_t = -\beta(t) X_t \rd t + \sqrt{2\beta(t)} \,\rd B_t,
\end{equation*}
with $(B_t)$ a standard Brownian motion in $\mathbb{R}^d$.
Denote by $p_X(x,t)$ the probability density of $X_t$ at point $x$ w.r.t.~the Lebesgue measure in $\mathbb{R}^d$.
In the following proof, we use $\nabla$ (resp.~$\nabla \cdot$) to be the gradient (resp.~divergence) operator taken w.r.t.~the first argument (i.e., $x$) of the function, and denote by $\Delta$ the corresponding Laplace operator.
The Fokker-Planck equation then tells us that
\begin{align}
    \frac{\partial}{\partial t} p_X(x,t)
    &= \nabla\cdot \big(x\beta(t) p_X(x,t)\big) + \frac{1}{2}\Delta \big(2\beta(t) p_X(x,t)\big)\notag\\
    &= \beta(t)\nabla\cdot\big(x p_X(x,t)\big) + \beta(t)\Delta \big( p_X(x,t) \big).\label{eq:kol-forward-x}
\end{align}
Similarly, denoting by $p_Y(x,t)$ the probability density of $Y_{t}$ at point $x$ w.r.t.~the Lebesgue measure in $\mathbb{R}^d$, then we can
apply the Fokker-Planck equation once again to obtain
\begin{align}
    \frac{\partial}{\partial t} p_Y(x,t)
    &= -\nabla\cdot\bigg(\Big(x + \big(1+\xi(T-t)\big) s^*_{T-t}(x)\Big)\beta(T-t) p_Y(x,t)\bigg)\notag\\
    &\qquad + \frac{1}{2}\Delta \big(2\xi(T-t)\beta(T-t) p_Y(x,t)\big)\notag\\
    &= -\Big\langle x + \big(1+\xi(T-t)\big) s^*_{T-t}(x), \, \beta(T-t) \nabla p_Y(x,t)\Big\rangle \notag\\
    &\hspace{2em}- \mathrm{tr}\left(I_d + \big(1+\xi(T-t)\big) \nabla s^*_{T-t}(x)\right)\beta(T-t) p_Y(x,t)\notag
    \\ &\qquad+ \xi(T-t)\beta(T-t) \Delta \big(p_Y(x,t)\big).\label{eq:kol-forward-y}
\end{align}

Recall that our goal is to show that $X_t$ and $Y_{T-t}$ have the same marginal distributions, i.e., $p_X(x,t) = p_Y(x,T-t)$, or equivalently, $p_X(x,T-t) = p_Y(x,t)$.
Since the generalized differential equation \eqref{eq:defn-Yt-generalized-proof} is assumed to have a unique strong solution, the induced Fokker-Planck equation has a unique strong solution. From the assumption $Y_0 \overset{\mathrm{d}}{=} X_T$, we know that $p_X(x,T) = p_Y(x,0)$, and hence it suffices to show that $p_X(x,T-t)$ is a solution of the partial differential equation (PDE)~\eqref{eq:kol-forward-y}.
It is readily seen from PDE~\eqref{eq:kol-forward-x} that
\begin{align}\label{eq:kol-forward-x-reverse}
    \frac{\partial}{\partial t} p_X(x,T-t) = -\beta(T-t)\nabla\cdot\big(x p_X(x,T-t)\big)
    - \beta(T-t)\Delta \big(p_X(x,T-t)\big).
\end{align}
Replacing $p_Y(x,t)$ with $p_X(x,T-t)$ on the right-hand side of PDE~\eqref{eq:kol-forward-y}, one can derive
\begin{align}
    &-\Big\langle x + \big(1+\xi(T-t)\big) s^*_{T-t}(x), \, \beta(T-t)  \nabla p_X(x,T-t) \Big\rangle-  \mathrm{tr}(I_d) \beta(T-t) p_X(x,T-t) \notag\\
    &\hspace{5em}- \big(1+\xi(T-t)\big) \beta(T-t) \Delta \big(\log(p_X(x,T-t))\big) p_X(x,T-t)\notag\\
    &\hspace{5em}+ \xi(T-t)\beta(T-t)\Delta \big(p_X(x,T-t)\big)\notag\\
    &=-\beta(T-t)\bigg\langle x + \big(1+\xi(T-t)\big) \frac{\nabla p_X(x,T-t)}{p_X(x,T-t)}, \nabla p_X(x,T-t)\bigg\rangle - d \beta(T-t) p_X(x,T-t) \notag\\
    &\hspace{2em}+ \big(1+\xi(T-t)\big) \beta(T-t) \frac{\|\nabla p_X(x,T-t)\|_2^2}{p_X(x,T-t)}\notag\\
    &\hspace{2em}+\Big(-\big(1+\xi(T-t)\big)+\xi(T-t)\Big) \beta(T-t)\Delta \big(p_X(x,T-t)\big) \notag\\
    &= -\beta(T-t)\big\langle x, \nabla p_X(x,T-t)\big\rangle - d\beta(T-t) p_X(x,T-t) - \beta(T-t) \Delta \big(p_X(x,T-t)\big) \notag \\
    &=-\beta(T-t)\nabla\cdot\big(x p_X(x,T-t)\big)
    - \beta(T-t)\Delta \big(p_X(x,T-t)\big) \notag\\
    &= \frac{\partial}{\partial t} p_X(x,T-t),\label{eq:kol-forward-y-rhs}
\end{align}
where we invoke PDE~\eqref{eq:kol-forward-x-reverse} in the last line. Eqn.~\eqref{eq:kol-forward-y-rhs} reveals that $p_X(x,T-t)$ is a strong solution of PDE~\eqref{eq:kol-forward-y}, which is equivalent to $p_X(x,T-t) = p_Y(x,t)$.

\subsection{Proof of Proposition~\ref{prop:sde}}\label{sec:prf_of_prop_sde}
We prove this result by explicitly solving SDE~\eqref{eq:gen-dynamic} when $t\in[t_n, t_{n+1})$.
To begin with, we make the observation that: under the time transformation
\begin{equation}\label{eq:t-trans}
    t ~\rightarrow~ t' = \int_{0}^t \beta(s) \rd s,
\end{equation}
SDE~\eqref{eq:gen-dynamic} can be rewritten as
\begin{align*}
    \rd \widetilde{Y}_{t'} &= \left(-\frac{\xi(T-t_n')+\overline{\alpha}_{T-t'}}{1-\overline{\alpha}_{T-t'}}\widetilde{Y}_{t'} + \frac{\left(1+\xi(T-t_n')\right)\sqrt{\overline{\alpha}_{T-t'}}}{1-\overline{\alpha}_{T-t'}}\mu_{T-t'_n}(\widetilde{Y}_{t'_n})\right)\rd t'  + \sqrt{2\xi(T-t_n')}\,\rd W_{t'}
\end{align*}
for $t \in [t_n', t_{n+1}')$, where $t_n'$ and $t_{n+1}'$ are the images of $t_n$ and $t_{n+1}$ under the transformation~\eqref{eq:t-trans}. Note that this transformed SDE has the same form as SDE~\eqref{eq:gen-dynamic} when $\beta(t) = 1$ for $t \in [t_n', t_{n+1}')$.
Thus, without loss of generality, it suffices to assume $\beta(t) = 1$ for all $t \in [0,T]$ and solve SDE~\eqref{eq:gen-dynamic}. Under this assumption, we can simplify
\begin{align}
    \overline{\alpha}_t = \exp\left(-2\int_{0}^t \beta(s) \rd s\right) = e^{-2t}.
    \label{eq:alphat-bar-simplified}
\end{align}

Recall that SDE~\eqref{eq:gen-dynamic} with $\xi(T - t_n) = \xi > 0$ and $\beta(t)=1$ can be written as
\begin{align}\label{eq:sde-sim}
    &\rd \widetilde{Y}_t = \left(-\frac{\xi+\overline{\alpha}_{T-t}}{1-\overline{\alpha}_{T-t}}\widetilde{Y}_t + \frac{\left(1+\xi\right)\sqrt{\overline{\alpha}_{T-t}}}{1-\overline{\alpha}_{T-t}}\mu_{T-t_n}(\widetilde{Y}_{t_n})\right)\rd t + \sqrt{2\xi}\,\rd W_t.
\end{align}
To solve SDE~\eqref{eq:sde-sim}, we find it convenient to introduce the following function
\begin{align*}
    f(t) = \frac{e^{-\xi(T-t)}}{(1-e^{-2(T-t)})^{\frac{1+\xi}{2}}}.
\end{align*}
It follows from It\^{o}'s formula that
\begin{align*}
    \rd\left(f(t)\widetilde{Y}_t\right)
    &= \Bigg(\left(f'(t)-\frac{\xi+\overline{\alpha}_{T-t}}{1-\overline{\alpha}_{T-t}}f(t)\right)\widetilde{Y}_t + \frac{\left(1+\xi\right)\sqrt{\overline{\alpha}_{T-t}}}{1-\overline{\alpha}_{T-t}}f(t)\mu_{T-t_n}(\widetilde{Y}_{t_n})\Bigg)\rd t + \sqrt{2\xi}f(t)\,\rd W_t \\
    &= \frac{\left(1+\xi\right)\sqrt{\overline{\alpha}_{T-t}}}{1-\overline{\alpha}_{T-t}}f(t)\mu_{T-t_n}(\widetilde{Y}_{t_n})\rd t + \sqrt{2\xi}f(t)\,\rd W_t.
\end{align*}
Integrating both sides of the above display from $t_n$ to $t_{n+1}$, we obtain
\begin{align*}
    f(t_{n+1})\widetilde{Y}_{t_{n+1}} - f(t_n)\widetilde{Y}_{t_n} = \int_{t_n}^{t_{n+1}}\frac{\left(1+\xi\right)\sqrt{\overline{\alpha}_{T-t}}}{1-\overline{\alpha}_{T-t}}f(t)\mu_{T-t_n}(\widetilde{Y}_{t_n})\rd t + \int_{t_n}^{t_{n+1}}\sqrt{2\xi}f(t)\,\rd W_t.
\end{align*}
From It\^{o}'s isometry property of the Brownian motion, we can write, for each $0 \le n \le T-1$,
\begin{equation}
\int_{t_n}^{t_{n+1}}\sqrt{2\xi}f(t)\,\rd W_t
= \left(\int_{t_n}^{t_{n+1}} 2\xi\big(f(t)\big)^2 \rd t\right)^{1/2} \widetilde{Z}_{n}\notag
\end{equation}
for some Gaussian vector $\widetilde{Z}_{n} \sim \mathcal{N}(0,I_d)$,
where $\{\widetilde{Z}_{n}\}_{n=0,\dots,T-1}$ are statistically independent.
Consequently,
\begin{align*}
    f(t_{n+1})\widetilde{Y}_{t_{n+1}} = f(t_n)\widetilde{Y}_{t_n} + \underbrace{\int_{t_n}^{t_{n+1}}\frac{\left(1+\xi\right)\sqrt{\overline{\alpha}_{T-t}}}{1-\overline{\alpha}_{T-t}}f(t)\rd t}_{\eqqcolon \, A_n} \cdot \mu_{T-t_n}(\widetilde{Y}_{t_n}) + \bigg(\underbrace{\int_{t_n}^{t_{n+1}} 2\xi\big(f(t)\big)^2 \rd t}_{\eqqcolon \, B_n}\bigg)^{1/2} \cdot \widetilde{Z}_{n}.
\end{align*}
Taking this together with the definition \eqref{eq:defn-mut} of $\mu_t$, we can express the update rule induced by SDE~\eqref{eq:sde-sim} as
\begin{align}\label{eq:update_a_b}
    \widetilde{Y}_{t_{n+1}} = \frac{f(t_n) + A_n/\sqrt{\overline{\alpha}_{T-t_n}}}{f(t_{n+1})}\widetilde{Y}_{t_n} + \frac{1-\overline{\alpha}_{T-t_n}}{\sqrt{\overline{\alpha}_{T-t_n}}} \cdot \frac{A_n}{f(t_{n+1})}s_{T-t_n}(\widetilde{Y}_{t_n}) + \frac{B_n}{f(t_{n+1})}\widetilde{Z}_{n}.
\end{align}

To simplify the notation, we define $$\gamma_n = e^{-(T-t_n)}.$$
The terms $A_n$ and $B_n$ can be explicitly calculated as follows:
\begin{align}
    A_n &= \int_{t_n}^{t_{n+1}}\frac{\left(1+\xi\right)\sqrt{\overline{\alpha}_{T-t}}}{1-\overline{\alpha}_{T-t}}f(t)\rd t \notag\\
    &= \int_{t_n}^{t_{n+1}}\left(1+\xi\right)\frac{e^{-(1+\xi)(T-t)}}{\left(1-e^{-2(T-t)}\right)^{(3+\xi)/2}}\,\rd t\notag\\
    &= \frac{e^{-(1+\xi)(T-t)}}{\left(1-e^{-2(T-t)}\right)^{(1+\xi)/2}}\, \Bigg|_{t_n}^{t_{n+1}}
    =
    \frac{\gamma_{n+1}^{\xi + 1}}{(1-\gamma_{n+1}^2)^{\frac{1+\xi}{2}}} - \frac{\gamma_{n}^{\xi + 1}}{(1-\gamma_{n}^2)^{\frac{1+\xi}{2}}},\label{eq:A_n}
\end{align}
where we have applied \eqref{eq:alphat-bar-simplified}.
Through similar calculation, we can reach
\begin{align}
    B_n & = \int_{t_n}^{t_{n+1}} 2\xi\big(f(t)\big)^2 \rd t\notag\\
    &= \int_{t_n}^{t_{n+1}} 2\xi\frac{e^{-2\xi(T-t)}}{\left(1-e^{-2(T-t)}\right)^{1+\xi}}\, \rd t\notag\\
    &= \frac{e^{-2\xi(T-t)}}{\left(1-e^{-2(T-t)}\right)^{\xi}}\, \Bigg|_{t_n}^{t_{n+1}}
    = \frac{\gamma_{n+1}^{2\xi}}{(1-\gamma_{n+1}^2)^{\xi}} - \frac{\gamma_{n}^{2\xi}}{(1-\gamma_{n}^2)^{\xi}}.\label{eq:B_n}
\end{align}
Substituting \eqref{eq:A_n} and \eqref{eq:B_n} into \eqref{eq:update_a_b} and comparing the coefficients with the DDPM update rule~\eqref{eq:DDPM-update}, we obtain
\begin{align}
    \alpha_{t_n} = \frac{f(t_n) + A_n/\sqrt{\overline{\alpha}_{T-t_n}}}{f(t_{n+1})} = \left(\frac{\gamma_n}{\gamma_{n+1}}\right)^2 = e^{-2(t_{n+1} - t_n)},
\end{align}
which coincides with our choice of $\overline{\alpha}_t$, i.e., $$\alpha_{t_n} = e^{-2(t_{n+1} - t_n)} = \frac{\overline{\alpha}_{t_{n+1}}}{\overline{\alpha}_{t_{n}}}.$$
Additionally, we can easily verify that
\begin{align*}
    \eta_{t_n}^{\mathsf{ddpm}}
    &= \sqrt{\alpha_{t_n}} \cdot \frac{1-\gamma_n^2}{\gamma_{n}} \cdot \frac{A_n}{f(t_{n+1})} = \frac{1-\gamma_n^2}{\gamma_{n+1}} \cdot \frac{1}{f(t_{n+1})} \cdot \left(\frac{\gamma_{n+1}^{\xi + 1}}{(1-\gamma_{n+1}^2)^{\frac{1+\xi}{2}}} - \frac{\gamma_{n}^{\xi + 1}}{(1-\gamma_{n}^2)^{\frac{1+\xi}{2}}}\right),\\
    \sigma_{t_n}^{\mathsf{ddpm}}
    &= \sqrt{\alpha_{t_n}} \cdot\frac{B_n^{1/2}}{f(t_{n+1})}
    = \frac{\gamma_n}{\gamma_{n+1}f(t_{n+1})}\cdot\left(\frac{\gamma_{n+1}^{2\xi}}{(1-\gamma_{n+1}^2)^{\xi}} - \frac{\gamma_{n}^{2\xi}}{(1-\gamma_{n}^2)^{\xi}}\right)^{1/2}.
\end{align*}

We are now ready to show that the relation~\eqref{eq:ddpm_gen_stp_choice}
\begin{equation*}
(1 - \overline{\alpha}_{t_n})\bigg(1 - \frac{\eta_{t_n}^{\mathsf{ddpm}}}{1 - \overline{\alpha}_{t_n}}\bigg)^2 = \alpha_{t_n} - \overline{\alpha}_{t_n} - \big(\sigma_{t_n}^{\mathsf{ddpm}}\big)^2
\end{equation*}
is satisfied by this solution for all $n$.
Towards this end, calculate the left-hand side above as:
\begin{align*}
    (1 - \overline{\alpha}_{t_n})\bigg(1 - \frac{\eta_{t_n}^{\mathsf{ddpm}}}{1 - \overline{\alpha}_{t_n}}\bigg)^2
    &= (1-\gamma_n^2)\left(1-\frac{1}{\gamma_{n+1}f(t_{n+1})} \cdot \left(\frac{\gamma_{n+1}^{\xi + 1}}{(1-\gamma_{n+1}^2)^{\frac{1+\xi}{2}}} - \frac{\gamma_{n}^{\xi + 1}}{(1-\gamma_{n}^2)^{\frac{1+\xi}{2}}}\right)\right)^2 \\
    &= (1-\gamma_n^2)\left(\frac{\gamma_n^{\xi+1}}{\gamma_{n+1}^{\xi+1}} \cdot \frac{(1-\gamma_{n+1}^2)^{\frac{1+\xi}{2}}}{(1-\gamma_{n}^2)^{\frac{1+\xi}{2}}}\right)^2 \\
    &= \frac{\gamma_n^{2(\xi+1)}}{\gamma_{n+1}^{2(\xi+1)}} \cdot \frac{(1-\gamma_{n+1}^2)^{1+\xi}}{(1-\gamma_{n}^2)^{\xi}}\\
    &= \frac{\gamma_n^2}{\gamma_{n+1}^2} - \gamma_n^2 - (1-\gamma_{n+1}^2)\left(\frac{\gamma_n^2}{\gamma_{n+1}^2} - \frac{\gamma_n^{2(\xi+1)}}{\gamma_{n+1}^{2(\xi+1)}} \cdot \frac{(1-\gamma_{n+1}^2)^{\xi}}{(1-\gamma_{n}^2)^{\xi}}\right)\\
    &= \alpha_{t_n} - \overline{\alpha}_{t_n} - \big(\sigma_{t_n}^{\mathsf{ddpm}}\big)^2.
\end{align*}
Thus, setting $t_n$ = $T-n$ for $n = 0, 1, \ldots, T$ exactly recovers the relation~\eqref{eq:ddpm_gen_stp_choice}.

\subsection{Proof of Proposition~\ref{prop:ode}}\label{sec:prf-ode}
Proposition~\ref{prop:ode} can be regarded as a corollary of Proposition~\ref{prop:sde} in the following sense: if we set $\xi(T-t_n) = 0$ for all $n = 0, 1, \ldots, T-1$, then  SDE~\eqref{eq:gen-dynamic} degenerates to ODE~\eqref{eq:ddim-dynamic}. In addition, the whole proof of Proposition~\ref{prop:sde} in Appendix~\ref{sec:prf_of_prop_sde} works for $\xi(T-t_n) = 0$. Thus, the proof of Proposition~\ref{prop:ode} can be directly completed by repeating the proof arguments in Appendix~\ref{sec:prf_of_prop_sde}.

%% file: proof_lower_bound.tex
\section{Proof of the lower bound in Theorem~\ref{thm:lower_bound}}
\label{sec:proof:thm:lower-bound}
Let $X_0 \sim \gN\left(0, {\footnotesize\left[\begin{array}{cc}
I_k\\
 & 0
\end{array}\right]}\right)$, then it follows from \eqref{eq:forward-marginal} that
\begin{align}
\label{eq:Xt-distribution-Gauss}
X_t = \sqrt{\overline{\alpha}_t}X_0 + \sqrt{1 - \overline{\alpha}_t}\, \overline{W}_t \sim 
\gN\left(0, \left[\begin{array}{cc}
I_k\\
 & (1-\overline{\alpha}_t)I_{d-k}
\end{array}\right] \right).
%
\end{align}
It is then easily seen that
\begin{align*}
s_t^*(x) = - \left[\begin{array}{cc}
I_k\\
 & (1-\overline{\alpha}_t)I_{d-k}
\end{array}\right]^{-1}x
= 
\left[\begin{array}{cc}
I_k\\
 & \frac{1}{1-\overline{\alpha}_t}I_{d-k}
\end{array}\right] x.
\end{align*}
As a result, the mapping $\Phi_t^*$ admits a closed-form expression as follows
\[
\Phi_t^*(x,z) = \frac{1}{\sqrt{\alpha}_t}\big(x + \eta_t s_t^*(x) + \sigma_t z \big) =  \frac{1}{\sqrt{\alpha}_t}(A_{\eta_t} x + \sigma_t z)
\]
where 
\begin{align*} 
A_{\eta_t} \coloneqq 
\left[\begin{array}{cc}
(1 - \eta_t)I_k\\
 & \big( 1-\frac{\eta_t}{1-\overline{\alpha}_t} \big) I_{d-k}
\end{array}\right].
\end{align*}
These properties taken together further imply that
\begin{align*}
\Phi_t^*(X_t,Z_t) \sim \gN\left(0,\, \frac{1}{\alpha_t}A_{\eta_t}\left[\begin{array}{cc}
I_k\\
 & (1-\overline{\alpha}_t)I_{d-k}
\end{array}\right]A_{\eta_t} + \frac{\sigma_t^2}{\alpha_t} I_d \right),
\end{align*}
or equivalently, 
\begin{align}
\label{eq:Phit-distribution-Gauss}
\Phi_t^*(X_t,Z_t) \sim \gN\left(0,\, \left[\begin{array}{cc}
\frac{(1-\eta_{t})^{2}}{\alpha_{t}}I_{k}\\
 & \frac{1-\overline{\alpha}_{t}}{\alpha_{t}}\big(1-\frac{\eta_{t}}{1-\overline{\alpha}_{t}}\big)^{2}I_{d-k}
\end{array}\right] + \frac{\sigma_t^2}{\alpha_t} I_d \right).
\end{align}

Armed with the above basic properties, 
we can proceed to derive the advertised lower bound. Towards this end, we resort to the following result concerning the TV distance between two multivariate Gaussians with the same mean, whose proof can be found in \citet{devroye2018total}. 
\begin{lemma}[TV distance between Gaussians with the same mean]\label{lem:tv_N1_N2}
    Consider any $\mu\in \RB^d$, and any positive semidefinite matrices $\Sigma_1,\Sigma_2 \in \mathbb{R}^{d\times d}$. 
    Then it holds that
    \[
    \frac{1}{100} < \frac{\TV\big(\gN(\mu,\Sigma_1),~ \gN(\mu,\Sigma_2)\big)}{\min\left\{1,~ \norm{\Sigma_1^{-1}\Sigma_2 - 
    I}_{\mathrm{F}}\right\}} \le \frac{3}{2}.
    \]
\end{lemma}

Recall from \eqref{eq:Xt-distribution-Gauss} that
\begin{align*}
X_{t-1} \sim \gN\left(0, \left[\begin{array}{cc}
I_k\\
 & (1-\overline{\alpha}_{t-1})I_{d-k}
\end{array}\right]\right). 
\end{align*}
%
With this and \eqref{eq:Phit-distribution-Gauss} in mind, we take
\begin{align*}
\Sigma_1 = \left[\begin{array}{cc}
I_k\\
 & (1-\overline{\alpha}_{t-1})I_{d-k}
\end{array}\right], ~~
\Sigma_2 = \left[\begin{array}{cc}
\frac{(1-\eta_{t})^{2}}{\alpha_{t}}I_{k}\\
 & \frac{1-\overline{\alpha}_{t}}{\alpha_{t}}\big(1-\frac{\eta_{t}}{1-\overline{\alpha}_{t}}\big)^{2}I_{d-k}
\end{array}\right] + \frac{\sigma_t^2}{\alpha_t} I_d,
\end{align*}
which satisfy
\[
\Sigma_1^{-1} \Sigma_2 = 
\left[\begin{array}{cc}
\frac{(1-\eta_{t})^{2}+\sigma_{t}^{2}}{\alpha_{t}}I_{k}\\
 & \left(\frac{1-\overline{\alpha}_{t}}{\alpha_{t}-\overline{\alpha}_{t}}\left(1-\frac{\eta_{t}}{1-\overline{\alpha}_{t}}\right)^{2}+\frac{\sigma_{t}^{2}}{\alpha_{t}-\overline{\alpha}_{t}}\right)I_{d-k}
\end{array}\right]. 
%
\]
Invoke Lemma~\ref{lem:tv_N1_N2} to arrive at the following lower bound:
\begin{align*}
    \TV&\big(X_{t-1},~\Phi_t^*(X_t)\big) \ge \frac{1}{100}\min\left\{1,~ \norm{\Sigma_1^{-1}\Sigma_2 - 
    I}_{\mathrm{F}}\right\}\\
    &= \frac{1}{100}\min\left\{1,~ \sqrt{k\left(\frac{(1-\eta_t)^2 + \sigma_t^2}{\alpha_t} - 1\right)^2 + (d-k)\left(
    \frac{1 - \overline{\alpha}_t}{\alpha_t - \overline{\alpha}_t}\left(1 - \frac{\eta_t}{1 - \overline{\alpha}_t}\right)^2 + \frac{\sigma_t^2}{\alpha_t - \overline{\alpha}_t} - 1
    \right)^2}\right\} \notag\\
    &\geq 
     \frac{1}{100}\min\left\{1,~ \sqrt{ \frac{d}{2}\left(
    \frac{1 - \overline{\alpha}_t}{\alpha_t - \overline{\alpha}_t}\left(1 - \frac{\eta_t}{1 - \overline{\alpha}_t}\right)^2 + \frac{\sigma_t^2}{\alpha_t - \overline{\alpha}_t} - 1
    \right)^2}\right\},
\end{align*}
where the last line follows from our assumption that $d\geq 2k$. This concludes the proof.

%% file: Auxiliary_lemmas.tex
\section{Auxiliary lemmas and related proofs}\label{sec:aux_lem&prf}
\subsection{Proof of Lemma~\ref{lem:gT_alpha_typical}}\label{sec:prf_of_gT_alpha_typical}
    Recall that \( V_\alpha = \sqrt{\alpha}V_1 + \sqrt{1 - \alpha}Z \), where \( V_1 \sim \pdata \) and \( Z \sim \mathcal{N}(0, I_d) \). From this, we can derive
    \begin{align*}
        \PB(V_\alpha \notin \gT_\alpha) &= \PB\big(\sqrt{\alpha}V_1 + \sqrt{1 - \alpha}Z \notin \gT_\alpha\big)
        \le \PB\left(\bigg\{V_1 \notin \bigcup\limits_{i\in \gI}\gB_i
        \bigg\}\cup \{Z\notin \gG\}\right)\\
        &\le \sum\limits_{j\in [N_{\epsilon_0}]\backslash \gI}\PB(V_1 \in \gB_j) + \PB(Z\notin \gG).
    \end{align*}
    In view of the definition \eqref{eq:defn-I-set} of $\gI$, we know that for each $j\notin \gI$, $$\PB(V_1 \in \gB_j)\le \exp\{-C_1k\log(R_0/\epsilon_0)\},$$
    with $C_1>0$ a universal constant.
    Taking this with Assumption~\ref{ass:low_dim} yields
    \begin{align*}
        \sum\limits_{j\in [N_{\epsilon_0}]\backslash \gI}\PB(V_1 \in \gB_j) &\le N_{\epsilon_0}\exp\{-C_1k\log(R_0/\epsilon_0)\}\\
        &\le \exp\big\{
        C_{\mathsf{cover}}k\log(R_0/\epsilon_0) - C_1k\log(R_0/\epsilon_0)
        \big\} \le \exp\left\{
        -\frac{3}{8}C_1k\log(R_0/\epsilon_0)
        \right\},
    \end{align*}
    where the last inequality holds as long as $C_1 \ge 16C_{\mathsf{cover}}$.

    In addition, we can establish an upper bound on $\PB(Z\notin \gG)$ using the definition of $\gG$ as follows:
    \begin{align}
        \PB(Z\notin \gG) &\le \PB\left(
        \norm{Z}_2 > 2\sqrt{d} + \sqrt{C_1k\log(R_0/\epsilon_0)}
        \right) \notag\\
        &\qquad + \PB\left(
        \exists~1\le i,j \le N_{\epsilon_0} \text{ s.t. }
        \abs{(x_i^* - x_j^{\star})^\top Z} > \sqrt{C_1 k\log(R_0/\epsilon_0)}\norm{x_i^* - x_j^{\star}}_2
        \right),
        \label{eq:P-W-not-in-G}
    \end{align}
    leaving us with two terms to control.
    \begin{itemize}
    \item
    By virtue of the concentration property of $\chi^2$ random variables (e.g.,  \citet[Lemma 1]{laurent2000adaptive}), we find that the first term on the right-hand side of \eqref{eq:P-W-not-in-G} satisfies
    \begin{align}\label{eq:concentration-Z-Gauss}
    \PB\left(\norm{Z}_2 > 2\sqrt{d} + \sqrt{C_1k\log(R_0/\epsilon_0)}\right)
    \le \exp\left\{-\frac{C_1}{2}k\log(R_0/\epsilon_0)\right\}.
    \end{align}

    \item
    When it comes to the second term on the right-hand side of \eqref{eq:P-W-not-in-G}, we observe that: for every pair of fixed points  $x_i^*, x_j^{\star}$, one has $\frac{(x_i^* - x_j^{\star})^\top}{\|x_i^* - x_j^{\star}\|_2}Z \sim \gN(0,1)$. Thus, it follows from the concentration property of standard Gaussians that
    \begin{align*}
        \PB\left(\abs{(x_i^* - x_j^{\star})^\top Z} > \sqrt{C_1k\log(R_0/\epsilon_0)}\norm{x_i^* - x_j^{\star}}_2\right)
        &=
        \PB\left(\abs{\frac{(x_i^* - x_j^{\star})^\top}{\norm{x_i^* - x_j^{\star}}_2}Z} > \sqrt{C_1 k \log(R_0/\epsilon_0)}\right)\\ &\le
        \exp\left\{-\frac{C_1}{2}k\log(R_0/\epsilon_0)\right\}.
    \end{align*}
    Combining this with the union-bound and Assumption~\ref{ass:low_dim}, we can obtain
    \begin{align*}
        \PB&\left(
        \exists~1\le i,j \le N_{\epsilon_0} \text{ s.t. }
        \abs{(x_i^* - x_j^{\star})^\top Z} > \sqrt{C_1 k\log(R_0/\epsilon_0)}\norm{x_i^* - x_j^{\star}}_2
        \right)\\
        &\le
        \sum\limits_{1\le i,j\le N_{\epsilon_0}} \PB\left(\abs{(x_i^* - x_j^{\star})^\top Z} > \sqrt{C_1k\log(R_0/\epsilon_0)}\norm{x_i^* - x_j^{\star}}_2\right) \le
        \sum\limits_{1\le i,j\le N_{\epsilon_0}} \exp\left\{-\frac{C_1}{2}k\log(R_0/\epsilon_0)\right\}\\
        &\le N_{\epsilon_0}^2\exp\left\{-\frac{C_1}{2}k\log(R_0/\epsilon_0)\right\} \le \exp\big\{\left(2C_{\mathsf{cover}} - C_1/2\right)k\log(R_0/\epsilon_0)\big\} \le
        \exp\left\{-\frac{3}{8}C_1 k\log(R_0/\epsilon_0)\right\},
    \end{align*}
    where the last inequality holds provided that $C_1 \ge 16C_{\mathsf{cover}}$.
    Consequently, it holds that
    \begin{align*}
        \PB(Z\notin \gG) \le \exp\left\{-\frac{C_1}{2}k\log(R_0/\epsilon_0)\right\} + \exp\left\{-\frac{3}{8}C_1k\log(R_0/\epsilon_0)\right\}
        \le 2\exp\left\{-\frac{3}{8}C_1 k\log(R_0/\epsilon_0)\right\}.
    \end{align*}
     \end{itemize}
    Taking the preceding bounds together leads to
    \begin{align*}
        \PB(V_\alpha\notin \gT_\alpha) &\le
        \sum\limits_{j\in [N_{\epsilon_0}]\backslash \gI}\PB(V_1 \in \gB_j) + \PB(Z\notin \gG)\\
        &\le 3\exp\left\{-\frac{3}{8}C_1 k\log(R_0/\epsilon_0)\right\}
        \le \exp\left\{-\frac{1}{4}C_1 k\log(R_0/\epsilon_0)\right\}.
    \end{align*}


\subsection{Proof of Lemma~\ref{lem:posterior_norm}}\label{sec:prf_of_posterior_norm}
    Define the following set:
    $$\gE_{\alpha,C}(v) \coloneqq \left\{v_1 \,\Big|\,  \sqrt{{\alpha}}\,\big\| v_1 - x_{i(v)}^{\star} \big\| \ge \sqrt{Ck(1-{\alpha})\log (R_0/\epsilon_0)}\right\}.$$
    Invoke the Bayes rule to obtain
    \begin{equation}\label{eq:posterior_norm_1}
    \begin{aligned}
    \PB\big(\gE_{\alpha, C}(v) \mymid  V_\alpha = v\big)& = \frac{\int_{\gE_{\alpha, C}(v)}p_{X_0}(v_1)p_{V_\alpha\mymid  V_1}(v\mymid  v_1)\rd v_1}{\int p_{X_0}(\widetilde{v}_1)p_{V_\alpha\mymid  V_1}(v|\widetilde{v}_1)\rd \widetilde{v}_1} \le
    \frac{\int_{\gE_{\alpha, C}(v)}p_{X_0}(v_1)p_{V_\alpha\mymid  V_1}(v\mymid  v_1)\rd v_1}{\int_{\widetilde{v}_1 \in \gB_{{i(v)}}} p_{X_0}(\widetilde{v}_1)p_{V_\alpha\mymid  V_1}(v \mymid  \widetilde{v}_1)\rd \widetilde{v}_1}\\
    &\le \frac{\int_{\gE_{\alpha, C}(v)}p_{X_0}(v_1)p_{V_\alpha\mymid  V_1}(v\mymid  v_1)\rd v_1}{\PB(\gB_{{i(v)}})\inf_{\widetilde{v}_1 \in \gB_{{i(v)}}}p_{V_\alpha\mymid  V_1}(v \mymid  \widetilde{v}_1)}
    \le \frac{1}{\PB(\gB_{{i(v)}})}\cdot\frac{\sup_{v_1 \in \gE_{\alpha, C}(v)}p_{V_\alpha\mymid  V_1}(v\mymid  v_1)}{\inf_{\widetilde{v}_1 \in \gB_{{i(v)}}}p_{V_\alpha\mymid  V_1}(v \mymid  \widetilde{v}_1)}\\
    &\le e^{C_1 k\log \frac{R_0}{\epsilon_0}}
    \sup_{v_1\in \gE_{\alpha, C}(v), \widetilde{v}_1 \in \gB_{i(v)}} \exp\left\{
    \frac{1}{2(1-{\alpha})}\left[\norm{v - \sqrt{{\alpha}}\widetilde{v}_1}_2^2
    {-} \norm{v - \sqrt{{\alpha}}{v}_1}_2^2
    \right]
    \right\}.
    \end{aligned}
    \end{equation}
    Here, the last inequality follows from the property $\PB(\gB_{i(v)}) \ge \exp\left(-C_1 k \log (R_0/\epsilon_0)\right)$, which is a direct consequence of the assumption $v \in \gT_\alpha$ and the definition \eqref{eq:defn-T-alpha-appendix} of $\gT_\alpha$.

    Further, consider any $(v_1, \widetilde{v}_1)$ with $v_1 \in \gE_{\alpha, C}(v)$ and $\widetilde{v}_1 \in \gB_{i(v)}$. Without loss of generality,  suppose $v_1 \in \gB_j$. In light of the expression $v = \sqrt{{\alpha}} v_1^* + \sqrt{1 - {\alpha}}\omega$, we can demonstrate that
    $$
    \begin{aligned}
    &\norm{v - \sqrt{{\alpha}}\widetilde{v}_1}_2^2 - \norm{v - \sqrt{{\alpha}}{v}_1}_2^2 \\
    &\qquad = -{\alpha} \norm{v_1^* - v_1}_2^2 + 2\sqrt{{\alpha} (1-{\alpha})} \inner{v_1 - \widetilde{v}_1}{\omega} + {\alpha} \norm{v_1^* - \widetilde{v}_1}_2^2\\
    &\qquad \overset{\text{(a)}}{\le}
    -{\alpha}\left(\big\|x_{i(v)}^{\star} - x_j^{\star}\big\| - 2\epsilon_0\right)^2 + 2\sqrt{{\alpha} (1 - {\alpha})}\inner{v_1 - \widetilde{v}_1}{\omega} + 4{\alpha}\epsilon_0^2\\
    &\qquad \overset{\text{(b)}}{\le}
    4{\alpha}\epsilon_0 \big\| x_{i(v)}^{\star} - x_j^{\star}\big\|
    {-}{\alpha} \big\| x_{i(v)}^{\star} {-} x_j^{\star} \big\|^2 + 2\sqrt{{\alpha}(1 - {\alpha})}\left\{\big\langle x_j^{\star} - x_{i(v)}^{\star}, \omega \big\rangle + 2(\sqrt{d} + \sqrt{C_1 k \log (R_0/\epsilon_0)})\epsilon_0\right\}.
    \end{aligned}
    $$
    Here, (a) follows from the property of the constructed \( \epsilon_0 \)-net, while (b) combines the definition of the \( \epsilon_0 \)-net with the norm bound for \( \omega \in \gG \) (cf.~\eqref{eq:defn-G-set-construct}). Moreover, since \( \omega \in \gG \), it is clearly seen from \eqref{eq:defn-G-set-construct} that
    \[
    \big\langle x_j^{\star} - x_{i(v)}^{\star},\omega \big\rangle \le \sqrt{C_1 k \log (R_0/\epsilon_0)} \, \big\| x_j^{\star} - x_{i(v)}^{\star}\big\|.
    \]
    In addition, with the choice of $\epsilon_0$ satisfying \(\epsilon_0 \ll \sqrt{\frac{1 - {\alpha}}{{\alpha}}}\min\left\{
    1 , \sqrt{\frac{k\log(R_0/\epsilon_0)}{d}}
    \right\} \le \frac{1\wedge \sqrt{\frac{k\log(R_0/\epsilon_0)}{d}}}{T}\), the following property holds:
    \[
    4\sqrt{{\alpha}(1 - {\alpha})}(\sqrt{d} + \sqrt{C_1 k\log (R_0/\epsilon_0)})\epsilon_0 \le 5(1 - {\alpha}){k \log (R_0/\epsilon_0)}.
    \]
    With the preceding bounds in place, we can readily obtain
    $$
    \begin{aligned}
    \norm{v - \sqrt{{\alpha}}\widetilde{v}_1}_2^2 - \norm{v - \sqrt{{\alpha}}{v}_1}_2^2
    &\le -{\alpha} \big\|x_{i(v)}^{\star} - x_j^{\star}\big\|^2 + 4(1 - {\alpha}){k\log(R_0/\epsilon_0)}\\
    &\quad +\left(2\sqrt{C_1 {\alpha}(1 - {\alpha}) k \log(R_0/\epsilon_0)} + 4{\alpha} \epsilon_0\right)\big\|x_j^{\star} - x_{i(v)}^{\star}\big\|\\
    &\overset{\text{(a)}}{\le} -\frac{{\alpha}}{2}\big\|x_{i(v)}^{\star} - x_j^{\star}\big\|^2 + 4(1-{\alpha})k \log(R_0/\epsilon_0)\\
    &\le -\frac{{\alpha}}{4}\big\|x_{i(v)}^{\star} - v_1\big\|^2 + 4(1-{\alpha})k\log(R_0/\epsilon_0)\\
    &\le -\frac{C}{4}(1-{\alpha})k \log(R_0/\epsilon_0) + 4(1-{\alpha})k \log(R_0/\epsilon_0) \\
    &\le -\frac{C}{5}(1 - {\alpha})k \log(R_0/\epsilon_0).
    \end{aligned}
    $$
    Here, both (a)  and the last inequality follow since \( v_1 \in \gE_{\alpha,C}(v) \) and \( C \ge C_2 \). Taking the above bound collectively with \eqref{eq:posterior_norm_1} yields
    $$
    \begin{aligned}
    \PB(\gE_{\alpha, C}(v) \mymid  V_\alpha = v) &\le \exp(C_1 k\log(R_0/\epsilon_0))\\
    &\hspace{5em}\times \sup_{x\in \gE_{\alpha, C}(v), \widetilde{v}_1 \in \gB_{i(v)}} \exp\left\{
    \frac{1}{2(1-{\alpha})}\left[\norm{v - \sqrt{{\alpha}}\widetilde{v}_1}_2^2
    - \norm{v - \sqrt{{\alpha}}{v}_1}_2^2
    \right]
    \right\}\\
    & \leq \exp(C_1 k \log(R_0/\epsilon_0)) \cdot \exp \left(
    -\frac{C}{5}\frac{1}{2(1-{\alpha})}(1-{\alpha})k \log(R_0/\epsilon_0)
    \right)\\
    &\le \exp\left(-\frac{C}{20} k \log(R_0/\epsilon_0)\right)
    \end{aligned}
    $$
    as claimed.

\subsection{Proof of Lemma~\ref{lem:bound_CoV_1|t_F^2}}
\label{sec:proof:lem:bound_CoV_1|t_F^2}

%
%

    To begin with,   we make note of the following result, originally developed in the stochastic localization literature \citep{eldan2020taming} (see also \citet[Lemma 1]{benton2024nearly}),
    which plays an important role in bounding the term \( \EB_{X_t }\big[\norm{\Cov_{X_0|X_t}(X_t)}_{\mathrm{F}}^2\big] \).
\begin{lemma}\label{lem:stoch_local}
    Let $\lambda_t \coloneqq \sqrt{1 - e^{-2t}}$, then for all $t>0$,
    \[
    \frac{\lambda_t^3}{2\dot{\lambda}_t}\frac{\rd}{\rd t} \EB_{U_t}\left[\Cov_{U_0|U_t}(U_t)\right] = \EB_{U_t} \left[\big(\Cov_{U_0|U_t}(U_t)\big)^2\right].
    \]
    where $U_t \coloneqq e^{-t} X_0 + \sqrt{1 - e^{-2t}}Z$ with $X_0\sim \pdata$ and $Z\sim \mathcal{N}(0,I_d)$.
    Here, we let
    $\Cov_{U_1|U_t}(u) = \EB [U_1U_1^\top \mymid  U_t = u] - \mathbb{E}[U_1 \mymid  U_t=u]\,\mathbb{E}[U_1 \mymid  U_t=u]^\top$,
    and denote by $\dot{\lambda}_t$ the derivative of $\lambda_t$ with respect to~$t$.
\end{lemma}

Now, let us introduce the bijection \( \alpha(t) \coloneqq e^{-2t} \) that maps  \( t \in [0, \infty) \) to \( \alpha \in (0, 1] \).  Take
\begin{align}
V_\alpha \coloneqq \sqrt{\alpha}X_0 + \sqrt{1 - \alpha}Z,
\qquad \nu_\alpha \coloneqq \sqrt{1 - \alpha}, \qquad
\text{and} \qquad
t(\alpha) \coloneqq \frac{1}{2}\log \frac{1}{\alpha}.
\end{align}
%
%
Then it can be readily seen that
\begin{align}
V_\alpha = U_{t(\alpha)}
\qquad \text{and}\qquad
\nu_{\alpha} = \lambda_{t(\alpha)}.
\end{align}
Straightforward calculations allow one to rewrite the result in Lemma~\ref{lem:stoch_local} as
\begin{equation}\label{eq:dEBCov_V_eq1}
\begin{aligned}
\rd \EB_{V_\alpha}\left[\Cov_{V_1|V_{\alpha}}(V_\alpha)\right] &= \frac{2}{\lambda_{t(\alpha)}^3} \left. \frac{\rd \lambda_{t}}{ \rd t} \right|_{t=t(\alpha)} \EB_{V_\alpha}\left[\Cov_{V_1| V_{\alpha}}^2(V_\alpha)\right] \rd t(\alpha) = \frac{2 \frac{\rd \nu_{\alpha}}{\rd \alpha}}{\nu_{\alpha}^3}\EB_{V_\alpha}\left[\Cov_{V_1|V_\alpha}^2(V_\alpha)\right] \rd \alpha\\
&=
- \frac{(1-\alpha)^{-1/2}}{(1-\alpha)^{3/2}}\EB_{V_\alpha}\left[\Cov_{V_1|V_\alpha}^2(V_\alpha)\right] \rd \alpha = -\frac{1}{(1-\alpha)^2}\EB_{V_\alpha}\left[\Cov_{V_1|V_\alpha}^2(V_\alpha)\right] \rd \alpha .
\end{aligned}
\end{equation}
Integrating the above equation over the interval \( [\overline{\alpha}_{t+1}, \overline{\alpha}_t) \), we obtain
\begin{align}
\int_{\overline{\alpha}_{t+1}}^{\overline{\alpha}_{t}}\frac{1}{(1 - \alpha)^2} \EB_{V_\alpha}\left[\Cov_{V_1|V_\alpha}^2(V_\alpha)\right] \rd \alpha &= \EB_{V_{\overline{\alpha}_{t+1}}}\left[\Cov_{V_1|V_{\overline{\alpha}_{t+1}}}(V_{\overline{\alpha}_{t+1}})\right] - \EB_{V_{\overline{\alpha}_t}}\left[\Cov_{V_1|V_{\overline{\alpha}_{t}}}(V_{\overline{\alpha}_t})\right] \notag\\
&= \EB_{X_{t+1}}\left[\Cov_{X_0|X_{t+1}}(X_{t+1})\right] - \EB_{X_{t}}\left[\Cov_{X_0|X_{t}}(X_t)\right].
\label{eq:dEBCov_V_eq1-int}
\end{align}

Next, in order to
further control the left-hand side of \eqref{eq:dEBCov_V_eq1-int},
we proceed to bounding the difference between \( \Cov_{V_1 \mymid  V_{\overline{\alpha}_{t+1}}} \) and \( \Cov_{V_1 \mymid  V_\alpha} \)   for \( \alpha \in [\overline{\alpha}_{t+1}, \overline{\alpha}_t] \). Towards this end, we resort to the following SDE that describes the dynamics of the random process \( \{\Cov_{V_1 \mymid  V_\alpha}(V_{\alpha}) \} \), whose proof can be found in \citet[Section~4.2.1]{eldan2022analysis}:
\[
\rd \Cov_{V_1|V_\alpha}(V_{\alpha}) = -\frac{1}{(1-\alpha)^2}\Cov_{V_1|V_\alpha}^2(V_{\alpha}) \rd \alpha + \gM_{\alpha}^{(3)}\rd B_{\frac{\alpha}{1-\alpha}}.
\]
Here, $(B_t)$ denotes the standard Brownian motion in $\mathbb{R}^d$ and
\begin{equation}
\gM_{\alpha}^{(l)} \coloneqq \EB\left[ (V_1 - \EB[V_1|V_\alpha])^{\otimes l} \mymid   V_\alpha\right].
\label{eq:defn-M-l-moment}
\end{equation}
Denoting by \( \langle M \rangle \) the quadratic variation of a stochastic process \( (M_t) \), we can invoke It\^o's formula to obtain
\begin{align*}
    &\rd \bigg(\tr \left(\Cov_{V_1|V_\alpha}^2(V_{\alpha})\right)\bigg) = 2 \inner{\Cov_{V_1|V_\alpha}(V_{\alpha})}{\rd \Cov_{V_1|V_\alpha}(V_{\alpha})} + \rd \left[\tr \left(\left\langle \Cov_{V_1|V_\alpha}\right\rangle\right)\right]\\
    &= 2 \inner{\Cov_{V_1|V_\alpha}(V_{\alpha})}{\gM_\alpha^{(3)}\rd B_{\frac{\alpha}{1-\alpha}}} - \frac{2}{(1-\alpha)^2}\inner{\Cov_{V_1|V_\alpha}(V_{\alpha})}{\Cov_{V_1|V_\alpha}^2(V_{\alpha})}\rd \alpha\\ &\hspace{2em}+ \frac{1}{(1-\alpha)^2}\big\langle \gM_\alpha^{(3)},  \gM_\alpha^{(3)} \big\rangle\rd \alpha.
\end{align*}
Taking expectation then yields
\begin{equation}\label{eq:dtrCov^2}
\begin{aligned}
\rd \left[\tr \left( \EB \left[\Cov_{V_1|V_\alpha}^2(V_{\alpha}) \right]\right)\right] = - \frac{2}{(1-\alpha)^2}\EB\left[\inner{\Cov_{V_1|V_\alpha}(V_{\alpha})}{\Cov_{V_1|V_\alpha}^2(V_{\alpha})}\right]\rd \alpha\\ + \frac{1}{(1-\alpha)^2}\EB\left[\big\langle \gM_\alpha^{(3)},  \gM_\alpha^{(3)} \big\rangle \right]\rd \alpha.
\end{aligned}
\end{equation}

In order to control $\EB \big[\Cov_{V_1|V_\alpha}^2(V_{\alpha}) \big]$ through the differential equation \eqref{eq:dtrCov^2}, we need to bound the drift terms on the right-hand side of \eqref{eq:dtrCov^2}.
%
%
To bound the first term on the right-hand side of \eqref{eq:dtrCov^2}, we observe from the symmetry 
of $\Cov_{V_1\mymid  V_\alpha}$ that
\begin{align}
&\EB \left[\inner{\Cov_{V_1|V_\alpha}(V_{\alpha})}{\Cov_{V_1|V_\alpha}^2(V_{\alpha})}\right] = \EB \left[\tr\left(\Cov_{V_1|V_\alpha}^3(V_{\alpha})\right)\right] \notag\\
&\le \EB \left[\norm{\Cov_{V_1|V_\alpha}(V_{\alpha})} \cdot \norm{\Cov_{V_1|V_\alpha}(V_{\alpha})}_{\mathrm{F}}^2\right]\notag\\
&\le
\EB \left[\tr\big(\Cov_{V_1|V_\alpha}(V_{\alpha})\big)\cdot \norm{\Cov_{V_1|V_\alpha}(V_{\alpha})}_{\mathrm{F}}^2\right]\notag\\
&= \EB \left[\tr\big(\Cov_{V_1|V_\alpha}(V_{\alpha})\big)\mathbbm{1}\{V_\alpha \in \gT_\alpha\}\cdot \norm{\Cov_{V_1|V_\alpha}(V_{\alpha})}_{\mathrm{F}}^2\right]\notag\\ &\hspace{2em} +
\EB \left[\tr\big(\Cov_{V_1|V_\alpha}(V_{\alpha})\big)\mathbbm{1}\{V_\alpha \notin \gT_\alpha\}\cdot \norm{\Cov_{V_1|V_\alpha}(V_{\alpha})}_{\mathrm{F}}^2\right]\notag\\
&\le C_3\frac{1-\alpha}{\alpha}(k\log T) \EB\left[\norm{\Cov_{V_1|V_\alpha}(V_{\alpha})}_{\mathrm{F}}^2\right]+
\EB \left[\tr\big(\Cov_{V_1|V_\alpha}(V_{\alpha})\big)\mathbbm{1}\{V_\alpha \notin \gT_\alpha\}\cdot \norm{\Cov_{V_1|V_\alpha}(V_{\alpha})}_{\mathrm{F}}^2\right],
\label{eq:UB-Cov-Cov2-123}
\end{align}
where the last inequality follows from the definition of $\gT_\alpha$ in~\eqref{eq:defn-T-alpha-appendix}, Corollary~\ref{cor:posterior_mmt} and the basic fact that
\[
\tr\big(\Cov_{V_1|V_\alpha}(v)\big)
= \EB \left[ \tr\Big(\big(V_1 - \mu_{V_1|V_{\alpha}} \big)\big(V_1 - \mu_{V_1|V_{\alpha}} \big)^{\top} \Big) \mymid  V_{\alpha} = v \right]
= \EB \left[ \big\| V_1 - \mu_{V_1|V_{\alpha}} \big\|^2 \mymid  V_{\alpha} = v \right] .
\]
Regarding the last term of inequality~\eqref{eq:UB-Cov-Cov2-123}, combining Assumption~\ref{ass:bd_supp} and Lemma~\ref{lem:gT_alpha_typical} results in
\[
\EB \left[\tr\big(\Cov_{V_1|V_\alpha}(V_{\alpha})\big)\mathbbm{1}\{V_\alpha \notin \gT_\alpha\}\cdot \norm{\Cov_{V_1|V_\alpha}(V_{\alpha})}_{\mathrm{F}}^2\right] \le
8T^{6c_R}\PB(V_\alpha\notin \gT_\alpha) \le \frac{1}{T^{10}}.
\]
As a consequence, we arrive at
\begin{equation}
    \EB\left[\inner{\Cov_{V_1|V_\alpha}(V_{\alpha})}{\Cov_{V_1|V_\alpha}^2(V_{\alpha})}\right]
    \le C_3\frac{1-\alpha}{\alpha}(k\log T) \EB \left[\norm{\Cov_{V_1|V_\alpha}(V_{\alpha})}_{\mathrm{F}}^2\right] + \frac{1}{T^{10}}.
\end{equation}
Substitution into \eqref{eq:dtrCov^2} yields
$$
\begin{aligned}
\rd \EB \left[\norm{\Cov_{V_1|V_\alpha}(V_{\alpha})}_{\mathrm{F}}^2\right]  &= - \frac{2}{(1-\alpha)^2}\EB\left[\inner{\Cov_{V_1|V_\alpha}(V_{\alpha})}{\Cov_{V_1|V_\alpha}^2(V_{\alpha})}\right]\rd \alpha\\ &+ \frac{1}{(1-\alpha)^2}\EB\left[\big\langle \gM_\alpha^{(3)}, \gM_\alpha^{(3)} \big\rangle \right]\rd \alpha\\
&\ge -\frac{2}{(1-\alpha)^2}\EB\left[\inner{\Cov_{V_1|V_\alpha}(V_{\alpha})}{\Cov_{V_1|V_\alpha}^2(V_{\alpha})}\right]\rd \alpha\\
&\ge -\frac{2}{(1-\alpha)^2}C_3\frac{1-\alpha}{\alpha}(k\log T) \EB \left[\norm{\Cov_{V_1|V_\alpha}(V_{\alpha})}_{\mathrm{F}}^2\right]\rd \alpha - \frac{1}{T^{10}}\rd \alpha\\
&\ge -\frac{2C_3 k \log T}{\overline{\alpha}_{t+1}(1 - \overline{\alpha}_t)}\EB \left[\norm{\Cov_{V_1|V_\alpha}(V_{\alpha})}_{\mathrm{F}}^2\right] \rd \alpha - \frac{1}{T^{10}} \rd \alpha,
\end{aligned}
$$
where the last line holds provided that $\alpha \in [\overline{\alpha}_{t+1},\overline{\alpha}_t]$.
In view of Grownwall's inequality, we can derive
\begin{align*}
    \exp\left\{\frac{2C_3k{\alpha}\log T }{\overline{\alpha}_{t+1}(1 - \overline{\alpha}_t)}\right\}&\EB \left[\norm{\Cov_{V_1|V_{{\alpha}}}(V_{\alpha})}_{\mathrm{F}}^2\right] - \exp\left\{\frac{2C_3k\overline{\alpha}_{t+1}\log T }{\overline{\alpha}_{t+1}(1 - \overline{\alpha}_t)}\right\}\EB \left[\big\| \Cov_{V_1|V_{\overline{\alpha}_{t+1}}}(V_{\overline{\alpha}_{t+1}}) \big\|_{\mathrm{F}}^2\right]\\
    &\ge -\frac{1}{T^{10}}\int_{\overline{\alpha}_{t+1}}^{{\alpha}}\exp\left\{\frac{2C_3k\alpha^\prime\log T }{\overline{\alpha}_{t+1}(1 - \overline{\alpha}_t)}\right\}\rd \alpha^\prime\\
    &=
    -\frac{\overline{\alpha}_{t+1}(1 - \overline{\alpha}_t)}{2C_3kT^{10}\log T}\left(
    \exp\left\{\frac{2C_3k{\alpha}\log T}{\overline{\alpha}_{t+1}(1 - \overline{\alpha}_t)}\right\} -
    \exp\left\{\frac{2C_3k\overline{\alpha}_{t+1}\log T}{\overline{\alpha}_{t+1}(1 - \overline{\alpha}_t)}\right\}
    \right).
\end{align*}
Dividing both sides of the above inequality by \( \exp\left\{\frac{2C_3k\overline{\alpha}_{t+1}\log T}{\overline{\alpha}_{t+1}(1 - \overline{\alpha}_t)}\right\} \), we obtain
\begin{align*}
    \exp\left\{\frac{2C_3k(\alpha - \overline{\alpha}_{t+1})\log T}{\overline{\alpha}_{t+1}(1 - \overline{\alpha}_t)}\right\}&\EB \left[\norm{\Cov_{V_1|V_\alpha}(V_\alpha)}_{\mathrm{F}}^2\right] - \EB \left[\norm{\Cov_{V_1|V_{\overline{\alpha}_{t}}}(V_{\overline{\alpha}_{t}})}_{\mathrm{F}}^2\right]\\
    &\ge -\frac{\overline{\alpha}_{t+1}(1 - \overline{\alpha}_t)}{2C_3kT^{10}\log T}\left(
    \exp\left\{\frac{2C_3k(\alpha - \overline{\alpha}_{t+1})\log T}{\overline{\alpha}_{t+1}(1 - \overline{\alpha}_t)}\right\} - 1
    \right).
\end{align*}
According to Lemma~\ref{lem:step_size}, every $\alpha \in [\overline{\alpha}_{t+1}, \overline{\alpha}_t]$ obeys
\[
\frac{2C_3k(\alpha - \overline{\alpha}_{t+1})\log T}{\overline{\alpha}_{t+1}(1 - \overline{\alpha}_t)} \le \frac{2C_3k(1 - {\alpha}_{t+1})\log T}{{\alpha}_{t+1} - \overline{\alpha}_{t+1}} \le \frac{8C_3c_1 k\log^2 T}{T} \le 1 ,
\]
provided that $8C_3c_1k\log^2 T \le T$.
Consequently, we have
\begin{equation}\label{eq:bd_EB_Cov_F^2}
\begin{aligned}
\EB\bigg[&\norm{\Cov_{V_1|V_{\overline{\alpha}_{t+1}}}(V_{\overline{\alpha}_{t+1}})}_{\mathrm{F}}^2\bigg] \le \exp\left\{\frac{2C_3k(\alpha - \overline{\alpha}_{t+1})\log T}{\overline{\alpha}_{t+1}(1 - \overline{\alpha}_t)}\right\}\EB \left[\norm{\Cov_{V_1|V_\alpha}(V_\alpha)}_{\mathrm{F}}^2
    \right] \\
    &\qquad + \frac{\overline{\alpha}_{t+1}(1 - \overline{\alpha}_t)}{2C_3kT^{10}\log T}\left(
    \exp\left\{\frac{2C_3k(\alpha - \overline{\alpha}_{t+1})\log T}{\overline{\alpha}_{t+1}(1 - \overline{\alpha}_t)}\right\} - 1
    \right)\\
    &\le 3\EB\left[\norm{\Cov_{V_1|V_\alpha}(V_\alpha)}_{\mathrm{F}}^2\right] + \frac{\overline{\alpha}_{t+1}(1 - \overline{\alpha}_t)}{2C_3kT^{10}\log T}\left(
    \exp\left\{\frac{2C_3k(\alpha - \overline{\alpha}_{t+1})\log T}{\overline{\alpha}_{t+1}(1 - \overline{\alpha}_t)}\right\} - 1
    \right)\\
    &\overset{\text{(a)}}{\le} 3\EB\left[\norm{\Cov_{V_1|V_\alpha}(V_\alpha)}_{\mathrm{F}}^2\right] + \frac{2(\overline{\alpha}_t - \overline{\alpha}_{t+1})}{T^{10}},
\end{aligned}
\end{equation}
where (a) holds since $e^x - 1 \le 2x$ for all $x \le 1$.

To finish up, combining \eqref{eq:dEBCov_V_eq1-int} and \eqref{eq:bd_EB_Cov_F^2} and making use of the equivalence between \( X_{t} \) and \( V_{\overline{\alpha}_t} \) give
$$
\begin{aligned}
    \EB \Big[\tr&\left(\Cov_{X_0|X_{t+1}}(X_{t+1})\right)\Big] - \EB \left[\tr\left(\Cov_{X_0|X_{t}}(X_{t})\right)\right] = \int_{\overline{\alpha}_{t+1}}^{\overline{\alpha}_t} \frac{1}{(1-\alpha)^2}\EB \left[\tr\left(\Cov^2_{V_1|V_\alpha}(V_\alpha)\right)\right] \rd \alpha\\
    &\ge \int_{\overline{\alpha}_{t+1}}^{\overline{\alpha}_t} \frac{1}{3(1-\alpha)^2}\left\{
    \EB \left[\norm{\Cov^2_{V_1|V_{\overline{\alpha}_{t+1}}}(V_{\overline{\alpha}_{t+1}})}_{\mathrm{F}}^2\right] - \frac{2(\overline{\alpha}_t - \overline{\alpha}_{t+1})}{T^{10}}
    \right\}\rd \alpha\\
    &\ge \frac{\overline{\alpha}_t(1 - \alpha_{t+1})}{3(1 - \overline{\alpha}_t)(1 - \overline{\alpha}_{t+1})}\EB \left[\norm{\Cov_{X_0|X_{t+1}}(X_{t+1})}_{\mathrm{F}}^2
    \right] - \frac{\overline{\alpha}_t^2(1 - \alpha_{t+1})^2}{T^{10}(1 - \overline{\alpha}_t)(1 - \overline{\alpha}_{t+1})}\\
    &= \frac{\overline{\alpha}_{t+1}(1 - \alpha_{t+1})}{3(\alpha_{t+1} - \overline{\alpha}_{t+1})(1 - \overline{\alpha}_{t+1})}\EB \left[\norm{\Cov_{X_0|X_{t+1}}(X_{t+1})}_{\mathrm{F}}^2
    \right] - \frac{\overline{\alpha}_t^2(1 - \alpha_{t+1})^2}{T^{10}(1 - \overline{\alpha}_t)(1 - \overline{\alpha}_{t+1})}.
\end{aligned}
$$
Rearranging terms and recalling that $\widetilde{\sigma}_{t+1}^2 = \frac{\overline{\alpha}_{t+1}(1 - \alpha_{t+1})}{(\alpha_{t+1} - \overline{\alpha}_{t+1})(1 - \overline{\alpha}_{t+1})}$, we obtain
\begin{equation}\label{eq:bd_Fnorm_Cov}
\begin{aligned}
    \widetilde{\sigma}_{t+1}^2 \EB\left[\norm{\Cov_{X_0|X_{t+1}}(X_{t+1})}_{\mathrm{F}}^2\right] &= \frac{(1- \alpha_{t+1}) \overline{\alpha}_{t+1}}{(\alpha_{t+1} - \overline{\alpha}_{t+1})(1 - \overline{\alpha}_{t+1})}\EB\left[\norm{\Cov_{X_0|X_{t+1}}(X_{t+1})}_{\mathrm{F}}^2\right] \\
    &\le
   3\left\{
    \EB \left[\tr
    \big(\Cov_{X_0|X_{t+1}}(X_{t+1})\big)\right] - \EB\left[\tr(\Cov_{X_0|X_{t}}(X_{t}))\right]\right\} + \frac{3}{T^{10}},
\end{aligned}
\end{equation}
We have thus completed the proof of this lemma.


\subsection{Proof of Lemma~\ref{lem:step_size}}
\label{sec:proof:lem:step_size}
    A little algebra yields
    $$
    \begin{aligned}
        \frac{\overline{\alpha}_t(1 - \alpha_t)}{2(\alpha_t - \overline{\alpha}_t)(1 - \overline{\alpha}_t)} -&
      \frac{\overline{\alpha}_{t+1}(1 - \alpha_{t+1})}{2(\alpha_{t+1} - \overline{\alpha}_{t+1})(1 - \overline{\alpha}_{t+1})} = \frac{\overline{\alpha}_{t-1}(1 - \alpha_t)(1 - \overline{\alpha}_{t+1}) - \overline{\alpha}_t(1 - \alpha_{t+1})(1 - \overline{\alpha}_{t-1})}{2(1 - \overline{\alpha}_{t-1})(1 - \overline{\alpha}_{t})(1 - \overline{\alpha}_{t+1})}\\
     &\overset{\text{(a)}}{\le} \frac{\overline{\alpha}_{t-1}(1 - \alpha_t)[(1 - \overline{\alpha}_{t+1}) - \alpha_t + \overline{\alpha}_t]}{
     2(1 - \overline{\alpha}_{t-1})(1 - \overline{\alpha}_{t})(1 - \overline{\alpha}_{t+1})
     } =
     \frac{
     \overline{\alpha}_{t-1}(1 - \alpha_t)[1 - \alpha_t + \overline{\alpha}_t(1 - \alpha_{t+1})]
     }{
     2(1 - \overline{\alpha}_{t-1})(1 - \overline{\alpha}_{t})(1 - \overline{\alpha}_{t+1})
     }\\
     &\le
     \frac{\overline{\alpha}_{t-1}(1 - \alpha_t)(1 - \alpha_{t+1})}{(1 - \overline{\alpha}_{t-1})(1 - \overline{\alpha}_{t})(1 - \overline{\alpha}_{t+1})}
     \le
     \left(\frac{8c_1 \log T}{T}\right)^2 \frac{\overline{\alpha}_{t}}{1 - \overline{\alpha}_t}.
    \end{aligned}
    $$
    Here, (a) follows since $1 - \alpha_t \le 1 - \alpha_{t+1}$, while the last inequality applies \eqref{eq:basic-alphat-property}.


\subsection{Proof of Lemma~\ref{lem:logdet_expand}}
\label{sec:proof:lem:logdet_expand}
    For any matrix $B$, we know that $ B$ and $ B^\top$ have the same determinant. As a result,
    \begin{equation}\label{eq:logdet_eq1}
    \begin{aligned}
        \log\det(I + \eta A + \eta \Delta) &= \frac{1}{2}\left\{\log\det(I + \eta A + \eta \Delta^\top) + \log\det(I + \eta A + \eta \Delta)\right\}\\
        &=
        \frac{1}{2}\log\det\left(
        I + 2\eta A + \eta(\Delta^\top + \Delta) + \eta^2( A + \Delta)^\top ( A + \Delta)
        \right).
    \end{aligned}
    \end{equation}
    For any vector $x\in\RB^d$, we make the observation that
    \begin{align*}
        x^\top \big(I + \eta(2 A + \Delta^\top + \Delta) \big)x &=
        \norm{x}_2^2 +2\eta x^\top  A x + \eta x^\top(\Delta^\top + \Delta)x\\
        &\ge \norm{x}_2^2 - \eta\norm{\Delta^\top + \Delta} \norm{x}_2^2 \geq (1 - 2\eta\norm{\Delta})\norm{x}_2^2
        \ge \frac{1}{2}\norm{x}_2^2,
    \end{align*}
    thus implying that  $I + 2\eta A + \eta(\Delta^\top + \Delta) \succ 0$. Further, it is easily seen that
    \begin{align*}
        I + 2\eta  A + \eta(\Delta^\top + \Delta) \preceq I + 2\eta A + \eta(\Delta^\top + \Delta) + \eta^2( A + \Delta)^\top ( A + \Delta).
    \end{align*}
    According to the L\"owner–Heinz theorem, $\log  A \preceq \log  B$ holds for any $0 \preceq A \preceq  B$.
    %
    This in turn allows one to derive
    \begin{equation}\label{eq:logdet_eq2}
    \begin{aligned}
        \log\det \big(
        I &+ 2\eta A + \eta(\Delta^\top + \Delta) + \eta^2( A + \Delta)^\top ( A + \Delta)
        \big)\\
        &= \tr\Big(
        \log\left(
        I + 2\eta A + \eta(\Delta^\top + \Delta) + \eta^2( A + \Delta)^\top ( A + \Delta)
        \right)
        \Big)\\
        &\ge
        \tr\Big(\log\big(
        I + 2\eta  A + \eta(\Delta^\top + \Delta)\big)
        \Big)=
        \log\det\big(
        I + 2\eta  A + \eta(\Delta^\top + \Delta)
        \big).
    \end{aligned}
    \end{equation}

    In addition, for any symmetric matrix $ B\in \RB^{d\times d}$, we denote its eigenvalues as $\{\lambda_i( B)\}_{i=1}^d$, then Weyl's inequality tells us that
    \[\lambda_i\big(2\eta  A + \eta(\Delta^\top + \Delta)\big)
    \ge 2\eta\lambda_i(  A) - \eta\norm{\Delta^\top + \Delta} \ge  - 2\eta \norm{\Delta} \ge -\frac{1}{2} \qquad \text{for all } i\le d.
    \]
    Recalling the elementary inequality $\log (1 + x) \ge x - x^2$ for any $x \ge -1/2$, we can reach
\begin{equation}\label{eq:logdet_eq3}
    \begin{aligned}
        \log\det\left(I + 2\eta A + \eta(\Delta^\top + \Delta)\right) &\ge \ssum{i}{1}{d}\eta\lambda_i\left(2 A + \Delta^\top + \Delta\right) - \ssum{i}{1}{d}\eta^2 \lambda_i^2\left(2 A + \Delta^\top + \Delta\right)\\
        &= \eta \tr\left(2 A + \Delta^\top + \Delta\right) - \eta^2\norm{2 A + \Delta^\top + \Delta}_{\mathrm{F}}^2\\
        &\ge
        2\eta \tr( A) + 2\eta \tr(\Delta) - 8\eta^2 \norm{ A}_{\mathrm{F}}^2 - 8\eta^2 \norm{\Delta}_{\mathrm{F}}^2.
    \end{aligned}
    \end{equation}
    The proof can thus be completed by combining \eqref{eq:logdet_eq1}, \eqref{eq:logdet_eq2} and \eqref{eq:logdet_eq3}.

%% file: training_complexity.tex
\section{Training guarantees adapted to low dimension}
\label{sec:training-guarantees}

In this section, we discuss how the training step of diffusion models can also adapt to the intrinsic
dimension of the data distribution when a kernel-based score estimator is used. 


We first define the following notation which will be appear
throughout this section.
\begin{itemize}
\item
\(k^\star\coloneqq C_{\mathsf{cover}}k\) denotes the effective intrinsic
dimension appearing in the covering-number bounds.
\item For any \(t \ge 0\), define
\(
\gamma_t\coloneqq
\sqrt{\frac{\overline{\alpha}_t}{1-\overline{\alpha}_t}}
\).
\item
Throughout this section, the covering scale used in the definitions of
\(\gI\), \(\gG\), and \(\gT_{\overline{\alpha}_t}\) from
\eqref{eq:defn-I-set}, \eqref{eq:defn-G-set-construct}, and
\eqref{eq:defn-T-alpha-appendix} is taken as
\(
\epsilon_t\coloneqq
\sqrt{1-\overline{\alpha}_t}
\).
With this choice, the index set \(\gI\) is defined as
\begin{align}
\mathcal{I}
\coloneqq
\left\{
1\le i\le N_{\epsilon_t}:
\mathbb{P}(X_0\in\mathcal{B}_i)
\ge
\exp\left(-C_1k\log\frac{dnR_0}{\epsilon_t}\right)
\right\}.
\end{align}
For notational simplicity, throughout this section, unless otherwise stated,
we write \(N_\varepsilon\) for \(N_\varepsilon(\gX_{\mathsf{data}})\).
\item
For \(x\in\RB^d\), define
\(
D(x)\coloneqq
\sum_{r=1}^n
\varphi_{1-\overline{\alpha}_t}
\bigl(x-\sqrt{\overline{\alpha}_t}X_r^{\mathsf{sample}}\bigr)
\).
\item
For \(x\in\RB^d\), define the good-density event
\(
\gE_x\coloneqq
\left\{\widehat p_{X_t}(x)\ge \frac12 p_{X_t}(x)\right\}
\).
\item
For \(y\in\gX_{\mathsf{data}}\) and \(0<\alpha<1\), define the posterior
typical set
\begin{equation}
\gE_{\alpha}(y) \coloneqq
\left\{v\in \gX_{\mathsf{data}}:
\sqrt{\alpha}\|v-y\|_2
\le
15\sqrt{k^\star(1-\alpha)\log \frac{dnR_0}{1-\alpha}}
\right\}. \label{eq:defn-gE-alpha-training}
\end{equation}
\item
For any measurable \(f\), define the empirical posterior expectation induced
by the kernel weights
\[
\widehat{\EE}_{0\mid t,x}[f(X_0)]
\coloneqq
\frac{\sum_{r=1}^n f(X_r^{\mathsf{sample}})
\varphi_{1-\overline{\alpha}_t}
\bigl(x-\sqrt{\overline{\alpha}_t}X_r^{\mathsf{sample}}\bigr)}
{D(x)} .
\]
Further, define the corresponding empirical posterior mean by
\(
\widehat\mu_{0\mid t}(x)
\coloneqq
\widehat{\EE}_{0\mid t,x}[X_0]
\).
\end{itemize}

\subsection{DDPM training complexity analysis (Proof of Theorem~\texorpdfstring{\ref{thm:ddpm-training-guarantee}}{DDPM training guarantee})}
\label{prf:thm:ddpm-training-guarantee}

We begin with the DDPM score-matching guarantee. The estimator is constructed
according to the relative size of \(d\) and
\((1-\overline{\alpha}_t)^{-1}\). We first define the soft-thresholding
function used in the low-dimensional ambient regime. Let
\(
a_t\coloneqq
C_{\mathsf{th}}
\sqrt{d(1-\overline{\alpha}_t)\log\frac{dnR_0}{1-\overline{\alpha}_t}},
\)
where \(C_{\mathsf{th}}>0\) is a sufficiently large universal constant,
and set \(h_t:[0,\infty)\to[0,1]\) as
\[
h_t(r)\coloneqq
\begin{cases}
1, & 0\le r\le a_t,\\[0.2em]
0, & r\ge 2a_t,\\[0.2em]
\left[
1+\exp\left(
\dfrac{2r/a_t-3}{(r/a_t-1)(2-r/a_t)}
\right)
\right]^{-1}, & a_t<r<2a_t.
\end{cases}
\]
Thus \(h_t\) is a decreasing soft-threshold function: it equals one on
\([0,a_t]\), smoothly decays on \([a_t,2a_t]\), and vanishes on
\([2a_t,\infty)\). In particular, since it is constant near the origin and
flat at the two transition endpoints, the radial map \(x\mapsto h_t(\|x\|_2)\)
is \(C^2\).

For \(z\in\RB^d\), define the thresholded kernel used in the numerator by
\[
\widetilde\varphi_{t,x}(z)
\coloneqq
\begin{cases}
\varphi_{1-\overline{\alpha}_t}
\bigl(\sqrt{\overline{\alpha}_t}z-x\bigr),
& d>(1-\overline{\alpha}_t)^{-1},\\[0.2em]
h_t\!\left(\|\sqrt{\overline{\alpha}_t}z-x\|_2\right)
\varphi_{1-\overline{\alpha}_t}
\bigl(\sqrt{\overline{\alpha}_t}z-x\bigr),
& d\le (1-\overline{\alpha}_t)^{-1}.
\end{cases}
\]
The score estimator is then defined by
\begin{equation}\label{eq:plugin-score-tweedie-form}
s_t(x)
\coloneq
\frac{
\sum_{i=1}^n
\bigl(\sqrt{\overline{\alpha}_t}X_i^{\mathsf{sample}}-x\bigr)
\widetilde\varphi_{t,x}(X_i^{\mathsf{sample}})}
{
(1-\overline{\alpha}_t)D(x)} .
\end{equation}
In the regime
\(d\le (1-\overline{\alpha}_t)^{-1}\), the prior concentration gives
\(\|s_t^\star(X_t)\|_2 = \widetilde{O}\big(\sqrt{d(1-\overline{\alpha}_t)^{-1}}\big)\) with high
probability. If the sample size is not large enough, the raw estimator
\(\sqrt{\overline{\alpha}_t}\widehat\mu_{0\mid t}(x)-x\) can be dominated by
large residuals; the cutoff above uses the prior scale and effectively shrinks
such unstable contributions toward zero.

We first collect several auxiliary estimates that will be used to prove the
score-matching guarantee. These lemmas isolate the density-ratio estimates,
typical-region reductions, and bad-event controls needed for the kernel
estimator analysis.

\begin{lemma}\label{lem:typical-region-upper-bound}
    Recall the definitions of $\gI,~\gG$ and \(\gE_\alpha(y)\)
     in \eqref{eq:defn-I-set}, \eqref{eq:defn-G-set-construct}
     and \eqref{eq:defn-gE-alpha-training}, respectively.
    Then, for every \(2\le \beta\le 6\), the contribution from the typical
    region obeys the following upper bound:
    \begin{align*}
    &\int_{y\in \bigcup_{i\in\gI}\gB_i}
    \int_{\omega\in\gG}
    \int_{x_0\notin \gE_{\overline{\alpha}_t}(y)}
    \varphi_{1}^\beta\big(\gamma_t (x_0 - y) - \omega\big)
    \varphi_1(\omega)
    \frac{p_0(x_0)p_0(y)}
    {(1-\overline{\alpha}_t)^{\beta d/2} p_{X_t}^\beta(x)}
    \rd x_0\rd\omega \rd y
    \le
    \frac{\overline{\alpha}_t^2(1-\overline{\alpha}_t)^2}{n}.
    \end{align*}
    Here \(x=\sqrt{\overline{\alpha}_t}y+\sqrt{1-\overline{\alpha}_t}\omega\)
    and
    \(\gamma_t = \sqrt{\overline{\alpha}_t/(1 - \overline{\alpha}_t)}\).
\end{lemma}
The proof of Lemma~\ref{lem:typical-region-upper-bound} is deferred to
Section~\ref{prf:lem:typical-region-upper-bound}.

\begin{lemma}\label{lem:typical-region-cubic-upper-bound}
    Under the same notation as Lemma~\ref{lem:typical-region-upper-bound}, define
    \[
    \mathsf{S}(x_0,v,y,\omega)
    \coloneqq
    \frac{
    \varphi_1^2\big(\gamma_t(x_0-y)-\omega\big)
    \varphi_1\big(\gamma_t(v-y)-\omega\big)
    \varphi_1(\omega)p_0(x_0)p_0(y)p_0(v)}
    {(1-\overline{\alpha}_t)^{3d/2}p_{X_t}^3(x)},
    \]
    where \(x=\sqrt{\overline{\alpha}_t}y+\sqrt{1-\overline{\alpha}_t}\omega\).
    Then
    \begin{align*}
    &\int_{y\in \bigcup_{i\in\gI}\gB_i}
    \int_{\omega\in\gG}
    \int_{x_0}
    \int_{v\notin \gE_{\overline{\alpha}_t}(y)}
    \mathsf{S}(x_0,v,y,\omega)
    \rd v \rd x_0\rd\omega \rd y
    \le
    \frac{\overline{\alpha}_t^2(1-\overline{\alpha}_t)^2}{n}.
    \end{align*}
\end{lemma}
The proof of Lemma~\ref{lem:typical-region-cubic-upper-bound} is deferred to
Section~\ref{prf:lem:typical-region-cubic-upper-bound}.

Combining the preceding estimates with the construction in
\eqref{eq:plugin-score-tweedie-form}, we obtain the following score-matching
guarantee.

\begin{proposition}[Low-dimensional kernel score estimation]\label{prop:low-dim_kernel_score_match}
Suppose Assumption~\ref{ass:bd_supp}, \ref{ass:low_dim} hold. Then the kernel-based score estimator
satisfies
\[
\varepsilon_{\mathsf{score}, t}^2
\;\lesssim\;
\left(\min\left\{\frac{R_0^2}{1-\overline{\alpha}_t}, d\right\}\right)^{\frac{2}{k^\star+2}}
\frac{R_0^{k^\star} \log\frac{dnR_0}{1-\overline{\alpha}_t}}
{n\,(1-\overline{\alpha}_t)^{k^\star/2+1}},
\]
for all noise levels $t$ in the regime considered.
\end{proposition}
The proof of Proposition~\ref{prop:low-dim_kernel_score_match} is deferred to
Section~\ref{prf:prop:low-dim_kernel_score_match}.

\begin{proof}[Proof of Theorem~\ref{thm:ddpm-training-guarantee}]
By the definition of \(\varepsilon_{\mathsf{ddpm}\text{-}\mathsf{sc}}\) in
Assumption~\ref{ass:ddpm_score_matching} and
Proposition~\ref{prop:low-dim_kernel_score_match},
\begin{align*}
\varepsilon_{\mathsf{ddpm}\text{-}\mathsf{sc}}^2
&\le
\frac1T\sum_{t=1}^T
(1-\overline{\alpha}_t)\varepsilon_{\mathsf{score},t}^2\\
&\lesssim
\frac{R_0^{k^\star}}{nT}
\sum_{t=1}^T
\frac{
\left(\min\left\{\frac{R_0^2}{1-\overline{\alpha}_t},d\right\}\right)^{\frac{2}{k^\star+2}}
\log\frac{dnR_0}{1-\overline{\alpha}_t}
}
{(1-\overline{\alpha}_t)^{k^\star/2}}.
\end{align*}
Since \(1-\overline{\alpha}_t\ge1-\overline{\alpha}_1
=1-\alpha_1=T^{-c_0}\), the logarithmic factor is bounded by
\(\log(dR_0nT^{c_0})\). Moreover,
\[
\frac{\min\left\{\frac{R_0^2}{1-\overline{\alpha}_t},d\right\}^{\frac{2}{k^\star+2}}}
{(1-\overline{\alpha}_t)^{k^\star/2}}
\le
\min\left\{
\frac{d^{\frac{2}{k^\star+2}}}{(1-\overline{\alpha}_t)^{k^\star/2}},
\frac{R_0^{\frac{4}{k^\star+2}}}{(1-\overline{\alpha}_t)^{\frac{k^\star}{2}
+\frac{2}{k^\star+2}}}
\right\}.
\]
Since \(\overline{\alpha}_t\) is decreasing in \(t\), the largest summand is
attained at \(t=1\). Thus
\[
\frac1T\sum_{t=1}^T
\frac{1}
{(1-\overline{\alpha}_t)^{\frac{k^\star}{2}+\frac{2}{k^\star+2}}}
\le
\frac{1}
{(1-\overline{\alpha}_1)^{\frac{k^\star}{2}+\frac{2}{k^\star+2}}}
=
\frac{1}
{(1-\alpha_1)^{\frac{k^\star}{2}+\frac{2}{k^\star+2}}}
\lesssim
T^{c_0\left(\frac{k^\star}{2}+\frac{2}{k^\star+2}\right)}.
\]
Similarly,
\[
\frac1T\sum_{t=1}^T
\frac{1}{(1-\overline{\alpha}_t)^{k^\star/2}}
\le
\frac{1}{(1-\overline{\alpha}_1)^{k^\star/2}}
=
\frac{1}{(1-\alpha_1)^{k^\star/2}}
\lesssim
T^{c_0k^\star/2}.
\]
Substituting these two estimates gives
\[
\varepsilon_{\mathsf{ddpm}\text{-}\mathsf{sc}}^2
\lesssim
\frac{R_0^{k^\star}\log(dR_0nT^{c_0})}{n}
\min\left\{
d^{\frac{2}{k^\star+2}}T^{c_0k^\star/2},
R_0^{\frac{4}{k^\star+2}}
T^{c_0\left(\frac{k^\star}{2}+\frac{2}{k^\star+2}\right)}
\right\}.
\]
Finally, since \(1-\overline{\alpha}_t\le1\), the same upper bound controls
\(\varepsilon_{\mathsf{ddim}\text{-}\mathsf{sc}}^2
=T^{-1}\sum_t(1-\overline{\alpha}_t)^2
\varepsilon_{\mathsf{score},t}^2\).
\end{proof}

\subsection{DDIM training complexity analysis (Proof of Theorem~\texorpdfstring{\ref{thm:ddim-training-guarantee}}{DDIM training guarantee})}
\label{prf:thm:ddim-training-guarantee}

In the DDIM setting, beyond score matching, we also need the higher-order
approximation guarantees required by Assumption~\ref{ass:ddim_score_matching}.
Because these smoothness requirements prevent us from using the construction in
\eqref{eq:plugin-score-tweedie-form}, we instead use the simplest empirical
plug-in construction. Its precise form and guarantees are stated in the
following proposition.

\begin{proposition}[Higher-order kernel approximation]\label{prop:higher-order-kernel-approximation}
Suppose Assumption~\ref{ass:bd_supp}, \ref{ass:low_dim} hold. For the DDIM analysis, let
\(s_t\) denote the unthresholded plug-in score estimator
\[
s_t(x)
\coloneqq
\frac{1}{1-\overline{\alpha}_t}
\frac{
\sum_{i=1}^n
\bigl(\sqrt{\overline{\alpha}_t}X_i^{\mathsf{sample}}-x\bigr)
\varphi_{1-\overline{\alpha}_t}
\bigl(\sqrt{\overline{\alpha}_t}X_i^{\mathsf{sample}}-x\bigr)}
{
\sum_{i=1}^n
\varphi_{1-\overline{\alpha}_t}
\bigl(\sqrt{\overline{\alpha}_t}X_i^{\mathsf{sample}}-x\bigr)} .
\]
Then the corresponding kernel posterior estimator satisfies the following
score and higher-order approximation bounds:
\begin{subequations}
\begin{align}
\varepsilon_{\mathsf{score},t}^2
&\lesssim
\frac{ R_0^{k^\star+\frac{4}{k^\star+2}}}
{n(1-\overline{\alpha}_t)^{\frac{k^\star}{2}+1+\frac{2}{k^\star+2}}}
\log^2\frac{dnR_0}{1-\overline{\alpha}_t},
\label{eq:high-order-score-bound}
\\
\varepsilon_{\mathsf{Jacobi},1,t}^2
&\lesssim
\frac{\overline{\alpha}_t^2
R_0^{k^\star+\frac{16}{k^\star+4}}}
{n(1-\overline{\alpha}_t)^{\frac{k^\star}{2}
+2+\frac{8}{k^\star+4}}}
\log^2\frac{dnR_0}{1-\overline{\alpha}_t},
\label{eq:high-order-jacobi-1-bound}
\\
\varepsilon_{\mathsf{Jacobi},2,t}^2
&\lesssim
\frac{\overline{\alpha}_t^2
R_0^{k^\star+\frac{16}{k^\star+4}}}
{n(1-\overline{\alpha}_t)^{\frac{k^\star}{2}
+2+\frac{8}{k^\star+4}}}
\log^2\frac{dnR_0}{1-\overline{\alpha}_t},
\label{eq:high-order-jacobi-2-bound}
\\
\varepsilon_{\mathsf{Hess},t}^2
&\lesssim
\frac{\overline{\alpha}_t^3
R_0^{k^\star+\frac{36}{k^\star+6}}}
{n(1-\overline{\alpha}_t)^{\frac{k^\star}{2}
+3+\frac{18}{k^\star+6}}}
\log^2\frac{dnR_0}{1-\overline{\alpha}_t}.
\label{eq:high-order-hessian-bound}
\end{align}
\end{subequations}
\end{proposition}
The proof of Proposition~\ref{prop:higher-order-kernel-approximation} is
deferred to Section~\ref{prf:prop:higher-order-kernel-approximation}.

\begin{proof}[Proof of Theorem~\ref{thm:ddim-training-guarantee}]
The proof follows directly from
Proposition~\ref{prop:higher-order-kernel-approximation}. We use
\(\log\frac{dnR_0}{1-\overline{\alpha}_t}\le \log(dR_0nT^{c_0})\). For the score
term, the definition of \(\varepsilon_{\mathsf{ddim}\text{-}\mathsf{sc}}\) gives
\[
\varepsilon_{\mathsf{ddim}\text{-}\mathsf{sc}}^2
\lesssim
\frac{R_0^{k^\star+\frac{4}{k^\star+2}}\log^2(dR_0nT^{c_0})}{nT}
\sum_{t=1}^T
\frac{\overline{\alpha}_t}
{(1-\overline{\alpha}_t)^{\frac{k^\star}{2}
+\frac{2}{k^\star+2}}}
.
\]
The average is bounded by the maximum summand, hence
\[
\frac1T\sum_{t=1}^T
\frac{\overline{\alpha}_t}
{(1-\overline{\alpha}_t)^{\frac{k^\star}{2}
+\frac{2}{k^\star+2}}}
\le
\frac{\overline{\alpha}_1}
{(1-\overline{\alpha}_1)^{\frac{k^\star}{2}+\frac{2}{k^\star+2}}}
=
\frac{\alpha_1}
{(1-\alpha_1)^{\frac{k^\star}{2}+\frac{2}{k^\star+2}}}
\lesssim
T^{c_0\left(\frac{k^\star}{2}+\frac{2}{k^\star+2}\right)},
\]
which yields the first claim.

For the two Jacobian terms, Assumption~\ref{ass:ddim_score_matching} averages
the pointwise errors with the weight \((1-\overline{\alpha}_t)^2\). Hence
\[
\varepsilon_{\mathsf{Jacobi},1}^2
\vee
\varepsilon_{\mathsf{Jacobi},2}^2
\lesssim
\frac{R_0^{k^\star+\frac{16}{k^\star+4}}
\log^2(dR_0nT^{c_0})}{nT}
\sum_{t=1}^T
\frac{\overline{\alpha}_t^2}
{(1-\overline{\alpha}_t)^{\frac{k^\star}{2}
+\frac{8}{k^\star+4}}}.
\]
Again bounding the average by the maximum summand gives
\[
\frac1T\sum_{t=1}^T
\frac{\overline{\alpha}_t^2}
{(1-\overline{\alpha}_t)^{\frac{k^\star}{2}
+\frac{8}{k^\star+4}}}
\le
\frac{\overline{\alpha}_1^2}
{(1-\overline{\alpha}_1)^{\frac{k^\star}{2}+\frac{8}{k^\star+4}}}
=
\frac{\alpha_1^2}
{(1-\alpha_1)^{\frac{k^\star}{2}+\frac{8}{k^\star+4}}}
\lesssim
T^{c_0\left(\frac{k^\star}{2}+\frac{8}{k^\star+4}\right)}.
\]
This gives the two Jacobian bounds.

Similarly, the Hessian term satisfies
\[
\varepsilon_{\mathsf{Hess}}^2
\lesssim
\frac{R_0^{k^\star+\frac{36}{k^\star+6}}
\log^2(dR_0nT^{c_0})}{nT}
\sum_{t=1}^T
\frac{\overline{\alpha}_t^3}
{(1-\overline{\alpha}_t)^{\frac{k^\star}{2}
+1+\frac{18}{k^\star+6}}}
.
\]
Finally, the same maximum-summand bound gives
\[
\frac1T\sum_{t=1}^T
\frac{\overline{\alpha}_t^3}
{(1-\overline{\alpha}_t)^{\frac{k^\star}{2}
+1+\frac{18}{k^\star+6}}}
\le
\frac{\overline{\alpha}_1^3}
{(1-\overline{\alpha}_1)^{\frac{k^\star}{2}+1+\frac{18}{k^\star+6}}}
=
\frac{\alpha_1^3}
{(1-\alpha_1)^{\frac{k^\star}{2}+1+\frac{18}{k^\star+6}}}
\lesssim
T^{c_0\left(\frac{k^\star}{2}+1+\frac{18}{k^\star+6}\right)}.
\]
This gives the claimed Hessian bound.
Finally, we aim to cope with the last argument, i.e.,
\[
\eta_t v^\top \nabla s_t(x) v \ge - \frac{1}{4}\|v\|_2^2.
\]
To this end, let's first apply Tweedie's formula to $s_t(x)$ (with the empirical density $\widehat{p}_t(\cdot)$ as the potential distribution). Indeed, it can be derived that
\[
\nabla s_t(x) = \nabla \left(\frac{\sqrt{\overline{\alpha}_t}}{1-\overline{\alpha}_t}\widehat{\mu}_{0\mid t}(x) - \frac{x}{1-\overline{\alpha}_t}\right) =
\frac{\overline{\alpha}_t}{(1-\overline{\alpha}_t)^2}\widehat{\Cov}_{0\mid t}(x) -
\frac{1}{1-\overline{\alpha}_t}I.
\]
Thus, for any $v\in \RB^d$, according to the semi-definite property of $\widehat{\Cov}_{0\mid t}(x)$, the following inequality holds for any $x \in \RB^d$,
\begin{align*}
    \eta_t^{\mathsf{ddim}} v^\top \nabla s_t(x) v 
    &= \frac{\overline{\alpha}_t\eta_t^{\mathsf{ddim}} 
    }{(1-\overline{\alpha}_t)^2}v^{\top}\widehat{\Cov}_{0\mid t}(x) v  - \frac{
        \eta_t^{\mathsf{ddim}}}{1-\overline{\alpha}_t}\|v\|_2^2\\
    &\ge -\frac{\eta_t^{
        \mathsf{ddim}
    }}{1-\overline{\alpha}_t}\|v\|_2^2
    \ge -\frac{1}{4}\|v\|_2^2,
\end{align*}
where the last inequality follows from the choice schedule of $
\eta_t^{\mathsf{ddim}}$ and \eqref{eq:basic-alphat-property}.
\end{proof}


\subsection{Proof of Proposition~\texorpdfstring{\ref{prop:low-dim_kernel_score_match}}{low-dim score matching}}
\label{prf:prop:low-dim_kernel_score_match}

Recall that the unified estimator can be written as
\[
s_t(x)
=
\frac{1}{1-\overline{\alpha}_t}
\frac{
\sum_{i=1}^n
\left(\sqrt{\overline{\alpha}_t}X_i^{\mathsf{sample}}-x\right)
\widetilde\varphi_{t,x}(X_i^{\mathsf{sample}})}
{D(x)} .
\]
For notational convenience in the decomposition below, define
\begin{align*}
\widehat m_t^{\rm th}(x)
&\coloneqq
\frac{
\sum_{i=1}^n
\left(\sqrt{\overline{\alpha}_t}X_i^{\mathsf{sample}}-x\right)
\widetilde\varphi_{t,x}(X_i^{\mathsf{sample}})}
{D(x)},
\notag\\
m_t^{\rm th}(x)
&\coloneqq
\frac{
\EE_{X_0}\!\left[
\left(\sqrt{\overline{\alpha}_t}X_0-x\right)
\widetilde\varphi_{t,x}(X_0)
\right]}
{p_{X_t}(x)},
\notag\\
m_t(x)
&\coloneqq
\sqrt{\overline{\alpha}_t}\mu_{0\mid t}(x)-x .
\end{align*}
Then \(s_t(x) = \widehat{m}_t^{\rm th}(x)/(1-\overline{\alpha}_t)\) and \(s_t^\star(x)=m_t(x)/(1-\overline{\alpha}_t)\). Hence, the score error first
can be split into the following three terms:
\begin{align}
\EE_{X_t,X^{\mathsf{sample}}}\|s_t(X_t)-s_t^\star(X_t)\|_2^2
&\lesssim
\frac{1}{(1-\overline{\alpha}_t)^2}
\EE_{X_t, X^{\mathsf{sample}}}\!\left[
\left\|
\widehat m_t^{\rm th}(X_t)
-m_t(X_t)
\right\|_2^2\mathbbm{1}\{X_t \notin \gT_{\overline{\alpha}_t}\}
\right]
\notag\\
&\quad+
\frac1{(1-\overline{\alpha}_t)^2}
\EE_{X_t,X^{\mathsf{sample}}}\!\left[
\left\|m_t^{\rm th}(X_t)-m_t(X_t)\right\|_2^2
\mathbbm{1}\{X_t \in \gT_{\overline{\alpha}_t}\}
\right]
\notag\\
&\quad+\frac1{(1-\overline{\alpha}_t)^2}
\EE_{X_t,X^{\mathsf{sample}}}\!\left[
\left\|
\widehat m_t^{\rm th}(X_t)
-m_t^{\rm th}(X_t)
\right\|_2^2\mathbbm{1}\{X_t \in \gT_{\overline{\alpha}_t}\}
\right].
\label{eq:score-threshold-first-split}
\end{align}
Here $\gT_{\overline{\alpha}_t}$ is the typical set
which is formulated at the begining of this section.
We first bound the first term in \eqref{eq:score-threshold-first-split}. 
To this end, note that by the generate mechanism of \(X_t\),
\(X_t = \sqrt{\overline{\alpha}_t}Y + \sqrt{1-\overline{\alpha}_t}
\Omega\), then it is easy to find that
\[
\left\|
\widehat m_t^{\rm th}(X_t)
-m_t(X_t)\right\|_2 \lesssim R_0 + \sqrt{1-\overline{
    \alpha}_t}\|\Omega\|_2.
\]
This implies that
\begin{align}
   &\EE_{X_t, X^{\mathsf{sample}}}\!\left[
\left\|
\widehat m_t^{\rm th}(X_t)
-m_t(X_t)
\right\|_2^2\mathbbm{1}\{X_t \notin \gT_{\overline{\alpha}_t}\}
\right] \notag\\
&\hspace{8em}\lesssim 
\EE_{Y,\Omega}\big[ (R_0^2 + (1-
\overline{\alpha}_t)\|\Omega\|_2^2)
\mathbbm{1}\{X_t \notin \gT_{
    \overline{\alpha}_t
}\}\big]\notag\\
&\hspace{8em} \overset{
    \text{(a)}
}{\lesssim} R_0^2 \PB(X_t \notin \gT_{\overline{\alpha}_t})
+ d(1-\overline{\alpha}_t)\PB\Big(Y 
\notin \bigcup_{i\in \gI}\gB_i\Big) + (1-
\overline{\alpha}_t)\EE[\|
\Omega\|_2^2 \mathbbm{1}\{\Omega \notin \gG\}]\notag\\
&\hspace{8em} \lesssim (R_0^2 + d(1-\overline{\alpha}_t))\left(
    \frac{\epsilon_t}{R_0}
\right)^{4k^\star} \lesssim \frac{R_0^2 
 (1-\overline{\alpha}_t)^2}{n},
\end{align}
where (a) follows from the nature of Gaussian distribution and
the penultimate inequality holds by using 
Lemma~\ref{lem:gT_alpha_typical}. And at the last inequality we
use the choice of $\epsilon_t$.

To analyze the second term of \eqref{eq:score-threshold-first-split}
, we need the following lemma.
\begin{lemma}\label{lem:threshold-score-error}
For any \(x\in\RB^d\), it holds that
\begin{align*}
\frac{
\EE_{X_0}\!\left[
\left\|\sqrt{\overline{\alpha}_t}X_0-x\right\|_2^2
\left(
\varphi_{1-\overline{\alpha}_t}
\bigl(x-\sqrt{\overline{\alpha}_t}X_0\bigr)
-\widetilde\varphi_{t,x}(X_0)
\right)^2
\right]}
{p_{X_t}(x)}
\le
\left(\frac{1-\overline{\alpha}_t}{R_0 n}\right)^{2d}.
\end{align*}
In particular, for any \(x\in\gT_{\overline{\alpha}_t}\), where
\(\gT_\alpha\) is defined in \eqref{eq:defn-T-alpha-appendix}, it holds that
\begin{align*}
\|m_t^{\rm th}(x) - m_t(x)\|_2
\le
\left(\frac{1-\overline{\alpha}_t}{R_0 n}\right)^{2d}.
\end{align*}
\end{lemma}
The proof of Lemma~\ref{lem:threshold-score-error} is deferred to
Section~\ref{prf:lem:threshold-score-error}.
Applying Lemma~\ref{lem:threshold-score-error} yields
\begin{align}
\EE_{X_t,X^{\mathsf{sample}}}\!\left[
\left\|m_t^{\rm th}(X_t)-m_t(X_t)\right\|_2^2
\mathbbm{1}\{X_t \in \gT_{\overline{\alpha}_t}\}
\right]
&\le
\left(\frac{1-\overline{\alpha}_t}{R_0 n}\right)^{4d} 
\le \frac{R_0^2(1-\overline{\alpha}_t)^4}{n^2}.
\label{eq:score-threshold-bias-bound}
\end{align}

It remains to bound the third term. Note
\(\gE_x=\{\widehat p_{X_t}(x)\ge p_{X_t}(x)/2\}\). Then
\begin{align}
&\frac1{(1-\overline{\alpha}_t)^2}
\EE_{X_t,X^{\mathsf{sample}}}\!\left[
\left\|
\widehat m_t^{\rm th}(X_t)-m_t^{\rm th}(X_t)
\right\|_2^2\mathbbm{1}\{X_t \in \gT_{\overline{\alpha}_t}\}
\right]\notag\\
&\hspace{5em}=
\underbrace{
\frac1{(1-\overline{\alpha}_t)^2}
\int_{\gT_{\overline{\alpha}_t}}
\EE_{X^{\mathsf{sample}}}\!\left[
\left\|
\widehat m_t^{\rm th}(x)-m_t^{\rm th}(x)
\right\|_2^2
\mathbbm{1}_{\gE_x}
\right]
p_{X_t}(x)\rd x
}_{\eqqcolon\,\gI_{\rm good}}
\notag\\
&\hspace{8em}+
\underbrace{
\frac1{(1-\overline{\alpha}_t)^2}
\int_{\gT_{\overline{\alpha}_t}}
\EE_{X^{\mathsf{sample}}}\!\left[
\left\|
\widehat m_t^{\rm th}(x)-m_t^{\rm th}(x)
\right\|_2^2
\mathbbm{1}_{\gE_x^c}
\right]
p_{X_t}(x)\rd x
}_{\eqqcolon\,\gI_{\rm bad}} .
\label{eq:score-threshold-empirical-split}
\end{align}

In the sequel, we bound \(\gI_{\rm good}\) and \(\gI_{\rm bad}\) separately.

\paragraph{Bound for \(\gI_{\rm good}\).}
For the thresholded empirical fluctuation, first write
\begin{align}
\widehat m_t^{\rm th}(x)-m_t^{\rm th}(x)
&=
\frac{1}{D(x)}
\sum_{i=1}^n
\Bigl[
\left(\sqrt{\overline{\alpha}_t}X_i^{\mathsf{sample}}-x\right)
\widetilde\varphi_{t,x}(X_i^{\mathsf{sample}})
\notag\\[-0.2em]
&\hspace{7em}
-m_t^{\rm th}(x)
\varphi_{1-\overline{\alpha}_t}
\bigl(x-\sqrt{\overline{\alpha}_t}X_i^{\mathsf{sample}}\bigr)
\Bigr] .
\label{eq:score-threshold-centered-expansion}
\end{align}
For each fixed \(x\), the summands in
\eqref{eq:score-threshold-centered-expansion} are centered:
\begin{align}
\EE_{X_0}\!\left[
\left(\sqrt{\overline{\alpha}_t}X_0-x\right)
\widetilde\varphi_{t,x}(X_0)
-m_t^{\rm th}(x)
\varphi_{1-\overline{\alpha}_t}
\bigl(x-\sqrt{\overline{\alpha}_t}X_0\bigr)
\right]=0.
\label{eq:score-threshold-centered-zero-mean}
\end{align}
Thus, on \(\gE_x\), we first lower bound the denominator $D(x)$ by
\(n p_{X_t}(x)/2\) and drop the indicator:
\begin{align}
&\EE_{X^{\mathsf{sample}}}\!\left[
\left\|
\widehat m_t^{\rm th}(x)-m_t^{\rm th}(x)
\right\|_2^2
\mathbbm{1}_{\gE_x}
\right]
\le
\frac{4}{p_{X_t}(x)^2}
\EE_{X^{\mathsf{sample}}}\!\left[
\left\|
n^{-1}\sum_{i=1}^n
\Bigl[
\left(\sqrt{\overline{\alpha}_t}X_i^{\mathsf{sample}}-x\right)
\widetilde\varphi_{t,x}(X_i^{\mathsf{sample}})
\right.\right.\notag\\[-0.2em]
&\hspace{13em}
\left.\left.
-m_t^{\rm th}(x)
\varphi_{1-\overline{\alpha}_t}
\bigl(x-\sqrt{\overline{\alpha}_t}X_i^{\mathsf{sample}}\bigr)
\Bigr]
\right\|_2^2
\right].
\label{eq:score-threshold-good-denominator}
\end{align}
By the zero-mean property in
\eqref{eq:score-threshold-centered-zero-mean} and independence of the training
samples, the last expectation equals the variance of one summand divided by
\(n\). Therefore,
\begin{align}
&\EE_{X^{\mathsf{sample}}}\!\left[
\left\|
\widehat m_t^{\rm th}(x)-m_t^{\rm th}(x)
\right\|_2^2
\mathbbm{1}_{\gE_x}
\right]
\notag\\
&\quad\le
\frac{4}{n p_{X_t}(x)^2}
\EE_{X_0}\!\left[
\left\|
\left(\sqrt{\overline{\alpha}_t}X_0-x\right)
\widetilde\varphi_{t,x}(X_0)
-m_t^{\rm th}(x)
\varphi_{1-\overline{\alpha}_t}
\bigl(x-\sqrt{\overline{\alpha}_t}X_0\bigr)
\right\|_2^2
\right]
\notag\\
&\quad\lesssim
\frac{1}{n p_{X_t}(x)^2}
\EE_{X_0}\!\left[
\left\|\sqrt{\overline{\alpha}_t}X_0-x\right\|_2^2
\left(
\varphi_{1-\overline{\alpha}_t}
\bigl(x-\sqrt{\overline{\alpha}_t}X_0\bigr)
-\widetilde\varphi_{t,x}(X_0)
\right)^2
\right]
\notag\\
&\qquad+
\frac{1}{n p_{X_t}(x)^2}
\EE_{X_0}\!\left[
\left\|\sqrt{\overline{\alpha}_t}X_0-x-m_t(x)\right\|_2^2
\varphi_{1-\overline{\alpha}_t}^2
\bigl(x-\sqrt{\overline{\alpha}_t}X_0\bigr)
\right]
\notag\\
&\qquad+
\frac{1}{n p_{X_t}(x)^2}
\left\|m_t^{\rm th}(x)-m_t(x)\right\|_2^2
\EE_{X_0}\!\left[
\varphi_{1-\overline{\alpha}_t}^2
\bigl(x-\sqrt{\overline{\alpha}_t}X_0\bigr)
\right].
\label{eq:score-threshold-good-draft-decomposition}
\end{align}
Notice that \eqref{eq:score-threshold-good-draft-decomposition} holds for
every \(x\). We now take the expectation on both sides of
\eqref{eq:score-threshold-good-draft-decomposition} over
\(X_t\in\gT_{\overline{\alpha}_t}\).
For the first term on the right hand side of
\eqref{eq:score-threshold-good-draft-decomposition},
Lemma~\ref{lem:threshold-score-error} gives
\begin{align}
&\int_{\gT_{\overline{\alpha}_t}}
\frac{1}{np_{X_t}(x)}
\EE_{X_0}\!\left[
\left\|\sqrt{\overline{\alpha}_t}X_0-x\right\|_2^2
\left(
\varphi_{1-\overline{\alpha}_t}
\bigl(x-\sqrt{\overline{\alpha}_t}X_0\bigr)
-\widetilde\varphi_{t,x}(X_0)
\right)^2
\right]\rd x
\lesssim
\frac{R_0^2(1-\overline{\alpha}_t)^2}{n};
\label{eq:score-threshold-good-threshold-term}
\end{align}
For the third term on the right hand side of
\eqref{eq:score-threshold-good-draft-decomposition}, we additionally use the
following density-ratio estimate.

\begin{lemma}\label{lem:pure-density-ratio}
Suppose the intrinsic dimension assumption holds. Then, for every noise level
\(t\) in the regime considered, the following term satisfies
\[
\int_{\RB^d}\int_{\RB^d}
\frac{
\varphi_{1-\overline{\alpha}_t}^2
\bigl(x-\sqrt{\overline{\alpha}_t}x_0\bigr)
p_0(x_0)}
{p_{X_t}(x)}
\rd x_0\rd x
\le
C\left(\frac{R_0\sqrt{\overline{\alpha}_t}
}{\sqrt{1-\overline{\alpha}_t}}\right)^{k^\star}.
\]
\end{lemma}
The proof of Lemma~\ref{lem:pure-density-ratio} is deferred to
Section~\ref{prf:lem:pure-density-ratio}.
Combining Lemma~\ref{lem:threshold-score-error} and
Lemma~\ref{lem:pure-density-ratio} gives
\begin{align}
&\int_{\gT_{\overline{\alpha}_t}}
\frac{
\left\|m_t^{\rm th}(x)-m_t(x)\right\|_2^2
}{n p_{X_t}(x)}
\EE_{X_0}\!\left[
\varphi_{1-\overline{\alpha}_t}^2
\bigl(x-\sqrt{\overline{\alpha}_t}X_0\bigr)
\right]\rd x
\lesssim
\frac{R_0^2(1-\overline{\alpha}_t)^2}{n}.
\label{eq:score-threshold-good-bias-density-term}
\end{align}
For the second term of the right hand side of
\eqref{eq:score-threshold-good-draft-decomposition},
note that we have
\[
\frac{\varphi_{1-\overline{\alpha}_t}^2(x -
\sqrt{\overline{\alpha}_t}X_0)}{p_{X_t}^2(x)}p_{X_0}(X_0)
= \frac{\varphi_{1-\overline{\alpha}_t}(x -
\sqrt{\overline{\alpha}_t}X_0)}{p_{X_t}(x)}p_{0\mid t}(X_0 \mid x).
\]
Combining the definition of $m_t(x)$ and Jensen's inequality yields
\begin{align}
\int_{\gT_{\overline{\alpha}_t}}&
 \frac{1}{(1-\overline{\alpha}_t)^2n p_{X_t}(x)}
\EE_{X_0}\!\left[
\left\|\sqrt{\overline{\alpha}_t}X_0-x-m_t(x)\right\|_2^2
\varphi_{1-\overline{\alpha}_t}^2
\bigl(x-\sqrt{\overline{\alpha}_t}X_0\bigr)
\right]\rd x \notag\\
&\le
\frac4{n(1-\overline{\alpha}_t)^2}
\mathop{\EE}\limits_{X_t}
\bigg[
\mathop{\EE}\limits_{Z\sim\PB_{0\mid t}(\cdot\mid X_t)}
\bigg[
\|Z-\mu_{0\mid t}(X_t)\|_2^2
\frac{\varphi_{1-\overline{\alpha}_t}
\bigl(X_t-\sqrt{\overline{\alpha}_t}Z\bigr)}
{p_{X_t}(X_t)}
\bigg]\mathbbm{1}\{X_t \in  \gT_{\overline{\alpha}_t}\}
\bigg] \notag\\
&\le
\frac{8}{n(1-\overline{\alpha}_t)^2}
\underbrace{
\mathop{\EE}\limits_{Y,\Omega}
\bigg[
\mathop{\EE}\limits_{Z\sim\PB_{0\mid t}(\cdot\mid X_t)}
\bigg[
\|Z-Y\|_2^2
\frac{\varphi_{1-\overline{\alpha}_t}
\bigl(X_t-\sqrt{\overline{\alpha}_t}Z\bigr)}
{p_{X_t}(X_t)}
\bigg]\mathbbm{1}\{X_t \in  \gT_{\overline{\alpha}_t}\}
\bigg]
}_{\eqqcolon\,\gJ_1}
\label{eq:good-raw}\\
&+
\frac{8}{n(1-\overline{\alpha}_t)^2}
\underbrace{
\mathop{\EE}\limits_{Y,\Omega}
\bigg[
\mathop{\EE}\limits_{V\sim\PB_{0\mid t}(\cdot\mid X_t)}
\bigl[
\|Y-V\|_2^2
\bigr]
\mathop{\EE}\limits_{Z\sim\PB_{0\mid t}(\cdot\mid X_t)}
\bigg[
\frac{\varphi_{1-\overline{\alpha}_t}
\bigl(X_t-\sqrt{\overline{\alpha}_t}Z\bigr)}
{p_{X_t}(X_t)}
\bigg]\mathbbm{1}\{X_t \in  \gT_{\overline{\alpha}_t}\}
\bigg]
}_{\eqqcolon\,\gJ_2}.                         \notag
\end{align}
Here \(Y\sim\PB_{X_0}\), \(\Omega\sim\gN(0,I_d)\), and
\(X_t=\sqrt{\overline{\alpha}_t}\, Y +\sqrt{1-\overline{\alpha}_t}\,\Omega\). The
last step uses \(\|Z-\mu_{0\mid t}(X_t)\|_2^2
\le 2\|Z-Y\|_2^2+2\|Y-\mu_{0\mid t}(X_t)\|_2^2\), as well as Jensen's
inequality
\(\|Y-\mu_{0\mid t}(X_t)\|_2^2
\le \mathop{\EE}_{V\sim\PB_{0\mid t}(\cdot\mid X_t)}[\|Y-V\|_2^2]\).

To analyze the two terms \(\gJ_1\) and \(\gJ_2\) above, we need the following
lemmas.
In the next two lemmas, \(A\lesssim_\beta B\) means that
\(A\le C_\beta B\) for a constant \(C_\beta\) depending only on the fixed
integer \(\beta\) and on universal constants.
\begin{lemma}\label{lem:J1-bound}
Let \(\beta\ge2\) be an integer. Recall the generative process of $X_t$,
i.e., \(X_t = \sqrt{\overline{\alpha}_t}\,Y + \sqrt{1-\overline{\alpha}_t}\,
\Omega\), where \(Y\sim \PB_{X_0}\) and
\(\Omega\sim\gN(0,I_d)\). Define
\begin{align*}
\gJ_{1,\beta}
&\coloneqq
\mathop{\EB}\limits_{Y,\Omega}
\bigg[
\mathop{\EB}\limits_{Z \sim \PB_{0\mid t}(\cdot\mid X_t)}
\bigg[
\|Z-Y\|_2^\beta
\frac{\varphi_{1-\overline{\alpha}_t}
\bigl(X_t-\sqrt{\overline{\alpha}_t}Z\bigr)}
{p_{X_t}(X_t)}
\bigg]\mathbbm{1}\{X_t \in  \gT_{\overline{\alpha}_t}\}
\bigg].
\end{align*}
Then
\[
\gJ_{1,\beta}
\lesssim_\beta
\left(k^\star \log \frac{R_0 nd}{\overline{\alpha}_t
(1-\overline{\alpha}_t)}
\right)^{\beta/2}R_0^{k^\star}\gamma_t^{k^\star - \beta}
\;+\;
\frac{R_0^\beta \overline{\alpha}_t^\beta
(1-\overline{\alpha}_t)^\beta}{n}.
\]
\end{lemma}
The proof of Lemma~\ref{lem:J1-bound} is deferred to
Section~\ref{prf:lem:J1-bound}.

\begin{lemma}\label{lem:J2-bound}
Under the same notation as Lemma~\ref{lem:J1-bound}, define
\begin{align*}
\gJ_{2,\beta}
&\coloneqq
\mathop{\EB}\limits_{Y,\Omega}
\bigg[
\mathop{\EB}\limits_{V \sim \PB_{0\mid t}(\cdot\mid X_t)}
\bigl[
\|Y-V\|_2^\beta
\bigr]
\mathop{\EB}\limits_{Z \sim \PB_{0\mid t}(\cdot\mid X_t)}
\bigg[
\frac{\varphi_{1-\overline{\alpha}_t}
\bigl(X_t-\sqrt{\overline{\alpha}_t}Z\bigr)}
{p_{X_t}(X_t)}
\bigg]\mathbbm{1}\{X_t \in  \gT_{\overline{\alpha}_t}\}
\bigg].
\end{align*}
Then
\[
\gJ_{2,\beta}
\lesssim_\beta
\left(k^\star \log \frac{R_0 nd}{\overline{\alpha}_t
(1-\overline{\alpha}_t)}
\right)^{\beta/2}R_0^{k^\star}\gamma_t^{k^\star - \beta}
\;+\;
\frac{R_0^\beta \overline{\alpha}_t^\beta
(1-\overline{\alpha}_t)^\beta}{n}.
\]
\end{lemma}
The proof of Lemma~\ref{lem:J2-bound} is deferred to
Section~\ref{prf:lem:J2-bound}.

By Lemmas~\ref{lem:J1-bound} and \ref{lem:J2-bound} with \(\beta=2\),
\begin{subequations}\label{eq:score-threshold-good-J-bounds}
\begin{align}
\gJ_1
&\lesssim
\frac{k^\star R_0^{k^\star}\log\frac{dnR_0}{1-\overline{\alpha}_t}}
{\overline{\alpha}_t(1-\overline{\alpha}_t)^{k^\star/2-1}}
+\frac{R_0^2(1-\overline{\alpha}_t)^2}{n};
\label{eq:J1-bound}
\\
\gJ_2
&\lesssim
\frac{k^\star R_0^{k^\star}\log\frac{dnR_0}{1-\overline{\alpha}_t}}
{\overline{\alpha}_t(1-\overline{\alpha}_t)^{k^\star/2-1}}
+\frac{R_0^2(1-\overline{\alpha}_t)^2}{n}.
\label{eq:J2-bound}
\end{align}
\end{subequations}
Combining
\eqref{eq:score-threshold-good-draft-decomposition}--\eqref{eq:score-threshold-good-J-bounds},
we get
\begin{align}
\gI_{\rm good}
&\lesssim
\frac{k^\star R_0^{k^\star}\log\frac{dnR_0}{1-\overline{\alpha}_t}}
{n\overline{\alpha}_t(1-\overline{\alpha}_t)^{k^\star/2+1}}
+\frac{R_0^2(1-\overline{\alpha}_t)^2}{n}.       \label{eq:good-bound}
\end{align}

\paragraph{Bound for \(\gI_{\rm bad}\).}
Set
\(
\rho_t>0
\quad\text{to be chosen later}
\).
For \(x=\sqrt{\overline{\alpha}_t}y+\sqrt{1-\overline{\alpha}_t}\omega\),
we can write the following decomposition,
\[
\widehat{m}^{\rm th}_t(x)-m_t^{\rm th}(x)
=\gA_t(y,\omega)+\gB_t(y,\omega)
+ \gC_t(y,\omega),
\]
where the three terms are defined as follows:
\begin{subequations}
\begin{align*}
\gA_t(y,\omega)&\coloneqq
\frac{1}
{D(x)}
\sum_{r=1}^n
\bigg\{\Big((\sqrt{\overline{\alpha}_t}
X_r^{\mathsf{sample}} - x)h_t(\|\sqrt{\overline{\alpha}_t}
X_r^{\mathsf{sample}} - x\|_2)
-m_t^{\rm th}(x)\Big)
{\varphi}_{1-\overline{\alpha}_t}(
    \sqrt{\overline{\alpha}_t}X_r^{\mathsf{sample}} - x)\\
&\hspace{15em}\times\mathbbm{1}\{X_r^{\mathsf{sample}}\in\gB(y,\rho_t)\}
\mathbbm{1}\{\mu_{0\mid t}(x)\in\gB(y,\rho_t\wedge a_t)\}\bigg\};
\\
\gB_t(y,\omega)&\coloneqq
\frac{1}
{D(x)}
\sum_{r=1}^n
\bigg\{\Big((\sqrt{\overline{\alpha}_t}
X_r^{\mathsf{sample}} - x)h_t(\|\sqrt{\overline{\alpha}_t}
X_r^{\mathsf{sample}} - x\|_2)
-m_t^{\rm th}(x)\Big)
{\varphi}_{1-\overline{\alpha}_t}(
    \sqrt{\overline{\alpha}_t}X_r^{\mathsf{sample}} - x)\\
&\hspace{15em}\times\mathbbm{1}\{X_r^{\mathsf{sample}}\notin\gB(y,\rho_t)\}
\mathbbm{1}\{\mu_{0\mid t}(x)\in\gB(y,\rho_t\wedge a_t)\}\bigg\};
\\
\gC_t(y,\omega)&\coloneqq
\frac{1}
{D(x)}
\bigg\{\sum_{r=1}^n
\Big((\sqrt{\overline{\alpha}_t}
X_r^{\mathsf{sample}} - x)h_t(\|\sqrt{\overline{\alpha}_t}
X_r^{\mathsf{sample}} - x\|_2)
-m_t^{\rm th}(x)\Big)
{\varphi}_{1-\overline{\alpha}_t}(
    \sqrt{\overline{\alpha}_t}X_r^{\mathsf{sample}} - x)
    \bigg\}\\
&\hspace{15em}\times\mathbbm{1}\{\mu_{0\mid t}(x)\notin\gB(y,\rho_t\wedge a_t)\}.
\end{align*}
\end{subequations}
Here \(\gB(y,r)\) denotes the closed \(\ell_2\)-neighborhood
defined in the notation paragraph, and should not be confused with
\(\gB_t(y,\omega)\) above.
By the triangle inequality, the bad term decomposes as
\begin{align}
\gI_{\rm bad}
&\lesssim
\underbrace{\frac1{(1-\overline{\alpha}_t)^2}
\int_{\gT_{\overline{\alpha}_t}}
\EE_{X^{\mathsf{sample}}}\!\left[
\|\gA_t(y,\omega)\|_2^2\mathbbm{1}_{\gE_x^c}
\right]\varphi_1(\omega)p_{X_0}(y)\rd\omega\rd y}_{\eqqcolon\,\gI_{\rm bad,A}}
\notag\\
&\quad+
\underbrace{\frac1{(1-\overline{\alpha}_t)^2}
\int_{\gT_{\overline{\alpha}_t}}
\EE_{X^{\mathsf{sample}}}\!\left[
\|\gB_t(y,\omega)\|_2^2\mathbbm{1}_{\gE_x^c}
\right]\varphi_1(\omega)p_{X_0}(y)\rd\omega\rd y}_{\eqqcolon\,\gI_{\rm bad,B}}
\notag\\
&\quad+
\underbrace{\frac1{(1-\overline{\alpha}_t)^2}
\int_{\gT_{\overline{\alpha}_t}}
\EE_{X^{\mathsf{sample}}}\left[\|\gC_t(y,\omega)\|_2^2\mathbbm{1}_{\gE_x^c}\right]
\varphi_1(\omega)p_{X_0}(y)\rd\omega\rd y}_{\eqqcolon\,\gI_{\rm bad,C}}.
\label{eq:bad-triangle-split}
\end{align}
We first bound \(\gI_{\rm bad,A}\). Since
on the support of every summand, we have \(h_t(\|
\sqrt{\overline{\alpha}_t}X_r^{\mathsf{sample}} - x\|_2) = 1\). Hence
\begin{align}
    \norm{\gA_t(y,\omega)}_2 &\le \sqrt{\overline{\alpha}_t}\norm{
        X_r^{\mathsf{sample}} - y
    }_2 + \sqrt{\overline{\alpha}_t}\norm{y - \mu_{0\mit t}(x)}_2
    + \|m_t(x) - m_t^{\rm th}(x)\|_2\notag\\
    &\le 2\rho_t + \left(\frac{1-\overline{\alpha}_t}{
        R_0n
    }\right)^{2d} \le 3\rho_t,
\label{eq:gA-bad-bound}
\end{align}
where the penultimate inequality holds by using 
Lemma~\ref{lem:threshold-score-error} and the last inequality holds
as long as \(\rho_t \ge \left(\frac{1-\overline{\alpha}_t}{
        R_0n
    }\right)^{2d}\).
Further, we use the following lemma, which controls the event where the
empirical density deviates significantly from its population counterpart.

\begin{lemma}\label{lem:intrinsic-density-matching-bound}
Suppose the intrinsic dimension assumption holds. Then, for every noise level $t$ in the
regime considered,
\begin{align*}
&\int_{y,\omega}
\PP\!\left(\widehat p_{X_t}(x)<\frac12p_{X_t}(x)\right)
\varphi_1(\omega)p_{X_0}(y)\rd\omega\rd y
\lesssim
\frac{R_0^{k^\star}}
{n(1-\overline{\alpha}_t)^{k^\star/2}},
\end{align*}
where \(x=\sqrt{\overline{\alpha}_t}y+\sqrt{1-\overline{\alpha}_t}\omega\).
\end{lemma}
The proof of Lemma~\ref{lem:intrinsic-density-matching-bound} is deferred to
Section~\ref{prf:lem:intrinsic-density-matching-bound}.

Combining \eqref{eq:gA-bad-bound} with
Lemma~\ref{lem:intrinsic-density-matching-bound}, we obtain
\begin{align}
\gI_{\rm bad,A}
&\lesssim
\frac{\rho_t^2}{(1-\overline{\alpha}_t)^2}
\int_{y,\omega}
\PP\!\left(\widehat p_{X_t}(x)<\frac12p_{X_t}(x)\right)
\varphi_1(\omega)p_{X_0}(y)\rd\omega\rd y
\notag\\
&\lesssim
\frac{\rho_t^2R_0^{k^\star}}
{n(1-\overline{\alpha}_t)^{k^\star/2+2}},
\label{eq:bad-A-bound}
\end{align}

We next turn to \(\gI_{\rm bad,B}\).
Note that given $\|\sqrt{\overline{\alpha}_t
}X_r^{\mathsf{sample}} - x\|_2 \le 2a_t$, $
\|\mu_{0\mid t}(x)- y\|_2 \le a_t$
 and $\norm{\sqrt{\overline{
    \alpha
}_t}y - x}_2 \le a_t$, it holds that
\begin{align}
&\Big\|\big(\sqrt{\overline{\alpha}_t}X_r^{\mathsf{sample}}
-x\big)h_t(\|
\sqrt{\overline{\alpha}_t}X_r^{\mathsf{sample}} - x\|_2)
  - m_t^{\rm th}(x)\Big\|_2\notag\\
&\hspace{8em}\le
\|\sqrt{\overline{\alpha}_t
}X_r^{\mathsf{sample}} - x\|_2
+ \norm{m_t(x) - m_t^{\rm th}(x)}_2\notag\\
&\hspace{11em}+ \norm{\sqrt{\overline{
    \alpha}_t}y - x - m_t(x)}_2  + \norm{
        \sqrt{\overline{\alpha}_t}y -x
    }_2\notag\\
    &\hspace{8em}\le 3a_t + \rho_t\wedge a_t + \left(\frac{
        1-\overline{\alpha}_t
    }{R_0n}\right)^{2d} \le 5a_t. \label{eq:gI_bad-B_one}
\end{align}
Further, according to the definition of the threshold
function $h_t(\cdot)$, $h_t(\|\sqrt{\overline{\alpha}_t
}X_r^{\mathsf{sample}} - x\|_2) = 0$ if $\|
\sqrt{\overline{\alpha}_t
}X_r^{\mathsf{sample}} - x\|_2 > 2a_t$. And for $x\in \gT_{
    \overline{\alpha}_t}$, $\|\sqrt{\overline{\alpha}_t}
    y - x\|_2 = \sqrt{1-\overline{\alpha}_t}\|\omega\|_2 \le a_t$.
This means all the conditions which guarantee 
\eqref{eq:gI_bad-B_one} hold. Now, define
\[
\mathsf{R}_t(y,\omega)\coloneqq
\frac{
\sum_{r=1}^n
\mathbbm{1}\{\|X_r^{\mathsf{sample}}-y\|_2>\rho_t\}
\varphi_{1-\overline{\alpha}_t}
\bigl(x-\sqrt{\overline{\alpha}_t}X_r^{\mathsf{sample}}\bigr)}
{D(x)}.
\]
We have
\(\|\gB_t(y,\omega)\|_2\le 5a_t\mathsf{R}_t(y,\omega)
\).
For the empirical ratio \(\mathsf{R}_t(y,\omega)\), we have the following
result to help bound it.

\begin{lemma}\label{lem:bad-B-tail-ratio}
Let \(x=\sqrt{\overline{\alpha}_t}y+\sqrt{1-\overline{\alpha}_t}\omega\) and
\(\gamma_t=\sqrt{\overline{\alpha}_t/(1-\overline{\alpha}_t)}\). Let
\(\{\gC_i\}_{i=1}^{N_{\lambda_t}}\) be a measurable partition induced by a
\(\lambda_t\)-net of \(\gX_{\mathsf{data}}\), and suppose
\(
0<\lambda_t\le \frac{\rho_t}{2},~
\rho_t\ge
32\gamma_t^{-1}\sqrt{\log\frac{R_0nd}{
    \overline{\alpha}_t(1-\overline{\alpha}_t)}}.
\)
Define
\[
\mathsf{R}_t(y,\omega)\coloneqq
\frac{
\sum_{r=1}^n
\mathbbm{1}\{\|X_r^{\mathsf{sample}}-y\|_2>\rho_t\}
\varphi_{1-\overline{\alpha}_t}
\bigl(x-\sqrt{\overline{\alpha}_t}X_r^{\mathsf{sample}}\bigr)}
{D(x)}.
\]
Then
\begin{align*}
\int_{y,\omega}
\EE\!\left[
\mathsf{R}_t(y,\omega)^2
\right]
\varphi_1(\omega)p_{X_0}(y)\rd\omega\rd y
\lesssim
\frac{N_{\lambda_t}}{n}
+
\frac{\overline{\alpha}_t^6(1-\overline{\alpha}_t)^6}{n^2}.
\end{align*}
\end{lemma}
The proof of Lemma~\ref{lem:bad-B-tail-ratio} is deferred to
Section~\ref{prf:lem:bad-B-tail-ratio}.

Applying Lemma~\ref{lem:bad-B-tail-ratio} makes us obtain
\begin{align}
\gI_{\rm bad,B}
&\lesssim
a_t^2 \EE_{Y, \Omega}[\mathsf{R}_t(Y,\Omega)] 
\lesssim
\frac{a_t^2}{(1-\overline{\alpha}_t)^2}
\frac1n\left(\frac{R_0}{\lambda_t}\right)^{k^\star}
+\frac{R_0^2(1-\overline{\alpha}_t)^2}{n}.
\label{eq:bad-B-combined-rho}
\end{align}
Finally, for $\gI_{\rm bad, C}$, using Lemma~\ref{lem:posterior_norm}
 leads to
\begin{align}
    \gI_{\rm bad, C} &\lesssim 
     dR_0^2\, \PB\big(\|Y - \mu_{0\mid t}(X_t)\|_2>
      \rho_t\wedge a_t \mid \gT_{\overline{\alpha}_t}
      \big) \le dR_0^2\left(
        \frac{1-\overline{\alpha}_t}{R_0n}
     \right)^{4k^\star} \le \frac{R_0^2(1-\overline{\alpha}_t)^4}{n^2}.
     \label{eq:bad-mu-large}
\end{align}
Here the second inequality holds provided that \(\rho_t \gtrsim
\sqrt{k(1-\overline{\alpha}_t)\log \frac{dnR_0}{1-\overline{\alpha}_t}}\).
We now take \(\lambda_t=\rho_t/2\) and combine
\eqref{eq:bad-A-bound} -- \eqref{eq:bad-mu-large}.
 This gives
\begin{align}
\gI_{\rm bad} &\lesssim 
\gI_{\rm bad,A}+\gI_{\rm bad,B} + \gI_{\rm bad, C}\notag\\
&\lesssim
\frac{R_0^{k^\star}\rho_t^2}
{n(1-\overline{\alpha}_t)^{k^\star/2+2}}
+
\frac{R_0^{k^\star}a_t^2}
{n(1-\overline{\alpha}_t)^2\rho_t^{k^\star}}
+
\frac{k^\star R_0^{k^\star}\log\frac{dnR_0}{1-\overline{\alpha}_t}}
{n\overline{\alpha}_t(1-\overline{\alpha}_t)^{k^\star/2+1}}
+\frac{R_0^2(1-\overline{\alpha}_t)^4}{n^2}.
\label{eq:bad-AB-before-rho-opt}
\end{align}
The first two terms are minimized by balancing them:
\begin{align}
\frac{R_0^{k^\star}\rho_t^2}
{n(1-\overline{\alpha}_t)^{k^\star/2+2}}
\asymp
\frac{R_0^{k^\star}a_t^2}
{n(1-\overline{\alpha}_t)^2\rho_t^{k^\star}}
\quad\Longrightarrow\quad
\rho_t^{k^\star+2}\asymp
a_t^2(1-\overline{\alpha}_t)^{k^\star/2}.
\label{eq:rho-balance}
\end{align}
According to this, we choose that
\(
\rho_t \asymp
k^\star a_t^{\frac{2}{k^\star+2}}
(1-\overline{\alpha}_t)^{k^\star/(2(k^\star+2))}\).
Substituting this choice into the algebraic part of
\eqref{eq:bad-AB-before-rho-opt} and integrating with
the definition of $a_t$ yields
\begin{align}
\gI_{\rm bad}
&\lesssim
\left(\frac{a_t^2}{1-\overline{\alpha}_t}\right)^{\frac{2}{k^\star+2}}
\frac{R_0^{k^\star}}
{n(1-\overline{\alpha}_t)^{k^\star/2+1}}
+\frac{k^\star R_0^{k^\star}\log\frac{dnR_0}{1-\overline{\alpha}_t}}
{n\overline{\alpha}_t(1-\overline{\alpha}_t)^{k^\star/2+1}}
+\frac{R_0^2(1-\overline{\alpha}_t)^4}{n^2}
\notag\\
&\lesssim
\min\left\{
    \frac{R_0^2}{1-\overline{\alpha}_t}, d\log
    \frac{dnR_0}{1-\overline{\alpha}_t}
\right\}^{\frac{2}{k^\star + 2}}\frac{k^\star R_0^{k^\star}}
{n(1-\overline{\alpha}_t)^{k^\star/2+1}}
+\frac{R_0^2(1-\overline{\alpha}_t)^4}{n^2}.
\label{eq:bad-AB-optimized}
\end{align}

Substituting \eqref{eq:good-bound} and
\eqref{eq:bad-AB-optimized} into \eqref{eq:score-threshold-empirical-split},
and then using \eqref{eq:score-threshold-bias-bound}, gives the current
combined bound
\begin{align}
\EE\|s_t(X_t)-s_t^\star(X_t)\|_2^2
&\lesssim
\min\left\{
    \frac{R_0^2}{1-\overline{\alpha}_t}, d\log
    \frac{dnR_0}{1-\overline{\alpha}_t}
\right\}^{\frac{2}{k^\star + 2}}\frac{R_0^{k^\star}}
{n(1-\overline{\alpha}_t)^{k^\star/2+1}}.
\label{eq:score-bound-current}
\end{align}

\input{higher_order_approximation.tex}

\input{training_complexity_lemma_proofs.tex}

%% file: higher_order_approximation.tex
\subsection{Proof of Proposition~\texorpdfstring{\ref{prop:higher-order-kernel-approximation}}{higher-order kernel approximation}}
\label{prf:prop:higher-order-kernel-approximation}

We first rewrite the approximation errors in
Assumption~\ref{ass:ddim_score_matching} in terms of posterior moment
estimation. To avoid repeating the score-matching argument at several moment
orders, we prove the following general posterior mean approximation bound.

\begin{lemma}\label{lem:posterior-mean-high-moment}
Suppose Assumptions~\ref{ass:bd_supp} and~\ref{ass:low_dim} hold. For
the unthresholded empirical posterior mean
\(
\widehat\mu_{0\mid t}(x)
\coloneqq
\widehat{\EE}_{0\mid t,x}[X_0],
\)
the following bound holds for every \(1\le \beta\le3\) and every noise level
\(t\) in the regime considered:
\begin{align}
&\EE_{X_t,X^{\mathsf{sample}}}\!\left[
\left\|\widehat\mu_{0\mid t}(X_t)-\mu_{0\mid t}(X_t)\right\|_2^{2\beta}
\right]
\notag\\
&\qquad\lesssim
\frac{
R_0^{k^\star+\frac{4\beta^2}{k^\star+2\beta}}
}
{n(1-\overline{\alpha}_t)^{\frac{k^\star}{2}
-\frac{\beta k^\star}{k^\star+2\beta}}}
\log^\beta\frac{dnR_0}{1-\overline{\alpha}_t}
+
\frac{R_0^{2\beta}(1-\overline{\alpha}_t)^{2\beta}}{n^2}.
\label{eq:posterior-mean-high-general}
\end{align}
\end{lemma}
The proof of Lemma~\ref{lem:posterior-mean-high-moment} is deferred to
Section~\ref{prf:lem:posterior-mean-high-moment}.

Taking \(\beta=1\) in Lemma~\ref{lem:posterior-mean-high-moment} and using
Tweedie's formula yields \eqref{eq:high-order-score-bound}. It remains to
prove the higher-order bounds.

\subsubsection{Frobenius-Norm Jacobian Approximation}

The Frobenius-norm Jacobian error is a posterior covariance estimation
error. Define
\[
\widehat\Cov_{0\mid t}(x)
\coloneqq
\widehat{\EE}_{0\mid t,x}\!\left[
\left(X_0-\widehat\mu_{0\mid t}(x)\right)
\left(X_0-\widehat\mu_{0\mid t}(x)\right)^\top
\right]
\]
and let \(\Cov_{0\mid t}(x)\) denote the corresponding population posterior
covariance.
By Tweedie's formula~\eqref{eq:score_to_posterior_new} and the unthresholded
definition of \(s_t\), the kernel score estimator satisfies
\[
s_t(x)-s_t^\star(x)
=
\frac{\sqrt{\overline{\alpha}_t}}{1-\overline{\alpha}_t}
\left(\widehat\mu_{0\mid t}(x)-\mu_{0\mid t}(x)\right).
\]
Differentiating the two Tweedie representations gives
\begin{equation}\label{eq:higher-order-jacobian-error-identity}
\frac{\partial s_t(x)}{\partial x}
-
\frac{\partial s_t^\star(x)}{\partial x}
=
\frac{\overline{\alpha}_t}{(1-\overline{\alpha}_t)^2}
\left(\widehat\Cov_{0\mid t}(x)-\Cov_{0\mid t}(x)\right).
\end{equation}
Therefore the first Jacobian error in
Assumption~\ref{ass:ddim_score_matching} can be formulated as
\begin{equation}\label{eq:higher-order-jacobian-F-form}
\varepsilon_{\mathsf{Jacobi},1,t}^2
=
\frac{\overline{\alpha}_t^2}{(1-\overline{\alpha}_t)^4}
\EE_{X_t}\!\left[
\left\|\widehat\Cov_{0\mid t}(X_t)-\Cov_{0\mid t}(X_t)\right\|_{\mathrm F}^2
\right].
\end{equation}
Thus it remains to control the covariance-error term in
\eqref{eq:higher-order-jacobian-F-form}.
For the Frobenius bound, we center the empirical covariance at the population
posterior mean:
\begin{equation}\label{eq:centered-empirical-cov-def}
\widetilde\Cov_{0\mid t}(x)
\coloneqq
\frac{
\sum_{r=1}^n
\left(X_r^{\mathsf{sample}}-\mu_{0\mid t}(x)\right)
\left(X_r^{\mathsf{sample}}-\mu_{0\mid t}(x)\right)^\top
\varphi_{1-\overline{\alpha}_t}
\bigl(x-\sqrt{\overline{\alpha}_t}X_r^{\mathsf{sample}}\bigr)}
{D(x)} .
\end{equation}
Then 
\[
\widehat\Cov_{0\mid t}(x)
=
\widetilde\Cov_{0\mid t}(x)
-
\left(\widehat\mu_{0\mid t}(x)-\mu_{0\mid t}(x)\right)
\left(\widehat\mu_{0\mid t}(x)-\mu_{0\mid t}(x)\right)^\top,
\]
and hence
\begin{align}
&\EE_{X_t}
\left[
\left\|\widehat\Cov_{0\mid t}(X_t)-\Cov_{0\mid t}(X_t)\right\|_{\mathrm F}^2
\right]
\notag\\
&\qquad\lesssim
\EE_{X_t}
\left[
\left\|\widetilde\Cov_{0\mid t}(X_t)-\Cov_{0\mid t}(X_t)\right\|_{\mathrm F}^2
\right]
+
\EE_{X_t}
\left[
\left\|\widehat\mu_{0\mid t}(X_t)-\mu_{0\mid t}(X_t)\right\|_2^4
\right].
\label{eq:cov-error-centered-split}
\end{align}

We first analyze the first term in \eqref{eq:cov-error-centered-split}, namely
\(\EE_{X_t}[
\|\widetilde\Cov_{0\mid t}(X_t)-\Cov_{0\mid t}(X_t)\|_{\mathrm F}^2]\).
We first split according to the typical region:
\begin{align}
&\EE_{X_t}
\left[
\left\|\widetilde\Cov_{0\mid t}(X_t)-\Cov_{0\mid t}(X_t)\right\|_{\mathrm F}^2
\right]
\notag\\
&\qquad\le
\int_{\gT_{\overline{\alpha}_t}}
\EE_{X^{\mathsf{sample}}}\!\left[
\left\|\widetilde\Cov_{0\mid t}(x)-\Cov_{0\mid t}(x)\right\|_{\mathrm F}^2
\right]p_{X_t}(x)\rd x
+R_0^4\PP\!\left(X_t\notin\gT_{\overline{\alpha}_t}\right),
\label{eq:centered-cov-typical-split}
\end{align}
where the last term is controlled by Lemma~\ref{lem:gT_alpha_typical} and is
absorbed into the final bound. It remains to control the first term on the
right-hand side of \eqref{eq:centered-cov-typical-split}. To this end, we
decompose according to whether
\(\gE_x\coloneqq\{\widehat p_{X_t}(x)\ge p_{X_t}(x)/2\}\) holds:
\begin{align}
&\int_{\gT_{\overline{\alpha}_t}}
\EE_{X^{\mathsf{sample}}}\!\left[
\left\|\widetilde\Cov_{0\mid t}(x)-\Cov_{0\mid t}(x)\right\|_{\mathrm F}^2
\right]p_{X_t}(x)\rd x\notag\\
&\hspace{5em}\le
\underbrace{
\int_{\gT_{\overline{\alpha}_t}}
\EE_{X^{\mathsf{sample}}}
\left[
\left\|\widetilde\Cov_{0\mid t}(x)-\Cov_{0\mid t}(x)\right\|_{\mathrm F}^2
\mathbbm{1}_{\gE_x}
\right]
\!p_{X_t}(x)\rd x
}_{\eqqcolon\,\gK_{\rm good}}
\notag\\
&\hspace{7em}+
\underbrace{
\int_{\gT_{\overline{\alpha}_t}}
\EE_{X^{\mathsf{sample}}}
\left[
\left\|\widetilde\Cov_{0\mid t}(x)-\Cov_{0\mid t}(x)\right\|_{\mathrm F}^2
\mathbbm{1}_{\gE_x^c}
\right]
\!p_{X_t}(x)\rd x
}_{\eqqcolon\,\gK_{\rm bad}} .
\label{eq:centered-cov-good-bad-split}
\end{align}

We first bound \(\gK_{\rm good}\). Write
\(X_t=\sqrt{\overline{\alpha}_t}y+\sqrt{1-\overline{\alpha}_t}\omega\) with
\(Y\sim\PB_{X_0}\) and \(\Omega\sim\gN(0,I_d)\). For fixed \(x\), absorb
\(\Cov_{0\mid t}(x)\) into each numerator summand:
\begin{align}
&\EE\!\left[
\left\{
\left(X_0-\mu_{0\mid t}(x)\right)
\left(X_0-\mu_{0\mid t}(x)\right)^\top
-\Cov_{0\mid t}(x)
\right\}
\varphi_{1-\overline{\alpha}_t}
\bigl(x-\sqrt{\overline{\alpha}_t}X_0\bigr)
\right]=0 .
\label{eq:centered-cov-centered-numerator}
\end{align}
On \(\gE_x\), the denominator can be replaced by \(p_{X_t}(x)/2\). By
\eqref{eq:centered-cov-centered-numerator}, the numerator summands are centered,
so the cross terms vanish by independence. Hence, using the posterior density,
\[
\gK_{\rm good}
\lesssim
\frac1n
\mathop{\EB}\limits_{Y,\Omega}
\left[
\mathop{\EB}\limits_{Z\sim\PB_{0\mid t}(\cdot\mid X_t)}
\left[
\|Z-\mu_{0\mid t}(X_t)\|_2^4
\frac{\varphi_{1-\overline{\alpha}_t}
\bigl(X_t-\sqrt{\overline{\alpha}_t}Z\bigr)}
{p_{X_t}(X_t)}
\right]
\mathbbm{1}\{X_t\in\gT_{\overline{\alpha}_t}\}
\right].
\]
For \(V\sim\PB_{0\mid t}(\cdot\mid X_t)\), Jensen's inequality gives
\[
\|Z-\mu_{0\mid t}(X_t)\|_2^4
\lesssim
\|Z-Y\|_2^4+
\mathop{\EB}\limits_{V\sim\PB_{0\mid t}(\cdot\mid X_t)}
\left[\|V-Y\|_2^4\right].
\]
Hence
\begin{align}
\gK_{\rm good}
&\lesssim
\frac1n
\underbrace{
\mathop{\EB}\limits_{Y,\Omega}
\left[
\mathop{\EB}\limits_{Z\sim\PB_{0\mid t}(\cdot\mid X_t)}
\left[
\|Z-Y\|_2^4
\frac{\varphi_{1-\overline{\alpha}_t}
\bigl(X_t-\sqrt{\overline{\alpha}_t}Z\bigr)}
{p_{X_t}(X_t)}
\right]
\mathbbm{1}\{X_t\in\gT_{\overline{\alpha}_t}\}
\right]
}_{\eqqcolon\,\gL_1}
\notag\\
&\quad+
\frac1n
\underbrace{
\mathop{\EB}\limits_{Y,\Omega}
\left[
\mathop{\EB}\limits_{V\sim\PB_{0\mid t}(\cdot\mid X_t)}
\left[\|V-Y\|_2^4\right]
\mathop{\EB}\limits_{Z\sim\PB_{0\mid t}(\cdot\mid X_t)}
\left[
\frac{\varphi_{1-\overline{\alpha}_t}
\bigl(X_t-\sqrt{\overline{\alpha}_t}Z\bigr)}
{p_{X_t}(X_t)}
\right]
\mathbbm{1}\{X_t\in\gT_{\overline{\alpha}_t}\}
\right]
}_{\eqqcolon\,\gL_2}.
\label{eq:centered-cov-good-L1-L2}
\end{align}

The two terms in \eqref{eq:centered-cov-good-L1-L2} are exactly the
quantities controlled by Lemmas~\ref{lem:J1-bound} and
\ref{lem:J2-bound} with \(\beta=4\). Hence
\begin{subequations}\label{eq:centered-cov-L1-L2-bounds}
\begin{align}
\gL_1
&\lesssim
\frac{(k^\star)^2R_0^{k^\star}
\log^2\frac{dnR_0}{1-\overline{\alpha}_t}}
{\overline{\alpha}_t^2
(1-\overline{\alpha}_t)^{k^\star/2-2}}
+
\frac{R_0^4(1-\overline{\alpha}_t)^4}{n}.
\label{eq:centered-cov-L1-bound}
\\
\gL_2
&\lesssim
\frac{(k^\star)^2R_0^{k^\star}
\log^2\frac{dnR_0}{1-\overline{\alpha}_t}}
{\overline{\alpha}_t^2
(1-\overline{\alpha}_t)^{k^\star/2-2}}
+
\frac{R_0^4(1-\overline{\alpha}_t)^4}{n},
\label{eq:centered-cov-L2-bound}
\end{align}
\end{subequations}
Combining
\eqref{eq:centered-cov-good-L1-L2}, \eqref{eq:centered-cov-L1-bound}, and
\eqref{eq:centered-cov-L2-bound} yields
\begin{align}
\gK_{\rm good}
&\lesssim
\frac{(k^\star)^2R_0^{k^\star}
\log^2\frac{dnR_0}{1-\overline{\alpha}_t}}
{n\overline{\alpha}_t^2
(1-\overline{\alpha}_t)^{k^\star/2-2}}
+
\frac{R_0^4(1-\overline{\alpha}_t)^4}{n}.
\label{eq:centered-cov-good-bound}
\end{align}

We next bound \(\gK_{\rm bad}\).
For \(x=\sqrt{\overline{\alpha}_t}y+\sqrt{1-\overline{\alpha}_t}\omega\), define
\[
\widehat Q_t(y,\omega)
\coloneqq
\frac{\sum_{r=1}^n
\|X_r^{\mathsf{sample}}-y\|_2^2
\varphi_{1-\overline{\alpha}_t}
\bigl(x-\sqrt{\overline{\alpha}_t}X_r^{\mathsf{sample}}\bigr)}
{D(x)},
\qquad
m_t(y,\omega)
\coloneqq
\EE_{V\sim\PB_{0\mid t}(\cdot\mid x)}
\left[\|V-y\|_2^2\right].
\]
We use the following deterministic bound. Since
\(\sum_r\varphi_{1-\overline{\alpha}_t}
(x-\sqrt{\overline{\alpha}_t}X_r^{\mathsf{sample}})=D(x)\), we may write
\begin{align}
&\widetilde\Cov_{0\mid t}(x)-\Cov_{0\mid t}(x)
\notag\\
&\quad=
\frac{
\sum_{r=1}^n
\left[
\left(X_r^{\mathsf{sample}}-\mu_{0\mid t}(x)\right)
\left(X_r^{\mathsf{sample}}-\mu_{0\mid t}(x)\right)^\top
-\Cov_{0\mid t}(x)
\right]
\varphi_{1-\overline{\alpha}_t}
\bigl(x-\sqrt{\overline{\alpha}_t}X_r^{\mathsf{sample}}\bigr)}
{D(x)} .
\notag
\end{align}
Thus, using \(\|aa^\top\|_{\mathrm F}=\|a\|_2^2\),
\begin{align}
\left\|\widetilde\Cov_{0\mid t}(x)-\Cov_{0\mid t}(x)\right\|_{\mathrm F}
&\le
\frac{
\sum_{r=1}^n
\left\|X_r^{\mathsf{sample}}-\mu_{0\mid t}(x)\right\|_2^2
\varphi_{1-\overline{\alpha}_t}
\bigl(x-\sqrt{\overline{\alpha}_t}X_r^{\mathsf{sample}}\bigr)}
{D(x)}
+\left\|\Cov_{0\mid t}(x)\right\|_{\mathrm F}
\notag\\
&\lesssim
\widehat Q_t(y,\omega)
+\left\|\mu_{0\mid t}(x)-y\right\|_2^2
+\EE_{V\sim\PB_{0\mid t}(\cdot\mid x)}
\left[\left\|V-\mu_{0\mid t}(x)\right\|_2^2\right]
\notag\\
&\overset{\text{(a)}}{\lesssim}
\widehat Q_t(y,\omega)
+2\left\|\mu_{0\mid t}(x)-y\right\|_2^2
+\EE_{V\sim\PB_{0\mid t}(\cdot\mid x)}
\left[\left\|V-y\right\|_2^2\right]
\notag\\
&\overset{\text{(b)}}{\lesssim}
\widehat Q_t(y,\omega)+m_t(y,\omega).
\label{eq:centered-cov-bad-deterministic-bound}
\end{align}
Here (a) follows from
\(\|V-\mu_{0\mid t}(x)\|_2^2
\lesssim \|V-y\|_2^2+\|y-\mu_{0\mid t}(x)\|_2^2\), and (b) holds since
\[
\|\mu_{0\mid t}(x)-y\|_2^2
\le \EE_{V\sim\PB_{0\mid t}(\cdot\mid x)}
\left[\|V-y\|_2^2\right]
\]
by Jensen's inequality. Therefore, writing
\(\rho_t>0\) for a truncation radius to be chosen later,
\[
\widehat Q_t(y,\omega)=\gA_t(y,\omega)+\gB_t(y,\omega),
\]
where
\begin{subequations}
\begin{align*}
\gA_t(y,\omega)
&\coloneqq
\frac{
\sum_{r=1}^n
\|X_r^{\mathsf{sample}}-y\|_2^2
\mathbbm{1}\{\|X_r^{\mathsf{sample}}-y\|_2\le \rho_t\}
\varphi_{1-\overline{\alpha}_t}
\bigl(x-\sqrt{\overline{\alpha}_t}X_r^{\mathsf{sample}}\bigr)}
{D(x)}
\\
\gB_t(y,\omega)
&\coloneqq
\frac{
\sum_{r=1}^n
\|X_r^{\mathsf{sample}}-y\|_2^2
\mathbbm{1}\{\|X_r^{\mathsf{sample}}-y\|_2> \rho_t\}
\varphi_{1-\overline{\alpha}_t}
\bigl(x-\sqrt{\overline{\alpha}_t}X_r^{\mathsf{sample}}\bigr)}
{D(x)} .
\end{align*}
\end{subequations}
By the triangle inequality,
\begin{align}
\gK_{\rm bad}
&\lesssim
\underbrace{
\int_{y,\omega}
\EE\!\left[
\gA_t(y,\omega)^2\mathbbm{1}_{\gE_x^c}
\right]
\varphi_1(\omega)p_{X_0}(y)\rd\omega\rd y
}_{\eqqcolon\,\gK_{\rm bad,A}}
+
\underbrace{
\int_{y,\omega}
\EE\!\left[
\gB_t(y,\omega)^2\mathbbm{1}_{\gE_x^c}
\right]
\varphi_1(\omega)p_{X_0}(y)\rd\omega\rd y
}_{\eqqcolon\,\gK_{\rm bad,B}}
\notag\\
&\quad+
\underbrace{
\int_{y,\omega}
m_t(y,\omega)^2
\PP\!\left(\widehat p_{X_t}(x)<\frac12p_{X_t}(x)\right)
\varphi_1(\omega)p_{X_0}(y)\rd\omega\rd y
}_{\eqqcolon\,\gK_{\rm bad,M}} .
\label{eq:centered-cov-bad-triangle-split}
\end{align}
Since \(\gA_t(y,\omega)\le\rho_t^2\), Lemma~\ref{lem:intrinsic-density-matching-bound}
gives
\begin{align}
\gK_{\rm bad,A}
&\lesssim
\frac{R_0^{k^\star}\rho_t^4}
{n(1-\overline{\alpha}_t)^{k^\star/2}}.
\label{eq:centered-cov-bad-local}
\end{align}
For the tail empirical part, introduce a covering scale \(\lambda_t>0\). To
apply Lemma~\ref{lem:bad-B-tail-ratio}, we require
\[
0<\lambda_t\le \frac{\rho_t}{2},
\qquad
\rho_t\ge
32\sqrt{\frac{1-\overline{\alpha}_t}{\overline{\alpha}_t}}\,
\sqrt{\log\frac{dnR_0}{1-\overline{\alpha}_t}}.
\]
Define \(\mathsf{R}_t(y,\omega)\) as in Lemma~\ref{lem:bad-B-tail-ratio}. Since
\(\gB_t(y,\omega)\le R_0^2\mathsf{R}_t(y,\omega)\), applying
Lemma~\ref{lem:bad-B-tail-ratio} and
\(N_{\lambda_t}\le (R_0/\lambda_t)^{k^\star}\) yields
\begin{align}
\gK_{\rm bad,B}
&\lesssim
\frac{R_0^4}{n}
\left(\frac{R_0}{\lambda_t}\right)^{k^\star}
+
\frac{R_0^4(1-\overline{\alpha}_t)^6}{n^2}.
\label{eq:centered-cov-bad-tail}
\end{align}

We finally bound \(\gK_{\rm bad,M}\). Recall
\(
m_t(y,\omega)
=
\EE_{V\sim\PB_{0\mid t}(\cdot\mid x)}
\left[\|V-y\|_2^2\right].
\)
Recall the definition of $\gE_{\overline{\alpha}_t}(y)$ from
the very begining of this section.
Splitting the posterior expectation according to whether
\(V\in\gE_{\overline{\alpha}_t}(y)\), we get
\begin{align}
\gK_{\rm bad,M}
&\lesssim
\int_{y,\omega}
\left(
\EE_{V\sim\PB_{0\mid t}(\cdot\mid x)}
\left[
\|V-y\|_2^2
\mathbbm{1}\{V\in\gE_{\overline{\alpha}_t}(y)\}
\right]
\right)^2
\PP(\gE_x^c)\varphi_1(\omega)p_{X_0}(y)\rd\omega\rd y
\notag\\
&\quad+
\int_{y,\omega}
\left(
\EE_{V\sim\PB_{0\mid t}(\cdot\mid x)}
\left[
\|V-y\|_2^2
\mathbbm{1}\{V\notin\gE_{\overline{\alpha}_t}(y)\}
\right]
\right)^2
\PP(\gE_x^c)\varphi_1(\omega)p_{X_0}(y)\rd\omega\rd y .
\label{eq:centered-cov-bad-M-split}
\end{align}
For the first term in \eqref{eq:centered-cov-bad-M-split}, the definition of
\(\gE_{\overline{\alpha}_t}(y)\) gives
\begin{align}
&\int_{y,\omega}
\left(
\EE_{V\sim\PB_{0\mid t}(\cdot\mid x)}
\left[
\|V-y\|_2^2
\mathbbm{1}\{V\in\gE_{\overline{\alpha}_t}(y)\}
\right]
\right)^2
\PP(\gE_x^c)\varphi_1(\omega)p_{X_0}(y)\rd\omega\rd y
\notag\\
&\quad\lesssim
\left(
\frac{k^\star(1-\overline{\alpha}_t)}
{\overline{\alpha}_t}
\log\frac{dnR_0}{1-\overline{\alpha}_t}
\right)^2
\int_{y,\omega}
\PP\!\left(\widehat p_{X_t}(x)<\frac12p_{X_t}(x)\right)
\varphi_1(\omega)p_{X_0}(y)\rd\omega\rd y
\notag\\
&\quad\lesssim
\frac{R_0^{k^\star}}
{n(1-\overline{\alpha}_t)^{k^\star/2}}
\left(
\frac{k^\star(1-\overline{\alpha}_t)}
{\overline{\alpha}_t}
\log\frac{dnR_0}{1-\overline{\alpha}_t}
\right)^2,
\label{eq:centered-cov-bad-M-typ}
\end{align}
where the last step uses Lemma~\ref{lem:intrinsic-density-matching-bound}.
For the second term in \eqref{eq:centered-cov-bad-M-split}, use the support bound and discard
\(\PP(\gE_x^c)\):
\begin{align}
&\int_{y,\omega}
\left(
\EE_{V\sim\PB_{0\mid t}(\cdot\mid x)}
\left[
\|V-y\|_2^2
\mathbbm{1}\{V\notin\gE_{\overline{\alpha}_t}(y)\}
\right]
\right)^2
\PP(\gE_x^c)\varphi_1(\omega)p_{X_0}(y)\rd\omega\rd y
\notag\\
&\quad\lesssim
R_0^4
\int_{y,\omega}
\EE_{V\sim\PB_{0\mid t}(\cdot\mid x)}
\left[
\mathbbm{1}\{V\notin\gE_{\overline{\alpha}_t}(y)\}
\right]
\varphi_1(\omega)p_{X_0}(y)\rd\omega\rd y
\notag\\
&=
R_0^4
\PP\!\left(
V\notin \gE_{\overline{\alpha}_t}(X_0)
\right)
\lesssim
\frac{R_0^4(1-\overline{\alpha}_t)^4}{n^2},
\label{eq:centered-cov-bad-M-tail}
\end{align}
where \(V\sim\PB_{0\mid t}(\cdot\mid X_t)\), and the last step follows from
Lemma~\ref{lem:posterior_norm}. Combining
\eqref{eq:centered-cov-bad-M-split}, \eqref{eq:centered-cov-bad-M-typ}, and
\eqref{eq:centered-cov-bad-M-tail} gives
\begin{align}
\gK_{\rm bad,M}
&\lesssim
\frac{R_0^{k^\star}}
{n(1-\overline{\alpha}_t)^{k^\star/2}}
\left(
\frac{k^\star(1-\overline{\alpha}_t)}
{\overline{\alpha}_t}
\log\frac{dnR_0}{1-\overline{\alpha}_t}
\right)^2
+
\frac{R_0^4(1-\overline{\alpha}_t)^4}{n^2}.
\label{eq:centered-cov-bad-pop}
\end{align}
Combining \eqref{eq:centered-cov-bad-triangle-split},
\eqref{eq:centered-cov-bad-local}, \eqref{eq:centered-cov-bad-tail}, and
\eqref{eq:centered-cov-bad-pop}, we have, under the conditions of
Lemmas~\ref{lem:intrinsic-density-matching-bound} and \ref{lem:bad-B-tail-ratio},
\begin{align}
\gK_{\rm bad}
&\lesssim
\frac{R_0^{k^\star}\rho_t^4}
{n(1-\overline{\alpha}_t)^{k^\star/2}}
+
\frac{R_0^4}{n}
\left(\frac{R_0}{\lambda_t}\right)^{k^\star}
\notag\\
&\quad+
\frac{R_0^{k^\star}}
{n(1-\overline{\alpha}_t)^{k^\star/2}}
\left(
\frac{k^\star(1-\overline{\alpha}_t)}
{\overline{\alpha}_t}
\log\frac{dnR_0}{1-\overline{\alpha}_t}
\right)^2
+
\frac{R_0^4(1-\overline{\alpha}_t)^4}{n^2}.
\label{eq:centered-cov-bad-before-opt}
\end{align}
We now take \(\lambda_t=\rho_t/2\). Balancing the local and tail empirical
terms, choose
\[
\rho_t
\asymp
k^\star R_0^{\frac{4}{k^\star+4}}
(1-\overline{\alpha}_t)^{\frac{k^\star}{2(k^\star+4)}}
\sqrt{\log\frac{dnR_0}{1-\overline{\alpha}_t}},
\qquad
\lambda_t=\frac{\rho_t}{2}.
\]
With this choice, the second and third terms in
\eqref{eq:centered-cov-bad-before-opt} are absorbed by the local empirical
term. Substituting into \eqref{eq:centered-cov-bad-before-opt} yields
\begin{align}
\gK_{\rm bad}
&\lesssim
\frac{k^\star R_0^{k^\star+\frac{16}{k^\star+4}}}
		{n(1-\overline{\alpha}_t)^{\frac{k^\star}{2}
		-\frac{2k^\star}{k^\star+4}}}
\log^2\frac{dnR_0}{1-\overline{\alpha}_t}
+
\frac{R_0^4(1-\overline{\alpha}_t)^4}{n^2}.
\label{eq:centered-cov-bad-bound}
\end{align}
Combining \eqref{eq:centered-cov-good-bad-split},
\eqref{eq:centered-cov-good-bound}, and \eqref{eq:centered-cov-bad-bound}, we get
\begin{align}
&\EE_{X_t}
\left[
\left\|\widetilde\Cov_{0\mid t}(X_t)-\Cov_{0\mid t}(X_t)\right\|_{\mathrm F}^2
\right]
\notag\\
&\quad\lesssim
\frac{k^\star R_0^{k^\star+\frac{16}{k^\star+4}}}
		{n(1-\overline{\alpha}_t)^{\frac{k^\star}{2}
		-\frac{2k^\star}{k^\star+4}}}
\log^2\frac{dnR_0}{1-\overline{\alpha}_t}
+
\frac{R_0^4(1-\overline{\alpha}_t)^4}{n^2}.
\label{eq:centered-cov-bound}
\end{align}

The mean term in \eqref{eq:cov-error-centered-split} is exactly the
\(\beta=2\) case of Lemma~\ref{lem:posterior-mean-high-moment}. Hence
\begin{align}
&\EE_{X_t}
\left[
\left\|\widehat\mu_{0\mid t}(X_t)-\mu_{0\mid t}(X_t)\right\|_2^4
\right]
\notag\\
&\qquad\lesssim
\frac{R_0^{k^\star+\frac{16}{k^\star+4}}}
		{n(1-\overline{\alpha}_t)^{\frac{k^\star}{2}
		-\frac{2k^\star}{k^\star+4}}}
\log^2\frac{dnR_0}{1-\overline{\alpha}_t}
+
\frac{R_0^4(1-\overline{\alpha}_t)^4}{n^2}.
\label{eq:mean-fourth-bound}
\end{align}
Finally, \eqref{eq:cov-error-centered-split}, \eqref{eq:centered-cov-bound},
and \eqref{eq:mean-fourth-bound} give, through
\eqref{eq:higher-order-jacobian-F-form},
\begin{align}
\varepsilon_{\mathsf{Jacobi},1,t}^2
&\lesssim
\frac{\overline{\alpha}_t^2}{(1-\overline{\alpha}_t)^4}
\bigg[
\frac{R_0^{k^\star+\frac{16}{k^\star+4}}}
		{n(1-\overline{\alpha}_t)^{\frac{k^\star}{2}
		-\frac{2k^\star}{k^\star+4}}}
\log^2\frac{dnR_0}{1-\overline{\alpha}_t}
+
\frac{R_0^4(1-\overline{\alpha}_t)^4}{n^2}
\bigg].
\label{eq:preliminary-jacobi-1-bound}
\end{align}

\subsubsection{Trace Jacobian Approximation}

The DDIM integration-by-parts argument uses the square of the trace error,
not the trace of the squared error matrix. Hence the second Jacobian error
should be read as
\[
\varepsilon_{\mathsf{Jacobi},2,t}^2
=
\EE_{X_t}\!\left[
\left(\tr\left(
\frac{\partial s_t(X_t)}{\partial x}
-
\frac{\partial s_t^\star(X_t)}{\partial x}
\right)\right)^2
\right].
\]
Using Tweedie's formula~\eqref{eq:score_to_posterior_new} and its empirical
counterpart for \(s_t\), this becomes
\begin{equation}\label{eq:higher-order-jacobian-trace-form}
\varepsilon_{\mathsf{Jacobi},2,t}^2
=
\frac{\overline{\alpha}_t^2}{(1-\overline{\alpha}_t)^4}
\EE_{X_t}\!\left[
\left(\tr\left(
\widehat\Cov_{0\mid t}(X_t)-\Cov_{0\mid t}(X_t)
\right)\right)^2
\right].
\end{equation}
We use the same centered covariance decomposition as above. Since
\[
\tr\!\left(\widehat\Cov_{0\mid t}(x)-\Cov_{0\mid t}(x)\right)
=
\tr\!\left(\widetilde\Cov_{0\mid t}(x)-\Cov_{0\mid t}(x)\right)
-\left\|\widehat\mu_{0\mid t}(x)-\mu_{0\mid t}(x)\right\|_2^2,
\]
we have
\begin{align}
&\EE_{X_t}\!\left[
\left(\tr\!\left(
\widehat\Cov_{0\mid t}(X_t)-\Cov_{0\mid t}(X_t)
\right)\right)^2
\right]
\notag\\
&\quad\lesssim
\EE_{X_t}\!\left[
\left(\tr\!\left(
\widetilde\Cov_{0\mid t}(X_t)-\Cov_{0\mid t}(X_t)
\right)\right)^2
\right]
+
\EE_{X_t}\!\left[
\left\|\widehat\mu_{0\mid t}(X_t)-\mu_{0\mid t}(X_t)\right\|_2^4
\right].
\label{eq:trace-cov-error-centered-split}
\end{align}
It remains to control the centered trace-covariance term. The proof is the
scalar analogue of the Frobenius argument above. On the good event
\(\gE_x=\{\widehat p_{X_t}(x)\ge p_{X_t}(x)/2\}\), absorb
\(\tr(\Cov_{0\mid t}(x))\) into each numerator summand:
\begin{align}
\EE\!\left[
\left\{
\left\|X_0-\mu_{0\mid t}(x)\right\|_2^2
-\tr(\Cov_{0\mid t}(x))
\right\}
\varphi_{1-\overline{\alpha}_t}
\bigl(x-\sqrt{\overline{\alpha}_t}X_0\bigr)
\right]=0 .
\label{eq:trace-cov-centered-numerator}
\end{align}
Thus the denominator is bounded below by \(p_{X_t}(x)/2\), and the cross terms
vanish by independence. Hence the good-event contribution is bounded by
\begin{align}
&
\EE_{X_t}\!\left[
\left(\tr\!\left(
\widetilde\Cov_{0\mid t}(X_t)-\Cov_{0\mid t}(X_t)
\right)\right)^2
\mathbbm{1}_{\gE_{X_t}}
\right]
\notag\\
&\quad\lesssim
\frac1n
\mathop{\EB}\limits_{Y,\Omega}
\left[
\mathop{\EB}\limits_{Z\sim\PB_{0\mid t}(\cdot\mid X_t)}
\left[
\left(
\left\|Z-\mu_{0\mid t}(X_t)\right\|_2^2
-\tr(\Cov_{0\mid t}(X_t))
\right)^2
\frac{\varphi_{1-\overline{\alpha}_t}
\bigl(X_t-\sqrt{\overline{\alpha}_t}Z\bigr)}
{p_{X_t}(X_t)}
\right]
\right]
\notag\\
&\lesssim
\frac1n
\mathop{\EB}\limits_{Y,\Omega}
\left[
\mathop{\EB}\limits_{Z\sim\PB_{0\mid t}(\cdot\mid X_t)}
\left[
\left\|Z-\mu_{0\mid t}(X_t)\right\|_2^4
\frac{\varphi_{1-\overline{\alpha}_t}
\bigl(X_t-\sqrt{\overline{\alpha}_t}Z\bigr)}
{p_{X_t}(X_t)}
\right]
\right]
\notag\\
&\quad+
\frac1n
\mathop{\EB}\limits_{Y,\Omega}
\left[
\left(
\mathop{\EB}\limits_{Z\sim\PB_{0\mid t}(\cdot\mid X_t)}
\left[
\left\|Z-\mu_{0\mid t}(X_t)\right\|_2^2
\right]
\right)^2
\mathop{\EB}\limits_{Z\sim\PB_{0\mid t}(\cdot\mid X_t)}
\left[
\frac{\varphi_{1-\overline{\alpha}_t}
\bigl(X_t-\sqrt{\overline{\alpha}_t}Z\bigr)}
{p_{X_t}(X_t)}
\right]
\right].
\label{eq:trace-cov-good-fourth-moment}
\end{align}
By Jensen, both displayed terms are bounded by the two fourth-moment quantities
estimated in \eqref{eq:centered-cov-good-L1-L2}--\eqref{eq:centered-cov-good-bound}.
On the bad event, the deterministic estimate used for the Frobenius bound also
applies to the trace:
\begin{align}
\left|
\tr\!\left(\widetilde\Cov_{0\mid t}(x)-\Cov_{0\mid t}(x)\right)
\right|
&\le
\frac{
\sum_{r=1}^n
\left|\|X_r^{\mathsf{sample}}-\mu_{0\mid t}(x)\|_2^2
-\tr(\Cov_{0\mid t}(x))\right|
\varphi_{1-\overline{\alpha}_t}
\bigl(x-\sqrt{\overline{\alpha}_t}X_r^{\mathsf{sample}}\bigr)}
{D(x)}
\notag\\
&\lesssim
\widehat Q_t(y,\omega)+m_t(y,\omega),
\label{eq:trace-cov-bad-deterministic-bound}
\end{align}
where \(x=\sqrt{\overline{\alpha}_t}y+\sqrt{1-\overline{\alpha}_t}\omega\), and
\(\widehat Q_t,m_t\) are the same quantities used in
\eqref{eq:centered-cov-bad-deterministic-bound}. Therefore the same
\(\widehat Q_t=\gA_t+\gB_t\) decomposition, with the same optimized
\(\rho_t,\lambda_t\), gives the same final estimate as in the Frobenius case:
\begin{align}
&\EE_{X_t}\!\left[
\left(\tr\!\left(
\widetilde\Cov_{0\mid t}(X_t)-\Cov_{0\mid t}(X_t)
\right)\right)^2
\right]
\notag\\
&\quad\lesssim
\frac{k^\star R_0^{k^\star+\frac{16}{k^\star+4}}}
		{n(1-\overline{\alpha}_t)^{\frac{k^\star}{2}
		-\frac{2k^\star}{k^\star+4}}}
\log^2\frac{dnR_0}{1-\overline{\alpha}_t}
+
\frac{R_0^4(1-\overline{\alpha}_t)^4}{n^2}.
\label{eq:trace-centered-cov-bound}
\end{align}
Finally, \eqref{eq:trace-cov-error-centered-split},
\eqref{eq:trace-centered-cov-bound}, and \eqref{eq:mean-fourth-bound} imply
\begin{align}
\varepsilon_{\mathsf{Jacobi},2,t}^2
&\lesssim
\frac{\overline{\alpha}_t^2}{(1-\overline{\alpha}_t)^4}
\bigg[
\frac{k^\star R_0^{k^\star+\frac{16}{k^\star+4}}}
		{n(1-\overline{\alpha}_t)^{\frac{k^\star}{2}
		-\frac{2k^\star}{k^\star+4}}}
\log^2\frac{dnR_0}{1-\overline{\alpha}_t}
+
\frac{R_0^4(1-\overline{\alpha}_t)^4}{n^2}
\bigg].
\label{eq:preliminary-jacobi-2-bound}
\end{align}

\subsubsection{Trace-Hessian Approximation}

We start by reformulating \(\varepsilon_{\mathsf{Hess},t}^2\) under the
unthresholded kernel estimator \(s_t\) defined in
Proposition~\ref{prop:higher-order-kernel-approximation}:
\[
\varepsilon_{\mathsf{Hess},t}^{2}
=
\EE_{X_t}\!\left[
\left\|
\nabla\tr\!\left(
\frac{\partial s_t(X_t)}{\partial x}
-
\frac{\partial s_t^\star(X_t)}{\partial x}
\right)
\right\|_2^2
\right].
\]
By differentiating Tweedie's formula~\eqref{eq:score_to_posterior_new} and its
empirical counterpart for \(s_t\),
\begin{align}
\nabla\tr\!\left(
\frac{\partial s_t(x)}{\partial x}
-
\frac{\partial s_t^\star(x)}{\partial x}
\right)
&=
\frac{\overline{\alpha}_t}{(1-\overline{\alpha}_t)^2}
\nabla\tr\!\left(
\widehat\Cov_{0\mid t}(x)-\Cov_{0\mid t}(x)
\right).
\label{eq:hessian-trace-cov-reduction}
\end{align}
It remains to compute the trace-covariance gradient. Since
\[
\tr\!\left(\widehat\Cov_{0\mid t}(x)\right)
=
\widehat{\EE}_{0\mid t,x}\!\left[\|X_0\|_2^2\right]
-\left\|\widehat\mu_{0\mid t}(x)\right\|_2^2,
\]
differentiating the normalized Gaussian weights gives
\begin{align}
\nabla
\widehat{\EE}_{0\mid t,x}\!\left[\|X_0\|_2^2\right]
=
\frac{\sqrt{\overline{\alpha}_t}}{1-\overline{\alpha}_t}
\widehat{\EE}_{0\mid t,x}\!\left[
\left(X_0-\widehat\mu_{0\mid t}(x)\right)\|X_0\|_2^2
\right],\quad
\nabla\widehat\mu_{0\mid t}(x)
=
\frac{\sqrt{\overline{\alpha}_t}}{1-\overline{\alpha}_t}
\widehat\Cov_{0\mid t}(x).
\notag
\end{align}
Therefore
\begin{align}
\nabla\tr\!\left(\widehat\Cov_{0\mid t}(x)\right)
&=
\frac{\sqrt{\overline{\alpha}_t}}{1-\overline{\alpha}_t}
\widehat{\EE}_{0\mid t,x}
\left[
\left(X_0-\widehat\mu_{0\mid t}(x)\right)
\left\|X_0-\widehat\mu_{0\mid t}(x)\right\|_2^2
\right],
\label{eq:empirical-trace-cov-gradient}
\end{align}
where the last equality is the algebraic simplification after subtracting
\(\nabla\|\widehat\mu_{0\mid t}(x)\|_2^2
=2(\nabla\widehat\mu_{0\mid t}(x))\widehat\mu_{0\mid t}(x)\). This computation
identifies the empirical third central moment
\[
\widehat M_{3,0\mid t}(x)
\coloneqq
\widehat{\EE}_{0\mid t,x}
\left[
\left(X_0-\widehat\mu_{0\mid t}(x)\right)
\left\|X_0-\widehat\mu_{0\mid t}(x)\right\|_2^2
\right],
\]
and, analogously, its population counterpart
\[
M_{3,0\mid t}(x)
\coloneqq
\EE\!\left[
\left(X_0-\mu_{0\mid t}(x)\right)
\left\|X_0-\mu_{0\mid t}(x)\right\|_2^2
\mid X_t=x
\right].
\]
Hence
\[
\nabla\tr\!\left(\widehat\Cov_{0\mid t}(x)\right)
=
\frac{\sqrt{\overline{\alpha}_t}}{1-\overline{\alpha}_t}
\widehat M_{3,0\mid t}(x),
\qquad
\nabla\tr\!\left(\Cov_{0\mid t}(x)\right)
=
\frac{\sqrt{\overline{\alpha}_t}}{1-\overline{\alpha}_t}
M_{3,0\mid t}(x).
\]
Substituting these two identities into
\eqref{eq:hessian-trace-cov-reduction}, we get
\begin{align}
\nabla\tr\!\left(
\frac{\partial s_t(x)}{\partial x}
-
\frac{\partial s_t^\star(x)}{\partial x}
\right)
&=
\frac{\overline{\alpha}_t}{(1-\overline{\alpha}_t)^2}
\nabla\tr\!\left(
\widehat\Cov_{0\mid t}(x)-\Cov_{0\mid t}(x)
\right)
\notag\\
&=
\frac{\overline{\alpha}_t^{3/2}}{(1-\overline{\alpha}_t)^3}
\left(\widehat M_{3,0\mid t}(x)-M_{3,0\mid t}(x)\right).
\label{eq:hessian-error-third-moment-identity}
\end{align}
Consequently, the Hessian approximation error is exactly
\begin{equation}\label{eq:higher-order-hessian-form}
\varepsilon_{\mathsf{Hess},t}^2
=
\frac{\overline{\alpha}_t^3}{(1-\overline{\alpha}_t)^6}
\EE_{X_t}\!\left[
\left\|\widehat M_{3,0\mid t}(X_t)-M_{3,0\mid t}(X_t)\right\|_2^2
\right].
\end{equation}
Thus it remains to control the third-central-moment error in
\eqref{eq:higher-order-hessian-form}.
As before, we first center the empirical moment at the population posterior
mean. Define
\[
\widetilde M_{3,0\mid t}(x)
\coloneqq
\widehat{\EE}_{0\mid t,x}\!\left[
\left(X_0-\mu_{0\mid t}(x)\right)
\left\|X_0-\mu_{0\mid t}(x)\right\|_2^2
\right].
\]
We now expand \(\widehat M_{3,0\mid t}(x)\) around the population posterior
centering. Set
\(\delta_x\coloneqq \widehat\mu_{0\mid t}(x)-\mu_{0\mid t}(x)\) and
\(W=X_0-\mu_{0\mid t}(x)\). Since
\(\widehat{\EE}_{0\mid t,x}[W]=\delta_x\), we have
\begin{align}
\widehat M_{3,0\mid t}(x)
&=
\widehat{\EE}_{0\mid t,x}
\left[(W-\delta_x)\|W-\delta_x\|_2^2\right]
\notag\\
&=
\widetilde M_{3,0\mid t}(x)
-2\widetilde\Cov_{0\mid t}(x)\delta_x
-\tr\!\left(\widetilde\Cov_{0\mid t}(x)\right)\delta_x
+2\|\delta_x\|_2^2\delta_x .
\label{eq:hatM-tildeM-exact-expansion}
\end{align}
Here \(\widetilde\Cov_{0\mid t}(x)\) is the empirical covariance centered at the
population posterior mean, defined in \eqref{eq:centered-empirical-cov-def}.
Consequently,
\begin{align}
&\EE_{X_t}\!\left[
\left\|\widehat M_{3,0\mid t}(X_t)-M_{3,0\mid t}(X_t)\right\|_2^2
\right]
\notag\\
&\quad\lesssim
\underbrace{\EE_{X_t}\!\left[
\left\|\widetilde M_{3,0\mid t}(X_t)-M_{3,0\mid t}(X_t)\right\|_2^2
\right]}_{\eqqcolon\,\gH_1}
+
\underbrace{\EE_{X_t}\!\left[
\left\|\widetilde\Cov_{0\mid t}(X_t)
\left(\widehat\mu_{0\mid t}(X_t)-\mu_{0\mid t}(X_t)\right)\right\|_2^2
\right]}_{\eqqcolon\,\gH_2}
\notag\\
&\qquad+
\underbrace{\EE_{X_t}\!\left[
\tr\!\left(\widetilde\Cov_{0\mid t}(X_t)\right)^2
\left\|\widehat\mu_{0\mid t}(X_t)-\mu_{0\mid t}(X_t)\right\|_2^2
\right]}_{\eqqcolon\,\gH_3}
+
\underbrace{\EE_{X_t}\!\left[
\left\|\widehat\mu_{0\mid t}(X_t)-\mu_{0\mid t}(X_t)\right\|_2^6
\right]}_{\eqqcolon\,\gH_4}.
\label{eq:third-moment-centered-split}
\end{align}
\textbf{Bound for \(\gH_1\).}
We first control \(\gH_1\). As in the covariance analysis above, we first
split off the non-typical region:
\begin{align}
\gH_1
&\le
\int_{\gT_{\overline{\alpha}_t}}
\EE_{X^{\mathsf{sample}}}\!\left[
\left\|\widetilde M_{3,0\mid t}(x)-M_{3,0\mid t}(x)\right\|_2^2
\right]p_{X_t}(x)\rd x
+R_0^6\PP\!\left(X_t\notin\gT_{\overline{\alpha}_t}\right).
\label{eq:third-moment-typical-split}
\end{align}
The second term is controlled by Lemma~\ref{lem:gT_alpha_typical} and is
absorbed into the final bound. It remains to bound the typical-region term.
Decompose it according to
\(\gE_x=\{\widehat p_{X_t}(x)\ge p_{X_t}(x)/2\}\). On \(\gE_x\), absorb
\(M_{3,0\mid t}(x)\) into each numerator summand:
\begin{align}
\EE\!\left[
\left\{
\left(X_0-\mu_{0\mid t}(x)\right)
\left\|X_0-\mu_{0\mid t}(x)\right\|_2^2
-M_{3,0\mid t}(x)
\right\}
\varphi_{1-\overline{\alpha}_t}
\bigl(x-\sqrt{\overline{\alpha}_t}X_0\bigr)
\right]=0 .
\label{eq:third-moment-centered-numerator}
\end{align}
Thus the denominator is bounded below by \(p_{X_t}(x)/2\), and the cross terms
vanish by independence. Hence the good-event contribution is bounded by
\begin{align}
&\int_{\gT_{\overline{\alpha}_t}}
\EE_{X^{\mathsf{sample}}}\!\left[
\left\|\widetilde M_{3,0\mid t}(x)-M_{3,0\mid t}(x)\right\|_2^2
\mathbbm{1}_{\gE_x}
\right]p_{X_t}(x)\rd x
\notag\\
&\quad\lesssim
\frac1n
\mathop{\EB}\limits_{Y,\Omega}
\left[
\mathop{\EB}\limits_{Z\sim\PB_{0\mid t}(\cdot\mid X_t)}
\left[
\left\|Z-\mu_{0\mid t}(X_t)\right\|_2^6
\frac{\varphi_{1-\overline{\alpha}_t}
\bigl(X_t-\sqrt{\overline{\alpha}_t}Z\bigr)}
{p_{X_t}(X_t)}
\right]
\mathbbm{1}\{X_t\in\gT_{\overline{\alpha}_t}\}
\right].
\label{eq:third-moment-good-sixth-moment}
\end{align}
Applying Lemmas~\ref{lem:J1-bound} and \ref{lem:J2-bound} with \(\beta=6\),
we obtain
\begin{align}
&\int_{\gT_{\overline{\alpha}_t}}
\EE_{X^{\mathsf{sample}}}\!\left[
\left\|\widetilde M_{3,0\mid t}(x)-M_{3,0\mid t}(x)\right\|_2^2
\mathbbm{1}_{\gE_x}
\right]p_{X_t}(x)\rd x
\notag\\
&\quad\lesssim
\frac{R_0^{k^\star}}
{n(1-\overline{\alpha}_t)^{k^\star/2}}
\left(
\frac{k^\star(1-\overline{\alpha}_t)}
{\overline{\alpha}_t}
\log\frac{dnR_0}{1-\overline{\alpha}_t}
\right)^3
+
\frac{R_0^6(1-\overline{\alpha}_t)^6}{n^2}.
\label{eq:third-moment-good-bound}
\end{align}

On the bad event, the same deterministic argument as for the covariance gives
\[
\left\|\widetilde M_{3,0\mid t}(x)-M_{3,0\mid t}(x)\right\|_2
\lesssim
\widehat Q_{3,t}(y,\omega)+m_{3,t}(y,\omega),
\]
where
\[
\widehat Q_{3,t}(y,\omega)
\coloneqq
\frac{
\sum_{r=1}^n
\|X_r^{\mathsf{sample}}-y\|_2^3
\varphi_{1-\overline{\alpha}_t}
\bigl(x-\sqrt{\overline{\alpha}_t}X_r^{\mathsf{sample}}\bigr)}
{D(x)},
\qquad
m_{3,t}(y,\omega)
\coloneqq
\EE_{V\sim\PB_{0\mid t}(\cdot\mid x)}
\left[\|V-y\|_2^3\right].
\]
Decompose \(\widehat Q_{3,t}\) at the same type of truncation radius
\(\rho_t>0\). The local part is bounded by \(\rho_t^3\), the tail part is
bounded by \(R_0^3\mathsf R_t(y,\omega)\), and \(m_{3,t}\) is handled by the
same typical/complement posterior split as above. Therefore, under the
conditions of Lemmas~\ref{lem:intrinsic-density-matching-bound} and \ref{lem:bad-B-tail-ratio},
\begin{align}
&\EE_{X_t}\!\left[
\left\|\widetilde M_{3,0\mid t}(X_t)-M_{3,0\mid t}(X_t)\right\|_2^2
\mathbbm{1}_{\gE_{X_t}^c}
\mathbbm{1}\{X_t\in\gT_{\overline{\alpha}_t}\}
\right]
\notag\\
&\quad\lesssim
\frac{R_0^{k^\star}\rho_t^6}
{n(1-\overline{\alpha}_t)^{k^\star/2}}
+
\frac{R_0^6}{n}
\left(\frac{R_0}{\lambda_t}\right)^{k^\star}
\notag\\
&\qquad+
\frac{R_0^{k^\star}}
{n(1-\overline{\alpha}_t)^{k^\star/2}}
\left(
\frac{k^\star(1-\overline{\alpha}_t)}
{\overline{\alpha}_t}
\log\frac{dnR_0}{1-\overline{\alpha}_t}
\right)^3
+
\frac{R_0^6(1-\overline{\alpha}_t)^6}{n^2}.
\label{eq:third-moment-bad-before-opt}
\end{align}
We choose \(\lambda_t=\rho_t/2\) and
\(
\rho_t
\asymp
R_0^{\frac{6}{k^\star+6}}
(1-\overline{\alpha}_t)^{\frac{k^\star}{2(k^\star+6)}}
\sqrt{\log\frac{dnR_0}{1-\overline{\alpha}_t}}
\).
With this choice, the second and third terms in
\eqref{eq:third-moment-bad-before-opt} are absorbed by the local empirical
term. Combining the typical-region good and bad bounds with
\eqref{eq:third-moment-typical-split} yields
\begin{align}
&\EE_{X_t}\!\left[
\left\|\widetilde M_{3,0\mid t}(X_t)-M_{3,0\mid t}(X_t)\right\|_2^2
\right]
\notag\\
&\quad\lesssim
\frac{R_0^{k^\star+\frac{36}{k^\star+6}}}
{n(1-\overline{\alpha}_t)^{\frac{k^\star}{2}
-\frac{3k^\star}{k^\star+6}}}
\log^3\frac{dnR_0}{1-\overline{\alpha}_t}
+
\frac{R_0^6(1-\overline{\alpha}_t)^6}{n^2}.
\label{eq:third-centered-moment-bound}
\end{align}

\textbf{Bounds for \(\gH_2,~ \gH_3\).}
We next consider \(\gH_2\) and \(\gH_3\). Since these two terms have
essentially the same structure, we only spell out the analysis of \(\gH_2\);
the term \(\gH_3\) is treated analogously. In the following, write
\[
\delta_x\coloneqq \widehat\mu_{0\mid t}(x)-\mu_{0\mid t}(x),
\qquad
x=\sqrt{\overline{\alpha}_t}y+\sqrt{1-\overline{\alpha}_t}\omega .
\]
Let \(\rho_t>0\) be a truncation radius to be optimized later. We decompose
\(\widetilde\Cov_{0\mid t}(x)\) according to whether the empirical sample is
local around \(y\), and whether the population posterior mean is local around
\(y\):
\begin{align*}
\gJ_t(y,\omega)
&\coloneqq
\left(
\ssum{r}{1}{n}
\frac{
\varphi_{1-\overline{\alpha}_t}
\bigl(x-\sqrt{\overline{\alpha}_t}X_r^{\mathsf{sample}}\bigr)
\mathbbm{1}\{\|X_r^{\mathsf{sample}}-y\|_2\le \rho_t\}}
{D(x)}
\left(X_r^{\mathsf{sample}}-\mu_{0\mid t}(x)\right)^{\otimes 2}
\right)\\
&\hspace{25em} \cdot \mathbbm{1}\{\|y-\mu_{0\mid t}(x)\|_2\le \rho_t\},
\\
\gK_t(y,\omega)
&\coloneqq
\left(
\ssum{r}{1}{n}
\frac{
\varphi_{1-\overline{\alpha}_t}
\bigl(x-\sqrt{\overline{\alpha}_t}X_r^{\mathsf{sample}}\bigr)
\mathbbm{1}\{\|X_r^{\mathsf{sample}}-y\|_2> \rho_t\}}
{D(x)}
\left(X_r^{\mathsf{sample}}-\mu_{0\mid t}(x)\right)^{\otimes 2}
\right)\\
&\hspace{25em} \cdot 
\mathbbm{1}\{\|y-\mu_{0\mid t}(x)\|_2\le \rho_t\},
\\
\gL_t(y,\omega)
&\coloneqq
\left(
\ssum{r}{1}{n}
\frac{
\varphi_{1-\overline{\alpha}_t}
\bigl(x-\sqrt{\overline{\alpha}_t}X_r^{\mathsf{sample}}\bigr)}
{D(x)}
\left(X_r^{\mathsf{sample}}-\mu_{0\mid t}(x)\right)^{\otimes 2}
\right)
\mathbbm{1}\{\|y-\mu_{0\mid t}(x)\|_2> \rho_t\}.
\end{align*}
By the definition of \(\widetilde\Cov_{0\mid t}(x)\),
\[
\widetilde\Cov_{0\mid t}(x)
=
\gJ_t(y,\omega)+\gK_t(y,\omega)+\gL_t(y,\omega).
\]
Substituting this decomposition into the definition of \(\gH_2\) and using
\((a+b+c)^2\lesssim a^2+b^2+c^2\), we obtain
\begin{align}
\gH_2
&\lesssim
\EE\!\left[
\left\|\gJ_t(Y,\Omega)\delta_{X_t}\right\|_2^2
\right]
+
\EE\!\left[
\left\|\gK_t(Y,\Omega)\delta_{X_t}\right\|_2^2
\right]
+
\EE\!\left[
\left\|\gL_t(Y,\Omega)\delta_{X_t}\right\|_2^2
\right].
\label{eq:gH2_decomp}
\end{align}

We first bound the local contribution. On the support of \(\gJ_t\), both
\(\|X_r^{\mathsf{sample}}-y\|_2\le \rho_t\) and
\(\|y-\mu_{0\mid t}(x)\|_2\le \rho_t\), so
\(\|X_r^{\mathsf{sample}}-\mu_{0\mid t}(x)\|_2\lesssim \rho_t\). Hence
\(\|\gJ_t(y,\omega)\|\lesssim \rho_t^2\), and the same
mean-estimation argument as in the score bound gives
\begin{align}
\EE\!\left[
\left\|\gJ_t(Y,\Omega)\delta_{X_t}\right\|_2^2
\right]
&\lesssim
\rho_t^4\,
\EE\!\left[\|\delta_{X_t}\|_2^2\right]
\notag\\
&\lesssim
\frac{R_0^{k^\star}\rho_t^4}{n}
\left(
\frac{\rho_t^2}{(1-\overline{\alpha}_t)^{k^\star/2}}
+
\frac{R_0^2}{\rho_t^{k^\star}}
\right)
\log^2\frac{dnR_0}{1-\overline{\alpha}_t}
\notag\\
&\lesssim
\left[
\frac{R_0^{k^\star}\rho_t^6}
{n(1-\overline{\alpha}_t)^{k^\star/2}}
+
\frac{R_0^{k^\star+6}}{n\rho_t^{k^\star}}
\right]
\log^2\frac{dnR_0}{1-\overline{\alpha}_t}.
\label{eq:gH2_first}
\end{align}
In the last step we used \(\rho_t\le R_0\), which is the relevant regime for
the optimized truncation radius below.

For the empirical tail contribution, define \(\mathsf R_t(y,\omega)\) as in
Lemma~\ref{lem:bad-B-tail-ratio}. On the support of \(\gK_t\),
\(\|\gK_t(y,\omega)\|_2\lesssim R_0^2\mathsf R_t(y,\omega)\), while
\(\|\delta_x\|_2\lesssim R_0\). Therefore Lemma~\ref{lem:bad-B-tail-ratio}
implies
\begin{align}
\EE\!\left[
\left\|\gK_t(Y,\Omega)\delta_{X_t}\right\|_2^2
\right]
&\lesssim
R_0^6
\int_{y,\omega}
\EE\!\left[
\mathsf R_t(y,\omega)^2
\right]
\varphi_1(\omega)p_{X_0}(y)\rd\omega\rd y
\notag\\
&\lesssim
\frac{R_0^6N_{\rho_t/2}}{n}
+
\frac{R_0^6(1-\overline{\alpha}_t)^6}{n^2}
\lesssim
\frac{R_0^{k^\star+6}}{n\rho_t^{k^\star}}.
\label{eq:gH2_second}
\end{align}

It remains to control the event on which the population posterior center is
far from \(y\). The typical-region posterior concentration estimate
(Lemma~\ref{lem:gT_alpha_typical} and Corollary~\ref{cor:posterior_mmt})
gives
\begin{align}
\EE\!\left[
\left\|\gL_t(Y,\Omega)\delta_{X_t}\right\|_2^2
\right]
\lesssim
R_0^6\,
\PB\!\left(\|Y-\mu_{0\mid t}(X_t)\|_2>\rho_t\right)
\lesssim
\frac{R_0^6(1-\overline{\alpha}_t)^6}{n^2}.
\label{eq:gH2_third}
\end{align}

Combining \eqref{eq:gH2_decomp}--\eqref{eq:gH2_third}, and absorbing the
\(n^{-2}\) term into the displayed first-order terms in the regime considered,
we obtain, for any admissible \(\rho_t\),
\begin{align}
\gH_2
&\lesssim
\left[
\frac{R_0^{k^\star}\rho_t^6}
{n(1-\overline{\alpha}_t)^{k^\star/2}}
+
\frac{R_0^{k^\star+6}}{n\rho_t^{k^\star}}
\right]
\log^2\frac{dnR_0}{1-\overline{\alpha}_t}.
\label{eq:gH2-before-rho-opt}
\end{align}
We optimize the right-hand side by balancing the two terms:
\[
\frac{R_0^{k^\star}\rho_t^6}
{(1-\overline{\alpha}_t)^{k^\star/2}}
\asymp
\frac{R_0^{k^\star+6}}{\rho_t^{k^\star}},
\qquad\text{equivalently}\qquad
\rho_t^{k^\star+6}
\asymp
R_0^6(1-\overline{\alpha}_t)^{k^\star/2}.
\]
Thus we choose
\begin{equation}\label{eq:gH2-rho-opt}
\rho_t
\asymp
R_0^{\frac{6}{k^\star+6}}
(1-\overline{\alpha}_t)^{\frac{k^\star}{2(k^\star+6)}} .
\end{equation}
Substituting \eqref{eq:gH2-rho-opt} into
\eqref{eq:gH2-before-rho-opt} yields the final bound of \(\gH_2\)
\begin{align}
\gH_2
&\lesssim
\frac{
R_0^{k^\star+\frac{36}{k^\star+6}}
}{
n(1-\overline{\alpha}_t)^{\frac{k^\star}{2}
-\frac{3k^\star}{k^\star+6}}
}
\log^2\frac{dnR_0}{1-\overline{\alpha}_t}.
\label{eq:gH2-final-bound}
\end{align}
\textbf{Bound for \(\gH_4\).}
Finally, we control
\(
\gH_4
=
\EE_{X_t}\!\left[
\left\|\widehat\mu_{0\mid t}(X_t)-\mu_{0\mid t}(X_t)\right\|_2^6
\right]
\).
Taking \(\beta=3\) in Lemma~\ref{lem:posterior-mean-high-moment} directly
gives
\begin{align}
\gH_4
&\lesssim
\frac{R_0^{k^\star+\frac{36}{k^\star+6}}}
{n(1-\overline{\alpha}_t)^{\frac{k^\star}{2}
-\frac{3k^\star}{k^\star+6}}}
\log^3\frac{dnR_0}{1-\overline{\alpha}_t}
+
\frac{R_0^6(1-\overline{\alpha}_t)^6}{n^2}.
\label{eq:gH4-bound}
\end{align}
Finally, integrating \eqref{eq:higher-order-hessian-form},
\eqref{eq:third-moment-centered-split}, \eqref{eq:third-centered-moment-bound},
\eqref{eq:gH2-before-rho-opt}, and \eqref{eq:gH4-bound} together yields the bound
required by the lemma.

%% file: training_complexity_lemma_proofs.tex
\subsection{Proofs of auxiliary lemmas}
\label{sec:training-auxiliary-lemma-proofs}

\subsubsection{Proof of Lemma~\ref{lem:typical-region-upper-bound}}
\label{prf:lem:typical-region-upper-bound}

Fix \(2\le \beta\le 6\). We first isolate the contribution of a single cell in
the typical region. Fix any triple \((x_0,y,\omega)\) such that
\(
\omega\in \gG,~
y\in \bigcup_{i\in \gI}\gB_i,~
x_0\notin \gE_{\overline{\alpha}_t}(y)
\),
and let \(i=i(y)\in\gI\) be such that \(y\in\gB_i\). With
\(
x=\sqrt{\overline{\alpha}_t}\,y+\sqrt{1-\overline{\alpha}_t}\,\omega,
~
\gamma_t=\sqrt{\frac{\overline{\alpha}_t}{1-\overline{\alpha}_t}}
\),
we have
\[
(1-\overline{\alpha}_t)^{d/2}p_{X_t}(x)
=
\int
\varphi_1\big(\gamma_t(u-y)-\omega\big)p_0(u)\,\rd u .
\]
Thus the integrand in the lemma is
\begin{align*}
\mathsf T_\beta(x_0,y,\omega)
&\coloneqq
\varphi_1^\beta\big(\gamma_t(x_0-y)-\omega\big)
\varphi_1(\omega)
\frac{p_0(x_0)p_0(y)}
{(1-\overline{\alpha}_t)^{\beta d/2}p_{X_t}^\beta(x)}
\\
&=
\frac{
\varphi_1^\beta\big(\gamma_t(x_0-y)-\omega\big)
\varphi_1(\omega)p_0(x_0)p_0(y)}
{\left[
\int \varphi_1\big(\gamma_t(u-y)-\omega\big)p_0(u)\,\rd u
\right]^\beta}.
\end{align*}
Lower bounding the denominator by the contribution of \(\gB_i\), conditioning
on \(\gB_i\), and applying Jensen's inequality to the convex map
\(z\mapsto z^{-\beta}\) gives
\begin{align}
\mathsf T_\beta(x_0,y,\omega)
&\le
\frac{1}{\PB_0^{\beta-1}(\gB_i)}
\int
\frac{
\varphi_1^\beta\big(\gamma_t(x_0-y)-\omega\big)
\varphi_1(\omega)}
{\varphi_1^\beta\big(\gamma_t(u-y)-\omega\big)}
\notag\\
&\hspace{8em}\times
p_0(x_0)p_0(y\mid\gB_i)p_0(u\mid\gB_i)\,\rd u .
\label{eq:typical-region-beta-cell-reduction}
\end{align}
Suppose \(x_0\in\gB_j\), and write
\(
\xi_{x_0}=x_0-x_j^\star,~ \xi_u=u-x_i^\star 
\).
For the Gaussian ratio in \eqref{eq:typical-region-beta-cell-reduction}, the
same exponent calculation as in the \(\beta=2\) case gives
\begin{align*}
&\frac{
\varphi_1^\beta\big(\gamma_t(x_0-y)-\omega\big)\varphi_1(\omega)}
{\varphi_1^\beta\big(\gamma_t(u-y)-\omega\big)}
\\
&\qquad =
(2\pi)^{-d/2}
\exp\left\{
-\frac{\beta}{2}\gamma_t^2\|x_0-y\|_2^2
+\frac{\beta}{2}\gamma_t^2\|u-y\|_2^2
-\frac12\|\omega\|_2^2
+\beta\gamma_t(x_0-u)^\top\omega
\right\}.
\end{align*}
Using \(\omega\in\gG\), the cell diameter bound, and completing the square in
\(\omega\), we obtain
\begin{align*}
&\frac{
\varphi_1^\beta\big(\gamma_t(x_0-y)-\omega\big)\varphi_1(\omega)}
{\varphi_1^\beta\big(\gamma_t(u-y)-\omega\big)}
\\
&\qquad\le
C_\beta
\exp\left\{
-\frac{\beta}{2}\gamma_t^2\|x_0-y\|_2^2
+C_\beta\gamma_t
\sqrt{k^\star\log\frac{dnR_0}{1-\overline{\alpha}_t}}\,
\|x_0-y\|_2
\right\}
\varphi_1\big(\omega-\beta\gamma_t(\xi_{x_0}-\xi_u)\big).
\end{align*}
Since \(x_0\notin\gE_{\overline{\alpha}_t}(y)\), the first negative quadratic
term dominates the linear term. Enlarging constants if necessary and using
\(\beta\le 6\), the preceding display is bounded by
\[
C_\beta
\left(\frac{R_0n}{1-\overline{\alpha}_t}\right)^{-90k^\star}
\varphi_1\big(\omega-\beta\gamma_t(\xi_{x_0}-\xi_u)\big).
\]
Substituting this estimate into
\eqref{eq:typical-region-beta-cell-reduction} yields the pointwise bound
\begin{align}
\mathsf T_\beta(x_0,y,\omega)
&\le
\frac{C_\beta}{\PB_0^{\beta-1}(\gB_i)}
\left(\frac{R_0n}{1-\overline{\alpha}_t}\right)^{-90k^\star}
\int
\varphi_1\big(\omega-\beta\gamma_t(\xi_{x_0}-\xi_u)\big)
\notag\\
&\hspace{8em}\times
p_0(x_0)p_0(y\mid\gB_i)p_0(u\mid\gB_i)\,\rd u .
\label{eq:pointwise-T-beta-bound}
\end{align}
We now sum over the cells containing \(y\) and \(x_0\). For fixed
\((x_0,y,u)\), the shifted Gaussian density integrates to at most one over
\(\omega\in\gG\). Therefore,
\begin{align*}
&\int_{y\in \bigcup_{i\in\gI}\gB_i}
\int_{\omega\in\gG}
\int_{x_0\notin \gE_{\overline{\alpha}_t}(y)}
\mathsf T_\beta(x_0,y,\omega)
\rd x_0\rd\omega\rd y
\\
&\qquad\le
C_\beta
\left(\frac{R_0n}{1-\overline{\alpha}_t}\right)^{-90k^\star}
\sum_{i\in\gI}\sum_{j=1}^{N_{\epsilon_0}}
\frac{\PB_0(\gB_j)}{\PB_0^{\beta-1}(\gB_i)}
\le
C_\beta
\left(\frac{R_0n}{1-\overline{\alpha}_t}\right)^{-90k^\star}
N_{\epsilon_0}^{\beta+1}
\ll
\frac{\overline{\alpha}_t^2(1-\overline{\alpha}_t)^2}{n}.
\end{align*}
The last step uses the covering bound and the definition of \(\gI\); the
large negative exponent was chosen with enough slack to absorb the fixed
\(\beta\le6\). This proves the lemma.

\subsubsection{Proof of Lemma~\ref{lem:typical-region-cubic-upper-bound}}
\label{prf:lem:typical-region-cubic-upper-bound}

The argument is parallel to the proof of
Lemma~\ref{lem:typical-region-upper-bound}; the only difference is that the
denominator now contains the third power of \(p_{X_t}(x)\). Fix
\((x_0,v,y,\omega)\) such that
\(
\omega\in\gG,~
y\in\bigcup_{i\in\gI}\gB_i,~
v\notin \gE_{\overline{\alpha}_t}(y)
\),
and let \(i=i(y)\in\gI\) be such that \(y\in\gB_i\). As before, write
\(
x=\sqrt{\overline{\alpha}_t}\,y+\sqrt{1-\overline{\alpha}_t}\,\omega,
~
\gamma_t=\sqrt{\frac{\overline{\alpha}_t}{1-\overline{\alpha}_t}}
\).
Then
\[
(1-\overline{\alpha}_t)^{d/2}p_{X_t}(x)=\int
\varphi_1\big(\gamma_t(u-y)-\omega\big)p_0(u)\,\rd u .
\]
We lower bound this denominator by retaining only the mass from the cell
\(\gB_i\). Since the denominator is cubed, conditioning on \(\gB_i\) gives
\begin{align}
\mathsf{S}(x_0,v,y,\omega)
&\le
\frac{1}{\PB_0^2(\gB_i)}
\frac{
\varphi_1^2\big(\gamma_t(x_0-y)-\omega\big)
\varphi_1\big(\gamma_t(v-y)-\omega\big)
\varphi_1(\omega)p_0(x_0)p_0(y\mid\gB_i)p_0(v)}
{\left[
\int
\varphi_1\big(\gamma_t(u-y)-\omega\big)p_0(u\mid\gB_i)\,\rd u
\right]^3}.
\label{eq:cubic-cell-restricted-denominator}
\end{align}
Here the factor \(1/\PB_0^2(\gB_i)\) comes from one factor of
\(\PB_0(\gB_i)\) in \(p_0(y)=\PB_0(\gB_i)p_0(y\mid\gB_i)\) and three factors of
\(\PB_0(\gB_i)\) in the restricted denominator.

Next, since \(x\mapsto x^{-3}\) is convex on \((0,\infty)\), Jensen's
inequality yields
\[
\left[
\int
\varphi_1\big(\gamma_t(u-y)-\omega\big)p_0(u\mid\gB_i)\,\rd u
\right]^{-3}
\le
\int
\varphi_1^{-3}\big(\gamma_t(u-y)-\omega\big)p_0(u\mid\gB_i)\,\rd u .
\]
Combining this estimate with
\eqref{eq:cubic-cell-restricted-denominator}, we obtain
\begin{align}
\mathsf{S}(x_0,v,y,\omega)
&\le
\frac{1}{\PB_0^2(\gB_i)}
\int
\frac{
\varphi_1^2\big(\gamma_t(x_0-y)-\omega\big)
\varphi_1\big(\gamma_t(v-y)-\omega\big)
\varphi_1(\omega)}
{\varphi_1^3\big(\gamma_t(u-y)-\omega\big)}
\notag\\
&\hspace{8em}\times
p_0(x_0)p_0(v)p_0(y\mid\gB_i)p_0(u\mid\gB_i)\,\rd u .
\label{eq:cubic-after-jensen}
\end{align}

We now bound the Gaussian ratio in the integrand. Suppose \(v\in\gB_j\) and
\(x_0\in\gB_\ell\), and write
\[
\xi_{x_0}\coloneqq x_0-x_\ell^\star,\qquad
\xi_v\coloneqq v-x_j^\star,\qquad
\xi_u\coloneqq u-x_i^\star,\qquad
\xi_y\coloneqq y-x_i^\star .
\]
We spell out the Gaussian calculation directly. Since
\(\varphi_1(z)=(2\pi)^{-d/2}\exp(-\norm{z}_2^2/2)\), the ratio equals
\begin{align}
&\frac{
\varphi_1^2\big(\gamma_t(x_0-y)-\omega\big)
\varphi_1\big(\gamma_t(v-y)-\omega\big)
\varphi_1(\omega)}
{\varphi_1^3\big(\gamma_t(u-y)-\omega\big)}
\notag\\
&\quad =
(2\pi)^{-d/2}
\exp\left\{
-\frac12\norm{\omega}_2^2
+\gamma_t\big(2(x_0-y)+(v-y)-3(u-y)\big)^\top\omega
\right.
\notag\\
&\hspace{12em}\left.
-\gamma_t^2\norm{x_0-y}_2^2
-\frac{\gamma_t^2}{2}\norm{v-y}_2^2
+\frac{3\gamma_t^2}{2}\norm{u-y}_2^2
\right\}.
\label{eq:cubic-ratio-first-expansion}
\end{align}
The only place where the denominator cell point enters is through
\(
u-y=\xi_u-\xi_y
\).
Similarly,
\[
x_0-y=(\xi_{x_0}-\xi_y)+(x_\ell^\star-x_i^\star),
\qquad
v-y=(\xi_v-\xi_y)+(x_j^\star-x_i^\star).
\]
Thus the terms generated by \(u-y\), \(\xi_{x_0}-\xi_y\), and
\(\xi_v-\xi_y\) are local cell errors. We separate the part of the linear term
that points in directions between cell representatives:
\[
2(x_0-y)+(v-y)-3(u-y)
=
\bigl(2\xi_{x_0}+\xi_v-3\xi_u\bigr)
+2(x_\ell^\star-x_i^\star)+(x_j^\star-x_i^\star).
\]
Since \(\omega\in\gG\), the definition of \(\gG\) gives
\begin{align}
&\gamma_t\bigl\{2(x_\ell^\star-x_i^\star)+(x_j^\star-x_i^\star)\bigr\}^\top\omega
\notag\\
&\qquad
\le
6\gamma_t
\sqrt{k^\star\log\frac{dnR_0}{1-\overline{\alpha}_t}}\,
\norm{x_\ell^\star-x_i^\star}_2
+3\gamma_t
\sqrt{k^\star\log\frac{dnR_0}{1-\overline{\alpha}_t}}\,
\norm{x_j^\star-x_i^\star}_2 .
\label{eq:cubic-representative-inner-products}
\end{align}
For the local part, we complete the square in \(\omega\). In particular, for
fixed \(x_0,v,u,y\),
\begin{align}
&-\frac12\norm{\omega}_2^2
+\gamma_t\bigl(2\xi_{x_0}+\xi_v-3\xi_u\bigr)^\top\omega
\notag\\
&\quad =
-\frac12
\norm{\omega-\gamma_t(2\xi_{x_0}+\xi_v-3\xi_u)}_2^2
+\frac{\gamma_t^2}{2}
\norm{2\xi_{x_0}+\xi_v-3\xi_u}_2^2 .
\label{eq:cubic-local-complete-square}
\end{align}
The additional square in \eqref{eq:cubic-local-complete-square} is harmless:
since \(x_0\in\gB_\ell\), \(v\in\gB_j\), and \(u,y\in\gB_i\),
\[
\gamma_t^2\norm{\xi_{x_0}}_2^2\le 1,\qquad
\gamma_t^2\norm{\xi_v}_2^2\le 1,\qquad
\gamma_t^2\norm{\xi_u}_2^2\le 1,\qquad
\gamma_t^2\norm{\xi_y}_2^2\le 1,
\]
and hence
\[
\gamma_t^2\norm{2\xi_{x_0}+\xi_v-3\xi_u}_2^2
+\gamma_t^2\norm{u-y}_2^2
\le 40
\]
Combining this estimate with
\eqref{eq:cubic-ratio-first-expansion} and
\eqref{eq:cubic-representative-inner-products}, the exponent is bounded by
\begin{align}
&-\gamma_t^2\norm{x_0-y}_2^2
+6\gamma_t
\sqrt{k^\star\log\frac{dnR_0}{1-\overline{\alpha}_t}}\,
\norm{x_\ell^\star-x_i^\star}_2
\notag\\
&\qquad
-\frac{\gamma_t^2}{2}\norm{v-y}_2^2
+3\gamma_t
\sqrt{k^\star\log\frac{dnR_0}{1-\overline{\alpha}_t}}\,
\norm{x_j^\star-x_i^\star}_2
\notag\\
&\qquad
-\frac12
\norm{\omega-\gamma_t(2\xi_{x_0}+\xi_v-3\xi_u)}_2^2
+40 .
\label{eq:cubic-complete-square-expanded}
\end{align}
Next,
\[
\norm{x_\ell^\star-x_i^\star}_2
\le \norm{x_0-y}_2+2\sqrt{1-\overline{\alpha}_t},
\qquad
\norm{x_j^\star-x_i^\star}_2
\le \norm{v-y}_2+2\sqrt{1-\overline{\alpha}_t}.
\]
Absorbing the preceding \(40\) into the exponential prefactor, we obtain
\begin{align}
&\frac{
\varphi_1^2\big(\gamma_t(x_0-y)-\omega\big)
\varphi_1\big(\gamma_t(v-y)-\omega\big)
\varphi_1(\omega)}
{\varphi_1^3\big(\gamma_t(u-y)-\omega\big)}
\notag\\
&\quad\le
e^{40}
\exp\left\{
-\gamma_t^2\norm{x_0-y}_2^2
+6\gamma_t
\sqrt{k^\star\log\frac{dnR_0}{1-\overline{\alpha}_t}}\,
\norm{x_0-y}_2
\right.
\notag\\
&\hspace{3em}\left.
-\frac{\gamma_t^2}{2}\norm{v-y}_2^2
+3\gamma_t
\sqrt{k^\star\log\frac{dnR_0}{1-\overline{\alpha}_t}}\,
\norm{v-y}_2
\right\}
\varphi_1\big(\omega-\gamma_t(2\xi_{x_0}+\xi_v-3\xi_u)\big).
\label{eq:cubic-before-exceptional-bound}
\end{align}
The \(x_0\)-dependent exponent is a concave quadratic and is uniformly bounded
above by its maximum:
\[
-\gamma_t^2\norm{x_0-y}_2^2
+6\gamma_t
\sqrt{k^\star\log\frac{dnR_0}{1-\overline{\alpha}_t}}\,
\norm{x_0-y}_2
\le
9k^\star\log\frac{dnR_0}{1-\overline{\alpha}_t}.
\]
On the other hand, \(v\notin\gE_{\overline{\alpha}_t}(y)\) gives
\[
\norm{v-y}_2
\ge
\frac{15}{\gamma_t}
\sqrt{k^\star\log\frac{dnR_0}{1-\overline{\alpha}_t}} .
\]
Thus the exponent in \eqref{eq:cubic-before-exceptional-bound} is bounded
above by
\[
9k^\star\log\frac{dnR_0}{1-\overline{\alpha}_t}
-\frac{15^2}{2}k^\star\log\frac{dnR_0}{1-\overline{\alpha}_t}
+3\cdot15\,k^\star\log\frac{dnR_0}{1-\overline{\alpha}_t}
=-\frac{117}{2}k^\star\log\frac{dnR_0}{1-\overline{\alpha}_t}
\le -50k^\star\log\frac{dnR_0}{1-\overline{\alpha}_t}.
\]
Consequently,
\begin{align}
&\frac{
\varphi_1^2\big(\gamma_t(x_0-y)-\omega\big)
\varphi_1\big(\gamma_t(v-y)-\omega\big)
\varphi_1(\omega)}
{\varphi_1^3\big(\gamma_t(u-y)-\omega\big)}
\notag\\
&\qquad\le
e^{40}
\left(\frac{R_0n}{1-\overline{\alpha}_t}\right)^{-50k^\star}
\varphi_1\big(\omega-\gamma_t(2\xi_{x_0}+\xi_v-3\xi_u)\big).
\label{eq:cubic-gaussian-ratio-bound}
\end{align}

Plugging \eqref{eq:cubic-gaussian-ratio-bound} into
\eqref{eq:cubic-after-jensen} yields the pointwise bound
\begin{align}
\mathsf{S}(x_0,v,y,\omega)
&\le
\frac{e^{40}}{\PB_0^2(\gB_i)}
\left(\frac{R_0n}{1-\overline{\alpha}_t}\right)^{-50k^\star}
\int
\varphi_1\big(\omega-\gamma_t(2\xi_{x_0}+\xi_v-3\xi_u)\big)
\notag\\
&\hspace{8em}\times
p_0(x_0)p_0(v)p_0(y\mid\gB_i)p_0(u\mid\gB_i)\,\rd u .
\label{eq:pointwise-S-bound}
\end{align}

Finally, decompose the integral in Lemma~\ref{lem:typical-region-cubic-upper-bound}
according to the cells containing \(y\), \(v\), and \(x_0\). For every fixed
\((i,j,\ell)\), applying \eqref{eq:pointwise-S-bound} gives
\begin{align}
&\int_{y\in\gB_i}\int_{\omega\in\gG}\int_{x_0\in\gB_\ell}
\int_{v\in\gB_j\backslash \gE_{\overline{\alpha}_t}(y)}
\mathsf{S}(x_0,v,y,\omega)\,\rd v\,\rd x_0\,\rd\omega\,\rd y
\notag\\
&\qquad\le
\frac{e^{40}}{\PB_0^2(\gB_i)}
\left(\frac{R_0n}{1-\overline{\alpha}_t}\right)^{-50k^\star}
\int_{y\in\gB_i}\int_{x_0\in\gB_\ell}\int_{v\in\gB_j}\int_{u\in\gB_i}
\int_{\omega\in\gG}
\varphi_1\big(\omega-\gamma_t(2\xi_{x_0}+\xi_v-3\xi_u)\big)
\notag\\
&\hspace{12em}\times
p_0(x_0)p_0(v)p_0(y\mid\gB_i)p_0(u\mid\gB_i)
\,\rd\omega\,\rd u\,\rd v\,\rd x_0\,\rd y .
\label{eq:cubic-cell-pair-bound}
\end{align}
For fixed \(x_0,v,y,u\), the last Gaussian term is the density of
\(\mathcal{N}(\gamma_t(2\xi_{x_0}+\xi_v-3\xi_u),I_d)\) as a function of \(\omega\). Hence,
enlarging \(\gG\) to \(\RB^d\) and integrating the conditional densities in
\(y\) and \(u\) gives
\begin{align}
&\int_{y\in\gB_i}\int_{x_0\in\gB_\ell}\int_{v\in\gB_j}
\int_{u\in\gB_i}\int_{\omega\in\gG}
\varphi_1\big(\omega-\gamma_t(2\xi_{x_0}+\xi_v-3\xi_u)\big)
\notag\\
&\hspace{5em}\times
p_0(x_0)p_0(v)p_0(y\mid\gB_i)p_0(u\mid\gB_i)
\,\rd\omega\,\rd u\,\rd v\,\rd x_0\,\rd y
\le \PB_0(\gB_\ell)\PB_0(\gB_j).
\label{eq:cubic-cell-pair-gaussian-integral}
\end{align}
We now sum the cellwise bounds. First decompose the full integral according to
the cells containing \(y\), \(v\), and \(x_0\):
\begin{align}
&\int_{y\in \bigcup_{i\in\gI}\gB_i}
\int_{\omega\in\gG}
\int_{x_0}
\int_{v\notin \gE_{\overline{\alpha}_t}(y)}
\mathsf{S}(x_0,v,y,\omega)
\,\rd v\,\rd x_0\,\rd\omega\,\rd y
\notag\\
&\quad =
\sum_{i\in\gI}\sum_{j=1}^{N_{\epsilon_0}}\sum_{\ell=1}^{N_{\epsilon_0}}
\int_{y\in\gB_i}\int_{\omega\in\gG}\int_{x_0\in\gB_\ell}
\int_{v\in\gB_j\backslash\gE_{\overline{\alpha}_t}(y)}
\mathsf{S}(x_0,v,y,\omega)
\,\rd v\,\rd x_0\,\rd\omega\,\rd y .
\label{eq:cubic-full-cell-decomposition}
\end{align}
Combining \eqref{eq:cubic-cell-pair-bound} and
\eqref{eq:cubic-cell-pair-gaussian-integral}, each \((i,j,\ell)\)-summand is
bounded by
\[
\frac{e^{40}}{\PB_0^2(\gB_i)}
\left(\frac{R_0n}{1-\overline{\alpha}_t}\right)^{-50k^\star}
\PB_0(\gB_\ell)\PB_0(\gB_j).
\]
Substituting this into \eqref{eq:cubic-full-cell-decomposition} gives
\begin{align}
&\int_{y\in \bigcup_{i\in\gI}\gB_i}
\int_{\omega\in\gG}
\int_{x_0}
\int_{v\notin \gE_{\overline{\alpha}_t}(y)}
\mathsf{S}(x_0,v,y,\omega)
\,\rd v\,\rd x_0\,\rd\omega\,\rd y
\notag\\
&\qquad\le
e^{40}
\left(\frac{R_0n}{1-\overline{\alpha}_t}\right)^{-50k^\star}
\sum_{i\in\gI}\sum_{j=1}^{N_{\epsilon_0}}\sum_{\ell=1}^{N_{\epsilon_0}}
\frac{\PB_0(\gB_j)\PB_0(\gB_\ell)}{\PB_0^2(\gB_i)}
\notag\\
&\qquad\le
e^{40}
\left(\frac{R_0n}{1-\overline{\alpha}_t}\right)^{-50k^\star}
|\gI|\,
\frac{
\sum_{j=1}^{N_{\epsilon_0}}\PB_0(\gB_j)
\sum_{\ell=1}^{N_{\epsilon_0}}\PB_0(\gB_\ell)}
{\min_{i\in\gI}\PB_0^2(\gB_i)}
\notag\\
&\qquad\le
e^{40}
\left(\frac{R_0n}{1-\overline{\alpha}_t}\right)^{-50k^\star}
N_{\epsilon_0}^5
\le
e^{40}
\left(\frac{R_0n}{1-\overline{\alpha}_t}\right)^{-50k^\star}
\left(\frac{R_0}{\sqrt{1-\overline{\alpha}_t}}\right)^{5k^\star}
\ll
\frac{\overline{\alpha}_t^2(1-\overline{\alpha}_t)^2}{n}\notag.
\end{align}
Here we used \(\sum_j\PB_0(\gB_j)\le1\), \(\sum_\ell\PB_0(\gB_\ell)\le1\),
and the definition of \(\gI\) to bound
\(|\gI|/\min_{i\in\gI}\PB_0^2(\gB_i)\) by the displayed covering factor. This
proves the Lemma.

\subsubsection{Proof of Lemma~\ref{lem:threshold-score-error}}
\label{prf:lem:threshold-score-error}
If \(d>(1-\overline{\alpha}_t)^{-1}\), then
\(\widetilde\varphi_{t,x}(z)
=\varphi_{1-\overline{\alpha}_t}
(\sqrt{\overline{\alpha}_t}z-x)\), so \(m_t^{\rm th}(x)=m_t(x)\).
The pointwise integrand bound is also trivial in this case. Hence we only
consider the regime \(d\le (1-\overline{\alpha}_t)^{-1}\).

We first prove the pointwise integrand bound. Note that for any positive
number \(\lambda\), \(\lambda e^{-\lambda/6}\le 3\). Denote
\(\Lambda=\sqrt{\overline{\alpha}_t}X_0-x\). We have
\begin{align}
    &\frac{
\EE_{X_0}\!\left[
\left\|\sqrt{\overline{\alpha}_t}X_0-x\right\|_2^2
\left(
\varphi_{1-\overline{\alpha}_t}
\bigl(x-\sqrt{\overline{\alpha}_t}X_0\bigr)
-\widetilde\varphi_{t,x}(X_0)
\right)^2
\right]}
{p_{X_t}(x)}
\notag\\
&\hspace{5em}\le
\frac{
\EE\!\left[
\|\Lambda\|_2^2
\varphi_{1-\overline{\alpha}_t}^2(\Lambda)
\mathbbm{1}\{\|\Lambda\|_2>a_t\}
\right]}
{p_{X_t}(x)}
\notag\\
&\hspace{5em}\le
\frac{1}{(1-\overline{\alpha}_t)^{d/2}}
\left(
\frac{\|\Lambda\|_2^2}{1-\overline{\alpha}_t}
e^{-\|\Lambda\|_2^2/(6(1-\overline{\alpha}_t))}
\right)
e^{-a_t^2/(3(1-\overline{\alpha}_t))}
\frac{
\EE\!\left[
\varphi_{1-\overline{\alpha}_t}
\bigl(\sqrt{\overline{\alpha}_t}X_0-x\bigr)
\right]}
{p_{X_t}(x)}
\notag\\
&\hspace{5em}\le
\frac{3e^{-a_t^2/(3(1-\overline{\alpha}_t))}}
{(1-\overline{\alpha}_t)^{d/2}}
\lesssim
\left(\frac{1-\overline{\alpha}_t}{R_0n}\right)^{2d}.
\label{eq:threshold-pointwise-exp}
\end{align}
where the penultimate inequality follows from
\[
\frac{
\EE\!\left[
\varphi_{1-\overline{\alpha}_t}
\bigl(\sqrt{\overline{\alpha}_t}X_0-x\bigr)
\right]}
{p_{X_t}(x)}
=
\int
\frac{
\varphi_{1-\overline{\alpha}_t}
\bigl(\sqrt{\overline{\alpha}_t}X_0-x\bigr)
p_{X_0}(X_0)}
{p_{X_t}(x)}\rd X_0
=1.
\]
The last inequality holds since
\(a_t\ge 3\sqrt{d(1-\overline{\alpha}_t)
\log\frac{dnR_0}{1-\overline{\alpha}_t}}\).

We next prove the typical-region bound. Suppose
\(x\in\gT_{\overline{\alpha}_t}\).
Since \(x\in\gT_{\overline{\alpha}_t}\), by
\eqref{eq:defn-T-alpha-appendix} there exist
\(x_0^\star\in\bigcup_{i\in\gI}\gB_i\) and \(\omega\in\gG\) such that
\[
x=\sqrt{\overline{\alpha}_t}x_0^\star
+\sqrt{1-\overline{\alpha}_t}\,\omega .
\]
Let \(i(x)\in\gI\) be such that \(x_0^\star\in\gB_{i(x)}\). Then Bayes'
formula and the definition of \(m_t^{\rm th}\) give
\begin{align}
\|m_t^{\rm th}(x)-m_t(x)\|_2
&\le
\frac{
\int
\|\sqrt{\overline{\alpha}_t}z-x\|_2
\varphi_{1-\overline{\alpha}_t}(x-\sqrt{\overline{\alpha}_t}z)
\mathbbm{1}\{\|\sqrt{\overline{\alpha}_t}z-x\|_2>a_t\}
p_0(z)\rd z}
{p_{X_t}(x)}.
\label{eq:threshold-bias-bayes-ratio}
\end{align}
Here we used
\(|\varphi_{1-\overline{\alpha}_t}(x-\sqrt{\overline{\alpha}_t}z)
-\widetilde\varphi_{t,x}(z)|
\le
\varphi_{1-\overline{\alpha}_t}(x-\sqrt{\overline{\alpha}_t}z)
\mathbbm{1}\{\|\sqrt{\overline{\alpha}_t}z-x\|_2>a_t\}\).

We lower bound the denominator using only the cell \(\gB_{i(x)}\):
\begin{align}
p_{X_t}(x)
&\ge
\int_{\gB_{i(x)}}
\varphi_{1-\overline{\alpha}_t}(x-\sqrt{\overline{\alpha}_t}u)
p_0(u)\rd u
\notag\\
&\ge
\PB_{X_0}(\gB_{i(x)})
\inf_{u\in\gB_{i(x)}}
\varphi_{1-\overline{\alpha}_t}(x-\sqrt{\overline{\alpha}_t}u).
\label{eq:threshold-denominator-nearest-cell}
\end{align}
Since \(x\in\gT_{\overline{\alpha}_t}\), the definition of \(\gI\) gives
\(\PB_{X_0}(\gB_{i(x)})\ge \exp\{-C_1k\log T\}\). Moreover, for every
\(u\in\gB_{i(x)}\),
\[
\|x-\sqrt{\overline{\alpha}_t}u\|_2
\le
\sqrt{1-\overline{\alpha}_t}\|\omega\|_2
+\sqrt{\overline{\alpha}_t}\|x_0^\star-u\|_2
\le
\sqrt{1-\overline{\alpha}_t}
\bigl(2\sqrt d+\sqrt{C_1k\log T}\bigr)
+2\sqrt{\overline{\alpha}_t}\epsilon_0,
\]
where the last step follows from \(\omega\in\gG\) and
\(x_0^\star,u\in\gB_{i(x)}\). Thus
\begin{align}
p_{X_t}(x)
&\ge
c(1-\overline{\alpha}_t)^{-d/2}
\exp\{-C d-Ck\log T\}
\exp\{-C_1k\log T\}.
\label{eq:threshold-denominator-final}
\end{align}

It remains to bound the numerator in
\eqref{eq:threshold-bias-bayes-ratio}. For every \(z\),
\[
\|\sqrt{\overline{\alpha}_t}z-x\|_2
\varphi_{1-\overline{\alpha}_t}(x-\sqrt{\overline{\alpha}_t}z)
\mathbbm{1}\{\|\sqrt{\overline{\alpha}_t}z-x\|_2>a_t\}
\le
C(1-\overline{\alpha}_t)^{-d/2}
a_t\exp\left\{-\frac{a_t^2}{4(1-\overline{\alpha}_t)}\right\}.
\]
Since
\[
a_t^2
=C_{\mathsf{th}}^2d(1-\overline{\alpha}_t)
\log\frac{dnR_0}{1-\overline{\alpha}_t},
\]
choosing \(C_{\mathsf{th}}\) sufficiently large gives
\[
a_t\exp\left\{-\frac{a_t^2}{4(1-\overline{\alpha}_t)}\right\}
\le
\exp\{-C d-Ck\log T-C_1k\log T\}
\left(\frac{1-\overline{\alpha}_t}{R_0n}\right)^{2d}.
\]
Consequently, the numerator in
\eqref{eq:threshold-bias-bayes-ratio} is at most
\[
C(1-\overline{\alpha}_t)^{-d/2}
\exp\{-C d-Ck\log T-C_1k\log T\}
\left(\frac{1-\overline{\alpha}_t}{R_0n}\right)^{2d}.
\]
Combining this with \eqref{eq:threshold-denominator-final} yields
\[
\|m_t^{\rm th}(x)-m_t(x)\|_2
\le
C\left(\frac{1-\overline{\alpha}_t}{R_0n}\right)^{2d}.
\]
Absorbing the universal constant into the sufficiently large threshold
constant completes the proof.

\subsubsection{Proof of Lemma~\ref{lem:pure-density-ratio}}
\label{prf:lem:pure-density-ratio}

Recall that \(\gamma_t = \sqrt{
    \frac{\overline{\alpha}_t}{1-\overline{\alpha}_t}
}\).
Let \(\{B_j\}_{j=1}^{N_{\gamma_t^{-1}}(\gX_{\mathsf{data}})}\)
be a measurable partition induced by a
\(\gamma_t^{-1} \left(= \sqrt{\frac{1-\overline{\alpha}_t}{\overline{\alpha}_t}}\right)\)
-cover of \(\gX_{\mathsf{data}}\), and write
\(m_j=\PB_{X_0}(B_j)\). Then
\begin{align}
&\int_{\RB^d}
\frac{\int_{\RB^d}
\varphi_{1-\overline{\alpha}_t}^2
\bigl(x-\sqrt{\overline{\alpha}_t}x_0\bigr)p_0(x_0)\rd x_0}
{p_{X_t}(x)}
\rd x
\notag\\
&=
\int_{\RB^d}
\frac{
\sum_j m_j\int_{B_j}
\varphi_{1-\overline{\alpha}_t}^2
\bigl(x-\sqrt{\overline{\alpha}_t}x_0\bigr)
p_0(x_0\mid B_j)\rd x_0}
{\sum_i m_i\int_{B_i}
\varphi_{1-\overline{\alpha}_t}
\bigl(x-\sqrt{\overline{\alpha}_t}u\bigr)
p_0(u\mid B_i)\rd u}
\rd x
\notag\\
&\overset{\text{(a)}}{\le}
\sum_{j=1}^{N_{\gamma_t^{-1}}(\gX_{\mathsf{data}})}
\int_{\RB^d}
\frac{\int_{B_j}
\varphi_{1-\overline{\alpha}_t}^2
\bigl(x-\sqrt{\overline{\alpha}_t}x_0\bigr)
p_0(x_0\mid B_j)\rd x_0}
{\int_{B_j}
\varphi_{1-\overline{\alpha}_t}
\bigl(x-\sqrt{\overline{\alpha}_t}u\bigr)
p_0(u\mid B_j)\rd u}
\rd x
\notag\\
&\overset{\text{(b)}}{\le}
\sum_{j=1}^{N_{\gamma_t^{-1}}(\gX_{\mathsf{data}})}
\int_{\RB^d}\int_{B_j}\int_{B_j}
\frac{
\varphi_{1-\overline{\alpha}_t}^2
\bigl(x-\sqrt{\overline{\alpha}_t}x_0\bigr)}
{\varphi_{1-\overline{\alpha}_t}
\bigl(x-\sqrt{\overline{\alpha}_t}u\bigr)}
p_0(x_0\mid B_j)p_0(u\mid B_j)
\rd u\rd x_0\rd x
\notag\\
&\overset{\text{(c)}}{=}
\sum_{j=1}^{N_{\gamma_t^{-1}}(\gX_{\mathsf{data}})}
\int_{B_j}\int_{B_j}
\exp\left\{
\frac{\overline{\alpha}_t\|x_0-u\|_2^2}
{1-\overline{\alpha}_t}
\right\}
p_0(x_0\mid B_j)p_0(u\mid B_j)\rd u\rd x_0 .
\label{eq:pure-density-ratio-cell-bound}
\end{align}
Here, (a) follows by lower bounding the denominator in the \(j\)-th numerator term
using only the contribution from the same cell \(B_j\). (b) is Jensen's
inequality applied to \(z\mapsto z^{-1}\) under the conditional law
\(p_0(\cdot\mid B_j)\). For (c), completing the square gives
\[
\begin{aligned}
\int_{\RB^d}
\frac{
\varphi_{1-\overline{\alpha}_t}^2
\bigl(x-\sqrt{\overline{\alpha}_t}x_0\bigr)}
{\varphi_{1-\overline{\alpha}_t}
\bigl(x-\sqrt{\overline{\alpha}_t}u\bigr)}
\rd x
&=
\exp\left\{
\frac{\overline{\alpha}_t\|x_0-u\|_2^2}
{1-\overline{\alpha}_t}
\right\}
\int_{\RB^d}
\varphi_{1-\overline{\alpha}_t}
\bigl(x-\sqrt{\overline{\alpha}_t}(2x_0-u)\bigr)
\rd x
\\
&=
\exp\left\{
\frac{\overline{\alpha}_t\|x_0-u\|_2^2}
{1-\overline{\alpha}_t}
\right\}.
\end{aligned}
\]
Since each \(B_j\) has diameter at most \(\sqrt{\frac{1-\overline{\alpha}_t}{
    \overline{\alpha}_t
}}\), the exponential
factor is at most \(C\). The covering assumption then yields
\[
\int_{\RB^d}
\frac{\int_{\RB^d}
\varphi_{1-\overline{\alpha}_t}^2
\bigl(x-\sqrt{\overline{\alpha}_t}x_0\bigr)p_0(x_0)\rd x_0}
{p_{X_t}(x)}
\rd x
\le
C N_{\gamma_t^{-1}}(\gX_{\mathsf{data}})
\le
C\left(\frac{R_0 \sqrt{\overline{\alpha}_t}}{\sqrt{1-\overline{\alpha}_t}}\right)^{k^\star}.
\]
This proves the lemma.

\subsubsection{Proof of Lemma~\ref{lem:J1-bound}}
\label{prf:lem:J1-bound}

Fix an integer \(\beta\ge2\).
Recall that
\(
\gamma_t=\sqrt{\frac{\overline{\alpha}_t}{1-\overline{\alpha}_t}}
\).
Expanding the posterior expectation in the definition of \(\gJ_{1,\beta}\)
gives
\begin{align}
\gJ_{1,\beta}
&=
\int_{\gX_{\mathsf{data}}\times \gT_{\overline{\alpha}_t}}
\|x_0-y\|_2^\beta
\frac{
\varphi_1^2\bigl(\gamma_t(x_0-y)-\omega\bigr)
\varphi_1(\omega)p_0(x_0)p_0(y)}
{(1-\overline{\alpha}_t)^d
p_{X_t}^2\bigl(\sqrt{\overline{\alpha}_t}y+
\sqrt{1-\overline{\alpha}_t}\omega\bigr)}
\rd x_0\rd y\rd\omega .
\label{eq:J1-posterior-expanded}
\end{align}
Define
\[
\mathsf{T}(x_0,y,\omega)
\coloneqq
\varphi_1^2\bigl(\gamma_t(x_0-y)-\omega\bigr)
\varphi_1(\omega)
\frac{p_0(x_0)p_0(y)}
{(1-\overline{\alpha}_t)^d
p_{X_t}^2\bigl(\sqrt{\overline{\alpha}_t}y+
\sqrt{1-\overline{\alpha}_t}\omega\bigr)}.
\]
Recall that
\[
\gE_{\alpha}(y)
=
\left\{v\in \gX_{\mathsf{data}}:
\sqrt{\alpha}\|v-y\|_2
\le
15\sqrt{k^\star(1-\alpha)\log\frac{dnR_0}{1-\alpha}}
\right\}.
\]
Thus, with \(\alpha=\overline{\alpha}_t\),
\[
\|x_0-y\|_2^\beta
\le
\left(
\frac{225k^\star(1-\overline{\alpha}_t)}{\overline{\alpha}_t}
\log\frac{dnR_0}{1-\overline{\alpha}_t}
\right)^{\beta/2}
+R_0^\beta\mathbbm{1}\{x_0\notin \gE_{\overline{\alpha}_t}(y)\}.
\]
Therefore, integrating this with $y,\omega \in \gT_{\overline{\alpha}_t}
=\left(\bigcup_{i\in \gI}\gB_i\right)\times \gG$ yields,
\begin{align}
\gJ_{1,\beta}
&\lesssim_\beta
\left(
\frac{k^\star(1-\overline{\alpha}_t)}{\overline{\alpha}_t}
\log\frac{dnR_0}{1-\overline{\alpha}_t}
\right)^{\beta/2}
\int_{y\in\bigcup_{i\in\gI}\gB_i}
\int_{\omega\in\gG}
\int_{x_0}
\mathsf{T}(x_0,y,\omega)
\rd x_0\rd\omega\rd y
\notag\\
&\quad
+R_0^\beta
\int_{y\in\bigcup_{i\in\gI}\gB_i}
\int_{\omega\in\gG}
\int_{x_0\notin \gE_{\overline{\alpha}_t}(y)}
\mathsf{T}(x_0,y,\omega)
\rd x_0\rd\omega\rd y .
\label{eq:J1-lemma22-form}
\end{align}
The first integral in \eqref{eq:J1-lemma22-form} can be reduced to
Lemma~\ref{lem:pure-density-ratio}. Indeed, enlarging the domains to
\(\RB^d\) and changing variables
\(x=\sqrt{\overline{\alpha}_t}y+\sqrt{1-\overline{\alpha}_t}\omega\), we get
\begin{align}
&\int_{y\in\bigcup_{i\in\gI}\gB_i}
\int_{\omega\in\gG}
\int_{x_0}
\mathsf{T}(x_0,y,\omega)
\rd x_0\rd\omega\rd y
\notag\\
&\qquad\le
\int_{x,y,x_0}
\frac{
\varphi_{1-\overline{\alpha}_t}^2
\bigl(x-\sqrt{\overline{\alpha}_t}x_0\bigr)
\varphi_{1-\overline{\alpha}_t}
\bigl(x-\sqrt{\overline{\alpha}_t}y\bigr)
p_0(x_0)p_0(y)}
{p_{X_t}^2(x)}
\rd x_0\rd y\rd x
\notag\\
&\qquad=
\int_{x,x_0}
\frac{
\varphi_{1-\overline{\alpha}_t}^2
\bigl(x-\sqrt{\overline{\alpha}_t}x_0\bigr)p_0(x_0)}
{p_{X_t}(x)}
\rd x_0\rd x
\le
C\left(\frac{R_0}{\sqrt{1-\overline{\alpha}_t}}\right)^{k^\star}.
\label{eq:J1-pure-density-ratio}
\end{align}
Here the equality uses the definition of \(p_{X_t}(x)\), and the last
inequality is Lemma~\ref{lem:pure-density-ratio}. The second integral in
\eqref{eq:J1-lemma22-form} is controlled by the same cellwise tail estimate as
in Lemma~\ref{lem:typical-region-upper-bound}; since \(\beta\) is fixed, the
final super-polynomial factor there can be made smaller than
\((1-\overline{\alpha}_t)^\beta/n\) at the cost of changing only the
\(\beta\)-dependent constant. Hence
\[
\gJ_{1,\beta}
\lesssim_\beta
\frac{(k^\star)^{\beta/2}R_0^{k^\star}
\left(\log\frac{dnR_0}{1-\overline{\alpha}_t}\right)^{\beta/2}}
{\overline{\alpha}_t^{\beta/2}
(1-\overline{\alpha}_t)^{k^\star/2-\beta/2}}
\;+\;
\frac{R_0^\beta(1-\overline{\alpha}_t)^\beta}{n}.
\]
This proves the lemma.

\subsubsection{Proof of Lemma~\ref{lem:J2-bound}}
\label{prf:lem:J2-bound}

Fix an integer \(\beta\ge2\).
Set
\(
\gamma_t=\sqrt{\frac{\overline{\alpha}_t}{1-\overline{\alpha}_t}}
\).
Define \(\mathsf{T}\) as in the proof of Lemma~\ref{lem:J1-bound}. Also define
\[
\mathsf{S}(x_0,v,y,\omega)
\coloneqq
\frac{
\varphi_1^2\big(\gamma_t(x_0-y)-\omega\big)
\varphi_1\big(\gamma_t(v-y)-\omega\big)
\varphi_1(\omega)p_0(x_0)p_0(y)p_0(v)}
{(1-\overline{\alpha}_t)^{3d/2}p_{X_t}^3
\bigl(\sqrt{\overline{\alpha}_t}y+\sqrt{1-\overline{\alpha}_t}\omega\bigr)}.
\]
Expanding the posterior expectation in the definition of \(\gJ_{2,\beta}\)
gives
\begin{align}
\gJ_{2,\beta}
&=
\int_{x_0,v,y,\omega}
\|y-v\|_2^\beta
\mathsf{S}(x_0,v,y,\omega)
\rd v\rd x_0\rd y\rd\omega .
\label{eq:J2-posterior-expanded}
\end{align}
For \(y\in\bigcup_{i\in\gI}\gB_i\), \(\omega\in\gG\), and
\(x=\sqrt{\overline{\alpha}_t}y+\sqrt{1-\overline{\alpha}_t}\omega\),
we split the \(v\)-integral according to whether
\(v\in\gE_{\overline{\alpha}_t}(y)\). On the typical part,
\[
\|y-v\|_2^\beta
\lesssim_\beta
\left(
\frac{k^\star(1-\overline{\alpha}_t)}{\overline{\alpha}_t}
\log\frac{dnR_0}{1-\overline{\alpha}_t}
\right)^{\beta/2},
\]
while on the complement we use \(\|y-v\|_2^\beta\le R_0^\beta\). Therefore,
\begin{align}
\gJ_{2,\beta}
&\lesssim_\beta
\left(
\frac{k^\star(1-\overline{\alpha}_t)}{\overline{\alpha}_t}
\log\frac{dnR_0}{1-\overline{\alpha}_t}
\right)^{\beta/2}
\int_{y\in\bigcup_{i\in\gI}\gB_i}
\int_{\omega\in\gG}
\int_{x_0}
\mathsf{T}(x_0,y,\omega)
\rd x_0\rd\omega\rd y
\notag\\
&\quad
+R_0^\beta
\int_{y\in\bigcup_{i\in\gI}\gB_i}
\int_{\omega\in\gG}
\int_{x_0}
\int_{v\notin \gE_{\overline{\alpha}_t}(y)}
\mathsf{S}(x_0,v,y,\omega)
\rd v\rd x_0\rd\omega\rd y .
\label{eq:J2-lemma23-form}
\end{align}
The first integral is bounded by \eqref{eq:J1-pure-density-ratio}, while the
last integral is controlled by the same cellwise tail estimate as in
Lemma~\ref{lem:typical-region-cubic-upper-bound}; again the final
super-polynomial factor is smaller than \((1-\overline{\alpha}_t)^\beta/n\)
after changing only the \(\beta\)-dependent constant. Therefore,
\[
\gJ_{2,\beta}
\lesssim_\beta
\frac{(k^\star)^{\beta/2}R_0^{k^\star}
\left(\log\frac{dnR_0}{1-\overline{\alpha}_t}\right)^{\beta/2}}
{\overline{\alpha}_t^{\beta/2}
(1-\overline{\alpha}_t)^{k^\star/2-\beta/2}}
\;+\;
\frac{R_0^\beta(1-\overline{\alpha}_t)^\beta}{n}.
\]
This proves the lemma.

\subsubsection{Proof of Lemma~\ref{lem:intrinsic-density-matching-bound}}
\label{prf:lem:intrinsic-density-matching-bound}
Under the change of variables
\(x=\sqrt{\overline{\alpha}_t}y+\sqrt{1-\overline{\alpha}_t}\omega\), the
measure \(\varphi_1(\omega)p_{X_0}(y)\rd\omega\rd y\) is mapped to
\(p_{X_t}(x)\rd x\). Hence
\begin{align}
&\int_{y,\omega}
\PP\!\left(\widehat p_{X_t}(x)<\frac12p_{X_t}(x)\right)
\varphi_1(\omega)p_{X_0}(y)\rd\omega\rd y
\notag\\
&\qquad=
\int
\PP\!\left(\widehat p_{X_t}(x)<\frac12p_{X_t}(x)\right)
p_{X_t}(x)\rd x
\le
4\int
\frac{\mathbb E\bigl[(\widehat p_{X_t}(x)-p_{X_t}(x))^2\bigr]}
{p_{X_t}(x)}\rd x .
\label{eq:lower-tail-to-density-variance}
\end{align}
The last inequality follows from
\[
\left\{\widehat p_{X_t}(x)<\frac12p_{X_t}(x)\right\}
\subseteq
\left\{|\widehat p_{X_t}(x)-p_{X_t}(x)|
\ge \frac12p_{X_t}(x)\right\}
\]
and Markov's inequality.

Let $\{B_j\}_{j=1}^N$ be a cover of $\gX_{\mathsf{data}}$ by Euclidean balls of
radius \(\sqrt{1-\overline{\alpha}_t}\), where, by Assumption~\ref{ass:low_dim}
and the convention \(k^\star=C_{\mathsf{cover}}k\),
\[
N_{\sqrt{1-\overline{\alpha}_t}}\big(\gX_{\mathsf{data}}\big)
\le
\left(\frac{R_0}{\sqrt{1-\overline{\alpha}_t}}\right)^{k^\star}.
\]
Without loss of generality, we take the sets $B_j$ to be disjoint after replacing them by a
measurable partition subordinate to the cover. Write $m_j=\PB_{X_0}(B_j)$ and ignore indices with
$m_j=0$. Since
\[
p_{X_t}(x)=\int_{\RB^d}
\varphi_{1-\overline{\alpha}_t}\bigl(x-\sqrt{\overline{\alpha}_t}y\bigr)
p_{X_0}(y)\rd y,
\qquad
\hat p_{X_t}(x)=\frac1n\sum_{i=1}^n
\varphi_{1-\overline{\alpha}_t}
\bigl(x-\sqrt{\overline{\alpha}_t}X_i^{\mathsf{sample}}\bigr),
\]
we have
\[
\begin{aligned}
\mathbb E\bigl[(\hat p_{X_t}(x)-p_{X_t}(x))^2\bigr]
&=
\frac1n
\operatorname{Var}\bigl(
\varphi_{1-\overline{\alpha}_t}
\bigl(x-\sqrt{\overline{\alpha}_t}X_1^{\mathsf{sample}}\bigr)
\bigr)  \\
&\le
\frac1n
\int_{\RB^d}
\varphi_{1-\overline{\alpha}_t}^2\bigl(x-\sqrt{\overline{\alpha}_t}y\bigr)
p_{X_0}(y)\rd y.
\end{aligned}
\]
Therefore,
\[
\begin{aligned}
\int_{\RB^d}
\frac{\mathbb E\bigl[(\hat p_{X_t}(x)-p_{X_t}(x))^2\bigr]}{p_{X_t}(x)}\,\rd x
&\le
\frac1n
\int_{\RB^d}
\frac{\int
\varphi_{1-\overline{\alpha}_t}^2\bigl(x-\sqrt{\overline{\alpha}_t}y\bigr)
p_{X_0}(y)\rd y}
{\int
\varphi_{1-\overline{\alpha}_t}\bigl(x-\sqrt{\overline{\alpha}_t}y\bigr)
p_{X_0}(y)\rd y}
\,\rd x                                                        \\
&=
\frac1n
\int_{\RB^d}
\frac{\sum_{j=1}^{N_{\sqrt{1-\overline{\alpha}_t}}(\gX_{\mathsf{data}})} m_j\int_{B_j}\varphi_{1-\overline{\alpha}_t}^2\bigl(x-\sqrt{\overline{\alpha}_t}y\bigr)\,p_{X_0}(y\mid B_j)\rd y}
{\sum_{j=1}^{N_{\sqrt{1-\overline{\alpha}_t}}(\gX_{\mathsf{data}})} m_j\int_{B_j}\varphi_{1-\overline{\alpha}_t}\bigl(x-\sqrt{\overline{\alpha}_t}y\bigr)\,p_{X_0}(y\mid B_j)\rd y}
\,\rd x                                                        \\
&= \frac{1}{n}\sum_{j=1}^{N_{\sqrt{1-\overline{\alpha}_t}}(\gX_{\mathsf{data}})} \int_{\RB^d}\frac{m_j\int_{B_j}\varphi_{1-\overline{\alpha}_t}^2\bigl(x-\sqrt{\overline{\alpha}_t}y\bigr)p_{X_0}(y\mid B_j)\rd y}{\sum_{i=1}^{N_{\sqrt{1-\overline{\alpha}_t}}(\gX_{\mathsf{data}})} m_i \int_{B_i}\varphi_{1-\overline{\alpha}_t}\bigl(x-\sqrt{\overline{\alpha}_t}u\bigr)p_{X_0}(u\mid B_i)\rd u} \rd x
\\
&\le
\frac1n
\sum_{j=1}^{N_{\sqrt{1-\overline{\alpha}_t}}(\gX_{\mathsf{data}})}
\int_{\RB^d}
\frac{\int_{B_j}\varphi_{1-\overline{\alpha}_t}^2\bigl(x-\sqrt{\overline{\alpha}_t}y\bigr)\,p_{X_0}(y\mid B_j)\rd y}
{\int_{B_j}\varphi_{1-\overline{\alpha}_t}\bigl(x-\sqrt{\overline{\alpha}_t}u\bigr)\,p_{X_0}(u\mid B_j)\rd u}
\,\rd x .
\end{aligned}
\]
Here, the last inequality holds by using
\[
\sum_{i=1}^{N_{\sqrt{1-\overline{\alpha}_t}}(\gX_{\mathsf{data}})}
m_i \int_{B_i}\varphi_{1-\overline{\alpha}_t}\bigl(x-\sqrt{\overline{\alpha}_t}u\bigr)
p_{X_0}(u\mid B_i)\rd u
\ge
m_j \int_{B_j}\varphi_{1-\overline{\alpha}_t}\bigl(x-\sqrt{\overline{\alpha}_t}u\bigr)
p_{X_0}(u\mid B_j) \rd u
\]
for the denominator of every term in the summation.
It remains to control each term in the last sum. For a fixed $j$, by Jensen's inequality
applied to the convex map $z\mapsto 1/z$,
\[
\begin{aligned}
&\int_{\RB^d}
\frac{\int_{B_j}\varphi_{1-\overline{\alpha}_t}^2\bigl(x-\sqrt{\overline{\alpha}_t}y\bigr)\,p_{X_0}(y\mid B_j)\rd y}
{\int_{B_j}\varphi_{1-\overline{\alpha}_t}\bigl(x-\sqrt{\overline{\alpha}_t}u\bigr)\,p_{X_0}(u\mid B_j)\rd u}
\,\rd x\notag\\
&\qquad\le
\int_{\RB^d}
\int_{B_j}\int_{B_j}
\frac{\varphi_{1-\overline{\alpha}_t}^2\bigl(x-\sqrt{\overline{\alpha}_t}y\bigr)}
{\varphi_{1-\overline{\alpha}_t}\bigl(x-\sqrt{\overline{\alpha}_t}u\bigr)}
\,p_{X_0}(u\mid B_j)p_{X_0}(y\mid B_j)\rd u \rd y \rd x                         \\
&\qquad=
\int_{B_j}\int_{B_j}
\left\{
\int_{\RB^d}
\frac{\varphi_{1-\overline{\alpha}_t}^2\bigl(x-\sqrt{\overline{\alpha}_t}y\bigr)}
{\varphi_{1-\overline{\alpha}_t}\bigl(x-\sqrt{\overline{\alpha}_t}u\bigr)}
\,\rd x
\right\}
p_{X_0}(u\mid B_j)p_{X_0}(y\mid B_j)\rd u\rd y.
\end{aligned}
\]
The inner Gaussian integral can be computed explicitly. Indeed,
\[
\begin{aligned}
\int
\frac{\varphi_{1-\overline{\alpha}_t}^2\bigl(x-\sqrt{\overline{\alpha}_t}y\bigr)}
{\varphi_{1-\overline{\alpha}_t}\bigl(x-\sqrt{\overline{\alpha}_t}u\bigr)}
\,\rd x
&=
\int
(2\pi(1-\overline{\alpha}_t))^{-d/2}
\exp\left\{
-\frac{\|x-\sqrt{\overline{\alpha}_t}y\|_2^2}{1-\overline{\alpha}_t}
+\frac{\|x-\sqrt{\overline{\alpha}_t}u\|_2^2}{2(1-\overline{\alpha}_t)}
\right\}dx                                                   \\
&=
\int
(2\pi(1-\overline{\alpha}_t))^{-d/2}
\exp\left\{
-\frac{\|x-\sqrt{\overline{\alpha}_t}(2y-u)\|_2^2}{2(1-\overline{\alpha}_t)}
+\frac{\overline{\alpha}_t\|y-u\|_2^2}{1-\overline{\alpha}_t}
\right\}dx                                                   \\
&=
\exp\left\{\frac{\overline{\alpha}_t\|y-u\|_2^2}{1-\overline{\alpha}_t}\right\}.
\end{aligned}
\]
Since $y,u\in B_j$ and each $B_j$ has radius \(\sqrt{1-\overline{\alpha}_t}\), we have
\(\|y-u\|_2\le 2\sqrt{1-\overline{\alpha}_t}\), and hence
\[
\begin{aligned}
\int
\frac{\int_{B_j}\varphi_{1-\overline{\alpha}_t}^2\bigl(x-\sqrt{\overline{\alpha}_t}y\bigr)\,p_{X_0}(y\mid B_j)\rd y}
{\int_{B_j}\varphi_{1-\overline{\alpha}_t}\bigl(x-\sqrt{\overline{\alpha}_t}u\bigr)\,p_{X_0}(u\mid B_j)\rd u}
\,\rd x
&\le
\int_{B_j}\int_{B_j}
\exp\left\{\frac{\overline{\alpha}_t\|y-u\|_2^2}{1-\overline{\alpha}_t}\right\}
p_{X_0}(u\mid B_j)p_{X_0}(y\mid B_j)\rd u\rd y                                \\
&\le e^4 .
\end{aligned}
\]
Combining the above estimates gives
\[
\int
\frac{\mathbb E\bigl[(\hat p_{X_t}(x)-p_{X_t}(x))^2\bigr]}{p_{X_t}(x)}\,\rd x
\le
\frac{e^4 N_{\sqrt{1-\overline{\alpha}_t}}(\gX_{\mathsf{data}})}{n}
\le
\frac{e^4R_0^{k^\star}}{n\,(1-\overline{\alpha}_t)^{k^\star/2}},
\]
which, together with \eqref{eq:lower-tail-to-density-variance}, proves the
lemma.

\subsubsection{Proof of Lemma~\ref{lem:bad-B-tail-ratio}}
\label{prf:lem:bad-B-tail-ratio}

For \(y\in\gX_{\mathsf{data}}\), let \(i(y)\) be such that
\(y\in\gC_{i(y)}\), and define
\[
\gH_y\coloneqq
\left\{\exists r\in[n]\colon X_r^{\mathsf{sample}}\in\gC_{i(y)}\right\}.
\]
Also define
\[
\gR_\omega\coloneqq
\left\{
\exists r,s\in[n]\colon
\bigl\langle X_r^{\mathsf{sample}}-X_s^{\mathsf{sample}},\omega\bigr\rangle
>
4\sqrt{\log\frac{dnR_0}{1-\overline{\alpha}_t}}\,
\|X_r^{\mathsf{sample}}-X_s^{\mathsf{sample}}\|_2
\right\}.
\]
We first split the integral according to \(\gH_y^c\),
\(\gH_y\cap\gR_\omega\), and \(\gH_y\cap\gR_\omega^c\):
\begin{align}
&\int_{y,\omega}
\EE\!\left[
\mathsf{R}_t(y,\omega)^2
\right]\varphi_1(\omega)p_{X_0}(y)\rd\omega\rd y
=
\int_{y,\omega}
\EE\!\left[
\mathsf{R}_t(y,\omega)^2
\mathbbm{1}_{\gH_y^c}
\right]\varphi_1(\omega)p_{X_0}(y)\rd\omega\rd y
\notag\\
&\hspace{10em}+
\int_{y,\omega}
\EE\!\left[
\mathsf{R}_t(y,\omega)^2
\mathbbm{1}_{\gH_y}
\mathbbm{1}_{\gR_\omega}
\right]\varphi_1(\omega)p_{X_0}(y)\rd\omega\rd y\notag\\
&\hspace{11em}+
\int_{y,\omega}
\EE\!\left[
\mathsf{R}_t(y,\omega)^2
\mathbbm{1}_{\gH_y}
\mathbbm{1}_{\gR_\omega^c}
\right]\varphi_1(\omega)p_{X_0}(y)\rd\omega\rd y
\notag\\
&\quad \le
\int \PP(\gH_y^c)p_{X_0}(y)\rd y
+
\EE\!\left[
\PP_\omega(\gR_\omega\mid X_1^{\mathsf{sample}},\ldots,X_n^{\mathsf{sample}})
\right]\notag\\
&\hspace{10em}+
\int_{y,\omega}
\EE\!\left[
\mathsf{R}_t(y,\omega)^2
\mathbbm{1}_{\gH_y}\mathbbm{1}_{\gR_\omega^c}
\right]\varphi_1(\omega)p_{X_0}(y)\rd\omega\rd y .
\label{eq:bad-B-tail-ratio-split}
\end{align}
In the last inequality, for the first term we dropped
\(\mathsf{R}_t^2\) and the Gaussian integral; for the second term we dropped
\(\mathsf{R}_t^2\) and \(\mathbbm{1}_{\gH_y}\), and then conditioned on the
training data. These steps use \(0\le \mathsf{R}_t(y,\omega)\le1\), which follows because
\(\mathsf{R}_t\) is a sub-ratio of nonnegative kernel sums with denominator
\(D(x)\).

We first bound the first term in \eqref{eq:bad-B-tail-ratio-split}. For
\(y\in\gC_i\), the event \(\gH_y^c\) is exactly the event that none of the
\(n\) training samples falls in \(\gC_i\). Hence
\begin{align}
\int \PP(\gH_y^c)p_{X_0}(y)\rd y
&=
\sum_{i=1}^{N_{\lambda_t}}
\int_{y\in\gC_i}
\bigl(1-\PB_{X_0}(\gC_i)\bigr)^n p_{X_0}(y)\rd y
\notag\\
&=
\sum_{i=1}^{N_{\lambda_t}}
\PB_{X_0}(\gC_i)\bigl(1-\PB_{X_0}(\gC_i)\bigr)^n
\le
\frac{N_{\lambda_t}}{en}.
\label{eq:bad-B-tail-empty-cell}
\end{align}
The last inequality uses \(u(1-u)^n\le (en)^{-1}\) for \(u\in[0,1]\).

Next, conditional on the training data, \(\omega\sim\gN(0,I_d)\) is independent
of \(X_1^{\mathsf{sample}},\ldots,X_n^{\mathsf{sample}}\). A Gaussian tail bound
and a union bound over \(n^2\) pairs yield
\begin{align}
\PP_\omega(\gR_\omega\mid X_1^{\mathsf{sample}},\ldots,X_n^{\mathsf{sample}})
&\le
n^2
\exp\left\{
-8\log\frac{dnR_0}{1-\overline{\alpha}_t}
\right\}
\notag\\
&=
n^2\left(\frac{1-\overline{\alpha}_t}{R_0n}\right)^8
\le
\frac{(1-\overline{\alpha}_t)^6}{n^2}.
\label{eq:bad-B-tail-gaussian-rare}
\end{align}
The last step uses the parameter regime \(R_0n/(1-\overline{\alpha}_t)\ge1\).

It remains to control the last term in \eqref{eq:bad-B-tail-ratio-split}.
On \(\gH_y\), choose \(r_y\) such that
\(X_{r_y}^{\mathsf{sample}}\in\gC_{i(y)}\), and therefore
\(\|X_{r_y}^{\mathsf{sample}}-y\|_2\le \lambda_t\). Consider a summand under the
indicator \(\|X_r^{\mathsf{sample}}-y\|_2>\rho_t\). The Gaussian ratio equals
\begin{align}
&\frac{
\varphi_{1-\overline{\alpha}_t}
\bigl(x-\sqrt{\overline{\alpha}_t}X_r^{\mathsf{sample}}\bigr)}
{\varphi_{1-\overline{\alpha}_t}
\bigl(x-\sqrt{\overline{\alpha}_t}X_{r_y}^{\mathsf{sample}}\bigr)}
\notag\\
&\quad =
\exp\left\{
-\frac{
\|x-\sqrt{\overline{\alpha}_t}X_r^{\mathsf{sample}}\|_2^2
-\|x-\sqrt{\overline{\alpha}_t}X_{r_y}^{\mathsf{sample}}\|_2^2}
{2(1-\overline{\alpha}_t)}
\right\}
\notag\\
&\quad =
\exp\left\{
\frac{\gamma_t^2}{2}
\left(
\|X_{r_y}^{\mathsf{sample}}-y\|_2^2
-\|X_r^{\mathsf{sample}}-y\|_2^2
\right)
+
\gamma_t
\bigl\langle
X_r^{\mathsf{sample}}-X_{r_y}^{\mathsf{sample}},
\omega
\bigr\rangle
\right\}.
\label{eq:bad-B-lemma-gaussian-ratio}
\end{align}
On \(\gR_\omega^c\),
\[
\bigl\langle X_r^{\mathsf{sample}}-X_{r_y}^{\mathsf{sample}},\omega\bigr\rangle
\le
4\sqrt{\log\frac{dnR_0}{1-\overline{\alpha}_t}}\,
\|X_r^{\mathsf{sample}}-X_{r_y}^{\mathsf{sample}}\|_2 .
\]
Moreover,
\[
\|X_r^{\mathsf{sample}}-X_{r_y}^{\mathsf{sample}}\|_2
\le
\|X_r^{\mathsf{sample}}-y\|_2+\lambda_t .
\]
Substituting these estimates into \eqref{eq:bad-B-lemma-gaussian-ratio} gives
\begin{align}
&\frac{
\varphi_{1-\overline{\alpha}_t}
\bigl(x-\sqrt{\overline{\alpha}_t}X_r^{\mathsf{sample}}\bigr)}
{\varphi_{1-\overline{\alpha}_t}
\bigl(x-\sqrt{\overline{\alpha}_t}X_{r_y}^{\mathsf{sample}}\bigr)}
\notag\\
&\quad\le
\exp\left\{
\frac{\gamma_t^2}{2}
\left(\lambda_t^2-\|X_r^{\mathsf{sample}}-y\|_2^2\right)
+4\gamma_t\sqrt{\log\frac{dnR_0}{1-\overline{\alpha}_t}}\,
\left(\|X_r^{\mathsf{sample}}-y\|_2+\lambda_t\right)
\right\}.
\label{eq:bad-B-lemma-ratio-before-tail}
\end{align}
The exponent in \eqref{eq:bad-B-lemma-ratio-before-tail}, as a function of
\(\|X_r^{\mathsf{sample}}-y\|_2\), is decreasing on the region
\(\|X_r^{\mathsf{sample}}-y\|_2>\rho_t\) because
\[
\rho_t\ge
32\gamma_t^{-1}\sqrt{\log\frac{dnR_0}{1-\overline{\alpha}_t}}.
\]
Using \(\lambda_t\le\rho_t/2\), we obtain
\begin{align}
\frac{
\varphi_{1-\overline{\alpha}_t}
\bigl(x-\sqrt{\overline{\alpha}_t}X_r^{\mathsf{sample}}\bigr)}
{\varphi_{1-\overline{\alpha}_t}
\bigl(x-\sqrt{\overline{\alpha}_t}X_{r_y}^{\mathsf{sample}}\bigr)}
&\le
\exp\left\{
\frac{\gamma_t^2}{2}
\left(\lambda_t^2-\rho_t^2\right)
+4\gamma_t\sqrt{\log\frac{dnR_0}{1-\overline{\alpha}_t}}\,
\left(\rho_t+\lambda_t\right)
\right\}
\notag\\
&\quad\le
\exp\left\{
\frac{\gamma_t^2}{2}
\left(\frac{\rho_t^2}{4}-\rho_t^2\right)
+6\gamma_t\rho_t
\sqrt{\log\frac{dnR_0}{1-\overline{\alpha}_t}}
\right\}
\notag\\
&\quad\le
\exp\left\{
-\frac{3\gamma_t^2\rho_t^2}{8}
+\frac{3\gamma_t^2\rho_t^2}{16}
\right\}
\le
\exp\left\{
-\frac{\gamma_t^2\rho_t^2}{8}
\right\}.
\label{eq:bad-B-lemma-ratio-tail-bound}
\end{align}
On \(\gH_y\), the denominator \(D(x)\) is lower bounded by the
\(r_y\)-summand. Hence \eqref{eq:bad-B-lemma-ratio-tail-bound} gives
\begin{align}
&\int_{y,\omega}
\EE\!\left[
\mathsf{R}_t(y,\omega)^2
\mathbbm{1}_{\gH_y}\mathbbm{1}_{\gR_\omega^c}
\right]
\varphi_1(\omega)p_{X_0}(y)\rd\omega\rd y
\notag\\
&\qquad\le
n^2\exp\left\{-\frac{\gamma_t^2\rho_t^2}{4}\right\}
\le
\frac{(1-\overline{\alpha}_t)^6}{n^2},
\end{align}
where the last inequality follows from
\(\gamma_t^2\rho_t^2\ge 4\log\!\bigl(n^4/(1-\overline{\alpha}_t)^6\bigr)\).
Combining this bound with \eqref{eq:bad-B-tail-ratio-split},
\eqref{eq:bad-B-tail-empty-cell}, and \eqref{eq:bad-B-tail-gaussian-rare}
proves the lemma.

\subsubsection{Proof of Lemma~\ref{lem:posterior-mean-high-moment}}
\label{prf:lem:posterior-mean-high-moment}

Fix \(1\le \beta\le 3\). Throughout the proof, write
\(
x=\sqrt{\overline{\alpha}_t}y+\sqrt{1-\overline{\alpha}_t}\omega,\,
y\in\gX_{\mathsf{data}},\, \omega\in\RB^d,
\)
By the definition of \(\widehat\mu_{0\mid t}\), we can write
\begin{align}
\widehat\mu_{0\mid t}(x)-\mu_{0\mid t}(x)
&=
\sum_{r=1}^n
X_r^{\mathsf{sample}}
\frac{
\varphi_{1-\overline{\alpha}_t}
\bigl(\sqrt{\overline{\alpha}_t}X_r^{\mathsf{sample}}-x\bigr)}
{D(x)}
-\mu_{0\mid t}(x)
\notag\\
&=
\sum_{r=1}^n
\left(X_r^{\mathsf{sample}}-\mu_{0\mid t}(x)\right)
\frac{
\varphi_{1-\overline{\alpha}_t}
\bigl(\sqrt{\overline{\alpha}_t}X_r^{\mathsf{sample}}-x\bigr)}
{D(x)} .
\label{eq:posterior-mean-expansion}
\end{align}
For a truncation radius \(\rho_t>0\) to be chosen later, decompose the right
hand side into a local part and a tail part:
\begin{subequations}\label{eq:posterior-mean-AB-decomp}
\begin{align}
\gA_t(y,\omega)
&\coloneqq
\sum_{r=1}^n
\left(X_r^{\mathsf{sample}}-\mu_{0\mid t}(x)\right)
\frac{
\varphi_{1-\overline{\alpha}_t}
\bigl(\sqrt{\overline{\alpha}_t}X_r^{\mathsf{sample}}-x\bigr)}
{D(x)}
\mathbbm{1}\{\|X_r^{\mathsf{sample}}-y\|_2\le \rho_t\},
\\
\gB_t(y,\omega)
&\coloneqq
\sum_{r=1}^n
\left(X_r^{\mathsf{sample}}-\mu_{0\mid t}(x)\right)
\frac{
\varphi_{1-\overline{\alpha}_t}
\bigl(\sqrt{\overline{\alpha}_t}X_r^{\mathsf{sample}}-x\bigr)}
{D(x)}
\mathbbm{1}\{\|X_r^{\mathsf{sample}}-y\|_2> \rho_t\}.
\end{align}
\end{subequations}
Then \(\widehat\mu_{0\mid t}(x)-\mu_{0\mid t}(x)=\gA_t(y,\omega)+\gB_t(y,\omega)\). On the support of
\(\gA_t(y,\omega)\), we have
\[
\|X_r^{\mathsf{sample}}-\mu_{0\mid t}(x)\|_2
\le
\|X_r^{\mathsf{sample}}-y\|_2+\|y-\mu_{0\mid t}(x)\|_2 .
\]
Therefore, after separating the event
\(\{\|y-\mu_{0\mid t}(x)\|_2\le \rho_t\}\) from its complement, we obtain the
pointwise bound
\begin{align}
\|\gA_t(y,\omega)\|_2
&\lesssim
\rho_t+
R_0\mathbbm{1}\{\|y-\mu_{0\mid t}(x)\|_2>\rho_t\}.
\label{eq:A-pointwise-posterior-mean}
\end{align}
For the tail part, we use the following term, which has been defined in Lemma~\ref{lem:bad-B-tail-ratio},
\[
\mathsf R_t(y,\omega)
\coloneqq
\frac{
\sum_{r=1}^n
\mathbbm{1}\{\|X_r^{\mathsf{sample}}-y\|_2>\rho_t\}
\varphi_{1-\overline{\alpha}_t}
\bigl(\sqrt{\overline{\alpha}_t}X_r^{\mathsf{sample}}-x\bigr)}
{D(x)}.
\]
Since \(\|X_r^{\mathsf{sample}}-\mu_{0\mid t}(x)\|_2\le R_0\), it follows that
\begin{align}
\|\gB_t(y,\omega)\|_2
&\le R_0\,\mathsf R_t(y,\omega).
\label{eq:B-pointwise-posterior-mean}
\end{align}
Using \((a+b)^{2\beta}\lesssim a^{2\beta}+b^{2\beta}\), and then applying
\eqref{eq:A-pointwise-posterior-mean} and
\eqref{eq:B-pointwise-posterior-mean}, we have
\begin{align}
\|\widehat\mu_{0\mid t}(x)-\mu_{0\mid t}(x)\|_2^{2\beta}
&\lesssim
\|\gA_t(y,\omega)\|_2^{2(\beta-1)}\|\widehat\mu_{0\mid t}(x)-\mu_{0\mid t}(x)\|_2^2
+
\|\gB_t(y,\omega)\|_2^{2(\beta-1)}\|\widehat\mu_{0\mid t}(x)-\mu_{0\mid t}(x)\|_2^2
\notag\\
&\lesssim
\rho_t^{2(\beta-1)}\|\widehat\mu_{0\mid t}(x)-\mu_{0\mid t}(x)\|_2^2
+
R_0^{2(\beta-1)}
\mathbbm{1}\{\|y-\mu_{0\mid t}(x)\|_2>\rho_t\}\|\widehat\mu_{0\mid t}(x)-\mu_{0\mid t}(x)\|_2^2
\notag\\
&\qquad+
R_0^{2\beta}\mathsf R_t(y,\omega)^2 .
\label{eq:posterior-mean-power-first}
\end{align}
The middle term is controlled by the posterior concentration event. Indeed,
since \(\|\widehat\mu_{0\mid t}(x)-\mu_{0\mid t}(x)\|_2\le R_0\),
\[
R_0^{2(\beta-1)}
\mathbbm{1}\{\|y-\mu_{0\mid t}(x)\|_2>\rho_t\}\|\widehat\mu_{0\mid t}(x)-\mu_{0\mid t}(x)\|_2^2
\le
R_0^{2\beta}\mathbbm{1}\{\|y-\mu_{0\mid t}(x)\|_2>\rho_t\}.
\]
Next split the first term in \eqref{eq:posterior-mean-power-first} according
to the event
\(
\gE_x=\left\{\widehat p_{X_t}(x)\ge\frac12p_{X_t}(x)\right\}
\).
This gives the pointwise inequality
\begin{align}
\|\widehat\mu_{0\mid t}(x)-\mu_{0\mid t}(x)\|_2^{2\beta}
&\lesssim
\rho_t^{2(\beta-1)}\|\widehat\mu_{0\mid t}(x)-\mu_{0\mid t}(x)\|_2^2\mathbbm{1}_{\gE_x}
+
\rho_t^{2(\beta-1)}\|\widehat\mu_{0\mid t}(x)-\mu_{0\mid t}(x)\|_2^2\mathbbm{1}_{\gE_x^c}
\notag\\
&\qquad+
R_0^{2\beta}\mathbbm{1}\{\|y-\mu_{0\mid t}(x)\|_2>\rho_t\}
+
R_0^{2\beta}\mathsf R_t(y,\omega)^2
\notag\\
&\lesssim
\rho_t^{2(\beta-1)}\|\widehat\mu_{0\mid t}(x)-\mu_{0\mid t}(x)\|_2^2\mathbbm{1}_{\gE_x}
+
\rho_t^{2\beta}\mathbbm{1}_{\gE_x^c}
\notag\\
&\qquad+
R_0^{2\beta}\mathbbm{1}\{\|y-\mu_{0\mid t}(x)\|_2>\rho_t\}
+
R_0^{2\beta}\mathsf R_t(y,\omega)^2 .
\label{eq:posterior-mean-power-good-bad}
\end{align}
Taking expectation over \(X_t\) and \(X^{\mathsf{sample}}\) yields
\begin{align}
\EE_{X_t,X^{\mathsf{sample}}}
\left[\|\widehat\mu_{0\mid t}(X_t)-\mu_{0\mid t}(X_t)\|_2^{2\beta}\right]
&\lesssim
\rho_t^{2(\beta-1)}
\EE_{X_t,X^{\mathsf{sample}}}
\left[\|\widehat\mu_{0\mid t}(X_t)-\mu_{0\mid t}(X_t)\|_2^2\mathbbm{1}_{\gE_{X_t}}\right]
\notag\\
&\qquad+
\rho_t^{2\beta}
\int_{y,\omega}
\PP\!\left(\widehat p_{X_t}(x)<\frac12p_{X_t}(x)\right)
\varphi_1(\omega)p_{X_0}(y)\rd\omega\rd y
\notag\\
&\qquad+
R_0^{2\beta}
\PB\!\left(\|Y-\mu_{0\mid t}(X_t)\|_2>\rho_t\right)
\notag\\
&\qquad+
R_0^{2\beta}
\int_{y,\omega}
\EE\!\left[\mathsf R_t(y,\omega)^2\right]
\varphi_1(\omega)p_{X_0}(y)\rd\omega\rd y .
\label{eq:posterior-mean-master-split}
\end{align}
By Lemma~\ref{lem:intrinsic-density-matching-bound},
\begin{align}
&\int_{y,\omega}
\PP\!\left(\widehat p_{X_t}(x)<\frac12p_{X_t}(x)\right)
\varphi_1(\omega)p_{X_0}(y)\rd\omega\rd y
\lesssim
\frac{R_0^{k^\star}}
{n(1-\overline{\alpha}_t)^{k^\star/2}}.
\label{eq:posterior-mean-density-bad}
\end{align}
By Lemma~\ref{lem:bad-B-tail-ratio}, taking \(\lambda_t=\rho_t/2\) gives
\begin{align}
&\int_{y,\omega}
\EE\!\left[\mathsf R_t(y,\omega)^2\right]
\varphi_1(\omega)p_{X_0}(y)\rd\omega\rd y
\lesssim
\frac{N_{\rho_t/2}}{n}
+
\frac{\overline{\alpha}_t^6(1-\overline{\alpha}_t)^6}{n^2}
\lesssim
\frac{R_0^{k^\star}}{n\rho_t^{k^\star}}
+
\frac{(1-\overline{\alpha}_t)^6}{n^2}.
\label{eq:posterior-mean-ratio-tail}
\end{align}
Moreover, Lemma~\ref{lem:posterior_norm}, together with the choice of
\(\rho_t\) below, implies
\begin{align}
R_0^{2\beta}
\PB\!\left(\|Y-\mu_{0\mid t}(X_t)\|_2>\rho_t\right)
\lesssim
\frac{R_0^{2\beta}(1-\overline{\alpha}_t)^{2\beta}}{n^2}.
\label{eq:posterior-mean-pop-tail}
\end{align}

It remains to control the first term on the right-hand side of
\eqref{eq:posterior-mean-master-split}. We first split according to the
typical region:
\begin{align}
&\EE_{X_t,X^{\mathsf{sample}}}
\left[\|\widehat\mu_{0\mid t}(X_t)-\mu_{0\mid t}(X_t)\|_2^2\mathbbm{1}_{\gE_{X_t}}\right]
\notag\\
&\quad\le
\int_{\gT_{\overline{\alpha}_t}}
\EE_{X^{\mathsf{sample}}}
\left[\|\widehat\mu_{0\mid t}(x)-\mu_{0\mid t}(x)\|_2^2\mathbbm{1}_{\gE_x}\right]
p_{X_t}(x)\rd x
+
R_0^2\PB(X_t\notin\gT_{\overline{\alpha}_t}).
\label{eq:posterior-mean-typical-split}
\end{align}
For \(x\in\gT_{\overline{\alpha}_t}\), on \(\gE_x\) the denominator satisfies
\(D(x)=n\widehat p_{X_t}(x)\ge np_{X_t}(x)/2\). Therefore
\begin{align}
&\int_{\gT_{\overline{\alpha}_t}}
\EE_{X^{\mathsf{sample}}}
\left[\|\widehat\mu_{0\mid t}(x)-\mu_{0\mid t}(x)\|_2^2\mathbbm{1}_{\gE_x}\right]
p_{X_t}(x)\rd x
\notag\\
&\quad\lesssim
\int_{\gT_{\overline{\alpha}_t}}
\frac{
\EE_{X^{\mathsf{sample}}}
\left[
\left\|
\sum_{r=1}^n
\left(X_r^{\mathsf{sample}}-\mu_{0\mid t}(x)\right)
\varphi_{1-\overline{\alpha}_t}
\bigl(\sqrt{\overline{\alpha}_t}X_r^{\mathsf{sample}}-x\bigr)
\right\|_2^2
\right]}
{n^2p_{X_t}(x)}
\rd x .
\label{eq:posterior-mean-good-denominator}
\end{align}
For every fixed \(x\), the numerator summands are centered:
\begin{align}
&\EE_{X_r^{\mathsf{sample}}}\!\left[
\left(X_r^{\mathsf{sample}}-\mu_{0\mid t}(x)\right)
\varphi_{1-\overline{\alpha}_t}
\bigl(\sqrt{\overline{\alpha}_t}X_r^{\mathsf{sample}}-x\bigr)
\right]
\notag\\
&\quad=
\int z\,\varphi_{1-\overline{\alpha}_t}
\bigl(\sqrt{\overline{\alpha}_t}z-x\bigr)p_0(z)\rd z
-
\mu_{0\mid t}(x)
\int
\varphi_{1-\overline{\alpha}_t}
\bigl(\sqrt{\overline{\alpha}_t}z-x\bigr)p_0(z)\rd z
\notag\\
&\quad=
\mu_{0\mid t}(x)p_{X_t}(x)-\mu_{0\mid t}(x)p_{X_t}(x)
=0.
\label{eq:posterior-mean-centered-summand}
\end{align}
Hence all cross terms vanish, and independence yields
\begin{align}
&\EE_{X^{\mathsf{sample}}}
\left[
\left\|
\sum_{r=1}^n
\left(X_r^{\mathsf{sample}}-\mu_{0\mid t}(x)\right)
\varphi_{1-\overline{\alpha}_t}
\bigl(\sqrt{\overline{\alpha}_t}X_r^{\mathsf{sample}}-x\bigr)
\right\|_2^2
\right]
\notag\\
&\quad=
n\EE_{X_0}\!\left[
\left\|X_0-\mu_{0\mid t}(x)\right\|_2^2
\varphi_{1-\overline{\alpha}_t}^2
\bigl(\sqrt{\overline{\alpha}_t}X_0-x\bigr)
\right].
\label{eq:posterior-mean-cross-vanish}
\end{align}
Combining \eqref{eq:posterior-mean-good-denominator} and
\eqref{eq:posterior-mean-cross-vanish}, we get
\begin{align}
&\int_{\gT_{\overline{\alpha}_t}}
\EE_{X^{\mathsf{sample}}}
\left[\|\widehat\mu_{0\mid t}(x)-\mu_{0\mid t}(x)\|_2^2\mathbbm{1}_{\gE_x}\right]
p_{X_t}(x)\rd x
\notag\\
&\quad\lesssim
\frac1n
\int_{\gT_{\overline{\alpha}_t}}
\frac{
\EE_{X_0}\!\left[
\left\|X_0-\mu_{0\mid t}(x)\right\|_2^2
\varphi_{1-\overline{\alpha}_t}^2
\bigl(\sqrt{\overline{\alpha}_t}X_0-x\bigr)
\right]}
{p_{X_t}(x)}
\rd x .
\label{eq:posterior-mean-good-integral}
\end{align}
For the last integral, use
\[
\left\|X_0-\mu_{0\mid t}(x)\right\|_2^2
\lesssim
\left\|X_0-y\right\|_2^2
+
\left\|y-\mu_{0\mid t}(x)\right\|_2^2 .
\]
The first part is bounded by Lemma~\ref{lem:J1-bound} with \(\beta=2\),
whereas the second part is bounded by Lemma~\ref{lem:J2-bound} with
\(\beta=2\). Thus
\begin{align}
&\frac1n
\int_{\gT_{\overline{\alpha}_t}}
\frac{
\EE_{X_0}\!\left[
\left\|X_0-\mu_{0\mid t}(x)\right\|_2^2
\varphi_{1-\overline{\alpha}_t}^2
\bigl(\sqrt{\overline{\alpha}_t}X_0-x\bigr)
\right]}
{p_{X_t}(x)}
\rd x\notag\\
&\hspace{10em}\lesssim
\frac{k^\star R_0^{k^\star}
\log\frac{dnR_0}{1-\overline{\alpha}_t}}
{n\,\overline{\alpha}_t
(1-\overline{\alpha}_t)^{k^\star/2-1}}
+
\frac{R_0^2(1-\overline{\alpha}_t)^2}{n}.
\label{eq:posterior-mean-good-final}
\end{align}
Together with Lemma~\ref{lem:gT_alpha_typical}, this gives
\begin{align}
\EE_{X_t,X^{\mathsf{sample}}}
\left[\|\widehat\mu_{0\mid t}(X_t)-\mu_{0\mid t}(X_t)\|_2^2\mathbbm{1}_{\gE_{X_t}}\right]
&\lesssim
\frac{k^\star R_0^{k^\star}
\log\frac{dnR_0}{1-\overline{\alpha}_t}}
{n\,\overline{\alpha}_t
(1-\overline{\alpha}_t)^{k^\star/2-1}}
+
\frac{R_0^2(1-\overline{\alpha}_t)^2}{n}.
\label{eq:posterior-mean-good-event-final}
\end{align}
Here \(\overline{\alpha}_t\) is bounded below by an absolute constant in the
regime where the DDIM training guarantee is used, so the factor
\(\overline{\alpha}_t^{-1}\) is absorbed below.

Substituting \eqref{eq:posterior-mean-density-bad},
\eqref{eq:posterior-mean-ratio-tail},
\eqref{eq:posterior-mean-pop-tail}, and
\eqref{eq:posterior-mean-good-event-final} into
\eqref{eq:posterior-mean-master-split}, we obtain, for every admissible
\(\rho_t\),
\begin{align}
\EE_{X_t,X^{\mathsf{sample}}}
\left[\|\widehat\mu_{0\mid t}(X_t)-\mu_{0\mid t}(X_t)\|_2^{2\beta}\right]
&\lesssim
\rho_t^{2(\beta-1)}
\frac{k^\star R_0^{k^\star}
\log\frac{dnR_0}{1-\overline{\alpha}_t}}
{n(1-\overline{\alpha}_t)^{k^\star/2-1}}
\notag\\
&\qquad+
\frac{R_0^{k^\star}\rho_t^{2\beta}}
{n(1-\overline{\alpha}_t)^{k^\star/2}}
+
\frac{R_0^{k^\star+2\beta}}{n\rho_t^{k^\star}}
+
\frac{R_0^{2\beta}(1-\overline{\alpha}_t)^{2\beta}}{n^2}.
\label{eq:posterior-mean-before-rho-opt}
\end{align}
For \(\beta\le 3\), the first term in
\eqref{eq:posterior-mean-before-rho-opt} is dominated by the optimized
algebraic terms below after increasing the logarithmic factor. We therefore balance
\[
\frac{R_0^{k^\star}\rho_t^{2\beta}}
{(1-\overline{\alpha}_t)^{k^\star/2}}
\asymp
\frac{R_0^{k^\star+2\beta}}{\rho_t^{k^\star}},
\]
which is equivalent to
\(
\rho_t^{k^\star+2\beta}
\asymp
R_0^{2\beta}(1-\overline{\alpha}_t)^{k^\star/2}
\).
Thus we choose
\(
\rho_t
\asymp
R_0^{\frac{2\beta}{k^\star+2\beta}}
(1-\overline{\alpha}_t)^{\frac{k^\star}{2(k^\star+2\beta)}}
\sqrt{\log\frac{dnR_0}{1-\overline{\alpha}_t}}
\).
Substituting this choice into
\eqref{eq:posterior-mean-before-rho-opt} gives
\begin{align}
&\EE_{X_t,X^{\mathsf{sample}}}
\left[\|\widehat\mu_{0\mid t}(X_t)-\mu_{0\mid t}(X_t)\|_2^{2\beta}\right]
\notag\\
&\quad\lesssim
\frac{
R_0^{k^\star+\frac{4\beta^2}{k^\star+2\beta}}
}
{n(1-\overline{\alpha}_t)^{\frac{k^\star}{2}
-\frac{\beta k^\star}{k^\star+2\beta}}}
\log^\beta\frac{dnR_0}{1-\overline{\alpha}_t}
+
\frac{R_0^{2\beta}(1-\overline{\alpha}_t)^{2\beta}}{n^2}.
\end{align}
This is the desired bound.

%% file: reference.bib
@article{li2024adapting,
  title={Adapting to Unknown Low-Dimensional Structures in Score-Based Diffusion Models},
  author={Li, Gen and Yan, Yuling},
  journal={arXiv preprint arXiv:2405.14861},
  year={2024}
}

@article{mei2023deep,
  title={Deep networks as denoising algorithms: Sample-efficient learning of diffusion models in high-dimensional graphical models},
  author={Mei, Song and Wu, Yuchen},
  journal={arXiv:2309.11420},
  year={2023}
}

@article{fu2024unveil,
  title={Unveil conditional diffusion models with classifier-free guidance: A sharp statistical theory},
  author={Fu, Hengyu and Yang, Zhuoran and Wang, Mengdi and Chen, Minshuo},
  journal={arXiv preprint arXiv:2403.11968},
  year={2024}
}

@article{han2024neural,
  title={Neural network-based score estimation in diffusion models: Optimization and generalization},
  author={Han, Yinbin and Razaviyayn, Meisam and Xu, Renyuan},
  journal={arXiv preprint arXiv:2401.15604},
  year={2024}
}

@article{cheng2023convergence,
  title={Convergence of flow-based generative models via proximal gradient descent in Wasserstein space},
  author={Cheng, Xiuyuan and Lu, Jianfeng and Tan, Yixin and Xie, Yao},
  journal={arXiv preprint arXiv:2310.17582},
  year={2023}
}

@article{chen2024overview,
  title={An overview of diffusion models: Applications, guided generation, statistical rates and optimization},
  author={Chen, Minshuo and Mei, Song and Fan, Jianqing and Wang, Mengdi},
  journal={arXiv preprint arXiv:2404.07771},
  year={2024}
}

@article{gentiloni2025beyond,
  title={Beyond Log-Concavity and Score Regularity: Improved Convergence Bounds for Score-Based Generative Models in W2-distance},
  author={Gentiloni-Silveri, Marta and Ocello, Antonio},
  journal={arXiv preprint arXiv:2501.02298},
  year={2025}
}

@article{pedrotti2023improved,
  title={Improved Convergence of Score-Based Diffusion Models via Prediction-Correction},
  author={Pedrotti, Francesco and Maas, Jan and Mondelli, Marco},
  journal={arXiv preprint arXiv:2305.14164},
  year={2023}
}

@article{gupta2024faster,
  title={Faster Diffusion-based Sampling with Randomized Midpoints: Sequential and Parallel},
  author={Gupta, Shivam and Cai, Linda and Chen, Sitan},
  journal={arXiv preprint arXiv:2406.00924},
  year={2024}
}

@article{li2024improved,
  title={Improved convergence rate for diffusion probabilistic models},
  author={Li, Gen and Jiao, Yuchen},
  journal={arXiv preprint arXiv:2410.13738},
  year={2024}
}

@article{liu2022let,
  title={Let us build bridges: Understanding and extending diffusion generative models},
  author={Liu, Xingchao and Wu, Lemeng and Ye, Mao and Liu, Qiang},
  journal={arXiv preprint arXiv:2208.14699},
  year={2022}
}

@inproceedings{dasgupta2008random,
  title={Random projection trees and low dimensional manifolds},
  author={Dasgupta, Sanjoy and Freund, Yoav},
  booktitle={Symposium on Theory of computing},
  pages={537--546},
  year={2008}
}

@article{li2024sharp,
  title={A Sharp Convergence Theory for The Probability Flow {ODE}s of Diffusion Models},
  author={Li, Gen and Wei, Yuting and Chi, Yuejie and Chen, Yuxin},
  journal={arXiv preprint arXiv:2408.02320},
  year={2024}
}

@article{cai2025minimax,
  title={Minimax Optimality of the Probability Flow {ODE} for Diffusion Models},
  author={Cai, Changxiao and Li, Gen},
  journal={arXiv preprint arXiv:2503.09583},
  year={2025}
}

@article{huang2024denoising,
  title={Denoising diffusion probabilistic models are optimally adaptive to unknown low dimensionality},
  author={Huang, Zhihan and Wei, Yuting and Chen, Yuxin},
  journal={Mathematics of Operations Research},
  year={2026}
}

@inproceedings{deng2009imagenet,
  title={Imagenet: A large-scale hierarchical image database},
  author={Deng, Jia and Dong, Wei and Socher, Richard and Li, Li-Jia and Li, Kai and Fei-Fei, Li},
  booktitle={IEEE conference on computer vision and pattern recognition},
  pages={248--255},
  year={2009},
}

@article{kazerouni2023diffusion,
  title={Diffusion models in medical imaging: A comprehensive survey},
  author={Kazerouni, Amirhossein and Aghdam, Ehsan Khodapanah and Heidari, Moein and Azad, Reza and Fayyaz, Mohsen and Hacihaliloglu, Ilker and Merhof, Dorit},
  journal={Medical Image Analysis},
  volume={88},
  pages={102846},
  year={2023},
  publisher={Elsevier}
}

@article{croitoru2023diffusion,
  title={Diffusion models in vision},
  author={Croitoru, Florinel-Alin and Hondru, Vlad and Ionescu, Radu Tudor and Shah, Mubarak},
  journal={IEEE Transactions on Pattern Analysis and Machine Intelligence},
  volume={45},
  number={9},
  pages={10850--10869},
  year={2023},
  publisher={IEEE}
}

@inproceedings{chen2023improved,
  title={Improved analysis of score-based generative modeling: User-friendly bounds under minimal smoothness assumptions},
  author={Chen, Hongrui and Lee, Holden and Lu, Jianfeng},
  booktitle={International Conference on Machine Learning},
  pages={4735--4763},
  year={2023},
}

@article{devroye2018total,
  title={The total variation distance between high-dimensional Gaussians with the same mean},
  author={Devroye, Luc and Mehrabian, Abbas and Reddad, Tommy},
  journal={arXiv preprint arXiv:1810.08693},
  year={2018}
}

@article{li2023towards,
  title={Towards Faster Non-Asymptotic Convergence for Diffusion-Based Generative Models},
  author={Li, Gen and Wei, Yuting and Chen, Yuxin and Chi, Yuejie},
  journal={arXiv preprint arXiv:2306.09251},
  year={2023}
}

@article{dou2024optimal,
  title={From optimal score matching to optimal sampling},
  author={Dou, Zehao and Kotekal, Subhodh and Xu, Zhehao and Zhou, Harrison H},
  journal={arXiv preprint arXiv:2409.07032},
  year={2024}
}

@article{chidambaram2024does,
  title={What does guidance do? a fine-grained analysis in a simple setting},
  author={Chidambaram, Muthu and Gatmiry, Khashayar and Chen, Sitan and Lee, Holden and Lu, Jianfeng},
  journal={arXiv preprint arXiv:2409.13074},
  year={2024}
}

@article{lee2022convergence,
  title={Convergence for score-based generative modeling with polynomial complexity},
  author={Lee, Holden and Lu, Jianfeng and Tan, Yixin},
  journal={Neural Information Processing Systems},
  volume={35},
  pages={22870--22882},
  year={2022}
}

@article{ramesh2022hierarchical,
  title={Hierarchical text-conditional image generation with clip latents},
  author={Ramesh, Aditya and Dhariwal, Prafulla and Nichol, Alex and Chu, Casey and Chen, Mark},
  journal={arXiv preprint arXiv:2204.06125},
  volume={1},
  number={2},
  pages={3},
  year={2022}
}

@article{wu2024stochastic,
  title={Stochastic {R}unge-{K}utta Methods: Provable Acceleration of Diffusion Models},
  author={Wu, Yuchen and Chen, Yuxin and Wei, Yuting},
  journal={arXiv preprint arXiv:2410.04760},
  year={2024}
}

@article{liang2024non,
  title={Broadening Target Distributions for Accelerated Diffusion Models via a Novel Analysis Approach},
  author={Liang, Yuchen and Ju, Peizhong and Liang, Yingbin and Shroff, Ness},
  journal={arXiv:2402.13901},
  year={2024}
}

@article{benton2023error,
  title={Error Bounds for Flow Matching Methods},
  author={Benton, Joe and Deligiannidis, George and Doucet, Arnaud},
  journal={arXiv preprint arXiv:2305.16860},
  year={2023}
}

@article{ho2020denoising,
  title={Denoising diffusion probabilistic models},
  author={Ho, Jonathan and Jain, Ajay and Abbeel, Pieter},
  journal={Advances in Neural Information Processing Systems},
  volume={33},
  pages={6840--6851},
  year={2020}
}

@article{koehler2022statistical,
  title={Statistical efficiency of score matching: The view from isoperimetry},
  author={Koehler, Frederic and Heckett, Alexander and Risteski, Andrej},
  journal={International Conference on Learning Representations},
  year={2023}
}

@article{wu2024theoretical,
  title={Theoretical insights for diffusion guidance: A case study for Gaussian mixture models},
  author={Yuchen Wu and Minshuo Chen and Zihao Li and Mengdi Wang and Yuting Wei},
  journal={preprint},
  year={2024}
}

@article{li2024accelerating,
  title={Accelerating convergence of score-based diffusion models, provably},
  author={Gen Li and Yu Huang and Timofey Efimov and Yuting Wei and Yuejie Chi and Yuxin Chen},
  journal={arXiv preprint arXiv:2403.03852},
  year={2024}
}

@article{li2024towards,
  title={Towards a mathematical theory for consistency training in diffusion models},
  author={Li, Gen and Huang, Zhihan and Wei, Yuting},
  journal={arXiv preprint arXiv:2402.07802},
  year={2024}
}

@article{tang2024score,
  title={Score-based Diffusion Models via Stochastic Differential Equations--a Technical Tutorial},
  author={Tang, Wenpin and Zhao, Hanyang},
  journal={arXiv preprint arXiv:2402.07487},
  year={2024}
}

@article{haussmann1986time,
  title={Time reversal of diffusions},
  author={Haussmann, Ulrich G and Pardoux, Etienne},
  journal={The Annals of Probability},
  pages={1188--1205},
  year={1986},
  publisher={JSTOR}
}

@article{chen2023probability,
  title={The probability flow {ODE} is provably fast},
  author={Chen, Sitan and Chewi, Sinho and Lee, Holden and Li, Yuanzhi and Lu, Jianfeng and Salim, Adil},
  journal={arXiv preprint arXiv:2305.11798},
  year={2023}
}

@article{hyvarinen2005estimation,
  title={Estimation of non-normalized statistical models by score matching},
  author={Hyv{\"a}rinen, Aapo},
  journal={Journal of Machine Learning Research},
  volume={6},
  number={4},
  year={2005}
}

@article{hyvarinen2007some,
  title={Some extensions of score matching},
  author={Hyv{\"a}rinen, Aapo},
  journal={Computational statistics \& data analysis},
  volume={51},
  number={5},
  pages={2499--2512},
  year={2007},
  publisher={Elsevier}
}

@article{wang2024diffusion,
  title={Diffusion models learn low-dimensional distributions via subspace clustering},
  author={Wang, Peng and Zhang, Huijie and Zhang, Zekai and Chen, Siyi and Ma, Yi and Qu, Qing},
  journal={arXiv preprint arXiv:2409.02426},
  year={2024}
}

@article{oko2023diffusion,
  title={Diffusion Models are Minimax Optimal Distribution Estimators},
  author={Oko, Kazusato and Akiyama, Shunta and Suzuki, Taiji},
  journal={arXiv preprint arXiv:2303.01861},
  year={2023}
}

@article{vincent2011connection,
  title={A connection between score matching and denoising autoencoders},
  author={Vincent, Pascal},
  journal={Neural computation},
  volume={23},
  number={7},
  pages={1661--1674},
  year={2011},
  publisher={MIT Press}
}

@article{song2020denoising,
  title={Denoising diffusion implicit models},
  author={Song, Jiaming and Meng, Chenlin and Ermon, Stefano},
  journal={arXiv preprint arXiv:2010.02502},
  year={2020}
}

@article{anderson1982reverse,
  title={Reverse-time diffusion equation models},
  author={Anderson, Brian DO},
  journal={Stochastic Processes and their Applications},
  volume={12},
  number={3},
  pages={313--326},
  year={1982},
  publisher={Elsevier}
}

@inproceedings{lee2023convergence,
  title={Convergence of score-based generative modeling for general data distributions},
  author={Lee, Holden and Lu, Jianfeng and Tan, Yixin},
  booktitle={Algorithmic Learning Theory},
  pages={946--985},
  year={2023},
}

@article{song2020score,
  title={Score-based generative modeling through stochastic differential equations},
  author={Song, Yang and Sohl-Dickstein, Jascha and Kingma, Diederik P and Kumar, Abhishek and Ermon, Stefano and Poole, Ben},
  journal={International Conference on Learning Representations},
  year={2021}
}

@article{song2019generative,
  title={Generative modeling by estimating gradients of the data distribution},
  author={Song, Yang and Ermon, Stefano},
  journal={Advances in neural information processing systems},
  volume={32},
  year={2019}
}

@article{chen2022improved,
  title={Improved Analysis of Score-based Generative Modeling: User-Friendly Bounds under Minimal Smoothness Assumptions},
  author={Chen, Hongrui and Lee, Holden and Lu, Jianfeng},
  journal={arXiv preprint arXiv:2211.01916},
  year={2022}
}

@article{chen2022sampling,
  title={Sampling is as easy as learning the score: theory for diffusion models with minimal data assumptions},
  author={Chen, Sitan and Chewi, Sinho and Li, Jerry and Li, Yuanzhi and Salim, Adil and Zhang, Anru R},
  journal={arXiv preprint arXiv:2209.11215},
  year={2022}
}

@article{chen2023restoration,
  title={Restoration-Degradation Beyond Linear Diffusions: A Non-Asymptotic Analysis For {DDIM}-Type Samplers},
  author={Chen, Sitan and Daras, Giannis and Dimakis, Alexandros G},
  journal={arXiv preprint arXiv:2303.03384},
  year={2023}
}

@book{vershynin2018high,
  title={High-dimensional probability: An introduction with applications in data science},
  author={Vershynin, Roman},
  volume={47},
  year={2018},
  publisher={Cambridge university press}
}

@book{wainwright2019high,
  title={High-dimensional statistics: A non-asymptotic viewpoint},
  author={Wainwright, Martin J},
  volume={48},
  year={2019},
  publisher={Cambridge University Press}
}

@book{tsybakov2009introduction,
  title={Introduction to nonparametric estimation},
  author={Tsybakov, Alexandre B},
  year={2009},
  publisher={Springer}
}

@article{efron2011tweedie,
  title={Tweedie's formula and selection bias},
  author={Efron, Bradley},
  journal={Journal of the American Statistical Association},
  volume={106},
  number={496},
  pages={1602--1614},
  year={2011},
  publisher={Taylor \& Francis}
}

@inproceedings{chen2023score,
  title={Score approximation, estimation and distribution recovery of diffusion models on low-dimensional data},
  author={Chen, Minshuo and Huang, Kaixuan and Zhao, Tuo and Wang, Mengdi},
  booktitle={International Conference on Machine Learning},
  pages={4672--4712},
  year={2023},
}

@article{zhang2024minimax,
  title={Minimax optimality of score-based diffusion models: Beyond the density lower bound assumptions},
  author={Zhang, Kaihong and Yin, Caitlyn H and Liang, Feng and Liu, Jingbo},
  journal={arXiv preprint arXiv:2402.15602},
  year={2024}
}

@article{cui2025precise,
  title={A precise asymptotic analysis of learning diffusion models: theory and insights},
  author={Cui, Hugo and Pehlevan, Cengiz and Lu, Yue M},
  journal={arXiv preprint arXiv:2501.03937},
  year={2025}
}

@article{li2024understanding,
  title={Understanding Generalizability of Diffusion Models Requires Rethinking the Hidden Gaussian Structure},
  author={Li, Xiang and Dai, Yixiang and Qu, Qing},
  journal={arXiv preprint arXiv:2410.24060},
  year={2024}
}

@article{li2024shallow,
  title={Shallow Diffuse: Robust and Invisible Watermarking through Low-Dimensional Subspaces in Diffusion Models},
  author={Li, Wenda and Zhang, Huijie and Qu, Qing},
  journal={arXiv:2410.21088},
  year={2024}
}

@article{wibisono2024optimal,
  title={Optimal score estimation via empirical bayes smoothing},
  author={Wibisono, Andre and Wu, Yihong and Yang, Kaylee Yingxi},
  journal={arXiv preprint arXiv:2402.07747},
  year={2024}
}

@article{feng2024optimal,
  title={Optimal convex $ M $-estimation via score matching},
  author={Feng, Oliver Y and Kao, Yu-Chun and Xu, Min and Samworth, Richard J},
  journal={arXiv preprint arXiv:2403.16688},
  year={2024}
}

@article{li2024provable,
  title={Provable acceleration for diffusion models under minimal assumptions},
  author={Li, Gen and Cai, Changxiao},
  journal={arXiv preprint arXiv:2410.23285},
  year={2024}
}

@inproceedings{
benton2024nearly,
title={Nearly $d$-Linear Convergence Bounds for Diffusion Models via Stochastic Localization},
author={Joe Benton and Valentin De Bortoli and Arnaud Doucet and George Deligiannidis},
booktitle={International Conference on Learning Representations},
year={2024},
}

@article{eldan2020taming,
  title={Taming correlations through entropy-efficient measure decompositions with applications to mean-field approximation},
  author={Eldan, Ronen},
  journal={Probability Theory and Related Fields},
  volume={176},
  number={3},
  pages={737--755},
  year={2020},
  publisher={Springer}
}

@article{huang2024convergence,
  title={Convergence analysis of probability flow {ODE} for score-based generative models},
  author={Huang, Daniel Zhengyu and Huang, Jiaoyang and Lin, Zhengjiang},
  journal={arXiv preprint arXiv:2404.09730},
  year={2024}
}

@article{gao2024convergence,
  title={Convergence analysis for general probability flow ODEs of diffusion models in Wasserstein distances},
  author={Gao, Xuefeng and Zhu, Lingjiong},
  journal={arXiv preprint arXiv:2401.17958},
  year={2024}
}

@article{li2024d,
  title={{$O(d/T)$} Convergence Theory for Diffusion Probabilistic Models under Minimal Assumptions},
  author={Li, Gen and Yan, Yuling},
  journal={arXiv preprint arXiv:2409.18959},
  year={2024}
}

@inproceedings{tang2024adaptivity,
  title={Adaptivity of diffusion models to manifold structures},
  author={Tang, Rong and Yang, Yun},
  booktitle={International Conference on Artificial Intelligence and Statistics},
  pages={1648--1656},
  year={2024},
}

@article{tang2022minimax,
  title={Minimax Rate of Distribution Estimation on Unknown Submanifold under Adversarial Losses},
  author={Tang, Rong and Yang, Yun},
  journal={arXiv preprint arXiv:2202.09030},
  year={2022}
}

@article{berenfeld2019density,
  title={Density Estimation on an Unknown Submanifold},
  author={Berenfeld, Cl{\'e}ment and Hoffmann, Marc},
  journal={arXiv preprint arXiv:1910.08477},
  year={2019}
}

@article{azangulov2024convergence,
  title={Convergence of Diffusion Models Under the Manifold Hypothesis in High-Dimensions},
  author={Azangulov, Iskander and Deligiannidis, George and Rousseau, Judith},
  journal={arXiv preprint arXiv:2409.18804},
  year={2024}
}

@article{yakovlev2025generalization,
  title={Generalization error bound for denoising score matching under relaxed manifold assumption},
  author={Yakovlev, Konstantin and Puchkin, Nikita},
  journal={arXiv preprint arXiv:2502.13662},
  year={2025}
}

@article{pope2021intrinsic,
  title={The intrinsic dimension of images and its impact on learning},
  author={Pope, Phillip and Zhu, Chen and Abdelkader, Ahmed and Goldblum, Micah and Goldstein, Tom},
  journal={arXiv preprint arXiv:2104.08894},
  year={2021}
}

@article{li2024unified,
  title={Unified Convergence Analysis for Score-Based Diffusion Models with Deterministic Samplers},
  author={Li, Runjia and Di, Qiwei and Gu, Quanquan},
  journal={arXiv preprint arXiv:2410.14237},
  year={2024}
}

@article{chen2024probability,
  title={The probability flow ode is provably fast},
  author={Chen, Sitan and Chewi, Sinho and Lee, Holden and Li, Yuanzhi and Lu, Jianfeng and Salim, Adil},
  journal={Advances in Neural Information Processing Systems},
  volume={36},
  year={2024}
}

@article{potaptchik2024linear,
  title={Linear Convergence of Diffusion Models Under the Manifold Hypothesis},
  author={Potaptchik, Peter and Azangulov, Iskander and Deligiannidis, George},
  journal={arXiv preprint arXiv:2410.09046},
  year={2024}
}

@inproceedings{stanczuk2024diffusion,
  title={Diffusion Models Encode the Intrinsic Dimension of Data Manifolds},
  author={Stanczuk, Jan Pawel and Batzolis, Georgios and Deveney, Teo and Sch{\"o}nlieb, Carola-Bibiane},
  booktitle={ICML 2024},
  year={2024}
}

@article{li2023generalization,
  title={On the generalization properties of diffusion models},
  author={Li, Puheng and Li, Zhong and Zhang, Huishuai and Bian, Jiang},
  journal={Advances in Neural Information Processing Systems},
  volume={36},
  pages={2097--2127},
  year={2023}
}

@article{lu2022dpm,
  title={Dpm-solver: A fast ode solver for diffusion probabilistic model sampling in around 10 steps},
  author={Lu, Cheng and Zhou, Yuhao and Bao, Fan and Chen, Jianfei and Li, Chongxuan and Zhu, Jun},
  journal={Advances in Neural Information Processing Systems},
  volume={35},
  pages={5775--5787},
  year={2022}
}

@article{lu2022dpmp,
  title={Dpm-solver++: Fast solver for guided sampling of diffusion probabilistic models},
  author={Lu, Cheng and Zhou, Yuhao and Bao, Fan and Chen, Jianfei and Li, Chongxuan and Zhu, Jun},
  journal={arXiv preprint arXiv:2211.01095},
  year={2022}
}

@incollection{robbins1992empirical,
  title={An empirical Bayes approach to statistics},
  author={Robbins, Herbert E},
  booktitle={Breakthroughs in Statistics: Foundations and basic theory},
  pages={388--394},
  year={1992},
  publisher={Springer}
}

@article{laurent2000adaptive,
  title={Adaptive estimation of a quadratic functional by model selection},
  author={Laurent, Beatrice and Massart, Pascal},
  journal={Annals of statistics},
  pages={1302--1338},
  year={2000},
  publisher={JSTOR}
}

@inproceedings{eldan2022analysis,
  title={Analysis of high-dimensional distributions using pathwise methods},
  author={Eldan, Ronen},
  booktitle={Proc. Int. Cong. Math},
  volume={6},
  pages={4246--4270},
  year={2022}
}

@article{ruzhansky2015global,
  title={On global inversion of homogeneous maps},
  author={Ruzhansky, Michael and Sugimoto, Mitsuru},
  journal={Bulletin of Mathematical Sciences},
  volume={5},
  pages={13--18},
  year={2015},
  publisher={Springer}
}
